\documentclass{article}

\usepackage[final]{neurips_2025}

\usepackage[utf8]{inputenc} 
\usepackage[T1]{fontenc}    
\usepackage{hyperref}       
\usepackage{url}            
\usepackage{booktabs}       
\usepackage{amsfonts}       
\usepackage{nicefrac}       
\usepackage{microtype}      
\usepackage{xcolor}         
\usepackage{graphicx} 
\usepackage{subcaption}
\usepackage{amsmath}
\usepackage{pifont}
\usepackage[dvipsnames]{xcolor}
\usepackage{tabularx}
\usepackage{amsthm}
\newtheorem{theorem}{Theorem}
\usepackage[framemethod=TikZ]{mdframed}
\usepackage{multirow}

\title{Impact of Layer Norm on Memorization and Generalization in Transformers}

\author{%
  Rishi Singhal\\
  Department of Computer Science\\
  North Carolina State University\\
  \And
  Jung-Eun Kim\thanks{Corresponding author} \\
  Department of Computer Science\\
  North Carolina State University\\
}

\begin{document}

\maketitle

\begin{abstract}
Layer Normalization (LayerNorm) is one of the fundamental components in transformers that stabilizes training and improves optimization. In recent times, Pre-LayerNorm transformers have become the preferred choice over Post-LayerNorm transformers due to their stable gradient flow. However, the impact of LayerNorm on learning and memorization across these architectures remains unclear. In this work, we investigate how LayerNorm influences memorization and learning for Pre- and Post-LayerNorm transformers. We identify that LayerNorm serves as a key factor for stable learning in Pre-LayerNorm transformers, while in Post-LayerNorm transformers, it impacts memorization. Our analysis reveals that eliminating LayerNorm parameters in Pre-LayerNorm models exacerbates memorization and destabilizes learning, while in Post-LayerNorm models, it effectively mitigates memorization by restoring genuine labels. We further precisely identify that early layers LayerNorm are the most critical over middle/later layers and their influence varies across Pre and Post LayerNorm models. We have validated it through 13 models across 6 Vision and Language datasets. These insights shed new light on the role of LayerNorm in shaping memorization and learning in transformers\footnote{Code available at: \url{https://github.com/JEKimLab/NeurIPS2025_LayernormMemorization}}.
\end{abstract}

\vspace{-1ex}
\section{Introduction}

\begin{figure}[t]
    \centering
    \begin{subfigure}{0.4\textwidth}
        \centering
        \includegraphics[width=\textwidth]{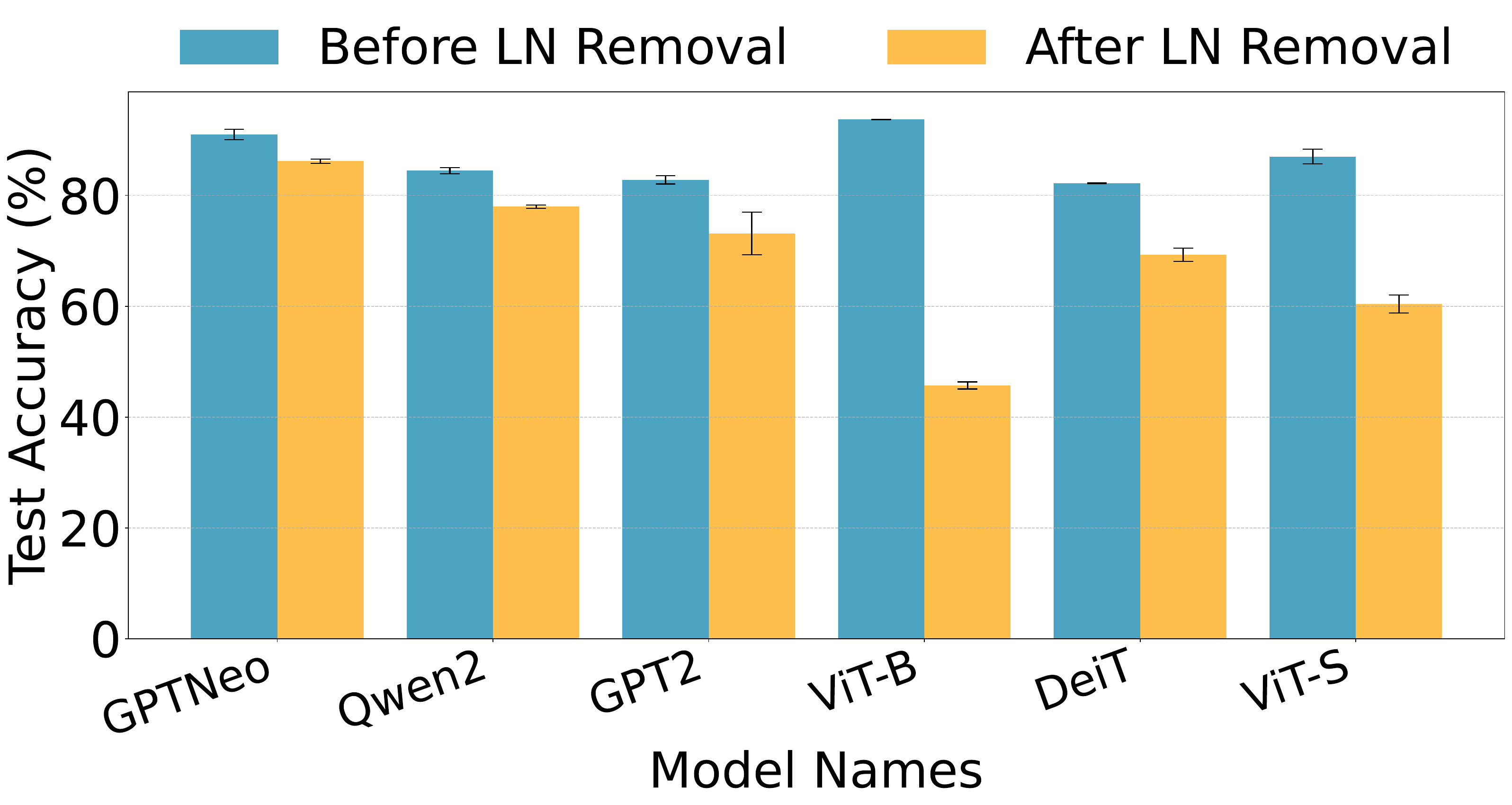}
        \caption{Learning (test) accuracy for Pre-LN models}
        \label{fig:pre_ln_learning}
        
    \end{subfigure}
    \hspace{5pt}
    \begin{subfigure}{0.38\textwidth}
        \centering
        \includegraphics[width=\textwidth]{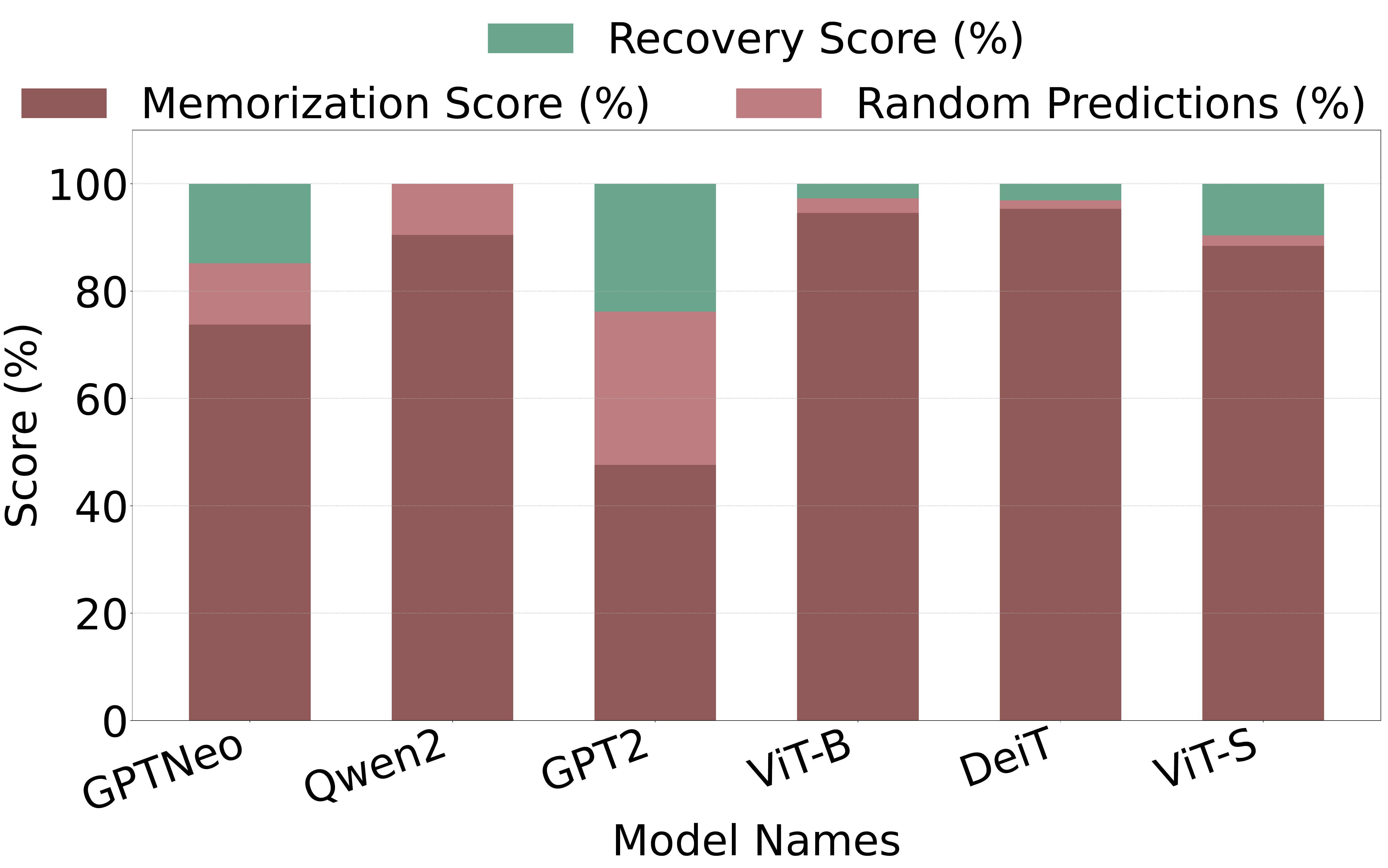}
        \caption{Memorization, Recovery \& Random prediction scores for Pre-LN models}
        \label{fig:pre_ln_memorization}
    \end{subfigure}

    \vspace{5pt} 

    \begin{subfigure}{0.4\textwidth}
        \centering
        \includegraphics[width=\textwidth]{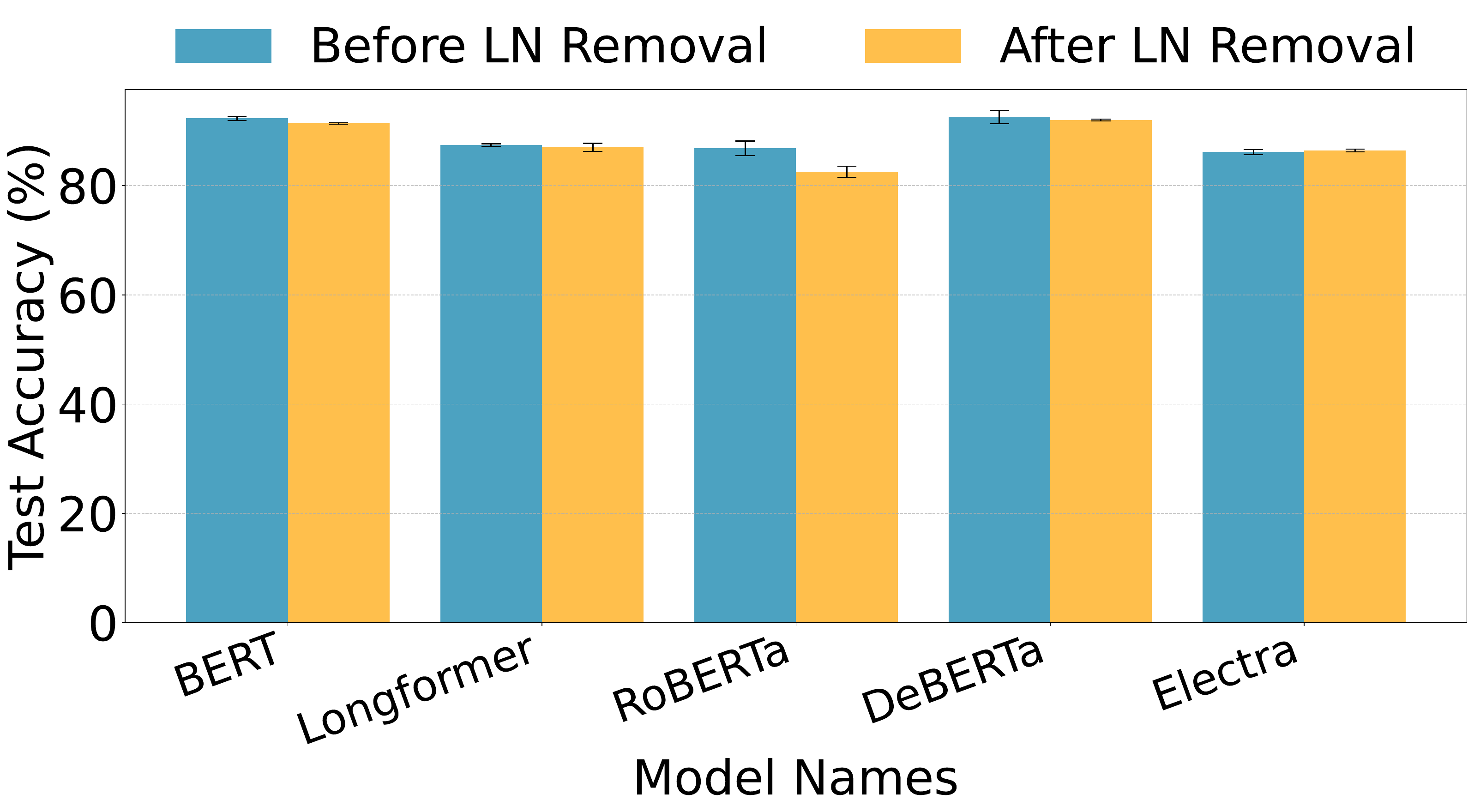}
        \caption{Learning (test) accuracy for Post-LN models}
        \label{fig:post_ln_learning}
    \end{subfigure}
    \hspace{5pt}
    \begin{subfigure}{0.38\textwidth}
        \centering
        \includegraphics[width=\textwidth]{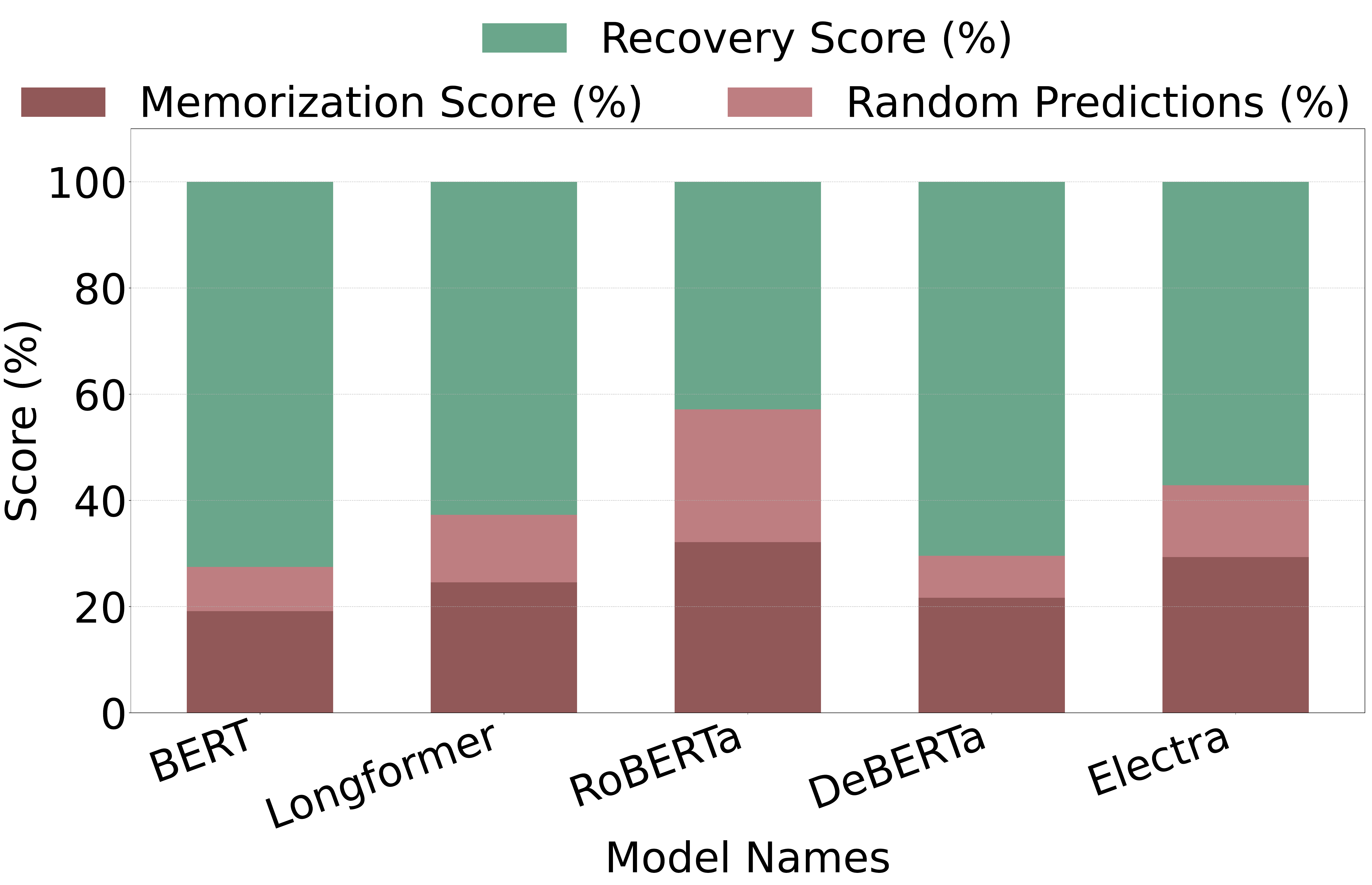}
        \caption{Memorization, Recovery \& Random prediction scores for Post-LN models}
        \label{fig:post_ln_memorization}
    \end{subfigure}

    \caption{\textbf{Impact of LN layer on memorization and learning of Pre- and Post-LN models.} (a) shows a clear impact of LN in Pre-LN models, whereas (c) shows no impact of the removal of LN parameters in Post-LN models for learning. (b) exhibits that, without LN layers, the Pre-LN models struggle with high memorization and random predictions (\textcolor{BrickRed}{red}-color-family bars), while (d) exhibits that in Post-LN models, removing LN parameters recovers a significant portion of correct predictions (\textcolor{Green}{green} bars).}
    \label{fig:learning_mem_recovery}
\end{figure}
\vspace{-1.5ex}

Layer Normalization \citep{lei2016layer} greatly contributes to stabilizing training and optimizing performance in deep learning models, especially in transformers. It works by normalizing the activations at each layer, ensuring a more consistent gradient flow during training. In recent years, transformers are primarily designed with two options by LayerNorm (LN) placements: \emph{Pre-LN} and \emph{Post-LN}. Post-LN transformer, introduced by \citet{vaswani2017attention}, applies normalization after the addition of the layer's output with the residual connection's output and has been showing competent performance in language modeling and machine translation. However, due to the issue of unstable gradient flow \citep{liu2020understanding} in Post-LN models, Pre-LN transformers \citep{xiong2020layer} were introduced, where normalization is applied before self-attention and feed-forward layers. This configuration stabilized training and achieved faster convergence by improving the gradient flow, making it the preferred choice in modern architectures such as GPT, Llama, and Vision Transformers.

Even though transformers have demonstrated remarkable capabilities in learning rich representations from data, they exhibit a strong tendency to memorize some samples due to their complex nature, which is commonly known as \emph{Label Memorization} \citep{feldman2020does, feldman2020neural}, where a model memorizes labels in training without learning the relevant patterns that generalize to unseen data, leading to overfitting. Recent studies have explored whether memorization can be localized to specific layers \citep{maini2023memorization, baldock2021deep} or components such as attention heads and the feed-forward network (FFN) \citep{haviv2023understanding, geva2023dissecting, yu2023characterizing}. Apart from the attention heads and the feed-forward network (FFN), LayerNorm stands as a pivotal component in the transformer architecture that further shapes its optimization dynamics and performance. 
Outlier neurons in LN layers have been shown to impair transformer performance and hinder quantization \citep{kovaleva2021bert, puccetti2022outliers, bondarenko2023quantizable, he2024understanding}. \citep{xu2019understanding} hints that LN may contribute to increased overfitting in Pre-LN models.

In our paper, we identify that in Pre-LN transformers, LN is critical to \emph{learning} and removal of its learnable parameters exacerbates overfitting and disrupts learning. On the contrary, in Post-LN transformers, LN plays a significant role in \emph{memorization}, where by eliminating LN parameters, memorization is suppressed by recovering the true genuine labels without affecting the model's learning capability. We rigorously validate our claims through various models - BERT, Longformer, RoBERTa, DeBERTa, ELECTRA, DistilBERT, GPTNeo, GPT2, Qwen2, RoBERTa-PreLayerNorm, ViT-Base, ViT-Small, and DeiT, across both Vision and Language tasks.
In summary, the core findings of our paper regarding the impact of LN on memorization and learning in transformers are as follows:
\vspace{-3.5ex}
\begin{itemize}
    \item \textbf{Learning Stability, Memorization Suppression \& Label Recovery:} We identify that LN is crucial for learning in Pre-LN models, unlike Post-LN models. For Post-LN models, LN learnable parameters removal suppresses memorization and recovers genuine labels, whereas in Pre-LN models, LN removal exacerbates overfitting, with persistent memorization.
    \item \textbf{Early LNs are Critical:} We uncover that removal of LNs parameters in early layers is most impactful in mitigating memorization for Post-LN models, and destabilizing the learning in Pre-LN architectures.
    \item \textbf{Gradients Explain LN's Impact:} We explain the divergent impacts of LN in Pre- and Post-LN models by comparing learning and memorization gradients, which reveal why LN parameter removal causes learning disruption and memorization suppression in Pre- and Post-LN models, respectively.
\vspace{-2ex}
\end{itemize}
\vspace{-1ex}
\section{Related Works}
\vspace{-1ex}

\textbf{Memorization \& Learning:} Transformers excel at learning general, simple patterns \citep{arpit2017closer, shah2020pitfalls,zhou2023samples}, but also tend to memorize rare, mislabeled, or complex examples \citep{stephenson2021geometry,baldock2021deep,agarwal2022estimating}. \citet{feldman2020neural, feldman2020does} formally define label memorization, while \citet{baldock2021deep} proposes prediction depth to capture example difficulty. Other works \citep{jiang2020characterizing,ravikumar2024unveiling, garg2023memorization} associate high curvature and consistency with long-tailed or mislabeled samples. Beyond identifying memorization, several studies \citep{haviv2023understanding, geva2023dissecting, dai2021knowledge} investigate how self-attention and feedforward layers contribute to factual recall across transformer layers. More recent work \citep{yin2023outlier, lad2024remarkable, men2024shortgpt, li2024mix, sun2025curse} highlights the limited effectiveness of deeper layers on learning in Pre-LN transformers. Despite these insights, the distinctive impact of LayerNorm in shaping memorization and learning across both Pre- and Post-LN architectures remains poorly understood.

\textbf{Understanding LayerNorm (LN) in Transformers:} In addition to self-attention and feedforward networks (FFNs), Layer Normalization (LN) plays a critical role in transformer models. Prior work \citep{brody2023expressivity, wu2024role} has demonstrated that LN is essential to the overall expressivity of transformers. Beyond its utility, LN has been found to contain outlier neurons \citep{kovaleva2021bert, puccetti2022outliers}, whose removal severely degrades model performance. These outliers have also been shown to hinder the quantization of transformer models \citep{bondarenko2023quantizable, he2024understanding}. Moreover, several studies \citep{xiong2020layer, liu2020understanding, takase2022b2t, xie2023residual, kim2025peri} have highlighted that Post-LN architectures can cause gradient instability during training, while Pre-LN configurations may lead to exploding gradients in early layers—prompting the development of techniques \citep{shleifer2021normformer,wang2022foundation,kumar2023dual, qi2023lipsformer,jiang2023pre} to address them. Additionally, \citet{xu2019understanding} suggested that LN parameters may contribute to overfitting in Pre-LN models.

However, we provide a far more nuanced understanding of LN's role: in Pre-LN transformers, LN is essential for learning but not memorization, whereas in Post-LN models, LN is crucial for memorization but not learning. This distinction offers a novel contribution to understanding the function of LN in transformers for learning and memorization.
\vspace{-1ex}
\section{Prelimnaries}
\vspace{-1ex}

\subsection{Understanding LayerNorm in Transformers and Defining Memorization \& Learning}
\underline{\textbf{LayerNorm Operation.}} Let \( x = (x_1, x_2, \dots, x_d) \) be the input of size $d$ to the LayerNorm function $\text{LN}(x)$ which first normalizes the input \(x\) as \( N(x) \) using mean \( \mu \) and standard deviation \( \sigma \). Then it re-scales and re-centers \( N(x) \) using the learnable parameters \( w \) (weight) and \( b \) (bias). The output of the LayerNorm layer is then given by:
\begin{equation}
\begin{aligned}
\text{LN}(x) &= w \odot N(x) + b, \quad 
\mu = \frac{1}{d} \sum_{i=1}^{d} x_i, \quad
\sigma = \sqrt{\frac{1}{d} \sum_{i=1}^{d} (x_i - \mu)^2}, \quad
N(x) = \frac{x - \mu}{\sigma},
\end{aligned}
\end{equation}
where \( \odot \) denotes the dot product operation.

\underline{\textbf{Pre-LN \& Post-LN Transformers.}} In the Pre-LN Transformer, LayerNorm is applied before each sub-layer - Multi-Head Self Attention (MHSA) and Feed-Forward Network (FFN). On the other hand, in the Post-LN Transformer, LN is applied after the residual connection. We represent the key difference in the architectural design of the two configurations as follows:
\begin{equation}
\begin{aligned}
\textbf{Pre-LN:} \quad &x' = x + \text{MHSA}(\text{LN}_1(x)) \quad &\textbf{Post-LN:} \quad &x' = \text{LN}_1(x + \text{MHSA}(x)) \\
&y = x' + \text{FFN}(\text{LN}_2(x')) & &y = \text{LN}_2(x' + \text{FFN}(x'))
\end{aligned}
\end{equation}

\vspace{-1ex}
\underline{\textbf{Understanding Learning and Label Memorization (LM)}.}
Deep neural network models, like transformers, learn meaningful relationships between features and labels during training and generalize the learned representations to unseen test data - the phenomenon is well understood as \textbf{\emph{learning/generalization}}. At the same time, these models also have the tendency to memorize training data points which are complex in nature, commonly known as \textbf{\emph{label memorization (LM)}} \citep{feldman2020does, feldman2020neural}, where the model just memorizes the labels during training without capturing meaningful patterns that generalize to new data, resulting in overfitting. Label memorization is known to occur due to multiple factors such as complex, ambiguous features, and noisy labels \citep{baldock2021deep}, which makes it difficult for the model to learn any meaningful relationship. 

In this work, we specifically focus on introducing \emph{noisy labels} as a way to study memorization, where we change the label of a particular class sample to a randomly chosen class label that is different from its original label. To ensure that the noisy label samples are memorized, we train the model until it achieves 100\% training accuracy. Throughout our experiments, we introduce random label noise in all datasets by modifying 1\% of the training set labels, maintaining consistency across evaluations.

\vspace{-1ex}
\subsection{Investigating LayerNorm (LN) Impact on Memorization and Learning} \label{subssec:ln_analysis}

\underline{\textbf{Removing LN parameters.}} To examine the role of LayerNorm (LN) in memorization and learning within transformers, we analyze the effects of omitting its learnable parameters, during training. This provides insights into how LN influences the balance between learning and memorization for both Pre- and Post-LN models. We precisely analyze the impact of LN on memorization and learning in Sec. \ref{sec:discrepancy_post_pre_ln}. Please note that we use \emph{LN removal} and \emph{LN parameters removal} interchangeably in our paper. They both refer to removal of learnable parameters of the LN layer, while keeping the normalization operation, $N(x)$, intact.

\underline{\textbf{Effect of LN Removal across Layers.}} To further understand the impact of LN at different stages of the model, we categorize the layers into - early, middle, and later layers (described in detail in Appendix~\ref{sec:grouping_early_middle_later}). We then selectively remove LN parameters from one set at a time to analyze how their absence affects learning and memorization. This analysis reveals which set of LNs is the most influential towards memorization and learning behaviors in Pre- and Post-LN transformers. The experiments and results are discussed in Sec. \ref{sec:early_layers_imp}.

\underline{\textbf{LN Gradients Analysis across Layers.}} To support our observations on the influence of LN, we compute the gradient of the loss function ($\mathcal{L}$) with respect to the input of LN, $x$, represented as \(\nabla_{x} \mathcal{L} \text{ or } g_x = \frac{\partial \mathcal{L}}{\partial x}\). This measure quantifies how much the input to the LN layer affects the model's loss, thereby its learning and memorization ability.

To understand the sensitivity of each layer's LN towards memorization and learning, we compute the L2-norm of this gradient (i.e., $\|g_x\|_2$). Specifically, to quantify sensitivity towards \emph{learning}, i.e., how the model generalizes the patterns to the test set, we compute $\|g_x\|_2$ for every test-set sample and average it across all of them, obtaining \textbf{\emph{learning gradient norm}}, denoted by $\|g_x^{\text{learn}}\|_2
$. For \emph{memorization}, we compute $||g_x||_2$ for each of the \emph{noisy labels} samples that we injected into the train set and then averaged across all such noisy samples to obtain \textbf{\emph{memorization gradient norm}}, denoted by $\|g_x^{\text{mem}}\|_2$. 

A higher gradient norm indicates that the layer’s LN significantly influences the model's ability to memorize or learn, while a lower gradient suggests minimal impact. The discussion of memorization and learning gradients and their significance is shown in Sec.~\ref{sec:learn_mem_grad}.

\vspace{-1ex}
\subsection{Key Metrics: Learning Accuracy,  Memorization, Recovery \& Random Predictions Score}
\vspace{-1ex}

To evaluate the impact of LN on the learning and memorization ability of the transformer models, we focus on several key metrics that provide insights into their behavior and effectiveness in the presence of noisy labels during training.

\underline{\textbf{Learning (Test) Accuracy (\%)}} refers to the model's performance on the test set, depicting how well it generalizes the learnt relationships to unseen data, marking it as a core indicator of the model's learning progress ($\tfrac{\# \text{Correct predictions on test set}}{\# \text{Total test set samples}} \times 100$). A high learning accuracy signifies that the model has learned meaningful patterns and is able to generalize well to unseen data. On the contrary, a low learning accuracy depicts poor generalizability. 

\underline{\textbf{Memorization Score (\%)}} serves as an indicator of the model's tendency to memorize noisy labels that are irrelevant or erroneous, rather than genuinely learning the true underlying relationships ($\tfrac{\# \text{Noisy label samples memorized}}{\# \text{Total noisy label samples}} \times 100$). A high memorization score indicates that the model has overfit the noisy labels, effectively ``memorizing'' them. 

\underline{\textbf{Recovery Score (\%)}} is a crucial metric that helps in understanding the impact of LayerNorm (LN) on memorization of noisy labels. It measures the model's ability to recover the genuine, true labels after the removal of LN parameters ($\tfrac{\# \text{Recovered noisy label samples as true labels}}{\# \text{Total noisy label samples}} \times 100
$). A high recovery score indicates that the model can successfully recover the original, correct labels by suppressing memorization.

\underline{\textbf{Random Prediction Score (\%)}} measures the percentage of noisy label samples whose predictions were changed to random labels after the removal of LN parameters. These predicted random labels are neither genuine nor the noisy label ($\tfrac{\# \text{Random predictions of noisy label samples}}{\# \text{Total noisy label samples}} \times 100$). Although this is not ideal, it provides a complete picture of the impact of LN parameters removal and indicates the extent to which the model can recover the true labels. A high percentage of random predictions suggests that the model struggles to recover the true labels effectively.

\subsection{Datasets \& Models Used}

We empirically verify all claims and show extensive results against both language and vision modalities, including 3 language and 3 vision classification datasets, and 7 Pre-LN and 6 Post-LN transformers architectures,  as follows:\\
\underline{\textbf{Datasets: }} CIFAR10 \citep{krizhevsky2009learning}, NICO++ \citep{zhang2023nico++}, UTK-Face \citep{zhang2017age}, Emotions \citep{saravia-etal-2018-carer}, News \citep{okite97_news_data}, and TweetTopic \citep{dimosthenis-etal-2022-twitter}\\
\underline{\textbf{Post-LN Models: }} BERT \citep{devlin2019bert}, RoBERTa \citep{yinhan2019roberta}, DistilBERT \citep{sanh2019distilbert}, DeBERTa \citep{he2020deberta}, ELECTRA \citep{clark2020electra}, and Longformer \citep{beltagy2020longformer}\\
\underline{\textbf{Pre-LN Models: }} ViT-B \citep{alexey2020image}, ViT-S \citep{assran2022masked}, DeiT \citep{touvron2021training}, GPT2 \citep{radford2019language}, GPT-Neo \citep{black2022gpt}, Qwen2 \citep{yang2024qwen2}, and RoBERTa-PreLayerNorm \citep{ott-etal-2019-fairseq}.

It needs to be acknowledged that only language modality is available for the Post-LN architecture in practice/literature. We provide a thorough discussion of the datasets, models, and training configurations in Appendix~\ref{sec:training_details}. All our experiments are run across 3 random seeds.

\section{Impact of LN on Memorization and Learning} \label{sec:discrepancy_post_pre_ln}

\begin{table}[t]
    \centering
    \small
    \caption{Summary of the impact of LN layer in Pre- and Post-LN Models.}
    \label{tab:ln_impact}
    \begin{tabular}{|c|c|c|c|}
        \hline
        \textbf{If LN Removed} & \textbf{Learning Intact?} & \textbf{Memorization Mitigated?} & \textbf{Recovery Happens?} \\
        \hline
        \textbf{Pre-LN Model} 
        & \textcolor{red}{\ding{55}} \emph{Learning Disrupted} 
        & \textcolor{red}{\ding{55}} \emph{Memorization Still Present} 
        & \textcolor{red}{\ding{55}} \emph{Negligible Recovery} \\
        \hline
        \textbf{Post-LN Model} 
        & \textcolor{green}{\ding{51}} \emph{Stable Learning} 
        & \textcolor{green}{\ding{51}} \emph{Memorization Mitigated} 
        & \textcolor{green}{\ding{51}} \emph{Genuine Labels Inferred} \\
        \hline
    \end{tabular}
\end{table}
\vspace{-1.5ex}

\begin{figure}[t]
    \centering
    \begin{subfigure}[t]{0.35\textwidth}
        \centering
        \includegraphics[width=\textwidth]{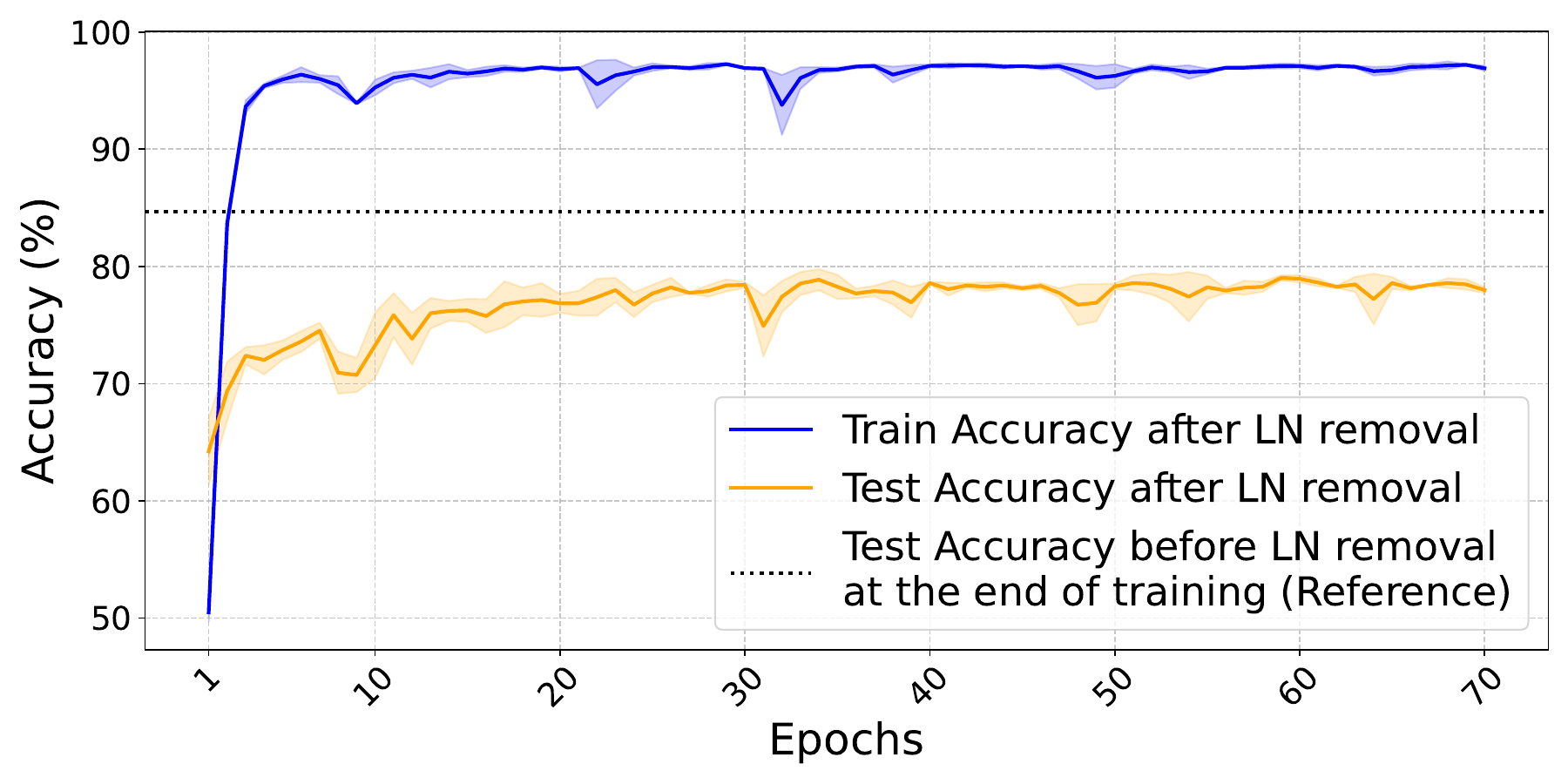}
        \caption{Learning (Test) accuracy over epochs for Pre-LN Model (Qwen2)}
        \label{fig:pre_ln_learning_epochs_qwen2}
    \end{subfigure}
    \hspace{5pt}
    \begin{subfigure}[t]{0.35\textwidth}
        \centering
        \includegraphics[width=\textwidth]{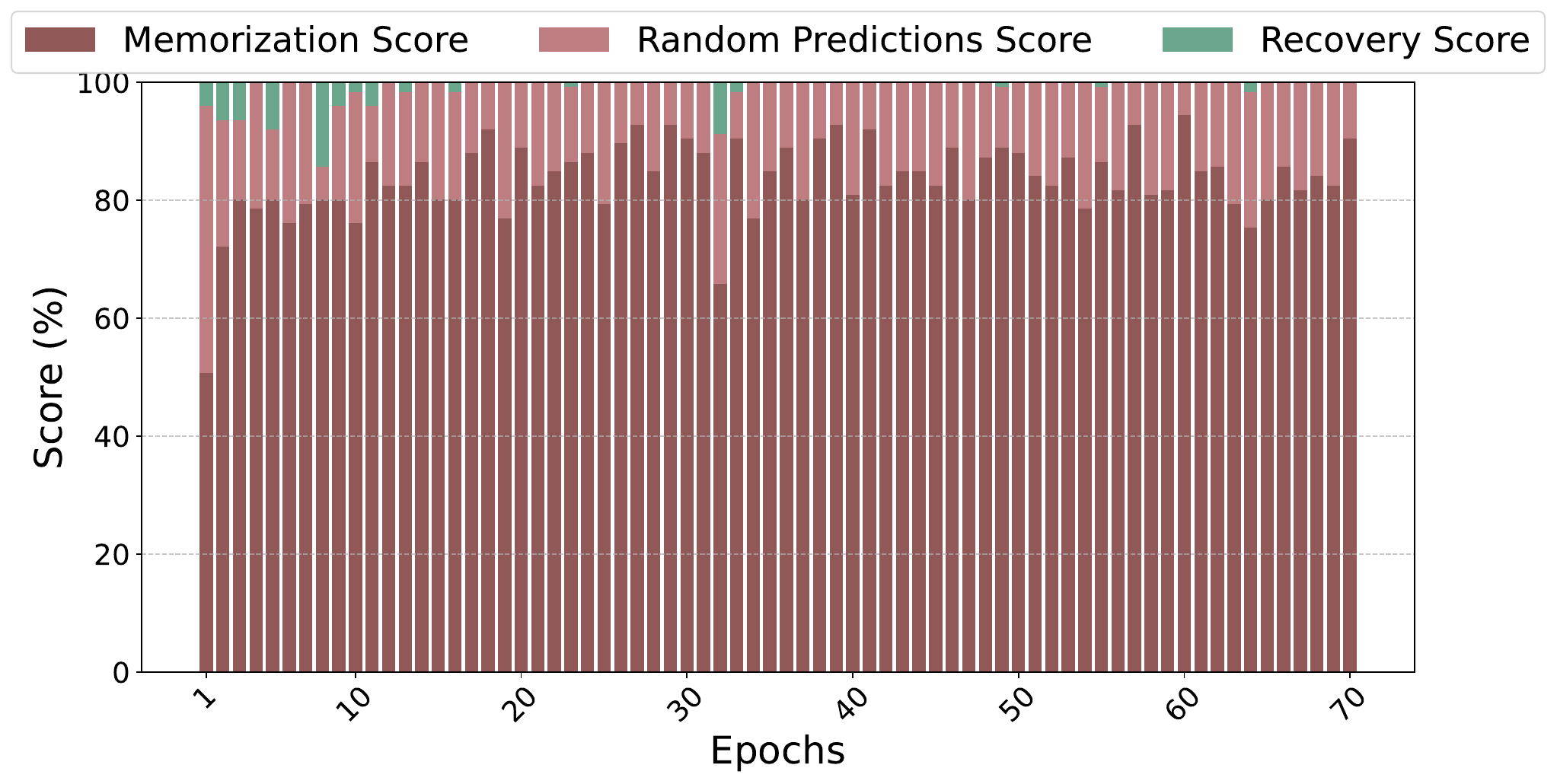}
        \caption{Memorization, Recovery and Random Predictions over epochs for Pre-LN Model (Qwen2)}
        \label{fig:pre_ln_mem_epochs_qwen2}
    \end{subfigure}
    \hspace{5pt}
    \begin{subfigure}[t]{0.25\textwidth}
        \centering
        \includegraphics[width=\textwidth]{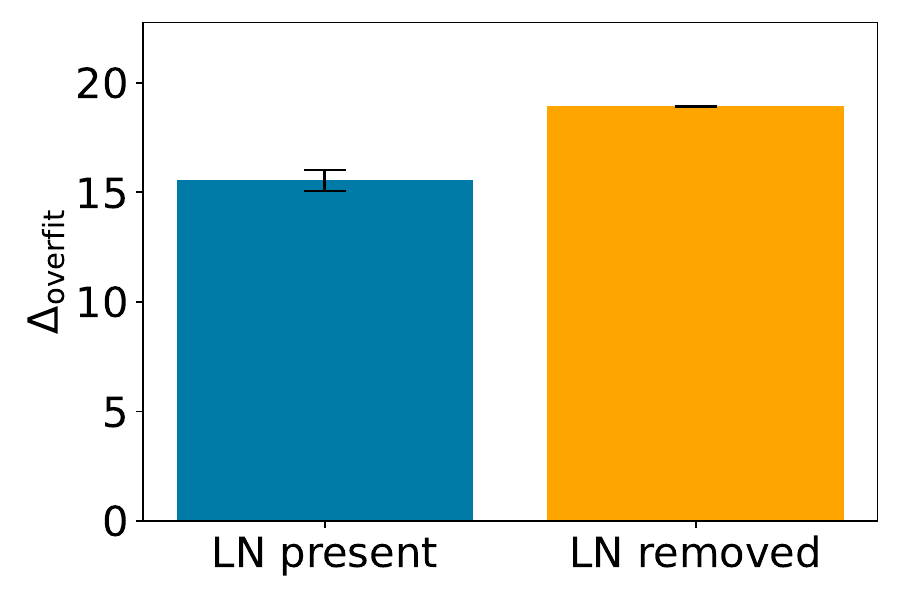}
        \caption{Overfitting gap for Pre-LN Model (Qwen2)}
        \label{fig:pre_ln_overfit_gap_qwen2}
    \end{subfigure}
    
    \vspace{10pt} 

    \begin{subfigure}[t]{0.35\textwidth}
        \centering
        \includegraphics[width=\textwidth]{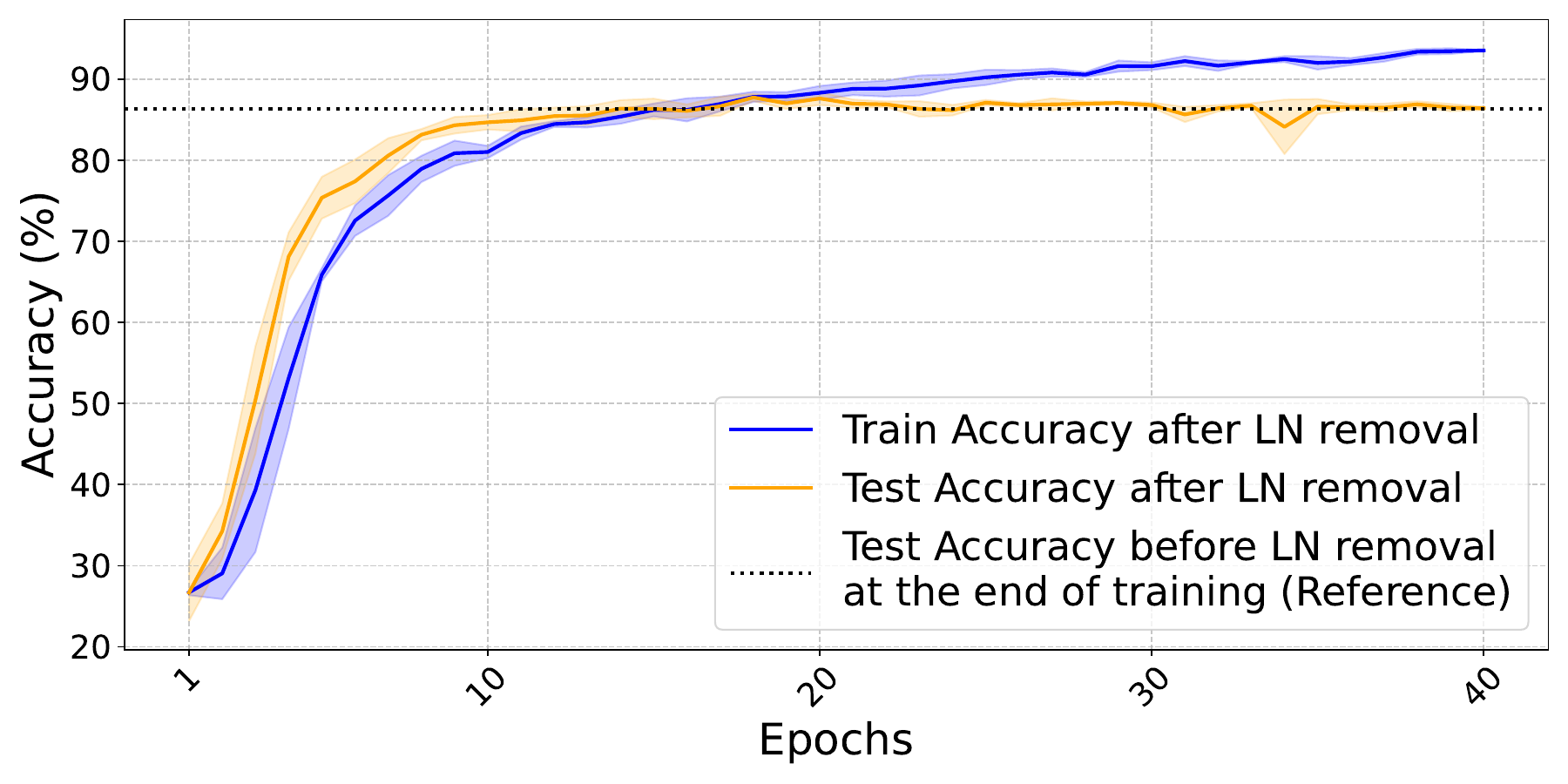}
        \caption{Learning (Test) accuracy over epochs for Post-LN Model (ELECTRA)}
        \label{fig:post_ln_learning_epochs_electra}
    \end{subfigure}
    \hspace{5pt}
    \begin{subfigure}[t]{0.35\textwidth}
        \centering
        \includegraphics[width=\textwidth]{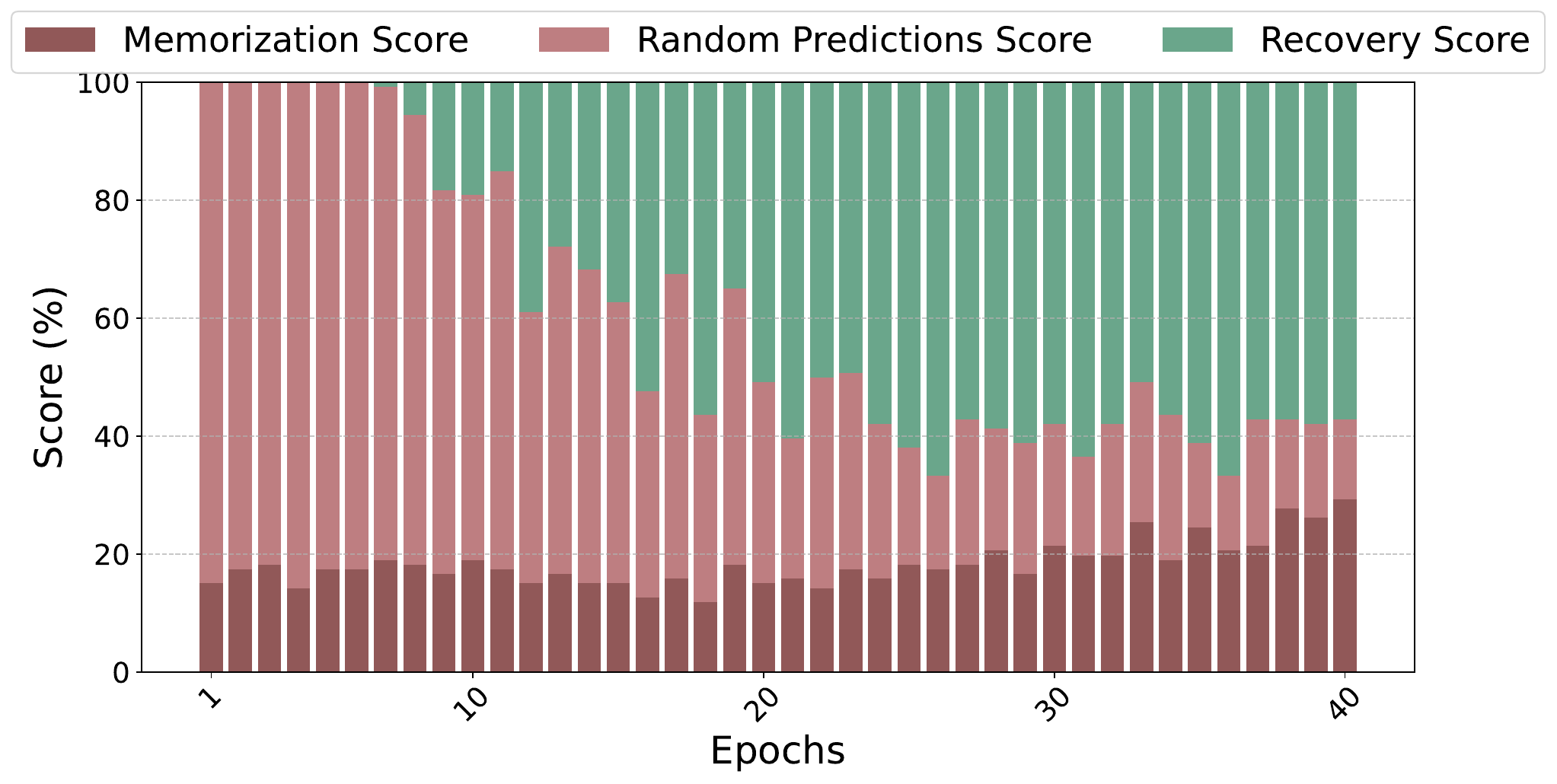}
        \caption{Memorization, Recovery and Random Predictions over epochs for Post-LN Model (ELECTRA)}
        \label{fig:post_ln_mem_epochs_electra}
    \end{subfigure}
    \hspace{5pt}
    \begin{subfigure}[t]{0.25\textwidth}
        \centering
        \includegraphics[width=\textwidth]{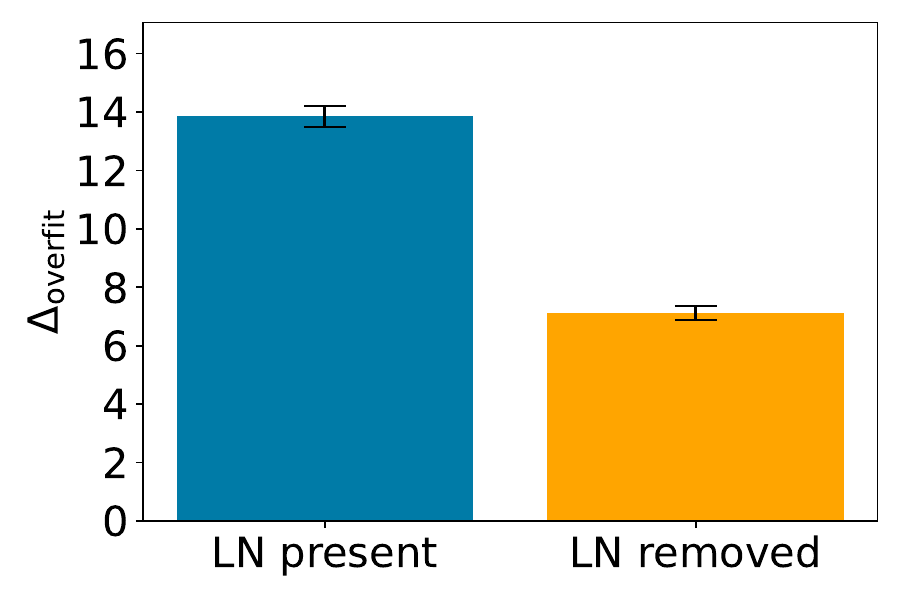}
        \caption{Overfitting gap for Post-LN Model (ELECTRA)}
        \label{fig:post_ln_overfit_gap_electra}
    \end{subfigure}
    \caption{\textbf{LN removal destabilizes learning in Pre-LN models, while mitigates memorization in Post-LN models (News Dataset):} LN removal in Pre-LN models critically affects learning (accuracy gap in (a)) while Post-LN models remain robust (negligible gap in (d)); LN removal helps in effective mitigation of memorization and high recovery in Post-LN models (\textcolor{Green}{green} bars in (e)), while memorization/random predictions still persist in Pre-LN models (\textcolor{BrickRed}{red}-color-family bars in (e)); LN removal in Pre-LN models exacerbates overfitting explained by increasing train-test accuracy gap in (c), and for Post-LN models it decreases due to memorization mitigation (see (f)).}
    \label{fig:learning_mem_recovery_epochs}
\end{figure}

In this section, we examine the distinct impact of Layer Normalization (LN) on memorization and learning in Pre-LN and Post-LN transformers. To assess its influence, we train two versions of the model — one with LN parameters removed and one with them intact — and compare their performance using learning accuracy, memorization, recovery, and random prediction scores.

\vspace{-1ex}
\subsection{Learning Stability}
\vspace{-1ex}

From Figs. \ref{fig:pre_ln_learning} \& \ref{fig:post_ln_learning}, we observe that removing LN parameters in Pre-LN transformers significantly disrupts learning, whereas Post-LN transformers remain robust, maintaining their learning accuracy even after LN parameter removal.

This discrepancy becomes even more evident when analyzing the progression of learning in epochs, as depicted for Qwen2 (Pre-LN) in  Fig.~\ref{fig:pre_ln_learning_epochs_qwen2} \& ELECTRA (Post-LN) in Fig.~\ref{fig:post_ln_learning_epochs_electra}. For Qwen2, once learning is disrupted by LN parameters removal, it does not recover till the end of training, indicating a fundamental instability. However, ELECTRA maintains stable learning throughout training, showing no signs of degradation, further highlighting its resilience to LN parameters removal. Similar results are observed for other Post-LN (BERT, DeBERTa, Longformer, RoBERT) and Pre-LN models (GPT2, GPTNeo, ViT-B, DeiT, ViT-S), as shown in Appendix~\ref{sec:ln_impact_mem_learn}. 

\vspace{-1ex}
\subsection{Memorization Suppression \& Label Recovery}
\vspace{-1ex}

We now examine the role of LN in memorization and label recovery. From Figs. \ref{fig:pre_ln_memorization} \& \ref{fig:post_ln_memorization}, we observe that in Post-LN models, LN governs memorization, its parameter removal mitigates memorization and enhances true label recovery, reflected in lower memorization scores and higher recovery scores. In contrast, for Pre-LN models, LN parameters removal does not mitigate memorization, as indicated by persistently high memorization and random prediction scores. 

This effect is even clearer when analyzing memorization over epochs, as shown for ELECTRA (Fig. \ref{fig:post_ln_mem_epochs_electra}) and Qwen2 (Fig. \ref{fig:pre_ln_mem_epochs_qwen2}). In ELECTRA, memorization decreases over time, with label recovery improving as training progresses. Conversely, in Qwen2, memorization persists throughout training, and label recovery remains poor, indicating that LN parameter removal does not suppress memorization or aid label recovery in Pre-LN models. Similar patterns are observed in other Post-LN (BERT, DeBERTa, Longformer, RoBERT) and Pre-LN models (GPT2, GPTNeo, ViT-B, DeiT, ViT-S), as shown in Appendix~\ref{sec:ln_impact_mem_learn}. 
These findings offer a nuanced perspective on LN's role: it is crucial for learning in Pre-LN models but does not influence memorization, contrary to prior work \citep{xu2019understanding}, which suggested that LN in Pre-LN models can contribute to overfitting.

In summary, \textbf{LN is essential for stable learning in Pre-LN models}, hence its parameters removal significantly destabilizes learning and widens the train-test accuracy gap ($\Delta_{\text{overfit}}^{\text{Pre}}$), i.e., exacerbating overfitting/memorization as illustrated in Fig.~\ref{fig:pre_ln_overfit_gap_qwen2}. In contrast, in \textbf{Post-LN models, LN parameters removal suppresses memorization and enhances true label recovery}, thereby narrowing the train-test accuracy gap ($\Delta_{\text{overfit}}^{\text{Post}}$) as shown in Fig. \ref{fig:post_ln_overfit_gap_electra}. This distinction is further illustrated in Table~\ref{tab:ln_impact}, which provides a comparative overview of LN’s role in learning and memorization across Pre-LN and Post-LN model. Similar observations are observed in other Pre-LN (GPT2, GPTNeo, ViT-B, ViT-S, DeiT) and Post-LN (BERT, RoBERTa, DeBERTa, Longformer) models as reported in Appendix~\ref{sec:ln_impact_mem_learn}.

\section{The Pivotal Impact of LN in Early Layers} \label{sec:early_layers_imp}

\begin{figure}[t]
    \centering
    \begin{subfigure}{1\textwidth}
        \centering
        \includegraphics[width=\textwidth]{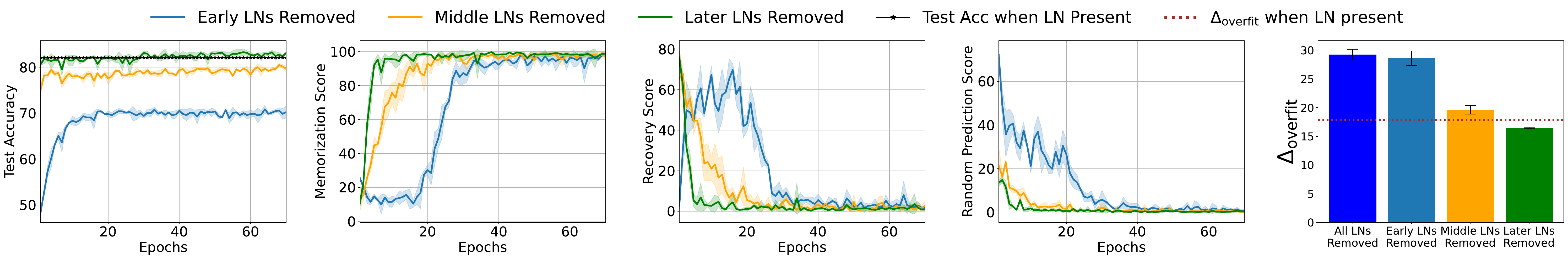}
        \caption{Impact of early, middle, later LNs on learning (test) accuracy, memorization, recovery and random predicitons scores for Pre-LN models (DeiT, UTK-Face)}
        \label{fig:Pre_LN_deit_early_middle_later_LNs}
    \end{subfigure}
    
    \begin{subfigure}{1\textwidth}
        \centering
        \includegraphics[width=\textwidth]{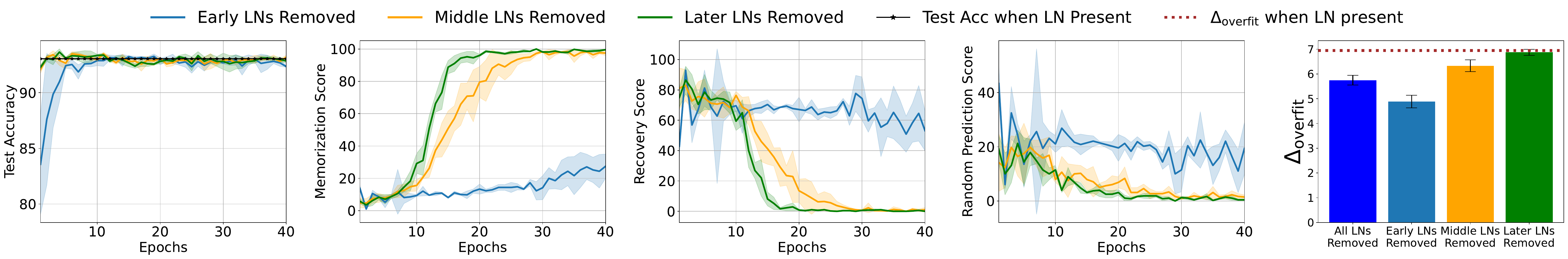} 
        \caption{Impact of early, middle, later LNs on learning (test) accuracy, memorization, recovery and random predicitons scores for Post-LN models (DeBERTa, Emotions)}
        \label{fig:Post_LN_deberta_early_middle_later_LNs}
    \end{subfigure}
    
    \caption{\textbf{Pivotal impact of early LNs for learning and memorization across Pre- and Post-LN models.} (a) clearly shows impact of early LNs removal on destabilizing learning in Pre-LN models, accompanied with higher train-test-accuracy gap, $\Delta_{\text{overfit}}^{\text{Pre, early}}$, than later layers, whereas (b) shows early LNs removal help in suppressing memorization and improving recovery in Post-LN models, alongwith lower train-test-accuracy gap, $\Delta_{\text{overfit}}^{\text{Post, early}}$, than later layers.}
    \label{fig:pre-post-ln-early-middle-later}
\end{figure}
\vspace{-1.5ex}

Building on the observation that LN has distinctive impacts on learning \& memorization for Pre- and Post-LN models, respectively, we now precisely investigate the impact of the early, middle, and later LN layers on Pre- and Post-LN models. Fig.~\ref{fig:pre-post-ln-early-middle-later} depicts that early LNs are more significant than middle/later LNs in driving learning and memorization in both Pre- and Post-LN models.

In the Pre-LN model (DeiT, Fig. \ref{fig:Pre_LN_deit_early_middle_later_LNs}), the removal of early LNs parameters significantly disrupts the learning process, highlighting their importance in learning for Pre-LN models. In the Post-LN model (DeBERTa, Fig. \ref{fig:Post_LN_deberta_early_middle_later_LNs}), the removal of early LNs parameters mitigates memorization and enhances true label recovery most significantly compared to the cases of middle or later layers. This contrast highlights the pivotal impact of early LNs in shaping learning and memorization dynamics, positively in Post-LN and negatively in Pre-LN models. Aligned trends are observed for other multiple Pre- and Post-LN models, as presented in Appendix~\ref{sec:significance_early_layers_ln_appendix}. Prior studies \citep{gromov2024unreasonable, li2024mix, lad2024remarkable, men2024shortgpt} highlighted the limited effectiveness of deeper layers for learning in Pre-LN models. Our observations take this a step further by precisely identifying that \textbf{LN in the early layers} is a critical factor in \textbf{memorization in Post-LN models}, presenting a \emph{novel} and \emph{distinctive} observation.

The distinctive effect of early LNs parameters removal—disrupting learning in Pre-LN models while mitigating memorization in Post-LN models—is further supported by the train-test accuracy gap ($\Delta_{\text{overfit}}$). Specifically, in Pre-LN models, removing them leads to a more pronounced increase in $\Delta_{\text{overfit}}^{\text{Pre, early}}$ compared to middle or later LNs, whereas in Post-LN models, removing early LNs parameters results in a sharper decrease in $\Delta_{\text{overfit}}^{\text{Post, early}}$. This trend is shown in Fig.~\ref{fig:pre-post-ln-early-middle-later} (bar plots) and formalized as follows:
\vspace{-0.5ex}
\begin{equation}
\begin{aligned}
\Delta_{\text{overfit}}^{\text{Pre, early}} &> \Delta_{\text{overfit}}^{\text{Pre, middle}} > \Delta_{\text{overfit}}^{\text{Pre, later}}, \quad \text{and} \quad 
\Delta_{\text{overfit}}^{\text{Post, early}} &< \Delta_{\text{overfit}}^{\text{Post, middle}} < \Delta_{\text{overfit}}^{\text{Post, later}}
\end{aligned}
\end{equation}

\vspace{-0.5ex}
In summary, we observe that early layers LN are more significant than later layers LN, where their removal disrupts learning, explained by high $\Delta_{\text{overfit}}^{\text{Pre, early}} $  in Pre-LN models. On the other hand, their removal suppresses memorization while recovering true labels in Post-LN models, illustrated by low $\Delta_{\text{overfit}}^{\text{Post, early}}$. Similar trends are observed in other Pre-LN (GPTNeo, Qwen2, GPT2, ViT-B, ViT-S) and Post-LN (BERT, RoBERTa, ELECTRA, Longformer) models as illustrated in Appendix~\ref{sec:significance_early_layers_ln_appendix}.

\section{Gradients Explain LN's Impact} \label{sec:learn_mem_grad}

\begin{figure}[h]
    \centering
    \begin{subfigure}{0.47\textwidth}
        \centering
        \includegraphics[width=\textwidth]{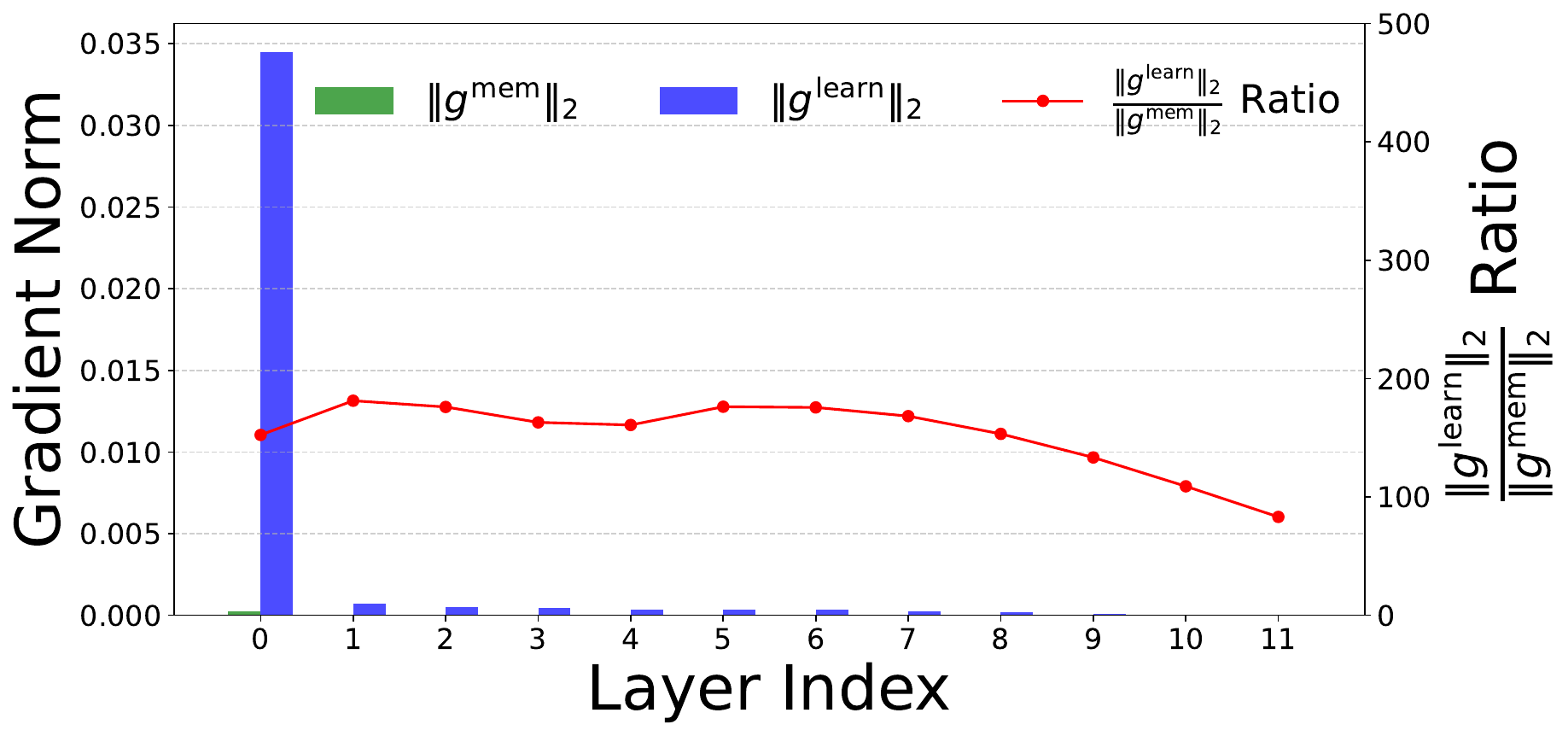}
        \caption{Pre-LN (GPTNeo) $\left\lVert g_x^{\text{learn}} \right\rVert_2$ \& $\left\lVert g_x^{\text{mem}} \right\rVert_2$ analysis across layers}
        \label{fig:Pre_LN_gradients_gpt_neo_lhs}
    \end{subfigure}
    \hspace{5mm}
    \begin{subfigure}{0.47\textwidth}
        \centering
        \includegraphics[width=\textwidth]{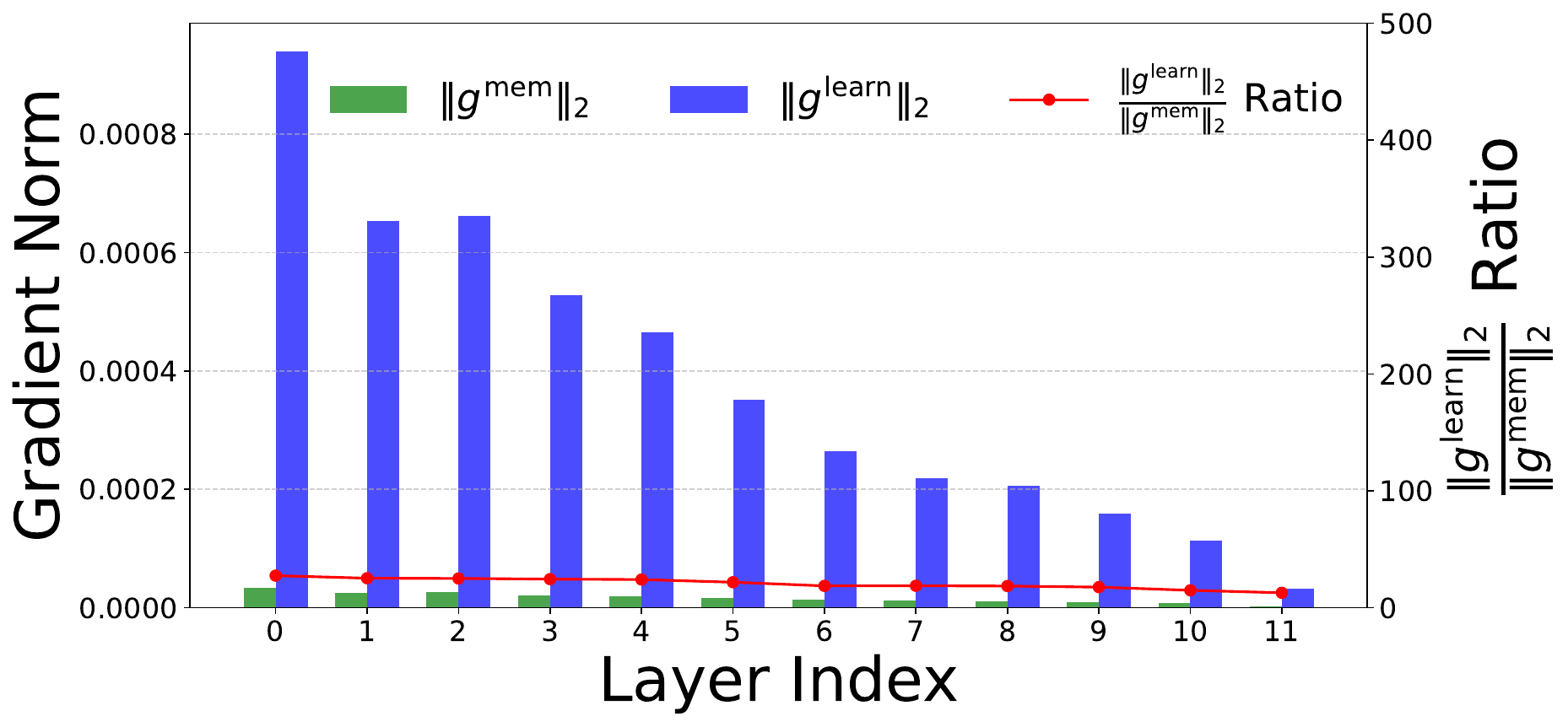}
        \caption{Post-LN (DeBERTa) $\left\lVert g_x^{\text{learn}} \right\rVert_2$ \& $\left\lVert g_x^{\text{mem}} \right\rVert_2$ analysis across layers}
        \label{fig:Post_LN_gradients_deberta_lhs}
    \end{subfigure}
    \caption{\textbf{Learning vs. Memorization Gradients in Pre- and Post-LN Models:} (in Emotions Dataset) Results clearly exhibit high gradient norms of early layers LNs than later layers for both learning and memorization in Pre-LN (GPTNeo) and Post-LN (DeBERTa) models. Importantly, the learning gradient norm ($\left\lVert g_x^{\text{learn}} \right\rVert_2$) is consistently stronger than the memorization gradient norm ($\left\lVert g_x^{\text{mem}} \right\rVert_2$) across all layers. Furthermore, the ratio $\left\lVert g_x^{\text{learn}} \right\rVert_2 \big/ \left\lVert g_x^{\text{mem}} \right\rVert_2$ is significantly higher in Pre-LN models compared to Post-LN models.}
    \label{fig:Pre_Post_Gradients_emotions_deberta_gpt_neo}
\end{figure}
\vspace{-1.5ex}

To better understand the role of layer normalization (LN) in learning and memorization, we compute the gradient norms associated with both processes across different layers ($g_x$). Specifically, we measure the norms for learning ($g_x^{\text{learn}}$) and memorization ($g_x^{\text{mem}}$) gradients separately, allowing us to quantify their relative contributions throughout the network.

\vspace{0.2in}
\begin{mdframed}[
    linewidth=0.5pt,
    roundcorner=4pt,
    backgroundcolor=gray!8,
    linecolor=gray!50,
    innertopmargin=2pt,
    innerbottommargin=2pt,
    innerleftmargin=6pt,
    innerrightmargin=6pt,
    skipabove=6pt,
    skipbelow=6pt
]
\begin{theorem}[\textbf{Learning Gradient Norm, $\|\boldsymbol{g}_{\boldsymbol{x}}^{\boldsymbol{\text{learn}}}\|_2$ is greater than or equal to Memorization Gradient Norm, $\|\boldsymbol{g}_{\boldsymbol{x}}^{\boldsymbol{\text{mem}}}\|_2$ across all layers}]
\label{theorem:mem-learn-grads-theorem}

It is formally represented as follows:
\emph{
\begin{equation}
    \left\lVert g_x^{\text{learn}} \right\rVert_2 \geq \left\lVert g_x^{\text{mem}} \right\rVert_2, \quad \quad \text{across all layers}
\end{equation}}
\end{theorem}
\end{mdframed}
\vspace{-0.2cm}
A proof of Theorem~\ref{theorem:mem-learn-grads-theorem} is provided in Appendix~\ref{theorem:mem-learn-grads-theorem-app}.

From Theorem 1, we observe that the learning gradient norms are generally greater than the memorizing gradient norms across all layers for both Pre- and Post-LN models. This observation is further validated empirically from the trend as seen in Fig.~\ref{fig:Pre_Post_Gradients_emotions_deberta_gpt_neo}.

\vspace{-2ex}
\subsection{Understanding the Distinctive Impact of LN in Pre and Post-LN Architectures}
\vspace{-1ex}

Having identified the importance of early layers LNs (Sec.~\ref{sec:early_layers_imp}), we now focus on explaining why the removal of Layer Normalization (LN) in Pre-LN models hinders learning, while in Post-LN models, it mitigates memorization without disrupting learning. To do so, we focus on the ratio of learning-to-memorization gradients norms ($\frac{\|g_x^{\text{learn}}\|_2}{\|g_x^{\text{mem}}\|_2}$, \textcolor{red}{red}-color line plots in Fig. \ref{fig:Pre_LN_gradients_gpt_neo_lhs} \& \ref{fig:Post_LN_gradients_deberta_lhs}) across layers. Based on the results, we uncover the following phenomenon:
\begin{equation} 
\frac{\|g_x^{\text{learn}}\|_2}{\|g_x^{\text{mem}}\|_2} \Big|_{\text{Pre-LN}} \gg \frac{\|g_x^{\text{learn}}\|_2}{\|g_x^{\text{mem}}\|_2}\Big|_{\text{Post-LN}} , \quad \text{across all layers}
\end{equation}
This indicates that in Pre-LN models, LayerNorm primarily facilitates learning, as evidenced by the dominance of \( \|g_x^{\text{learn}}\|_2 \) over \( \|g_x^{\text{mem}}\|_2 \). Consequently, the removal of its parameters disrupts learning and exacerbates overfitting. In contrast, in Post-LN models, \( \|g_x^{\text{learn}}\|_2 \) and \( \|g_x^{\text{mem}}\|_2 \) are of comparable magnitudes. As a result, removing LN parameters effectively mitigates memorization by restoring genuine labels without disturbing learning. Consistent trends are observed for other Pre-LN (GPT2, Qwen2, ViT-B, DeiT, ViT-S) and Post-LN (RoBERTa, BERT, Longformer, ELECTRA) models, illustrated in Appendix~\ref{sec:gradients_explain_appendix}.

\vspace{-1ex}
\subsection{Why are LNs in Early Layers Important for Memorization and Learning?}
\vspace{-1ex}
In this section, we explain why the early layers LN are pivotal in governing memorization and learning across Post and Pre-LN models, through the lens of gradient analysis.

\vspace{1.0in}
\begin{mdframed}[
    linewidth=0.5pt,
    roundcorner=4pt,
    backgroundcolor=gray!8,
    linecolor=gray!50,
    innertopmargin=2pt,
    innerbottommargin=2pt,
    innerleftmargin=6pt,
    innerrightmargin=6pt,
    skipabove=6pt,
    skipbelow=6pt
]
\begin{theorem}[\textbf{Gradient norm of loss $\mathcal{L}$ w.r.t input of LN is upper bounded}]
\label{theorem:upper_bound_l2_norm_theorem}

~\\
\textbf{Post-LN: Let $z_i$ denote the input to $\text{LN}_1$ of the $i^\text{th}$ Post-LN model layer. Then,}
\emph{
\begin{equation}
\begin{aligned}
    \| g_{z_{i}} \|_2 = \left\| \frac{\partial \mathcal{L}}{\partial z_{i}} \right\|_2 \leq & s_\text{max}(P_1) \cdot  \left( \frac{1}{\prod_{j=i}^{N} \left|1 - \sqrt{\text{Var}(\text{FFN}(x_{j}'))}\right| \left|1 - \sqrt{\text{Var}(\text{MHSA}(x_{j}))}\right|} \right) \cdot \\ & \cdot \prod_{j=i}^{N} \left( 1 + s_{\text{max}}(J_{\text{FFN}}^{x_{j}'}) \right) \cdot \prod_{j=i+1}^{N} \left( 1 + s_{\text{max}}(J_{\text{MHSA}}^{x_{j}}) \right)
\end{aligned}
\end{equation}}

\textbf{Pre-LN: Let $x_i$ denote the input to $\text{LN}_1$ of the $i^\text{th}$ Pre-LN model layer. Then,}
\emph{
\begin{equation}
\begin{aligned}
\|g_{x_{i}}\|_2 = \left\|\frac{\partial \mathcal{L}}{\partial x_{i}}\right\|_2 \leq s_\text{max}(P_2) \cdot \prod_{j=i}^{N} \left(1 + s_\text{max}(J_{\text{FFN}}^{\text{LN}_2(x_{j}')} J_{\text{LN}_2}^{x_{j}'})\right) \cdot \prod_{j=i}^{N} \left(1 + s_\text{max}(J_{\text{MHSA}}^{\text{LN}_1(x_{j})} J_{\text{LN}_1}^{x_{j}})\right)
\end{aligned}
\end{equation}}

\end{theorem}
\end{mdframed}
\vspace{-1ex}

A proof of Theorem~\ref{theorem:upper_bound_l2_norm_theorem}, along with the expressions for $\text{LN}_2$ in both Pre- and Post-LN setup, is provided in Appendix~\ref{theorem:l2_norm_bound_thm_app}.

~\\
\begin{mdframed}[
    linewidth=0.5pt,
    roundcorner=4pt,
    backgroundcolor=gray!8,
    linecolor=gray!50,
    innertopmargin=2pt,
    innerbottommargin=2pt,
    innerleftmargin=6pt,
    innerrightmargin=6pt,
    skipabove=6pt,
    skipbelow=6pt
]
\begin{theorem}[\textbf{Upper bound of the gradient norm of Early Layers LN are higher than those of Later layers LN}]
\label{theorem:early_layer_gradients_higher}

\textbf{} It is formally represented as follows:
\emph{
\begin{equation} 
\text{UB}(\|g_{x_{1}}\|_2) \geq \text{UB}(\|g_{x_{2}}\|_2) \geq \dots \geq \text{UB}(\|g_{x_{N}}\|_2) \; ; \; \text{for both Pre- and Post-LN models}
\end{equation}}

where $\text{UB}(\|g_{x_{i}}\|_2)$ denotes the upper bound of $\| g_{x_{i}} \|_2$, and $x_i$ is the input to the $i^\text{th}$ layer's LN.

\end{theorem}
\end{mdframed}
\vspace{-1ex}
A proof of Theorem~\ref{theorem:early_layer_gradients_higher} is provided in Appendix~\ref{theorem:l2_norm_bound_early_layers_app}.

\medskip

The results depicted in Fig.\ref{fig:Pre_Post_Gradients_emotions_deberta_gpt_neo} empirically confirm the trend established in Theorem \ref{theorem:early_layer_gradients_higher}. Specifically, we observe that both $\|g_x^{\text{learn}}\|_2$ and $\|g_x^{\text{mem}}\|_2$ are significantly higher in the earlier layers compared to the later ones. This gradient decay trend is consistent across both Pre-LN (GPTNeo, Fig.\ref{fig:Pre_LN_gradients_gpt_neo_lhs}) and Post-LN (DeBERTa, Fig.\ref{fig:Post_LN_gradients_deberta_lhs}) architectures. Similar trends are observed for other Pre-LN (GPT2, Qwen2, ViT-B, DeiT, ViT-S) and Post-LN (RoBERTa, BERT, Longformer, ELECTRA) models, as shown in Appendix~\ref{sec:gradients_explain_appendix}. 

Thus, the theoretical upper bounds not only provide an analytical explanation for the gradient magnitude behavior but also align closely with the empirical patterns observed across a wide range of transformer variants. This alignment helps explain why the removal of early layers LN parameters leads to disruption of learning in Pre-LN models and mitigation of memorization in Post-LN models, highlighting their predominant role in the entire network because of their higher gradient norm.

In addition to the isolation of learning and memorization in early layers, we observe another interesting pattern. For Post-LN models (Fig. \ref{fig:Post_LN_gradients_deberta_lhs}), both $\|g_x^{\text{learn}}\|_2$ and $\|g_x^{\text{mem}}\|_2$ decrease gradually over layers LN. However, for Pre-LN models (Fig. \ref{fig:Pre_LN_gradients_gpt_neo_lhs}), the gradient norms are predominantly high in the first layer, with the following layers having almost negligible norms. This observation explains why the removal of early layers LN parameters did not significantly affect learning for Post-LN models. That is because, in Post-LN models, later LNs can compensate for the absence of the early ones, recovering learning, while mitigating memorization, due to their comparable gradient norms. However, this does not hold for Pre-LN models, where the high gradient norms in the early layers LN are critical, and their absence severely disrupts learning. 

Similar observations are observed in other Pre-LN (GPT2, Qwen2, ViT-B, DeiT, ViT-S) and Post-LN (RoBERTa, BERT, Longformer, ELECTRA) models, as shown in Appendix~\ref{sec:gradients_explain_appendix}.

In summary, gradient analysis highlights why removal of LN parameters significantly affects both learning and memorization: (1) \textbf{disrupts learning in Pre-LN models and mitigates memorization in Post-LN models} due to the distinct behavior of their gradient norms ratio; and (2) \textbf{it reveals the particular significance of early layer LNs}, which exhibit stronger gradient norms and thus play a more influential role in both learning and memorization processes.
\section{Conclusion}
In conclusion, our study highlights the pivotal role of Layer Normalization (LN) in governing both memorization and learning across two different Transformer configurations: Pre-LN and Post-LN. We identified that the removal of LN parameters in Pre-LN models significantly destabilizes the learning process, leading to persistent overfitting. In contrast, removing LN parameters from Post-LN architectures effectively mitigates memorization and enables the recovery of genuine labels. More precisely, we find that LNs in the early layers are especially critical—removing them has the strongest impact in disrupting learning in Pre-LN models and mitigating memorization in Post-LN models. By analyzing the learning and memorization gradient norms, we further reveal how LN distinctively influences these two mechanisms across Pre- and Post-LN models. We show that this distinctive behavior across a wide range of model architectures, spanning several vision and language datasets. Overall, our findings uncover a crucial connection on how layer normalization impacts learning and memorization in transformer models, with its broader impacts discussed in Appendix~\ref{sec:broader_impacts}.

\bibliographystyle{unsrtnat}
\bibliography{references}

@article{vaswani2017attention,
  title={Attention is all you need},
  author={Ashish Vaswani and
                  Noam Shazeer and
                  Niki Parmar and
                  Jakob Uszkoreit and
                  Llion Jones and
                  Aidan N. Gomez and
                  Lukasz Kaiser and
                  Illia Polosukhin},
  journal={Advances in Neural Information Processing Systems},
  year={2017}
}

@article{lei2016layer,
  title={Layer normalization},
  author={Lei Ba, Jimmy and Kiros, Jamie Ryan and Hinton, Geoffrey E},
  journal={ArXiv e-prints},
  pages={arXiv--1607},
  year={2016}
}

@inproceedings{xiong2020layer,
  title={On layer normalization in the transformer architecture},
  author={Xiong, Ruibin and Yang, Yunchang and He, Di and Zheng, Kai and Zheng, Shuxin and Xing, Chen and Zhang, Huishuai and Lan, Yanyan and Wang, Liwei and Liu, Tieyan},
  booktitle={International Conference on Machine Learning},
  pages={10524--10533},
  year={2020},
  organization={PMLR}
}

@article{feldman2020neural,
  title={What neural networks memorize and why: Discovering the long tail via influence estimation},
  author={Feldman, Vitaly and Zhang, Chiyuan},
  journal={Advances in Neural Information Processing Systems},
  volume={33},
  pages={2881--2891},
  year={2020}
}

@inproceedings{feldman2020does,
  title={Does learning require memorization? a short tale about a long tail},
  author={Feldman, Vitaly},
  booktitle={Proceedings of the 52nd Annual ACM SIGACT Symposium on Theory of Computing},
  pages={954--959},
  year={2020}
}

@inproceedings{maini2023memorization,
  title={Can Neural Network Memorization Be Localized?},
  author={Maini, Pratyush and Mozer, Michael C and Sedghi, Hanie and Lipton, Zachary C and Kolter, J Zico and Zhang, Chiyuan},
  booktitle={International Conference on Machine Learning},
  year={2023}
}

@inproceedings{haviv2023understanding,
  title={Understanding Transformer Memorization Recall Through Idioms},
  author={Haviv, Adi and Cohen, Ido and Gidron, Jacob and Schuster, Roei and Goldberg, Yoav and Geva, Mor},
  booktitle={Proceedings of the 17th Conference of the European Chapter of the Association for Computational Linguistics},
  pages={248--264},
  year={2023}
}

@article{geva2023dissecting,
  title={Dissecting recall of factual associations in auto-regressive language models},
  author={Geva, Mor and Bastings, Jasmijn and Filippova, Katja and Globerson, Amir},
  journal={arXiv preprint arXiv:2304.14767},
  year={2023}
}

@article{yu2023characterizing,
  title={Characterizing mechanisms for factual recall in language models},
  author={Yu, Qinan and Merullo, Jack and Pavlick, Ellie},
  journal={arXiv preprint arXiv:2310.15910},
  year={2023}
}

@article{he2024understanding,
  title={Understanding and minimising outlier features in transformer training},
  author={He, Bobby and Noci, Lorenzo and Paliotta, Daniele and Schlag, Imanol and Hofmann, Thomas},
  journal={Advances in Neural Information Processing Systems},
  volume={37},
  pages={83786--83846},
  year={2024}
}

@article{kovaleva2021bert,
  title={BERT busters: Outlier dimensions that disrupt transformers},
  author={Kovaleva, Olga and Kulshreshtha, Saurabh and Rogers, Anna and Rumshisky, Anna},
  journal={arXiv preprint arXiv:2105.06990},
  year={2021}
}

@article{bondarenko2023quantizable,
  title={Quantizable transformers: Removing outliers by helping attention heads do nothing},
  author={Bondarenko, Yelysei and Nagel, Markus and Blankevoort, Tijmen},
  journal={Advances in Neural Information Processing Systems},
  volume={36},
  pages={75067--75096},
  year={2023}
}

@article{liu2020understanding,
  title={Understanding the difficulty of training transformers},
  author={Liu, Liyuan and Liu, Xiaodong and Gao, Jianfeng and Chen, Weizhu and Han, Jiawei},
  journal={arXiv preprint arXiv:2004.08249},
  year={2020}
}

@article{xu2019understanding,
  title={Understanding and improving layer normalization},
  author={Xu, Jingjing and Sun, Xu and Zhang, Zhiyuan and Zhao, Guangxiang and Lin, Junyang},
  journal={Advances in neural information processing systems},
  volume={32},
  year={2019}
}

@article{baldock2021deep,
  title={Deep learning through the lens of example difficulty},
  author={Baldock, Robert and Maennel, Hartmut and Neyshabur, Behnam},
  journal={Advances in Neural Information Processing Systems},
  volume={34},
  pages={10876--10889},
  year={2021}
}

@article{puccetti2022outliers,
  title={Outliers dimensions that disrupt transformers are driven by frequency},
  author={Puccetti, Giovanni and Rogers, Anna and Drozd, Aleksandr and Dell'Orletta, Felice},
  journal={arXiv preprint arXiv:2205.11380},
  year={2022}
}

@article{krizhevsky2009learning,
  title={Learning multiple layers of features from tiny images},
  author={Krizhevsky, Alex and Hinton, Geoffrey and others},
  year={2009},
  publisher={Toronto, ON, Canada}
}

@inproceedings{zhang2017age,
  title={Age progression/regression by conditional adversarial autoencoder},
  author={Zhang, Zhifei and Song, Yang and Qi, Hairong},
  booktitle={Proceedings of the IEEE conference on computer vision and pattern recognition},
  pages={5810--5818},
  year={2017}
}

@inproceedings{saravia-etal-2018-carer,
    title = "{CARER}: Contextualized Affect Representations for Emotion Recognition",
    author = "Saravia, Elvis  and
      Liu, Hsien-Chi Toby  and
      Huang, Yen-Hao  and
      Wu, Junlin  and
      Chen, Yi-Shin",
    editor = "Riloff, Ellen  and
      Chiang, David  and
      Hockenmaier, Julia  and
      Tsujii, Jun{'}ichi",
    booktitle = "Proceedings of the 2018 Conference on Empirical Methods in Natural Language Processing",
    month = oct # "-" # nov,
    year = "2018",
    address = "Brussels, Belgium",
    publisher = "Association for Computational Linguistics",
    url = "https://aclanthology.org/D18-1404/",
    doi = "10.18653/v1/D18-1404",
    pages = "3687--3697"
}

@inproceedings{dimosthenis-etal-2022-twitter,
    title = "{T}witter {T}opic {C}lassification",
    author = "Antypas, Dimosthenis  and
    Ushio, Asahi  and
    Camacho-Collados, Jose  and
    Neves, Leonardo  and
    Silva, Vitor  and
    Barbieri, Francesco",
    booktitle = "Proceedings of the 29th International Conference on Computational Linguistics",
    month = oct,
    year = "2022",
    address = "Gyeongju, Republic of Korea",
    publisher = "International Committee on Computational Linguistics"
}

@misc{okite97_news_data,
  author = {Okite97},
  title = {News Data Dataset},
  year = {2024},
  howpublished = {\url{https://huggingface.co/datasets/okite97/news-data}},
  note = {Accessed: 2024-03-03}
}

@article{sanh2019distilbert,
  title={DistilBERT, a distilled version of BERT: Smaller, faster, cheaper and lighter. arXiv 2019},
  author={Sanh, Victor and Debut, L and Chaumond, J and Wolf, T},
  journal={arXiv preprint arXiv:1910.01108},
  year={2019}
}

@article{yinhan2019roberta,
  title={RoBERTa: A robustly optimized BERT pretraining approach (2019)},
  author={Yinhan, Liu and Myle, Ott and Naman, Goyal and Jingfei, Du and Mandar, Joshi and Danqi, Chen and Omer, Levy and Mike, Lewis},
  journal={arXiv preprint arXiv:1907.11692},
  pages={1--13},
  year={2019},
  publisher={CoRR}
}

@article{he2020deberta,
  title={Deberta: Decoding-enhanced bert with disentangled attention},
  author={He, Pengcheng and Liu, Xiaodong and Gao, Jianfeng and Chen, Weizhu},
  journal={arXiv preprint arXiv:2006.03654},
  year={2020}
}

@article{clark2020electra,
  title={Electra: Pre-training text encoders as discriminators rather than generators},
  author={Clark, K},
  journal={arXiv preprint arXiv:2003.10555},
  year={2020}
}

@article{beltagy2020longformer,
  title={Longformer: The long-document transformer},
  author={Beltagy, Iz and Peters, Matthew E and Cohan, Arman},
  journal={arXiv preprint arXiv:2004.05150},
  year={2020}
}

@article{alexey2020image,
  title={An image is worth 16x16 words: Transformers for image recognition at scale},
  author={Alexey, Dosovitskiy},
  journal={arXiv preprint arXiv: 2010.11929},
  year={2020}
}

@inproceedings{touvron2021training,
  title={Training data-efficient image transformers \& distillation through attention},
  author={Touvron, Hugo and Cord, Matthieu and Douze, Matthijs and Massa, Francisco and Sablayrolles, Alexandre and J{\'e}gou, Herv{\'e}},
  booktitle={International conference on machine learning},
  pages={10347--10357},
  year={2021},
  organization={PMLR}
}

@inproceedings{ott-etal-2019-fairseq,
    title = "fairseq: A Fast, Extensible Toolkit for Sequence Modeling",
    author = "Ott, Myle  and
      Edunov, Sergey  and
      Baevski, Alexei  and
      Fan, Angela  and
      Gross, Sam  and
      Ng, Nathan  and
      Grangier, David  and
      Auli, Michael",
    editor = "Ammar, Waleed  and
      Louis, Annie  and
      Mostafazadeh, Nasrin",
    booktitle = "Proceedings of the 2019 Conference of the North {A}merican Chapter of the Association for Computational Linguistics (Demonstrations)",
    month = jun,
    year = "2019",
    address = "Minneapolis, Minnesota",
    publisher = "Association for Computational Linguistics",
    url = "https://aclanthology.org/N19-4009/",
    doi = "10.18653/v1/N19-4009",
    pages = "48--53"
}

@article{radford2019language,
  title={Language models are unsupervised multitask learners},
  author={Radford, Alec and Wu, Jeffrey and Child, Rewon and Luan, David and Amodei, Dario and Sutskever, Ilya and others},
  journal={OpenAI blog},
  volume={1},
  number={8},
  pages={9},
  year={2019}
}

@article{black2022gpt,
  title={Gpt-neo: Large scale autoregressive language modeling with mesh-tensorflow, 2021},
  author={Black, Sid and Gao, Leo and Wang, Phil and Leahy, Connor and Biderman, Stella},
  journal={URL: https://doi. org/10.5281/zenodo},
  volume={5297715},
  year={2022}
}

@article{yang2024qwen2,
  title={Qwen2. 5 technical report},
  author={Yang, An and Yang, Baosong and Zhang, Beichen and Hui, Binyuan and Zheng, Bo and Yu, Bowen and Li, Chengyuan and Liu, Dayiheng and Huang, Fei and Wei, Haoran and others},
  journal={arXiv preprint arXiv:2412.15115},
  year={2024}
}

@article{gromov2024unreasonable,
  title={The unreasonable ineffectiveness of the deeper layers},
  author={Gromov, Andrey and Tirumala, Kushal and Shapourian, Hassan and Glorioso, Paolo and Roberts, Daniel A},
  journal={arXiv preprint arXiv:2403.17887},
  year={2024}
}

@article{li2024mix,
  title={Mix-ln: Unleashing the power of deeper layers by combining pre-ln and post-ln},
  author={Li, Pengxiang and Yin, Lu and Liu, Shiwei},
  journal={arXiv preprint arXiv:2412.13795},
  year={2024}
}

@inproceedings{agarwal2022estimating,
  title={Estimating example difficulty using variance of gradients},
  author={Agarwal, Chirag and D'souza, Daniel and Hooker, Sara},
  booktitle={Proceedings of the IEEE/CVF Conference on Computer Vision and Pattern Recognition},
  pages={10368--10378},
  year={2022}
}

@article{dai2021knowledge,
  title={Knowledge neurons in pretrained transformers},
  author={Dai, Damai and Dong, Li and Hao, Yaru and Sui, Zhifang and Chang, Baobao and Wei, Furu},
  journal={arXiv preprint arXiv:2104.08696},
  year={2021}
}

@article{sun2025curse,
  title={The Curse of Depth in Large Language Models},
  author={Sun, Wenfang and Song, Xinyuan and Li, Pengxiang and Yin, Lu and Zheng, Yefeng and Liu, Shiwei},
  journal={arXiv preprint arXiv:2502.05795},
  year={2025}
}

@article{garg2023memorization,
  title={Memorization through the lens of curvature of loss function around samples},
  author={Garg, Isha and Ravikumar, Deepak and Roy, Kaushik},
  journal={arXiv preprint arXiv:2307.05831},
  year={2023}
}

@article{ravikumar2024unveiling,
  title={Unveiling privacy, memorization, and input curvature links},
  author={Ravikumar, Deepak and Soufleri, Efstathia and Hashemi, Abolfazl and Roy, Kaushik},
  journal={arXiv preprint arXiv:2402.18726},
  year={2024}
}

@article{brody2023expressivity,
  title={On the expressivity role of LayerNorm in transformers' attention},
  author={Brody, Shaked and Alon, Uri and Yahav, Eran},
  journal={arXiv preprint arXiv:2305.02582},
  year={2023}
}

@article{wu2024role,
  title={On the role of attention masks and layernorm in transformers},
  author={Wu, Xinyi and Ajorlou, Amir and Wang, Yifei and Jegelka, Stefanie and Jadbabaie, Ali},
  journal={arXiv preprint arXiv:2405.18781},
  year={2024}
}

@article{kim2025peri,
  title={Peri-LN: Revisiting Layer Normalization in the Transformer Architecture},
  author={Kim, Jeonghoon and Lee, Byeongchan and Park, Cheonbok and Oh, Yeontaek and Kim, Beomjun and Yoo, Taehwan and Shin, Seongjin and Han, Dongyoon and Shin, Jinwoo and Yoo, Kang Min},
  journal={arXiv preprint arXiv:2502.02732},
  year={2025}
}

@article{xie2023residual,
  title={Residual: Transformer with dual residual connections},
  author={Xie, Shufang and Zhang, Huishuai and Guo, Junliang and Tan, Xu and Bian, Jiang and Awadalla, Hany Hassan and Menezes, Arul and Qin, Tao and Yan, Rui},
  journal={arXiv preprint arXiv:2304.14802},
  year={2023}
}

@inproceedings{zhang2023nico++,
  title={Nico++: Towards better benchmarking for domain generalization},
  author={Zhang, Xingxuan and He, Yue and Xu, Renzhe and Yu, Han and Shen, Zheyan and Cui, Peng},
  booktitle={Proceedings of the IEEE/CVF conference on computer vision and pattern recognition},
  pages={16036--16047},
  year={2023}
}

@inproceedings{assran2022masked,
  title={Masked siamese networks for label-efficient learning},
  author={Assran, Mahmoud and Caron, Mathilde and Misra, Ishan and Bojanowski, Piotr and Bordes, Florian and Vincent, Pascal and Joulin, Armand and Rabbat, Mike and Ballas, Nicolas},
  booktitle={European conference on computer vision},
  pages={456--473},
  year={2022},
  organization={Springer}
}

@article{takase2022b2t,
  title={B2t connection: Serving stability and performance in deep transformers},
  author={Takase, Sho and Kiyono, Shun and Kobayashi, Sosuke and Suzuki, Jun},
  journal={arXiv preprint arXiv:2206.00330},
  year={2022}
}

@book{Horn_Johnson_1991, place={Cambridge}, title={Topics in Matrix Analysis}, publisher={Cambridge University Press}, author={Horn, Roger A. and Johnson, Charles R.}, year={1991}}

@article{hossjer2022sharp,
  title={Sharp lower and upper bounds for the covariance of bounded random variables},
  author={H{\"o}ssjer, Ola and Sj{\"o}lander, Arvid},
  journal={Statistics \& Probability Letters},
  volume={182},
  pages={109323},
  year={2022},
  publisher={Elsevier}
}

@article{men2024shortgpt,
  title={Shortgpt: Layers in large language models are more redundant than you expect},
  author={Men, Xin and Xu, Mingyu and Zhang, Qingyu and Wang, Bingning and Lin, Hongyu and Lu, Yaojie and Han, Xianpei and Chen, Weipeng},
  journal={arXiv preprint arXiv:2403.03853},
  year={2024}
}

@article{yin2023outlier,
  title={Outlier weighed layerwise sparsity (owl): A missing secret sauce for pruning llms to high sparsity},
  author={Yin, Lu and Wu, You and Zhang, Zhenyu and Hsieh, Cheng-Yu and Wang, Yaqing and Jia, Yiling and Li, Gen and Jaiswal, Ajay and Pechenizkiy, Mykola and Liang, Yi and others},
  journal={arXiv preprint arXiv:2310.05175},
  year={2023}
}

@article{lad2024remarkable,
  title={The Remarkable Robustness of LLMs: Stages of Inference?},
  author={Lad, Vedang and Gurnee, Wes and Tegmark, Max},
  journal={arXiv preprint arXiv:2406.19384},
  year={2024}
}

@article{jiang2020characterizing,
  title={Characterizing structural regularities of labeled data in overparameterized models},
  author={Jiang, Ziheng and Zhang, Chiyuan and Talwar, Kunal and Mozer, Michael C},
  journal={arXiv preprint arXiv:2002.03206},
  year={2020}
}

@article{zhou2023samples,
  title={Which samples should be learned first: Easy or hard?},
  author={Zhou, Xiaoling and Wu, Ou},
  journal={IEEE Transactions on Neural Networks and Learning Systems},
  year={2023},
  publisher={IEEE}
}

@inproceedings{arpit2017closer,
  title={A closer look at memorization in deep networks},
  author={Arpit, Devansh and Jastrz{\k{e}}bski, Stanis{\l}aw and Ballas, Nicolas and Krueger, David and Bengio, Emmanuel and Kanwal, Maxinder S and Maharaj, Tegan and Fischer, Asja and Courville, Aaron and Bengio, Yoshua and others},
  booktitle={International conference on machine learning},
  pages={233--242},
  year={2017},
  organization={PMLR}
}

@article{shah2020pitfalls,
  title={The pitfalls of simplicity bias in neural networks},
  author={Shah, Harshay and Tamuly, Kaustav and Raghunathan, Aditi and Jain, Prateek and Netrapalli, Praneeth},
  journal={Advances in Neural Information Processing Systems},
  volume={33},
  pages={9573--9585},
  year={2020}
}

@article{kumar2023dual,
  title={Dual patchnorm},
  author={Kumar, Manoj and Dehghani, Mostafa and Houlsby, Neil},
  journal={arXiv preprint arXiv:2302.01327},
  year={2023}
}

@article{qi2023lipsformer,
  title={Lipsformer: Introducing lipschitz continuity to vision transformers},
  author={Qi, Xianbiao and Wang, Jianan and Chen, Yihao and Shi, Yukai and Zhang, Lei},
  journal={arXiv preprint arXiv:2304.09856},
  year={2023}
}

@article{wang2022foundation,
  title={Foundation transformers},
  author={Wang, Hongyu and Ma, Shuming and Huang, Shaohan and Dong, Li and Wang, Wenhui and Peng, Zhiliang and Wu, Yu and Bajaj, Payal and Singhal, Saksham and Benhaim, Alon and others},
  journal={arXiv preprint arXiv:2210.06423},
  year={2022}
}

@article{shleifer2021normformer,
  title={Normformer: Improved transformer pretraining with extra normalization},
  author={Shleifer, Sam and Weston, Jason and Ott, Myle},
  journal={arXiv preprint arXiv:2110.09456},
  year={2021}
}

@article{stephenson2021geometry,
  title={On the geometry of generalization and memorization in deep neural networks},
  author={Stephenson, Cory and Padhy, Suchismita and Ganesh, Abhinav and Hui, Yue and Tang, Hanlin and Chung, SueYeon},
  journal={arXiv preprint arXiv:2105.14602},
  year={2021}
}

@article{zhang2019root,
  title={Root mean square layer normalization},
  author={Zhang, Biao and Sennrich, Rico},
  journal={Advances in Neural Information Processing Systems},
  volume={32},
  year={2019}
}

@article{jiang2023pre,
  title={Pre-rmsnorm and pre-crmsnorm transformers: equivalent and efficient pre-ln transformers},
  author={Jiang, Zixuan and Gu, Jiaqi and Zhu, Hanqing and Pan, David},
  journal={Advances in Neural Information Processing Systems},
  volume={36},
  pages={45777--45793},
  year={2023}
}

@inproceedings{devlin2019bert,
  title={Bert: Pre-training of deep bidirectional transformers for language understanding},
  author={Devlin, Jacob and Chang, Ming-Wei and Lee, Kenton and Toutanova, Kristina},
  booktitle={Proceedings of the 2019 conference of the North American chapter of the association for computational linguistics: human language technologies, volume 1 (long and short papers)},
  pages={4171--4186},
  year={2019}
}

\newpage
\appendix
\section*{Appendix}
 
\begin{table}[th]
\centering
\caption{Comparison of Post-Layer Normalization (Post-LN) and Pre-Layer Normalization (Pre-LN) Transformer Setup. MHSA = multi-head self attention, FFN = feed-forward network, LN: LayerNorm, N = number of transformer layers, $x_i$ = input to $i^\text{th}$ layer, $y_i$ = output of $i^\text{th}$ layer, $z_i$ \& $z_{i'}$ = intermediate vectors in Post-LN model, which are inputs to $\text{LN}_1$ \& $\text{LN}_2$ respectively.}
\label{tab:post_pre_eqns}
\vspace{0.1in}
\begin{tabular}{|c|c|}
\hline
\textbf{Post-LN Transformer Setup} & \textbf{Pre-LN Transformer Setup} \\
\hline
$ z_{i} = \text{MHSA}(x_i) + x_i $ & $ x_{i}'= \text{MHSA}(\text{LN}_1(x_{i})) + x_{i}$ \\
$ x_{i}' = \text{LN}_1(z_{i}) $ & $ y_{i} = \text{FFN}(\text{LN}_2(x_{i}')) + x_{i}'$ \\
$ z_{i}' = \text{FFN}(x_{i}') + x_{i}'$ & \\
$ y_i = \text{LN}_2(z_{i}') $ & \\
\hline
\multicolumn{2}{|c|}{$ y_{\text{out}} = \text{classification-head}(y_{N}) $} \\
\multicolumn{2}{|c|}{$ \mathcal{L} = \text{CrossEntropyloss}(y_{\text{out}}, y_{\text{true}}) $} \\
\hline
\end{tabular}
\end{table}

\section{Problem Setup for Gradients Analysis:}\label{prob_setup_appendix}
Consider a training dataset \(D_{\text{train}}\) consisting of \(C\) classes. We make an assumption that all samples in class c well represent class c. Then, we introduce a single noisy label by selecting a sample \((x_1, y_c)\) from class \(c\), where \(c \in C\), and modify its label to a different (incorrect) label, \(y_{\text{NL}}\), where \(\text{NL} (\in C) \neq c\). The noisy sample is now represented as \((x_1, y_{\text{NL}})\). At the same time, we do not modify any other training samples from class \(c\) to ensure that the model has access to correctly labeled examples for effective learning as well. Next, we train a transformer model on this modified dataset until it reaches 100\% training accuracy. At this stage, the model has fully memorized the noisy-labeled sample \((x_1, y_{\text{NL}})\) while also learning class \(c\) features from correctly labeled training samples. Now to measure the notion of memorization and learning, we compute \textbf{\emph{memorization gradient norm}} ($\|g_x^\text{mem}\|_2$) and \textbf{\emph{learning gradient norm}} ($\|g_x^\text{learn}\|_2$) as discussed in Sec.\ref{subssec:ln_analysis}, and compare them.

\begin{mdframed}[
    linewidth=0.5pt,
    roundcorner=4pt,
    backgroundcolor=gray!8,
    linecolor=gray!50,
    innertopmargin=2pt,
    innerbottommargin=2pt,
    innerleftmargin=6pt,
    innerrightmargin=6pt,
    skipabove=6pt,
    skipbelow=6pt
]
\section{Theorem 1: Learning Gradient Norm, \texorpdfstring{$\|\boldsymbol{g}_{\boldsymbol{x}}^{\boldsymbol{\text{learn}}}\|_2$}{||gₓ^learn||} is greater than or equal to Memorization Gradient Norm, \texorpdfstring{$\|\boldsymbol{g}_{\boldsymbol{x}}^{\boldsymbol{\text{mem}}}\|_2$}{||gₓ^mem||} across all layers.}
\label{theorem:mem-learn-grads-theorem-app}

\emph{It is formally represented as follows:}
\begin{equation}
    \begin{aligned}
        \left\lVert g_x^{\text{learn}} \right\rVert_2 \geq \left\lVert g_x^{\text{mem}} \right\rVert_2, \quad \quad \text{across all layers}
    \end{aligned}
\end{equation}
\end{mdframed}
\emph{\textbf{Proof:}}

Based on the Problem Setup as discussed in Sec.~\ref{prob_setup_appendix}, we prove that $\left\lVert g_x^{\text{learn}} \right\rVert_2 \geq \left\lVert g_x^{\text{mem}} \right\rVert_2$ across all layers, for both Pre- and Post-LN models as follows:

\subsection{\emph{For Post-LN model:}} Firstly, we elucidate the architecture of Post-LN transformer in Tab.~\ref{tab:post_pre_eqns}. Based on the Post-LN architecture, during backpropagation, we compute derivatives of loss w.r.t input of LN for every $i^{th}$-layer. Since there are two LNs in every layer, we compute and compare the learning and memorization gradients corresponding to the inputs of both LNs, i.e., $g_{z_{i'}}$ and $g_{z_{i}}$ (refer to Tab.~\ref{tab:post_pre_eqns}).

\subsubsection{Backpropagating gradient analysis for $\text{LN}_2$ \(( g_{z_{i}'}) \):}\label{para:g_x_i,6_discussion} 
The backpropagating gradient for \( g_{z_{i'}} \) can be expressed as follows:

\begin{equation}
\begin{aligned}
g_{z_{i}'} = \frac{\partial \mathcal{L}}{\partial z_{i}'}= \frac{\partial \mathcal{L}}{\partial y_{\text{out}}} \cdot \frac{\partial y_{\text{out}}}{\partial y_N} \cdot \prod_{j=i+1}^{N} \left( \frac{\partial y_j}{\partial z_{j}'} \cdot \frac{\partial z_{j}'}{\partial x_{j}'} \cdot \frac{\partial x_{j}'}{\partial z_{j}} \cdot \frac{\partial z_{j}}{\partial x_{j}} \right) \cdot \frac{\partial y_i}{\partial z_{i}'}
\end{aligned}
\label{eq:zjprimeGradient}
\end{equation}
where $x_{i+1} = y _{i}$ because $i^{\text{th}}$ layer's output $y_{i}$ is the input of $(i+1)^{\text{th}}$ layer, $x_{i+1}$. To measure memorization and learning, we compute these gradients for both \((x_1, y_{\text{NL}})\) and \((x_2, y_c)\), denoted as $g_{z_{i'}}^{\text{mem}}$ and $g_{z_{i'}}^{\text{learn}}$, respectively. In both gradients, $y_\text{out}, y_j, z_{j}', x_{j}', z_j, x_j$ are only dependent on the input samples $x_1$ and $x_2$, respectively, and not on their labels, where both $x_1$ and $x_2$ genuinely represent class c. Therefore, $\frac{\partial y_{\text{out}}}{\partial y_N}, \frac{\partial y_j}{\partial z_{j}'}, \frac{\partial z_{j}'}{\partial x_{j}'}, \frac{\partial x_{j}'}{\partial z_{j}}, \frac{\partial z_{j}}{\partial x_{j}}, \frac{\partial y_i}{\partial z_i}$ will not be significantly different for both $g_x^\text{mem}$ and $g_x^\text{learn}$, thus we regard the term, $\frac{\partial y_{\text{out}}}{\partial y_N} \cdot \prod_{j=i+1}^{N} \left( \frac{\partial y_j}{\partial z_{j}'} \cdot \frac{\partial z_{j}'}{\partial x_{j}'} \cdot \frac{\partial x_{j}'}{\partial z_{j}} \cdot \frac{\partial z_{j}}{\partial x_{j}} \right) \cdot \frac{\partial y_i}{\partial z_i'}$, as $A_1$ for both cases (as it does not vary across inputs). The only difference between the two gradients is due to $\frac{\partial \mathcal{L}}{\partial y_{\text{out}}}$, because the loss $\mathcal{L}$ is dependent on the label of both samples, i.e., $y_c$ and $y_\text{NL}$, as follows:
\begin{equation}
\mathcal{L} = - \sum_{k_i=1}^{C} y_{k_i} \log(\hat{y}_{k_i})
\label{eqn:cross_entropy_loss}
\end{equation}

where \( y_{k_i} = 1 \) if \( k_i \) is the ground truth class, otherwise 0, and $\hat{y}_{k_i}$ is the predicted softmax probability of class $k_i$. As a result, $g_{z_{i'}}^{\text{mem}}$ and $g_{z_{i'}}^{\text{learn}}$ can be respectively represented as follows:

\begin{equation}
g_{z_{i'}}^{\text{mem}} = \frac{\partial \mathcal{L^{\text{mem}}}}{\partial y^{\text{mem}}_{\text{out}}} \cdot A_1 
\quad \& \quad 
g_{z_{i'}}^{\text{learn}} = \frac{\partial \mathcal{L^{\text{learn}}}}{\partial y_{\text{out}}^{\text{learn}}} \cdot A_1
\label{eqn:g_x_i,6,mem_learn}
\end{equation}
\paragraph{Comparing learning and memorization gradients} Now the problem boils down to comparing the L2-norms of $\frac{\partial \mathcal{L^{\text{mem}}}}{\partial y_{\text{out}}^{\text{mem}}}$ and $\frac{\partial \mathcal{L^{\text{learn}}}}{\partial y_{\text{out}}^{\text{learn}}}$. To do so, we need to first define $\frac{\partial \mathcal{L}}{\partial y_{\text{out}}}$
We know that \(\mathcal{L}\) is the CrossEntropyLoss between the predicted softmax probability vector $\hat{y} = \text{Softmax}(y_{out})$ and the ground truth $y$ as defined in Eq.~\eqref{eqn:cross_entropy_loss}. Hence, $\frac{\partial \mathcal{L}}{\partial y_{\text{out}}}$ can be written as follows:
\begin{equation}
    \frac{\partial \mathcal{L}}{\partial y_{\text{out}}} = \hat{y} - y
    \label{eqn:dL/dy_out eqn}
\end{equation}

Now we can understand what it really means for memorization and learning. During inference of ($x_1, y_{\text{NL}}$), $\hat{y_{\text{mem}}}$ will be a vector which consists of very high probability ($\approx 1$) for class $y_{\text{NL}}$ while assigning extremely low probabilities ($\approx 0$) to remaining classes due to overfitting. Therefore, $\hat{y_{\text{mem}}} - y_{\text{NL}}$ will almost be a 0-vector, i.e., all the elements in the vector would be almost 0. This phenomenon is formally represented as follows:

\begin{equation}
\begin{aligned}
    \hat{y}_{\text{mem}} &\approx [0, \dots, 1, \dots, 0], \\
    y_{\text{NL}} &= [0, \dots, 1, \dots, 0], \\
    \hat{y}_{\text{mem}} - y_{\text{NL}} &\approx [0, \dots, 0, \dots, 0].
\end{aligned}
\label{eqn:mem_diff_vector_zero}
\end{equation}
where the index corresponding to class $y_{\text{NL}}$ is 1, and all other elements are (close to) 0. 

Now, we can take the L2-norm of both sides in Eq.~\eqref{eqn:dL/dy_out eqn} for the memorizing sample. Since $\hat{y_{\text{mem}}} - y_{\text{NL}}$ is close to a 0-vector, its L2-norm $\approx 0$. Therefore,

\begin{equation}
    \begin{aligned}
        \left\lVert \frac{\partial \mathcal{L}^{\text{mem}}}{\partial y^{\text{mem}}_{\text{out}}} \right\rVert_2 \approx 0
    \end{aligned}
    \label{eqn:mem_gradient_zero}
\end{equation}

On the other hand, for ($x_1, y_{\text{c}}$), even though the model has learned generalizable features for class \(c\), it has also been overfitted on noisy label samples. Hence, $\hat{y_{\text{learn}}}$ will not assign a very high probability to $y_c$ but instead will distribute some probability mass across multiple classes. Therefore, $\hat{y_{\text{learn}}} - y_{\text{c}}$ contains non-trivial, not near-zero values. This behavior is formally presented as follows:
\begin{equation}
\begin{aligned}
    \hat{y}_{\text{learn}} &= [p_1, p_2, \dots, p_{y_c}, \dots, p_C], \\
    y_c &= [0, \dots, 1, \dots, 0], \\
    \hat{y}_{\text{learn}} - y_c &= [p_1, p_2, \dots, p_{y_c}-1, \dots, p_C].
\end{aligned}
\label{eqn:learn_diff_vector_not_zero}
\end{equation}

$\text{where } p_{y_c} \text{ is not close to } 1, \text{ and other probabilities } p_i \text{ are non-negligible.}$

Now, by taking the L2-norm on both sides of Eq.~\eqref{eqn:dL/dy_out eqn} for the learning sample. Since $\hat{y_{\text{learn}}} - y_{\text{c}}$ contains non-trivial, non zero values, its L2-norm $\gg 0$. Therefore,

\begin{equation}
    \begin{aligned}
        \left\lVert \frac{\partial \mathcal{L}^{\text{learn}}}{\partial y^{\text{learn}}_{\text{out}}} \right\rVert_2 \gg 0
    \end{aligned}
    \label{eqn:learn_gradient_not_zero}
\end{equation}

\textbf{Therefore, by comparing Eq.~\eqref{eqn:mem_gradient_zero} \& ~\eqref{eqn:learn_gradient_not_zero}, we can establish the following relationship:}  
\begin{equation}
    \left\lVert \frac{\partial \mathcal{L}^{\text{learn}}}{\partial y^{\text{learn}}_{\text{out}}} \right\rVert_2 \geq \left\lVert \frac{\partial \mathcal{L}^{\text{mem}}}{\partial y^{\text{mem}}_{\text{out}}} \right\rVert_2,
    \label{eqn:mem_gen_gradients_l2_norm}
\end{equation} 
because overfitting on noisy samples, causes the memorizing gradients to be smaller than learning gradients.
However, note that an ideal case of perfect learning, where 100\% memorization and 100\% learning co-exist, is not achievable in practice as memorization inherently hinders generalization. Hence, the equality from the inequality can be disregarded in almost all cases.

Now, substituting the relation found in Eq.~\eqref{eqn:mem_gen_gradients_l2_norm} to the L2-norms of $g_{z_{i}'}^{\text{learn}}$ and $g_{z_{i}'}^{\text{mem}}$ in Eq.~\eqref{eqn:g_x_i,6,mem_learn}, we can formally conclude that:

\begin{equation}
    \left\lVert g_{z_{i}'}^{\text{learn}} \right\rVert_2 \geq \left\lVert g_{z_{i}'}^{\text{mem}} \right\rVert_2
    \label{g_x_i,6_l2_norm}
\end{equation}

This proof explains why $\left\lVert g_{z_{i}'}^{\text{mem}} \right\rVert_2$ is lower than  $\left\lVert g_{z_{i}'}^{\text{learn}} \right\rVert_2$ across all layers, as also empirically observed in Fig.~\ref{fig:Pre_Post_Gradients_emotions_deberta_gpt_neo}.

\subsubsection{Backpropagating gradient analysis for $\text{LN}_1$ \( (g_{z_{i}}) \):} Similar to \( g_{z_{i}'} \), we can express \( g_{z_{i}} \) as follows:
\begin{equation}
\begin{aligned}
g_{z_{i}} = \frac{\partial \mathcal{L}}{\partial z_{i}} = \frac{\partial \mathcal{L}}{\partial y_{\text{out}}} \cdot \frac{\partial y_{\text{out}}}{\partial y_N} \cdot \prod_{j=i+1}^{N} \left( \frac{\partial y_j}{\partial z_{j}'} \cdot \frac{\partial z_{j}'}{\partial x_{j}'} \cdot \frac{\partial x_{j}'}{\partial z_{j}} \cdot \frac{\partial z_{j}}{\partial x_{j}} \right) \cdot \frac{\partial y_i}{\partial z_{i}'} \cdot \frac{\partial z_{i}'}{\partial x_{i}'} \cdot \frac{\partial x_{i}'}{\partial z_i}
\end{aligned}
\end{equation}
Here, $x_{i+1} = y _{i}$ because $i^{\text{th}}$ layer's output $y_{i}$ is the input for $(i+1)^{\text{th}}$ layer, $x_{i+1}$.

We compute $g_{z_{i}}^{\text{mem}}$ and $g_{z_{i}}^{\text{learn}}$ for both memorization ($x_1, y_{\text{NL}}$) and learning ($x_2, y_{\text{c}}$) samples, respectively. Based on the discussion in Sec.~\ref{para:g_x_i,6_discussion} on the similarity of $x_1$ and $x_2$, since they both originally belong to the same class c, we can write $g_{z_{i'}}^{\text{mem}}$ and $g_{z_{i'}}^{\text{learn}}$ as follows:
\begin{equation}
g_{z_{i}}^{\text{mem}} = \frac{\partial \mathcal{L^{\text{mem}}}}{\partial y^{\text{mem}}_{\text{out}}} \cdot A_2 
\quad \& \quad 
g_{z_{i}}^{\text{learn}} = \frac{\partial \mathcal{L^{\text{learn}}}}{\partial y_{\text{out}}^{\text{learn}}} \cdot A_2
\label{eqn:g_x_i,3,mem_learn}
\end{equation}
where, $A_2 = \frac{\partial y_{\text{out}}}{\partial y_N} \cdot \prod_{j=i+1}^{N} \left( \frac{\partial y_j}{\partial z_{j}'} \cdot \frac{\partial z_{j}'}{\partial x_{j}'} \cdot \frac{\partial x_{j}'}{\partial z_{j}} \cdot \frac{\partial z_{j}}{\partial x_{j}} \right) \cdot \frac{\partial y_i}{\partial z_{i}'} \cdot \frac{\partial z_{i}'}{\partial x_{i}'} \cdot \frac{\partial x_{i}'}{\partial z_i}$, which does not vary across inputs.

To compare $g_{z_i}^{\text{mem}}$ and $g_{z_i}^{\text{learn}}$, we use the argument made in Eq.~\eqref{eqn:mem_diff_vector_zero} \& ~\eqref{eqn:learn_diff_vector_not_zero}, which explains why $\left\lVert \frac{\partial \mathcal{L}^{\text{learn}}}{\partial y^{\text{learn}}_{\text{out}}} \right\rVert_2 \geq \left\lVert \frac{\partial \mathcal{L}^{\text{mem}}}{\partial y^{\text{mem}}_{\text{out}}} \right\rVert_2$.

Using the above results and subsituting the relation in the L2-norms of $g_{z_{i}}^{\text{learn}}$ and $g_{z_{i}}^{\text{mem}}$ in Eq.~\eqref{eqn:g_x_i,3,mem_learn}, we can conclude that:
\begin{equation}
    \left\lVert g_{z_{i}}^{\text{learn}} \right\rVert_2 \geq \left\lVert g_{z_{i}}^{\text{mem}} \right\rVert_2
    \label{g_x_i,3_l2_norm}
\end{equation}

In conclusion, both Eq.~\eqref{g_x_i,6_l2_norm} \& ~\eqref{g_x_i,3_l2_norm} formally demonstrate that the L2-norm of \textbf{\emph{learning gradient}}, $g_x^\text{learn}$ is greater than or equal to \textbf{\emph{memorization gradient}}, $g_x^\text{mem}$ across all layers of a Post-LN model.

\subsection{\emph{For Pre-LN model:}} Firstly, we describe the architecture of the Pre-LN transformer in Tab.~\ref{tab:post_pre_eqns}. Based on the Pre-LN architecture, during backpropagation, we compute derivatives of loss wrt input of LN for every $i^{th}$ layer. Since there are two LNs in every layer, we compute and compare the learning and memorization gradients corresponding to the inputs of both LNs, i.e., $g_{x_{i'}}$ and $g_{x_{i}}$.

\subsubsection{Backpropagating gradient analysis for $\text{LN}_2$ \( (g_{x_{i}'}) \):}\label{para:g_x_i_discussion_pre} 
The backpropagating gradient for \( g_{x_{i'}} \) can be expressed as follows:

\begin{equation}
\begin{aligned}
g_{x_{i}'} = \frac{\partial \mathcal{L}}{\partial x_{i}'}= \frac{\partial \mathcal{L}}{\partial y_{\text{out}}} \cdot \frac{\partial y_{\text{out}}}{\partial y_N} \cdot \prod_{j=i+1}^{N} \left( \frac{\partial y_j}{\partial x_{j}'} \cdot \frac{\partial x_{j'}}{\partial x_{j}} \right) \cdot \frac{\partial y_i}{\partial x_{i}'}
\end{aligned}
\label{eq:xjprimeGradient}
\end{equation}
where $x_{i+1} = y _{i}$ because $i^{\text{th}}$ layer's output $y_{i}$ is the input of $(i+1)^{\text{th}}$ layer, $x_{i+1}$. To measure memorization and learning, we then compute these gradients for both \((x_1, y_{\text{NL}})\) and \((x_2, y_c)\). In both gradients, $y_\text{out}, y_j, x_{j}', x_j$ are only dependent on the input samples $x_1$ and $x_2$, respectively, and not on their labels, where both $x_1$ and $x_2$ genuinely represent class c. Therefore, $\frac{\partial y_{\text{out}}}{\partial y_N}, \frac{\partial y_j}{\partial x_{j}'}, \frac{\partial x_{j}'}{\partial x_{j}}, \frac{\partial y_i}{\partial x_i'}$ will not be significantly different for both $g_x^\text{learn}$ and $g_x^\text{mem}$, thus we regard the term, $\frac{\partial y_{\text{out}}}{\partial y_N} \cdot \prod_{j=i+1}^{N} \left( \frac{\partial y_j}{\partial x_{j}'} \cdot \frac{\partial x_{j}'}{\partial x_{j}} \right) \cdot \frac{\partial y_i}{\partial x_i'}$, as $B_1$ for both cases (as it does not vary across inputs.) The only difference between the two gradients is due to $\frac{\partial \mathcal{L}}{\partial y_{\text{out}}}$, because the loss $\mathcal{L}$ is dependent on the label of both samples, i.e., $y_c$ and $y_\text{NL}$ as shown in Eq.~\eqref{eqn:cross_entropy_loss}. As a result, $g_{x_{i'}}^{mem}$ and $g_{x_{i'}}^{learn}$ can be respectively represented as follows:

\begin{equation}
g_{x_{i'}}^{\text{mem}} = \frac{\partial \mathcal{L^{\text{mem}}}}{\partial y^{\text{mem}}_{\text{out}}} \cdot B_1 
\quad \& \quad 
g_{x_{i'}}^{\text{learn}} = \frac{\partial \mathcal{L^{\text{learn}}}}{\partial y_{\text{out}}^{\text{learn}}} \cdot B_1
\label{eqn:g_x_i',mem_learn_pre}
\end{equation}

\paragraph{Comparing learning and memorization gradient norms} Now the problem boils down to comparing the norms of $\frac{\partial \mathcal{L^{\text{mem}}}}{\partial y_{\text{out}}^{\text{mem}}}$ and $\frac{\partial \mathcal{L^{\text{learn}}}}{\partial y_{\text{out}}^{\text{learn}}}$.

\textbf{From Eq.~\eqref{eqn:mem_gen_gradients_l2_norm}, we know the following relation:}  
\begin{equation}
    \left\lVert \frac{\partial \mathcal{L}^{\text{learn}}}{\partial y^{\text{learn}}_{\text{out}}} \right\rVert_2 \geq \left\lVert \frac{\partial \mathcal{L}^{\text{mem}}}{\partial y^{\text{mem}}_{\text{out}}} \right\rVert_2,
\end{equation}

Now, substituting the relation found in Eq.~\eqref{eqn:mem_gen_gradients_l2_norm} to the l2-norms of $g_{x_{i}'}^{\text{learn}}$ and $g_{x_{i}'}^{\text{mem}}$ in Eq.~\eqref{eqn:g_x_i',mem_learn_pre}, we can formally conclude that:

\begin{equation}
    \left\lVert g_{x_{i}'}^{\text{learn}} \right\rVert_2 \geq \left\lVert g_{x_{i}'}^{\text{mem}} \right\rVert_2
    \label{g_x_i'_l2_norm_pre}
\end{equation}

This proof explains why $\left\lVert g_{x_{i}'}^{\text{mem}} \right\rVert_2$ is lower than  $\left\lVert g_{x_{i}'}^{\text{learn}} \right\rVert_2$ across all layers, as also empirically consistently observed in Fig.~\ref{fig:Pre_Post_Gradients_emotions_deberta_gpt_neo}.

\subsubsection{Backpropagating gradient analysis for $\text{LN}_1$ \( (g_{x_{i}}) \):} Similar to \( g_{x_{i'}} \), we can express \( g_{x_{i}} \) as follows:
\begin{equation}
\begin{aligned}
g_{x_{i}} = \frac{\partial \mathcal{L}}{\partial x_{i}}= \frac{\partial \mathcal{L}}{\partial y_{\text{out}}} \cdot \frac{\partial y_{\text{out}}}{\partial y_N} \cdot \prod_{j=i+1}^{N} \left( \frac{\partial y_j}{\partial x_{j}'} \cdot \frac{\partial x_{j'}}{\partial x_{j}} \right) \cdot \frac{\partial y_i}{\partial x_{i}'} \cdot \frac{\partial x_i'}{\partial x_{i}}
\end{aligned}
\label{eq:xjGradient}
\end{equation}
where $x_{i+1} = y _{i}$ because $i^{\text{th}}$ layer's output $y_{i}$ is the input for $(i+1)^{\text{th}}$ layer, $x_{i+1}$, and compute $g_{z_{i}}^{\text{mem}}$ and $g_{z_{i}}^{\text{learn}}$ to measure memorization and learning respectively. Based on the discussion in Sec.~\ref{para:g_x_i,6_discussion} on the similarity of $x_1$ and $x_2$ since they both genuinely belong to the same class c, we can write $g_{x_{i}}^{mem}$ and $g_{x_{i}}^{learn}$ as follows:
\begin{equation}
g_{x_{i}}^{\text{mem}} = \frac{\partial \mathcal{L^{\text{mem}}}}{\partial y^{\text{mem}}_{\text{out}}} \cdot B_2 
\quad \& \quad 
g_{x_{i}}^{\text{learn}} = \frac{\partial \mathcal{L^{\text{learn}}}}{\partial y_{\text{out}}^{\text{learn}}} \cdot B_2
\label{eqn:g_x_i,mem_learn_pre}
\end{equation}
where, $B_2 = \frac{\partial y_{\text{out}}}{\partial y_N} \cdot \prod_{j=i+1}^{N} \left( \frac{\partial y_j}{\partial x_{j}'} \cdot \frac{\partial x_{j'}}{\partial x_{j}} \right) \cdot \frac{\partial y_i}{\partial x_{i}'} \cdot \frac{\partial x_i'}{\partial x_{i}}$, which does not vary across inputs. To compare $g_{x_i}^{\text{mem}}$ and $g_{x_i}^{\text{learn}}$, we use Eq.~\eqref{eqn:mem_gen_gradients_l2_norm}, which states that $\left\lVert \frac{\partial \mathcal{L}^{\text{learn}}}{\partial y^{\text{learn}}_{\text{out}}} \right\rVert_2 \geq \left\lVert \frac{\partial \mathcal{L}^{\text{mem}}}{\partial y^{\text{mem}}_{\text{out}}} \right\rVert_2$.

Using the above results and substituting the relation in the L2-norms of $g_{x_{i}}^{\text{learn}}$ and $g_{x_{i}}^{\text{mem}}$ in Eq.~\eqref{eqn:g_x_i,mem_learn_pre}, we can conclude that:
\begin{equation}
    \left\lVert g_{x_{i}}^{\text{learn}} \right\rVert_2 \geq \left\lVert g_{x_{i}}^{\text{mem}} \right\rVert_2
    \label{g_x_i_l2_norm_pre}
\end{equation}

In conclusion, both Eq.~\eqref{g_x_i'_l2_norm_pre} \& ~\eqref{g_x_i_l2_norm_pre} formally demonstrate that the L2-norm of \textbf{\emph{learning gradient}}, $g_x^\text{learn}$ is greater than or equal to \textbf{\emph{memorization gradient}}, $g_x^\text{mem}$ across all layers of a Pre-LN model.

\hfill \qedsymbol{}

\bigskip

\begin{mdframed}[
    linewidth=0.5pt,
    roundcorner=4pt,
    backgroundcolor=gray!8,
    linecolor=gray!50,
    innertopmargin=2pt,
    innerbottommargin=2pt,
    innerleftmargin=6pt,
    innerrightmargin=6pt,
    skipabove=6pt,
    skipbelow=6pt
]
\section{Theorem 2: Gradient norm of loss \texorpdfstring{$\mathcal{L}$}{L} w.r.t input of LN is upper bounded.} \label{theorem:l2_norm_bound_thm_app}

\textbf{\emph{Post-LN: Let $z_i$ denote the input to $\text{LN}_1$ of the $i^\text{th}$ Post-LN model layer. Then, }}
\begin{equation}
\begin{aligned}
    \| g_{z_{i}} \|_2 = \left\| \frac{\partial \mathcal{L}}{\partial z_{i}} \right\|_2 \leq & s_\text{max}(P_1) \cdot  \left( \frac{1}{\prod_{j=i}^{N} \left|1 - \sqrt{\text{Var}(\text{FFN}(x_{j}')})\right| \left|1 - \sqrt{\text{Var}(\text{MHSA}(x_{j})})\right|} \right) \cdot \\ & \cdot \prod_{j=i}^{N} \left( 1 + s_{\text{max}}(J_{\text{FFN}}^{x_{j}'}) \right) \cdot \prod_{j=i+1}^{N} \left( 1 + s_{\text{max}}(J_{\text{MHSA}}^{x_{j}}) \right)
\end{aligned}
\end{equation}

\textbf{\emph{Pre-LN: Let $x_i$ denote the input to $\text{LN}_1$ of the $i^\text{th}$ Pre-LN model layer. Then, }}
\begin{equation}
\begin{aligned}
\|g_{x_{i}}\|_2 = \left\|\frac{\partial \mathcal{L}}{\partial x_{i}}\right\|_2 \leq s_\text{max}(P_2) \cdot \prod_{j=i}^{N} \left( (1 + s_\text{max}(J_{\text{FFN}}^{\text{LN}_2(x_{j}')} J_{\text{LN}_2}^{x_{j}'})\right) \cdot \prod_{j=i}^{N} \left((1 + s_\text{max}(J_{\text{MHSA}}^{\text{LN}_1(x_{j})} J_{\text{LN}_1}^{x_{j}})\right)
\end{aligned}
\end{equation}
\end{mdframed}

\textbf{\emph{Proof:}}

\subsection{For Post-LN model:}
The Post-LN model setup for \(i^{th}\) layer can be represented as follows:
\begin{equation}
\begin{aligned}
    x_{i}'&= \text{LN}_1(x_i + \text{MHSA}(x_i)) \\
    y_i &= \text{LN}_2(x_{i}'+ \text{FFN}(x_{i}'))
\end{aligned}
\end{equation}
where $x_i + \text{MHSA}(x_i)$ and $x_{i}' + \text{FFN}(x_{i}')$ are the inputs to $\text{LN}_1$ and $\text{LN}_2$, respectively. Later, we substitute and use them as
\begin{equation}
\begin{aligned}
z_{i} = x_i + \text{MHSA}(x_i)\\
z_{i'} = x_{i}' + \text{FFN}(x_{i}')
\end{aligned}
\label{eq:z_substitution}
\end{equation}

Since there are two LayerNorm (LN) operations in every layer, we prove it seperately for both of them.

\subsubsection{Backpropagation analysis for $\text{LN}_2$ $(g_{z_{i}'})$:} By applying Eq.~\eqref{eq:z_substitution}, we obtain $z_{j}' = x_{j}' + \text{FFN}(x_{j}')$, $z_{j} = x_{j} + \text{MHSA}(x_{j})$. Hence, we can write $g_{z_{i}'}$ (from Eq.~\eqref{eq:zjprimeGradient}) for the \(i^{th}\) layer as follows:

\begin{equation}
\begin{aligned}
g_{z_{i}'} = \frac{\partial \mathcal{L}}{\partial z_{i}'} = \frac{\partial \mathcal{L}}{\partial y_{\text{out}}} \cdot \frac{\partial y_{\text{out}}}{\partial y_N} \cdot \prod_{j=i+1}^{N} \left( \frac{\partial y_j}{\partial z_{j}'} \cdot \frac{\partial z_{j}'}{\partial x_j'} \cdot \frac{\partial x_j'}{\partial z_{j}} \cdot \frac{\partial z_{j}}{\partial x_j} \right) \cdot \frac{\partial y_i}{\partial z_{i}'}
\end{aligned}
\end{equation}

Here, $\frac{\partial \mathcal{L}}{\partial y_{\text{out}}} \cdot \frac{\partial y_{\text{out}}}{\partial y_N}$, is independent of the transformer's layers as they are computed using the classification head's output. Therefore, we can treat them as \(P_1\) (which does not vary across layers). We also compute the corresponding derivatives of $z_{j'}$ and $z_{j}$. Lastly, $x_{i+1}$ is same as $y_i$, because $y_i$ is the output of the $i^\text{th}$ layer which becomes input $x_{i+1}$ of the $(i+1)^\text{th}$ layer. By applying all of these, we obtain the following equation:

\begin{equation}
\begin{aligned}
g_{z_{i}'} = \frac{\partial \mathcal{L}}{\partial z_{i}'} = P_1 \cdot \prod_{j=i+1}^{N} \left( \frac{\partial y_j}{\partial z_{j}'} \cdot \frac{\partial (x_{j}'+ \text{FFN}(x_{j'}))}{\partial x_{j}'} \cdot \frac{\partial x_{j}'}{\partial z_{j}} \cdot \frac{\partial (x_{j} + \text{MHSA}(x_{j}))}{\partial x_j} \right) \cdot \frac{\partial y_i}{\partial z_{i}'}
\end{aligned}
\end{equation}

\begin{equation}
\begin{aligned}
\quad \quad = P_1 \cdot \prod_{j=i+1}^{N} \left( \frac{\partial y_j}{\partial z_{j}'} \cdot (\text{I} + \frac{\partial \text{FFN}(x_{j}')}{\partial x_{j}'}) \cdot \frac{\partial x_{j}'}{\partial z_{j}} \cdot (\text{I} + \frac{\partial \text{MHSA}(x_{j})}{\partial x_{j}}) \right) \cdot \frac{\partial y_i}{\partial z_{i}'}
\end{aligned}
\end{equation}

We also acknowledge that $\frac{\partial y_j}{\partial z_{j}}, \frac{\partial x_{j}'}{\partial z_{j}'},$  and $\frac{\partial y_i}{\partial z_{i}'}$, are derivatives of output of LN w.r.t their input, which can be simply represented as Jacobian matrices, $J_{\text{LN}_2}^{z_{j}}, J_{\text{LN}_1}^{z_{j}'}$, and  $J_{\text{LN}_2}^{z_{i}'}$, respectively. Likewise, $\frac{\partial \text{FFN}(x_{j}')}{\partial x_{j}'}$ and $\frac{\partial \text{MHSA}(x_{j})}{\partial x_{j}}$ are also derivatives of the output of FFN and MHSA w.r.t their inputs, and can be represented as Jacobian matrices, $J_{\text{FFN}}^{x_{j}'}$ and $J_{\text{MHSA}}^{x_{j}}$, respectively. After substituting these terms, we obtain the following:

\begin{equation}
\begin{aligned}
g_{z_{i}'} = \frac{\partial \mathcal{L}}{\partial z_{i}'} = P_1 \cdot \prod_{j=i+1}^{N} \left( J_{\text{LN}_2}^{z_{j}'} \cdot (\text{I} + J_{\text{FFN}}^{x_{j}'}) \cdot J_{\text{LN}_1}^{z_{j}} \cdot (\text{I} + J_{\text{MHSA}}^{x_{j}}) \right) \cdot J_{\text{LN}_2 }^{z_{i}'}
\end{aligned}
\label{eqn:post_ln_z2_grad_jacobian}
\end{equation}

Now, we take the L2-norm on both sides of Eq.~\eqref{eqn:post_ln_z2_grad_jacobian} as follows:
\begin{equation}
\begin{aligned}
\| g_{z_{i}'} \|_2 &= \left\| \frac{\partial \mathcal{L}}{\partial z_{i}'} \right\|_2
= \left\| P_1 \cdot \prod_{j=i+1}^{N} \left( J_{\text{LN}_2}^{z_{j}'} \cdot (\text{I} + J_{\text{FFN}}^{x_{j}'}) \cdot J_{\text{LN}_1}^{z_{j}} \cdot (\text{I} + J_{\text{MHSA}}^{x_{j}}) \right) \cdot J_{\text{LN}_2}^{z_{i}'} \right\|_2
\end{aligned}
\label{eqn:post_ln_z2_grad_jacobian_l2_norm}
\end{equation}

We know that L2-norm of a matrix is equivalent to its largest singular value \citep{Horn_Johnson_1991}. Hence, we can further write Eq.~\eqref{eqn:post_ln_z2_grad_jacobian_l2_norm} as follows:
\begin{equation}
\begin{aligned}
\| g_{z_{i}'} \|_2 &= \left\| \frac{\partial \mathcal{L}}{\partial z_{i}'} \right\|_2
= s_{\text{max}} (P_1 \cdot \prod_{j=i+1}^{N} \left( J_{\text{LN}_2}^{z_{j}'} \cdot (\text{I} + J_{\text{FFN}}^{x_{j}'}) \cdot J_{\text{LN}_1}^{z_{j}} \cdot (\text{I} + J_{\text{MHSA}}^{x_{j}}) \right) \cdot J_{\text{LN}_2}^{z_{i}'})
\end{aligned}
\label{eqn:post_ln_z2_grad_jacobian_l2_norm_step2}
\end{equation}
where $s_{\text{max}}$ outputs the largest singular value of $(P_1 \cdot \prod_{j=i+1}^{N} \left( J_{\text{LN}_2}^{z_{j}'} \cdot (\text{I} + J_{\text{FFN}}^{x_{j}'}) \cdot J_{\text{LN}_1}^{z_{j}} \cdot (\text{I} + J_{\text{MHSA}}^{x_{j}}) \right) \cdot J_{\text{LN}_2}^{z_{i}'})$. From the properties of singular values \citep{Horn_Johnson_1991}, we know that 
\begin{equation}
    \begin{aligned}
        s_{\text{max}} (A_{1} A_{2} \dots A_{n}) &\leq s_{\text{max}}(A_{1}) s_{\text{max}}(A_{2}) \dots s_{\text{max}}(A_{n}) \\
        s_{\text{max}}(A_1 + A_2) &\leq s_{\text{max}}(A_1) + s_{\text{max}}(A_2)
    \end{aligned}
    \label{eqn:singular_vals_properties}
\end{equation}
where $s_{\text{max}}(A_k)$ is the maximum singular value of matrix $A_k$. After applying these properties to Eq.~\eqref{eqn:post_ln_z2_grad_jacobian_l2_norm_step2}, we get the following:
\begin{equation}
\begin{aligned}
\| g_{z_{i}'} \|_2 &= \left\| \frac{\partial \mathcal{L}}{\partial z_{i}'} \right\|_2 \\
&\leq s_\text{max}(P_1) \cdot  
\prod_{j=i+1}^{N} \Bigg( 
s_{\text{max}}(J_{\text{LN}_2}^{z_{j}'}) \cdot 
\big( s_{\text{max}}(\text{I}) + s_{\text{max}}(J_{\text{FFN}}^{x_{j}'}) \big) \cdot \\
&\hspace{3cm} s_{\text{max}}(J_{\text{LN}_1}^{z_{j}}) \cdot 
\big( s_{\text{max}}(\text{I}) + s_{\text{max}}(J_{\text{MHSA}}^{x_{j}}) \big) 
\Bigg) \cdot s_{\text{max}}(J_{\text{LN}_2}^{z_{i}'})
\end{aligned}
\end{equation}

According to \cite{xiong2020layer}, we can rewrite the Jacobian of LNs as follows:
\begin{equation}
    \begin{aligned}
        J_{\text{LN}_1}^{z_{j}} = \frac{\text{I}}{\sigma_{z_{j}}}, \quad 
        J_{\text{LN}_2}^{z_{j}'} = \frac{\text{I}}{\sigma_{z_{j}'}}, \quad and \quad 
        J_{\text{LN}_2}^{z_{i}'} = \frac{\text{I}}{\sigma_{z_{i}'}}
    \end{aligned}
\end{equation}
where $\sigma_{z_{j}}$, $\sigma_{z_{j}'}$, and $\sigma_{z_{i}'}$ are the standard-deviations of $z_{j}$, $z_{j}'$, and $z_{i}'$, respectively. Therefore, we obtain the following equation:
\begin{equation}
\begin{aligned}
\| g_{z_{i}'} \|_2 &= \left\| \frac{\partial \mathcal{L}}{\partial z_{i}'} \right\|_2 \\
&\leq s_\text{max}(P_1) \cdot  
\prod_{j=i+1}^{N} \Bigg( 
s_{\text{max}}\left(\frac{\text{I}}{\sigma_{z_{j}'}}\right) \cdot 
\big( s_{\text{max}}(\text{I}) + s_{\text{max}}(J_{\text{FFN}}^{x_{j}'}) \big) \cdot \\
&\hspace{3cm} s_{\text{max}}\left(\frac{\text{I}}{\sigma_{z_{j}}}\right) \cdot 
\big( s_{\text{max}}(\text{I}) + s_{\text{max}}(J_{\text{MHSA}}^{x_{j}}) \big) 
\Bigg) \cdot s_{\text{max}}\left(\frac{\text{I}}{\sigma_{z_{i}'}}\right).
\end{aligned}
\label{eqn:post_ln_z2_grad_jacobian_l2_norm_step3}
\end{equation}

Another property of singular values states that all singular values of identity matrix $\text{I}$ are 1 \citep{Horn_Johnson_1991}, i.e., $s_k(\text{I}) = 1$. Therefore substituting with that in Eq.~\eqref{eqn:post_ln_z2_grad_jacobian_l2_norm_step3}, we obtain the following:
\begin{equation}
\begin{aligned}
\| g_{z_{i}'} \|_2 &= \left\| \frac{\partial \mathcal{L}}{\partial z_{i}'} \right\|_2
\leq s_\text{max}(P_1) \cdot \prod_{j=i+1}^{N} \left(\frac{1}{\sigma_{z_{j}'}} \cdot (1 + s_{\text{max}}(J_{\text{FFN}}^{x_{j}'})) \cdot \frac{1}{\sigma_{z_{j}}} \cdot (1 + s_{\text{max}}(J_{\text{MHSA}}^{x_{j}})) \right) \cdot \frac{1}{\sigma_{z_{i}'}}
\end{aligned}
\label{eqn:post_ln_z2_grad_jacobian_l2_norm_step4}
\end{equation}

By re-arranging the terms, we finally obtain the following equation:
\begin{equation}
\begin{aligned}
\| g_{z_{i}'} \|_2 &= \left\| \frac{\partial \mathcal{L}}{\partial z_{i}'} \right\|_2
\leq s_\text{max}(P_1) \cdot \left( \frac{1}{\prod_{j=i}^{N} \sigma_{z_{j}'}} \right) 
\cdot \left( \frac{1}{\prod_{j=i+1}^{N} \sigma_{z_{j}}} \right) \\
&\quad \cdot \prod_{j=i+1}^{N} \left( 
\left( 1 + s_{\text{max}}(J_{\text{FFN}}^{x_{j}'}) \right) 
\cdot \left( 1 + s_{\text{max}}(J_{\text{MHSA}}^{x_{j}}) \right) 
\right)
\end{aligned}
\label{eqn:post_ln_z2_grad_jacobian_l2_norm_step4_rearranged}
\end{equation}

Now, we need to investigate how $\sigma_{z_{j}}$ and $\sigma_{z_{j'}}$ behave. We know that $\sigma_{z_{j}} = \sqrt{\text{Var}(z_{j})}$ and $\sigma_{z_{j}'} = \sqrt{\text{Var}(z_{j}')}$, where Var($\cdot$) represents a variance. Therefore, we can instead focus on $\text{Var}(z_{j})$ and $\text{Var}(z_{j}')$. We know that $z_{j} = x_{j} + \text{MHSA}(x_{j})$ and $z_{j}' = x_{j}
'+ \text{FFN}(x_{j}')$. Therefore, their variances can be written as follows:

\begin{equation}
    \begin{aligned}
        \text{Var}(z_{j}) = \text{Var}(x_{j} + \text{MHSA}(x_{j}))\\
        \text{Var}(z_{j}') = \text{Var}(x_{j}' + \text{FFN}(x_{j}'))
    \end{aligned}
\end{equation}

Now to compute the upper bound of Eq.~\eqref{eqn:post_ln_z2_grad_jacobian_l2_norm_step4_rearranged}, we need to substitute the lower bound of $\sigma_{z_{j}}$ and $\sigma_{z_{j}'}$ as they are in the denominator. The lower bounds of $\sigma_{z_{j}}$ and $\sigma_{z_{j}'}$ would basically be lower bounds of $\text{Var}(z_{j})$ and $\text{Var}(z_{j}')$, respectively.

From \citet{hossjer2022sharp}, we know that for any two matrices A and B, 
\begin{equation}
    \begin{aligned}
        \text{Var}(\text{A} + \text{B}) \geq \left( \sqrt{\text{Var(A)}} - \sqrt{\text{Var(B)}} \right)^2
    \end{aligned}
\end{equation}

Therefore, the lower bounds of $\text{Var}(z_{j})$ and $\text{Var}(z_{j}')$ can be written as follows, 
\begin{equation}
    \begin{aligned}
        \text{Var}(z_{j}) \geq \left( \sqrt{\text{Var}(x_{j})} - \sqrt{\text{Var}(\text{MHSA}(x_{j}))} \right)^2 \\
        \text{Var}(z_{j}') \geq \left( \sqrt{\text{Var}(x_{j}')} - \sqrt{\text{Var}(\text{FFN}(x_{j}'))} \right)^2
    \end{aligned}
    \label{eqn:var_lower_bound_post}
\end{equation}

Since $x_{j}$ and $x_{j}'$ are the outputs of the two LN layers, their variance is 1. Therefore, we can rewrite Eq.~\eqref{eqn:var_lower_bound_post} as follows: 
\begin{equation}
    \begin{aligned}
        \text{Var}(z_{j}) &\geq \left( 1 - \sqrt{\text{Var}(\text{MHSA}(x_{j}))} \right)^2 
        \quad \Rightarrow \quad 
        \sigma_{z_j} \geq \left| 1 -\sqrt{\text{Var}( \text{MHSA}(x_j))} \right| \\
        \text{Var}(z_{j}') &\geq \left( 1 - \sqrt{\text{Var}(\text{FFN}(x_{j}'))} \right)^2 
        \quad \Rightarrow \quad 
        \sigma_{z'_j} \geq \left| 1 -\sqrt{\text{Var}( \text{FFN}(x_j'))} \right|
    \end{aligned}
    \label{eqn:var_lower_bound_post_rewrite}
\end{equation}

Hence, we can re-write Eq.~\eqref{eqn:post_ln_z2_grad_jacobian_l2_norm_step4_rearranged} as follows:

\begin{equation}
\begin{aligned}
    \| g_{z_{i}'} \|_2 = \left\| \frac{\partial \mathcal{L}}{\partial z_{i}'} \right\|_2 \leq & s_\text{max}(P_1) \cdot  \left( \frac{1}{\prod_{j=i}^{N} \left|1 - \sqrt{\text{Var}(\text{FFN}(x_{j}')}) \right|} \right) \cdot \left( \frac{1}{\prod_{j=i+1}^{N} \left| 1 - \sqrt{\text{Var}(\text{MHSA}(x_{j})}) \right|} \right) \\ & \cdot \prod_{j=i+1}^{N} \left( \left( 1 + s_{\text{max}}(J_{\text{FFN}}^{x_{j}'}) \right) \cdot \left( 1 + s_{\text{max}}(J_{\text{MHSA}}^{x_{j}}) \right) \right)
\end{aligned}
\label{eqn:post_ln_z2_grad_jacobian_l2_norm_step4_rearranged_var}
\end{equation}

\subsubsection{Backpropagation analysis for $\text{LN}_1$ $(g_{z_{i}})$:} Similar to $g_{z_{i}'}$, we can express $g_{z_{i}}$ as follows:

\begin{equation}
\begin{aligned}
g_{z_{i}} = \frac{\partial \mathcal{L}}{\partial z_{i}} = \frac{\partial \mathcal{L}}{\partial y_{\text{out}}} \cdot \frac{\partial y_{\text{out}}}{\partial y_N} \cdot \prod_{j=l+1}^{N} \left( \frac{\partial y_j}{\partial z_{j}'} \cdot \frac{\partial z_{j}'}{\partial x_{j}'} \cdot \frac{\partial x_{j}'}{\partial z_{j}} \cdot \frac{\partial z_{j}}{\partial x_j} \right) \cdot \frac{\partial y_i}{\partial z_{i}'} \cdot \frac{\partial z_{i}'}{\partial x_{i}'} \cdot \frac{\partial x_{i}'}{\partial z_{i}}
\end{aligned}
\label{eqn:post_z2_grad}
\end{equation}

Here, $\frac{\partial \mathcal{L}}{\partial y_{\text{out}}} \cdot \frac{\partial y_{\text{out}}}{\partial y_N}$, is independent of the transformer's layers as they are computed using the classification head's output. Therefore, we can treat them as \(P_1\) (which does not vary across layers). We also compute the corresponding derivatives of $z_{j'}$ and $z_{j}$. Lastly, $x_{i+1}$ is same as $y_i$, because $y_i$ is the output of the $i^\text{th}$ layer which becomes input $x_{i+1}$ of the $(i+1)^\text{th}$ layer. By applying all of these, we obtain the following equation:

\begin{equation}
\begin{aligned}
g_{z_{i}} = \frac{\partial \mathcal{L}}{\partial z_{i}} &= 
P_1 \cdot \prod_{j=i+1}^{N} \Bigg( 
\frac{\partial y_j}{\partial z_{j}'} 
\cdot \frac{\partial (x_{j}' + \text{FFN}(x_{j}'))}{\partial x_{j}'} 
\cdot \frac{\partial x_{j}'}{\partial z_{j}} 
\cdot \frac{\partial (x_{j} + \text{MHSA}(x_{j}))}{\partial x_j} 
\Bigg) \\
&\quad \cdot \frac{\partial y_{i}}{\partial z_{i}'} 
\cdot \frac{\partial (x_i' + \text{FFN}(x_i'))}{\partial x_{i}'} 
\cdot \frac{\partial x_{i}'}{\partial z_{i}}
\end{aligned}
\end{equation}

\begin{equation}
\begin{aligned}
\quad\quad\quad\quad 
=\, &P_1 \cdot \prod_{j=i+1}^{N} \Bigg( 
\frac{\partial y_j}{\partial z_{j}'} 
\cdot \left( \text{I}+ \frac{\partial\, \text{FFN}(x_{j}')}{\partial x_{j}'} \right) 
\cdot \frac{\partial x_{j}'}{\partial z_{j}} 
\cdot \left( \text{I}+ \frac{\partial\, \text{MHSA}(x_{j})}{\partial x_{j}} \right) 
\Bigg) \\
&\cdot \frac{\partial y_i}{\partial z_{i}'} 
\cdot \left( \text{I}+ \frac{\partial\, \text{FFN}(x_i')}{\partial x_{i}'} \right) 
\cdot \frac{\partial x_{i}'}{\partial z_{i}}
\end{aligned}
\end{equation}

Clearly, we can see that $\frac{\partial y_j}{\partial z_{j}}, \frac{\partial x_{j}'}{\partial z_{j}'},$ $\frac{\partial y_i}{\partial z_{i}'}$, and $\frac{\partial x_i'}{\partial z_{i}}$ are derivatives of output of LN w.r.t their input, which can be simply represented as Jacobian matrices, $J_{\text{LN}_2}^{z_{j}}, J_{\text{LN}_1}^{z_{j}'}$, $J_{\text{LN}_2}^{z_{i}'}$, and $J_{\text{LN}_1}^{z_{i}}$, respectively. Likewise, $\frac{\partial \text{FFN}(x_{j}')}{\partial x_{j}'}$ and $\frac{\partial \text{MHSA}(x_{j})}{\partial x_{j}}$ are also derivatives of the output of FFN and MHSA w.r.t their inputs, and can be represented as Jacobian matrices, $J_{\text{FFN}}^{x_{j}'}$ and $J_{\text{MHSA}}^{x_{j}}$, respectively. After substituting these terms, we obtain the following:

\begin{equation}
\begin{aligned}
g_{z_{i}} = \frac{\partial \mathcal{L}}{\partial z_{i}} = P_1 \cdot \prod_{j=i+1}^{N} \left( J_{\text{LN}_2}^{z_{j}'} \cdot (\text{I} + J_{\text{FFN}}^{x_{j}'}) \cdot J_{\text{LN}_1}^{z_{j}} \cdot (\text{I} + J_{\text{MHSA}}^{x_{j}}) \right) \cdot J_{\text{LN}_2}^{z_{i}'} \cdot (\text{I} + J_{\text{FFN}}^{x_{i}'}) \cdot J_{\text{LN}_1}^{z_{i}}
\end{aligned}
\label{eqn:post_ln_z1_grad_jacobian}
\end{equation}

Now, we take the L2-norm on both sides of Eq.~\eqref{eqn:post_ln_z1_grad_jacobian} as follows:
\begin{equation}
\begin{aligned}
\| g_{z_{i}} \|_2 &= \left\| \frac{\partial \mathcal{L}}{\partial z_{i}} \right\|_2
= \left\| P_1 \cdot \prod_{j=i+1}^{N} \left( J_{\text{LN}_2}^{z_{j}'} \cdot (\text{I} + J_{\text{FFN}}^{x_{j}'}) \cdot J_{\text{LN}_1}^{z_{j}} \cdot (\text{I} + J_{\text{MHSA}}^{x_{j}}) \right) \cdot J_{\text{LN}_2}^{z_{i}'} \cdot (\text{I} + J_{\text{FFN}}^{x_{i}'}) \cdot J_{\text{LN}_1}^{z_{i}} \right\|_2
\end{aligned}
\label{eqn:post_ln_z1_grad_jacobian_l2_norm}
\end{equation}

We know that the L2-norm of a matrix is equivalent to its largest singular value \citep{Horn_Johnson_1991}. Hence, we can further write Eq.~\eqref{eqn:post_ln_z1_grad_jacobian_l2_norm} as follows:
\begin{equation}
\begin{aligned}
\| g_{z_{i}} \|_2 &= \left\| \frac{\partial \mathcal{L}}{\partial z_{i}} \right\|_2
= s_{\text{max}} (P_1 \cdot \prod_{j=i+1}^{N} \left( J_{\text{LN}_2}^{z_{j}'} \cdot (\text{I} + J_{\text{FFN}}^{x_{j}'}) \cdot J_{\text{LN}_1}^{z_{j}} \cdot (\text{I} + J_{\text{MHSA}}^{x_{j}}) \right) \cdot J_{\text{LN}_2}^{z_{i}'} \cdot (\text{I} + J_{\text{FFN}}^{x_{i}'}) \cdot J_{\text{LN}_1}^{z_{i}})
\end{aligned}
\label{eqn:post_ln_z1_grad_jacobian_l2_norm_step2}
\end{equation}
where $s_{\text{max}}$ outputs the largest singular value of $(P_1 \cdot \prod_{j=i+1}^{N} \left( J_{\text{LN}_2}^{z_{j}'} \cdot (\text{I} + J_{\text{FFN}}^{x_{j}'}) \cdot J_{\text{LN}_1}^{z_{j}} \cdot (\text{I} + J_{\text{MHSA}}^{x_{j}}) \right) \cdot J_{\text{LN}_2}^{z_{i}'} \cdot (\text{I} + J_{\text{FFN}}^{x_{i}'}) \cdot J_{\text{LN}_1}^{z_{i}})$. Now from properties of singular values defined in Eq.~\eqref{eqn:singular_vals_properties}, we can further rewrite Eq.~\eqref{eqn:post_ln_z1_grad_jacobian_l2_norm_step2} as follows:

\begin{equation}
\begin{aligned}
\| g_{z_{i}} \|_2 &= \left\| \frac{\partial \mathcal{L}}{\partial z_{i}} \right\|_2 \\
&\leq s_\text{max}(P_1) \cdot  
\prod_{j=i+1}^{N} \Bigg( 
s_{\text{max}}(J_{\text{LN}_2}^{z_{j}'}) \cdot 
\big( s_{\text{max}}(\text{I}) + s_{\text{max}}(J_{\text{FFN}}^{x_{j}'}) \big) \cdot s_{\text{max}}(J_{\text{LN}_1}^{z_{j}}) \cdot 
\big( s_{\text{max}}(\text{I}) + s_{\text{max}}(J_{\text{MHSA}}^{x_{j}}) \big) 
\Bigg) \\
&\hspace{3cm}  \cdot s_{\text{max}}(J_{\text{LN}_2}^{z_{i}'}) \cdot \big( s_{\text{max}}(\text{I}) + s_{\text{max}}(J_{\text{FFN}}^{x_{i}'}) \big) \cdot s_{\text{max}}(J_{\text{LN}_1}^{z_{i}})
\end{aligned}
\end{equation}

From \cite{xiong2020layer}, we can rewrite the Jacobian of LNs as follows:
\begin{equation}
    \begin{aligned}
        J_{\text{LN}_1}^{z_{j}} = \frac{\text{I}}{\sigma_{z_{j}}}, \quad 
        J_{\text{LN}_2}^{z_{j}'} = \frac{\text{I}}{\sigma_{z_{j}'}}, \quad
        J_{\text{LN}_2}^{z_{i}'} = \frac{\text{I}}{\sigma_{z_{i}'}},
        \quad and \quad 
        J_{\text{LN}_1}^{z_{i}} = \frac{\text{I}}{\sigma_{z_{i}}}
    \end{aligned}
\end{equation}
where $\sigma_{z_{j}}$, $\sigma_{z_{j}'}$, $\sigma_{z_{i}'}$ and $\sigma_{z_{i}}$ are the standard-deviations of $z_{j}$, $z_{j}'$, $z_{i}'$, and $z_{i}$ respectively. Therefore, we obtain the following equation:

\begin{equation}
\begin{aligned}
\| g_{z_{i}} \|_2 &= \left\| \frac{\partial \mathcal{L}}{\partial z_{i}} \right\|_2 \\
&\leq s_\text{max}(P_1) \cdot  
\prod_{j=i+1}^{N} \Bigg( 
s_{\text{max}}\left(\frac{\text{I}}{\sigma_{z_{j}'}}\right) \cdot 
\big( s_{\text{max}}(\text{I}) + s_{\text{max}}(J_{\text{FFN}}^{x_{j}'}) \big) \cdot s_{\text{max}}\left(\frac{\text{I}}{\sigma_{z_{j}}}\right) \cdot 
\big( s_{\text{max}}(\text{I}) + s_{\text{max}}(J_{\text{MHSA}}^{x_{j}}) \big) 
\Bigg) \\
&\hspace{3cm} \cdot s_{\text{max}}\left(\frac{I}{\sigma_{z_{i}'}}\right) \cdot \big( s_{\text{max}}(\text{I}) + s_{\text{max}}(J_{\text{FFN}}^{x_{j}'}) \big) \cdot s_{\text{max}}\left(\frac{\text{I}}{\sigma_{z_{i}}}\right)
\end{aligned}
\label{eqn:post_ln_z1_grad_jacobian_l2_norm_step3}
\end{equation}

Another property of singular values states that all singular values of identity matrix $\text{I}$ are 1 \citep{Horn_Johnson_1991}, i.e., $s_k(\text{I}) = 1$. Therefore substituting with that in Eq.~\eqref{eqn:post_ln_z1_grad_jacobian_l2_norm_step3}, we obtain the following:

\begin{equation}
\begin{aligned}
\| g_{z_{i}} \|_2 &= \left\| \frac{\partial \mathcal{L}}{\partial z_{i}} \right\|_2 \\
&\leq s_\text{max}(P_1) \cdot  
\prod_{j=i+1}^{N} \Bigg( 
s_{\text{max}}\left(\frac{1}{\sigma_{z_{j}'}}\right) \cdot 
\big( 1 + s_{\text{max}}(J_{\text{FFN}}^{x_{j}'}) \big) \cdot  s_{\text{max}}\left(\frac{1}{\sigma_{z_{j}}}\right) \cdot 
\big( 1 + s_{\text{max}}(J_{\text{MHSA}}^{x_{j}}) \big) 
\Bigg) \\
&\hspace{3cm} \cdot s_{\text{max}}\left(\frac{1}{\sigma_{z_{i}'}}\right) \cdot \big( 1 + s_{\text{max}}(J_{\text{FFN}}^{x_{j}'}) \big) \cdot s_{\text{max}}\left(\frac{1}{\sigma_{z_{i}}}\right)
\end{aligned}
\label{eqn:post_ln_z1_grad_jacobian_l2_norm_step4}
\end{equation}

By re-arranging the terms, we finally obtain the following equation:
\begin{equation}
\begin{aligned}
    \| g_{z_{i}} \|_2 &= \left\| \frac{\partial \mathcal{L}}{\partial z_{i}} \right\|_2 \leq s_\text{max}(P_1) \cdot \left( \frac{1}{\prod_{j=i}^{N} \sigma_{z_{j}'} \sigma_{z_j}} \right) \cdot \prod_{j=i}^{N} \left( 1 + s_{\text{max}}(J_{\text{FFN}}^{x_{j}'}) \right) \cdot \prod_{j=i+1}^{N}\left( 1 + s_{\text{max}}(J_{\text{MHSA}}^{x_{j}}) \right)
\end{aligned}
\label{eqn:post_ln_z1_grad_jacobian_l2_norm_step4_rearranged}
\end{equation}

Now, based on Eq.~\eqref{eqn:var_lower_bound_post_rewrite}, we can re-write Eq.~\eqref{eqn:post_ln_z1_grad_jacobian_l2_norm_step4_rearranged} as follows:

\begin{equation}
\begin{aligned}
    \| g_{z_{i}} \|_2 = \left\| \frac{\partial \mathcal{L}}{\partial z_{i}} \right\|_2 \leq & s_\text{max}(P_1) \cdot  \left( \frac{1}{\prod_{j=i}^{N} \left|1 - \sqrt{\text{Var}(\text{FFN}(x_{j}')})\right| \left|1 - \sqrt{\text{Var}(\text{MHSA}(x_{j})})\right|} \right) \cdot \\ & \cdot \prod_{j=i}^{N} \left( 1 + s_{\text{max}}(J_{\text{FFN}}^{x_{j}'}) \right) \cdot \prod_{j=i+1}^{N} \left( 1 + s_{\text{max}}(J_{\text{MHSA}}^{x_{j}}) \right)
\end{aligned}
\label{eqn:post_ln_z1_grad_jacobian_l2_norm_step4_rearranged_var}
\end{equation}

\subsection{For Pre-LN model:}

The Pre-LN model setup for \(i^{th}\) layer can be represented as follows:
\begin{equation}
\begin{aligned}
    x_i' &= x_i + \text{MHSA}(\text{LN}_1(x_i)) \\
    y_i &= x_{i}' + \text{FFN}(\text{LN}_2(x_{i}'))
\end{aligned}
\label{eqn:pre_ln_setup}
\end{equation}
where $x_i$ and $x_{i}'$ are the inputs to $\text{LN}_1$ and $\text{LN}_2$, respectively.

Since there are two LayerNorm (LN) operations in every layer, we separately prove for both of them

\subsubsection{Backpropagation analysis for $\text{LN}_2$ $(g_{x_{i}'})$:} We can write $g_{x_{i}'}$ for the \(i^{th}\) layer as follows:

\begin{equation}
\begin{aligned}
g_{x_{i}'} = \frac{\partial \mathcal{L}}{\partial x_{i}'} = \frac{\partial \mathcal{L}}{\partial y_{\text{out}}} \cdot \frac{\partial y_{\text{out}}}{\partial y_N} \cdot \prod_{j=i+1}^{N} \left( \frac{\partial y_j}{\partial x_{j}'} \cdot \frac{\partial x_j'}{\partial x_{j}}\right) \cdot \frac{\partial y_i}{\partial x_{i}'}
\end{aligned}
\label{eqn:pre_xl'_grad}
\end{equation}

Here, $\frac{\partial \mathcal{L}}{\partial y_{\text{out}}} \cdot \frac{\partial y_{\text{out}}}{\partial y_N}$, is independent of the transformers layers as they are computed using the classification head's output. Therefore we can treat them as \(P_2\) (which does not vary across layers). Furthermore, we expand $y_{j} = x_{j}' + \text{FFN}(\text{LN}_1(x_{j}'))$, $x_{j}' = x_{j} + \text{MHSA}(\text{LN}_2(x_{j}))$, using Eq.~\eqref{eqn:pre_ln_setup}, and compute their corresponding derivatives in Eq.~\eqref{eqn:pre_xl'_grad}. Lastly, $x_{i+1}$ is same as $y_i$, because $y_i$ is the output of the $i^\text{th}$ layer which becomes input $x_{i+1}$ of the $(i+1)^\text{th}$ layer.

After substituting, we get the following equation:

\begin{equation}
\begin{aligned}
g_{x_{i}'} = \frac{\partial \mathcal{L}}{\partial x_{i}'} 
=\, &P_2 \cdot 
\prod_{j=i+1}^{N} 
\left( 
\frac{\partial (x_{j}' + \text{FFN}(\text{LN}_2(x_{j}')))}{\partial x_{j}'} 
\cdot 
\frac{\partial (x_{j} + \text{MHSA}(\text{LN}_1(x_{j})))}{\partial x_{j}} 
\right) 
\cdot 
\frac{\partial (x_{i}' + \text{FFN}(\text{LN}_2(x_{i}')))}{\partial x_{i}'}
\end{aligned}
\end{equation}

\begin{equation}
\begin{aligned}
=\, &P_2 \cdot 
\prod_{j=i+1}^{N} 
\left( 
(\text{I} + \frac{\partial \text{FFN}(\text{LN}_2(x_{j}'))}{\partial x_{j}'}) 
\cdot 
(\text{I} + \frac{\partial \text{MHSA}(\text{LN}_1(x_{j}))}{\partial x_{j}}) 
\right) 
\cdot 
(\text{I} + \frac{\partial \text{FFN}(\text{LN}_2(x_{i}'))}{\partial x_{i}'})
\end{aligned}
\end{equation}

\begin{equation}
\begin{aligned}
=\, &P_2 \cdot 
\prod_{j=i+1}^{N} 
\Big( 
(\text{I} + \frac{\partial \text{FFN}(\text{LN}_2(x_{j}'))}{\partial \text{LN}_2(x_{j}')} 
\cdot \frac{\partial \text{LN}_2(x_{j}')}{\partial x_{j}'}) 
\cdot 
(\text{I} + \frac{\partial \text{MHSA}(\text{LN}_1(x_{j}))}{\partial \text{LN}_1(x_{j})} 
\cdot \frac{\partial \text{LN}_1(x_{j})}{\partial x_{j}}) 
\Big) \\
&\cdot 
(\text{I} + \frac{\partial \text{FFN}(\text{LN}_2(x_{i}'))}{\partial \text{LN}_2(x_{i}')} 
\cdot \frac{\partial \text{LN}_2(x_{i}')}{\partial x_{i}'})
\end{aligned}
\label{eqn:pre_xl'_step1}
\end{equation}

Here, in Eq.~\eqref{eqn:pre_xl'_step1}, $\frac{\partial \text{LN}_1(x_{j})}{\partial x_{j}}, \frac{\partial \text{LN}_2(x_{j}')}{\partial x_{j}'}$, are both derivative of output of LN w.r.t their inputs, and hence can be represented as Jacobian matrices, $J_{\text{LN}_1}^{x_{j}}$ and $J_{\text{LN}_2}^{x_{j}'}$ respectively. Similarly, $\frac{ \partial \text{MHSA}(\text{LN}_1(x_{j}))}{\partial \text{LN}_1(x_{j})}$ and $\frac{ \partial \text{FFN}(\text{LN}_2(x_{j}'))}{\partial \text{LN}_2(x_{j}')}$, are derivatives of output of MHSA/FFN w.r.t their inputs, and can also be represented as Jacobian matrices, $J_{\text{MHSA}}^{\text{LN}_1(x_{j})}$ and $J_{\text{FFN}}^{\text{LN}_2(x_{j}')}$ respectively.

Using these relations, we can re-write Eq.~\eqref{eqn:pre_xl'_step1} as follows:

\begin{equation}
\begin{aligned}
g_{x_{i}'} = \frac{\partial \mathcal{L}}{\partial x_{i}'} = P_2 \cdot \prod_{j=i+1}^{N} \left( (\text{I} + J_{\text{FFN}}^{\text{LN}_2(x_{j}')} \cdot J_{\text{LN}_2}^{x_{j}'}) \cdot (\text{I} + J_{\text{MHSA}}^{\text{LN}_1(x_{j})} \cdot J_{\text{LN}_1}^{x_{j}})\right) \cdot (\text{I} + J_{\text{FFN}}^{\text{LN}_2(x_{i}')} \cdot J_{\text{LN}_2}^{x_{i}'})
\end{aligned}
\label{eqn:pre_xl'_step2}
\end{equation}

We can further re-arrange the terms in Eq.~\eqref{eqn:pre_xl'_step2} as follows:

\begin{equation}
\begin{aligned}
g_{x_{i}'} = \frac{\partial \mathcal{L}}{\partial x_{i}'} = P_2 \cdot \prod_{j=i}^{N} \left( \text{I} + J_{\text{FFN}}^{\text{LN}_2(x_{j}')} \cdot J_{\text{LN}_2}^{x_{j}'}\right) \cdot \prod_{j=i+1}^{N} \left(\text{I} + J_{\text{MHSA}}^{\text{LN}_1(x_{j})} \cdot J_{\text{LN}_1}^{x_{j}}\right)
\end{aligned}
\label{eqn:pre_xl'_step2_rearranged}
\end{equation}

Now, we take the L2-norm at both sides of Eq.~\eqref{eqn:pre_xl'_step2_rearranged} and since we know that L2-norm of a matrix is equivalent to its maximum singular value. Therefore, we get the following equation:

\begin{equation}
\begin{aligned}
\|g_{x_{i}'}\|_2 = \left\| \frac{\partial \mathcal{L}}{\partial x_{i}'} \right\|_2 = s_\text{max}(P_2 \cdot \prod_{j=i}^{N} \left( \text{I} + J_{\text{FFN}}^{\text{LN}_2(x_{j}')} \cdot J_{\text{LN}_2}^{x_{j}'}\right) \cdot \prod_{j=i+1}^{N} \left(\text{I} + J_{\text{MHSA}}^{\text{LN}_1(x_{j})} \cdot J_{\text{LN}_1}^{x_{j}}\right))
\end{aligned}
\label{eqn:pre_xl'_step3_l2_norm}
\end{equation}

where $s_\text{max}$ is the maximum singular value of $(P_2 \cdot \prod_{j=l}^{N} \left( \text{I} + J_{\text{FFN}}^{\text{LN}_2(x_{j}')} \cdot J_{\text{LN}_2}^{x_{j}'}\right) \cdot \prod_{j=l+1}^{N} \left(\text{I} + J_{\text{MHSA}}^{\text{LN}_1(x_{j})} \cdot J_{\text{LN}_1}^{x_{j}}\right))$.

From the singular values properties, discussed in Eq. \eqref{eqn:singular_vals_properties}, we can write the upper bound of $\|g_{x_{i}'}\|_2$ as follows:

\begin{equation}
\begin{aligned}
\|g_{x_{i}'}\|_2 = \left\| \frac{\partial \mathcal{L}}{\partial x_{i}'} \right\|_2 \leq s_\text{max}(P_2) \cdot \prod_{j=i}^{N} \left( s_\text{max}(\text{I}) + s_\text{max}(J_{\text{FFN}}^{\text{LN}_2(x_{j}')} \cdot J_{\text{LN}_2}^{x_{j}'})\right) \cdot \prod_{j=i+1}^{N} \left(s_\text{max}(\text{I}) + s_\text{max}(J_{\text{MHSA}}^{\text{LN}_1(x_{j})} \cdot J_{\text{LN}_1}^{x_{j}})\right)
\end{aligned}
\label{eqn:pre_xl'_step4_l2_norm_upper_bound}
\end{equation}

We know that all singular values of an Idenitiy matrix $\text{I}$ are 1, i.e., $s_k(\text{I}) = 1$. Thus,

\begin{equation}
\begin{aligned}
\|g_{x_{i}'}\|_2 = \left\| \frac{\partial \mathcal{L}}{\partial x_{i}'} \right\|_2 \leq s_\text{max}(P_2) \cdot \prod_{j=i}^{N} \left( 1 + s_\text{max}(J_{\text{FFN}}^{\text{LN}_2(x_{j}')} \cdot J_{\text{LN}_2}^{x_{j}'})\right) \cdot \prod_{j=i+1}^{N} \left(1 + s_\text{max}(J_{\text{MHSA}}^{\text{LN}_1(x_{j})} \cdot J_{\text{LN}_1}^{x_{j}})\right)
\end{aligned}
\label{eqn:pre_xl'_step5_l2_norm_upper_bound}
\end{equation}

\subsubsection{Backpropagation analysis for $\text{LN}_1$ $(g_{x_{i}})$:} We can write $g_{x_{i}}$ for the \(i^{th}\) layer as follows:

\begin{equation}
\begin{aligned}
g_{x_{i}} = \frac{\partial \mathcal{L}}{\partial x_{i}} = \frac{\partial \mathcal{L}}{\partial y_{\text{out}}} \cdot \frac{\partial y_{\text{out}}}{\partial y_N} \cdot \prod_{j=i+1}^{N} \left( \frac{\partial y_j}{\partial x_{j'}} \cdot \frac{\partial x_j'}{\partial x_{j}}\right) \cdot \frac{\partial y_i}{\partial x_{i'}} \cdot \frac{\partial x_{i'}}{\partial x_{i}}
\end{aligned}
\label{eqn:pre_xl_grad}
\end{equation}

Here, $\frac{\partial \mathcal{L}}{\partial y_{\text{out}}} \cdot \frac{\partial y_{\text{out}}}{\partial y_N}$, is independent of the transformers layers as they are computed using the classification head's output. Therefore we can treat them as \(P_2\) (which does not vary across layers). Furthermore, we expand $y_{j} = x_{j}' + \text{FFN}(\text{LN}_1(x_{j}'))$, $x_{j}' = x_{j} + \text{MHSA}(\text{LN}_2(x_{j}))$, using Eq.~\eqref{eqn:pre_ln_setup}, and compute their corresponding derivatives in Eq.~\eqref{eqn:pre_xl_grad}. Lastly, $x_{i+1}$ is same as $y_i$, because $y_i$ is the output of the $i^\text{th}$ layer which becomes input $x_{i+1}$ of the $(i+1)^\text{th}$ layer.

After substituting, we get the following equation:

\begin{equation}
\begin{split}
g_{x_{i}} = \frac{\partial \mathcal{L}}{\partial x_{i}} 
= P_2 \cdot \prod_{j=i+1}^{N} & 
\left( 
\frac{\partial (x_{j}' + \text{FFN}(\text{LN}_2(x_{j}')))}{\partial x_{j}'} 
\cdot 
\frac{\partial (x_{j} + \text{MHSA}(\text{LN}_1(x_{j})))}{\partial x_{j}} 
\right) \\
& \cdot 
\frac{\partial (x_{i}' + \text{FFN}(\text{LN}_2(x_{i}')))}{\partial x_{i}'} 
\cdot 
\frac{\partial (x_{i} + \text{MHSA}(\text{LN}_1(x_{i})))}{\partial x_{i}}
\end{split}
\end{equation}

\begin{equation}
\begin{split}
\quad \quad \quad \quad = P_2 \cdot \prod_{j=i+1}^{N} & 
\left( 
\text{I} + \frac{\partial \text{FFN}(\text{LN}_2(x_{j}'))}{\partial x_{j}'} 
\right) \cdot 
\left( 
\text{I} + \frac{\partial \text{MHSA}(\text{LN}_1(x_{j}))}{\partial x_{j}} 
\right) \\
& \cdot 
\left( 
\text{I} + \frac{\partial \text{FFN}(\text{LN}_2(x_{i}'))}{\partial x_{i}'} 
\right) 
\cdot 
\left( 
\text{I} + \frac{\partial \text{MHSA}(\text{LN}_1(x_{i}))}{\partial x_{i}} 
\right)
\end{split}
\end{equation}
\begin{equation}
\begin{aligned}
 \quad \quad \quad \quad \quad \quad = &P_2 \cdot \prod_{j=l+1}^{N} \left( (\text{I} + \frac{ \partial \text{FFN}(\text{LN}_2(x_{j}'))}{\partial \text{LN}_2(x_{j}')} \frac{\partial \text{LN}_2(x_{j}')}{\partial x_{j}'}) \cdot (\text{I} + \frac{ \partial \text{MHSA}(\text{LN}_1(x_{j}))}{\partial \text{LN}_1(x_{j})} \frac{\partial \text{LN}_1(x_{j})}{\partial x_{j}})\right) \\ &\cdot (\text{I} + \frac{ \partial \text{FFN}(\text{LN}_2(x_{i}'))}{\partial \text{LN}_2(x_{i}')} \frac{\partial \text{LN}_2(x_{i}')}{\partial x_{i}'}) \cdot (\text{I} + \frac{ \partial \text{MHSA}(\text{LN}_1(x_{i}))}{\partial \text{LN}_1(x_{i})} \frac{\partial \text{LN}_1(x_{i})}{\partial x_{i}})
\end{aligned}
\label{eqn:pre_xl_step1}
\end{equation}

Here, in Eq.~\eqref{eqn:pre_xl_step1}, $\frac{\partial \text{LN}_1(x_{j})}{\partial x_{j}}, \frac{\partial \text{LN}_2(x_{j}')}{\partial x_{j}'}$, are both derivative of output of LN w.r.t their inputs, and hence can be represented as Jacobian matrices, $J_{\text{LN}_1}^{x_{j}}$ and $J_{\text{LN}_2}^{x_{j}'}$ respectively. Similarly, $\frac{ \partial \text{MHSA}(\text{LN}_1(x_{j}))}{\partial \text{LN}_1(x_{j})}$ and $\frac{ \partial \text{FFN}(\text{LN}_2(x_{j}'))}{\partial \text{LN}_2(x_{j}')}$, are derivatives of output of MHSA/FFN w.r.t their inputs, and can also be represented as Jacobian matrices, $J_{\text{MHSA}}^{\text{LN}_1(x_{j})}$ and $J_{\text{FFN}}^{\text{LN}_2(x_{j}')}$ respectively.

Using these relations, we can re-write Eq.~\eqref{eqn:pre_xl_step1}, as follows

\begin{equation}
\begin{aligned}
g_{x_{i}} = \frac{\partial \mathcal{L}}{\partial x_{i}} 
=\, &P_2 \cdot 
\prod_{j=i+1}^{N} \Bigg( 
\left( \text{I} + J_{\text{FFN}}^{\text{LN}_2(x_{j}')} \cdot J_{\text{LN}_2}^{x_{j}'} \right) 
\cdot 
\left( \text{I} + J_{\text{MHSA}}^{\text{LN}_1(x_{j})} \cdot J_{\text{LN}_1}^{x_{j}} \right) 
\Bigg) \\
&\cdot 
\left( \text{I} + J_{\text{FFN}}^{\text{LN}_2(x_{i}')} \cdot J_{\text{LN}_2}^{x_{i}'} \right) 
\cdot 
\left( \text{I} + J_{\text{MHSA}}^{\text{LN}_1(x_{i})} \cdot J_{\text{LN}_1}^{x_{i}} \right)
\end{aligned}
\label{eqn:pre_xl_step2}
\end{equation}

We can further re-arrange the terms in Eq.~\eqref{eqn:pre_xl_step2} as follows:

\begin{equation}
\begin{aligned}
g_{x_{i}} = \frac{\partial \mathcal{L}}{\partial x_{i}} = P_2 \cdot \prod_{j=i}^{N} \left( \text{I} + J_{\text{FFN}}^{\text{LN}_2(x_{j}')} \cdot J_{\text{LN}_2}^{x_{j}'}\right) \cdot \prod_{j=i}^{N} \left(\text{I} + J_{\text{MHSA}}^{\text{LN}_1(x_{j})} \cdot J_{\text{LN}_1}^{x_{j}}\right)
\end{aligned}
\label{eqn:pre_xl_step2_rearranged}
\end{equation}

Now, we take the L2-norm at both sides of Eq.~\eqref{eqn:pre_xl_step2_rearranged} and since we know that L2-norm of a matrix is equivalent to its maximum singular value. Therefore, we get the following equation:

\begin{equation}
\begin{aligned}
\|g_{x_{i}}\|_2 = \left\| \frac{\partial \mathcal{L}}{\partial x_{i}} \right\|_2 = s_\text{max}(P_2 \cdot \prod_{j=i}^{N} \left( \text{I} + J_{\text{FFN}}^{\text{LN}_2(x_{j}')} \cdot J_{\text{LN}_2}^{x_{j}'}\right) \cdot \prod_{j=i}^{N} \left(\text{I} + J_{\text{MHSA}}^{\text{LN}_1(x_{j})} \cdot J_{\text{LN}_1}^{x_{j}}\right))
\end{aligned}
\label{eqn:pre_xl_step3_l2_norm}
\end{equation}

where $s_\text{max}$ is the maximum singular value of $(P_2 \cdot \prod_{j=l}^{N} \left( \text{I} + J_{\text{FFN}}^{\text{LN}_2(x_{j}')} \cdot J_{\text{LN}_2}^{x_{j}'}\right) \cdot \prod_{j=l+1}^{N} \left(\text{I} + J_{\text{MHSA}}^{\text{LN}_1(x_{j})} \cdot J_{\text{LN}_1}^{x_{j}}\right))$.

From the singular values properties, discussed in Eq. \eqref{eqn:singular_vals_properties}, we can write the upper bound of $\|g_{x_{i}'}\|_2$ as follows:

\begin{equation}
\begin{aligned}
\|g_{x_{i}}\|_2 = \left\| \frac{\partial \mathcal{L}}{\partial x_{i}} \right\|_2 \leq s_\text{max}(P_2) \cdot \prod_{j=i}^{N} \left( s_\text{max}(\text{I}) + s_\text{max}(J_{\text{FFN}}^{\text{LN}_2(x_{j}')} \cdot J_{\text{LN}_2}^{x_{j}'})\right) \cdot \prod_{j=i}^{N} \left(s_\text{max}(\text{I}) + s_\text{max}(J_{\text{MHSA}}^{\text{LN}_1(x_{j})} \cdot J_{\text{LN}_1}^{x_{j}})\right)
\end{aligned}
\label{eqn:pre_xl_step4_l2_norm_upper_bound}
\end{equation}

We know that all singular values of an Idenitiy matrix $\text{I}$ are 1, i.e., $s_k(\text{I}) = 1$. Thus,

\begin{equation}
\begin{aligned}
\|g_{x_{i}}\|_2 = \left\| \frac{\partial \mathcal{L}}{\partial x_{i}} \right\|_2 \leq s_\text{max}(P_2) \cdot \prod_{j=i}^{N} \left( 1 + s_\text{max}(J_{\text{FFN}}^{\text{LN}_2(x_{j}')} \cdot J_{\text{LN}_2}^{x_{j}'})\right) \cdot \prod_{j=i}^{N} \left(1 + s_\text{max}(J_{\text{MHSA}}^{\text{LN}_1(x_{j})} \cdot J_{\text{LN}_1}^{x_{j}})\right)
\end{aligned}
\label{eqn:pre_xl_step5_l2_norm_upper_bound}
\end{equation}

\hfill \qedsymbol{}

\begin{mdframed}[
    linewidth=0.5pt,
    roundcorner=4pt,
    backgroundcolor=gray!8,
    linecolor=gray!50,
    innertopmargin=2pt,
    innerbottommargin=2pt,
    innerleftmargin=6pt,
    innerrightmargin=6pt,
    skipabove=6pt,
    skipbelow=6pt
]
\section{Theorem 3: Upper bound of the gradient norm of Early Layers LN are higher than those of Later Layers LN.} \label{theorem:l2_norm_bound_early_layers_app}
\emph{It is formally represented as follows:}
\begin{equation} 
\text{UB}(\|g_{x_{1}}\|_2) \geq \text{UB}(\|g_{x_{2}}\|_2) \geq \dots \geq \text{UB}(\|g_{x_{N}}\|_2) \; ; \; \text{for both Pre- and Post-LN models}
\label{eqn:early_layers_grads_higher}
\end{equation}
\emph{where $\text{UB}(\|g_{x_{i}}\|_2)$ denotes the upper bound of $\| g_{x_{i}} \|_2$ and $x_i$ is the input to the $i^\text{th}$ layer's LN.}
\end{mdframed}

\textbf{\emph{Proof:}}

\subsection{For Post-LN model:}

\subsubsection{Analysis for $\text{LN}_2$ $(g_{z_{i}'})$:} We prove that $\text{UB}(\| g_{z_{i}'} \|_2) \geq \text{UB}(\| g_{z_{i+1}'} \|_2)$, corresponding to $i^\text{th}$ and $(i+1)^\text{th}$ layer. 

We can compute $\text{UB}(\| g_{z_{i}'} \|_2)$ and $\text{UB}(\| g_{z_{i+1}'} \|_2)$ using Eq.~\eqref{eqn:post_ln_z2_grad_jacobian_l2_norm_step4_rearranged_var} as follows:

\begin{equation}
\begin{aligned}
    \text{UB}(\| g_{z_{i}'} \|_2) = & s_\text{max}(P_1) \cdot  \left( \frac{1}{\prod_{j=i}^{N} \left|1 - \sqrt{\text{Var}(\text{FFN}(x_{j}')}) \right|} \right) \cdot \left( \frac{1}{\prod_{j=i+1}^{N} \left| 1 - \sqrt{\text{Var}(\text{MHSA}(x_{j})}) \right|} \right) \\ & \cdot \prod_{j=i+1}^{N} \left( \left( 1 + s_{\text{max}}(J_{\text{FFN}}^{x_{j}'}) \right) \cdot \left( 1 + s_{\text{max}}(J_{\text{MHSA}}^{x_{j}}) \right) \right)
\end{aligned}
\end{equation}

\begin{equation}
\begin{aligned}
    \text{UB}(\| g_{z_{i+1}'} \|_2)= & s_\text{max}(P_1) \cdot  \left( \frac{1}{\prod_{j=i+1}^{N} \left|1 - \sqrt{\text{Var}(\text{FFN}(x_{j}'))} \right|} \right) \cdot \left( \frac{1}{\prod_{j=i+2}^{N} \left| 1 - \sqrt{\text{Var}(\text{MHSA}(x_{j}))} \right|} \right) \\
    & \cdot \prod_{j=i+2}^{N} \left( \left( 1 + s_{\text{max}}(J_{\text{FFN}}^{x_{j}'}) \right) \cdot \left( 1 + s_{\text{max}}(J_{\text{MHSA}}^{x_{j}}) \right) \right)
\end{aligned}
\end{equation}

We then substitute these expressions in the inequality $\text{UB}(\| g_{z_{i}'} \|_2) \geq \text{UB}(\| g_{z_{i+1}'} \|_2)$, to prove that early layers have higher gradient norms in comparison to later layers.

\begin{equation}
\begin{split}
    s_\text{max}(P_1) &\cdot  \left( \frac{1}{\prod_{j=i}^{N} \left|1 - \sqrt{\text{Var}(\text{FFN}(x_{j}'))} \right|} \right) \cdot \left( \frac{1}{\prod_{j=i+1}^{N} \left| 1 - \sqrt{\text{Var}(\text{MHSA}(x_{j}))} \right|} \right) \\
    &\cdot \prod_{j=i+1}^{N} \left( \left( 1 + s_{\text{max}}(J_{\text{FFN}}^{x_{j}'}) \right) \cdot \left( 1 + s_{\text{max}}(J_{\text{MHSA}}^{x_{j}}) \right) \right) \\
    \geq \; &s_\text{max}(P_1) \cdot  \left( \frac{1}{\prod_{j=i+1}^{N} \left|1 - \sqrt{\text{Var}(\text{FFN}(x_{j}'))} \right|} \right) \cdot \left( \frac{1}{\prod_{j=i+2}^{N} \left| 1 - \sqrt{\text{Var}(\text{MHSA}(x_{j}))} \right|} \right) \\
    &\cdot \prod_{j=i+2}^{N} \left( \left( 1 + s_{\text{max}}(J_{\text{FFN}}^{x_{j}'}) \right) \cdot \left( 1 + s_{\text{max}}(J_{\text{MHSA}}^{x_{j}}) \right) \right)
\end{split}
\label{eqn:post_ln_z2_grad_jacobian_l2_norm_i_i+1}
\end{equation}

This can be further rewritten as follows:

\begin{equation}
        \frac{(1+s_\text{max}(J_\text{FFN}^{x_{i+1}'}))(1+s_\text{max}(J_\text{MHSA}^{x_{i+1}}))}{\left|1 - \sqrt{\text{Var}(\text{FFN}(x_{i}'))} \right| \left|1 - \sqrt{\text{Var}(\text{MHSA}(x_{i+1}))} \right|}  \geq 1
\label{eqn:post_ln_z2_grad_jacobian_l2_norm_i_i+1_rearranged}
\end{equation}

From \citet{Horn_Johnson_1991}, we know that every singular value of a matrix A is greater than or equal to 0, i.e., $s_k(A) \geq 0, \forall k$. Hence, for every transformer layer,
\begin{equation}
    \begin{aligned}
        (1 + s_{\text{max}}(J_{\text{MHSA}}^{x_{j}})) \geq 1 \quad \mbox{and} \quad (1 + s_{\text{max}}(J_{\text{FFN}}^{x_{j}'})) \geq 1
    \end{aligned}
    \label{eqn:post_ln_mhsa_ffn_bounds}
\end{equation}

Now, to prove Eq.~\eqref{eqn:post_ln_z2_grad_jacobian_l2_norm_i_i+1_rearranged} to be true, we need to prove that 

\begin{equation}
    0<\left|1 - \sqrt{\text{Var}(\text{FFN}(x_{i}'))} \right| \leq 1 \quad \& \quad 0<\left|1 - \sqrt{\text{Var}(\text{MHSA}(x_{i+1}))} \right| \leq 1
    \label{eqn:var_mhsa_ffn_bound}
\end{equation}

We do not consider the  scenario where either $\left|1 - \sqrt{\text{Var}(\text{FFN}(x_{i}'))} \right|=0$ or $\left|1 - \sqrt{\text{Var}(\text{MHSA}(x_{i+1}))} \right|=0$, because if either/both of them becomes 0 then the gradient norm would go infinity, which we do not observe in real-world models either.

We prove Eq.~\eqref{eqn:var_mhsa_ffn_bound} for the FFN component as follows (MHSA component will also have a similar proof):

Firstly, $\left|1 - \sqrt{\text{Var}(\text{FFN}(x_{i}'))} \right|$ can be rewritten as follows:

\begin{equation}
\left|1 - \sqrt{\text{Var}(\text{FFN}(x_i'))} \right| =
    \begin{cases}
    1 - \sqrt{\text{Var}(\text{FFN}(x_i'))}, & \text{if } \sqrt{\text{Var}(\text{FFN}(x_i'))} < 1 \\
    \sqrt{\text{Var}(\text{FFN}(x_i'))} - 1, & \text{if } \sqrt{\text{Var}(\text{FFN}(x_i'))} > 1
    \end{cases}
    \label{eqn:ffn_use_cases}
\end{equation}

We then apply the inequality described in Eq.~\eqref{eqn:var_mhsa_ffn_bound} and Eq.~\eqref{eqn:ffn_use_cases} together as follows:

\begin{equation}
0<\left|1 - \sqrt{\text{Var}(\text{FFN}(x_i'))} \right| \leq 1 \Rightarrow
    \begin{cases}
    0 < 1 - \sqrt{\text{Var}(\text{FFN}(x_i'))} \leq 1, & \text{if } \sqrt{\text{Var}(\text{FFN}(x_i'))} < 1 \\
    0 < \sqrt{\text{Var}(\text{FFN}(x_i'))} - 1 \leq 1, & \text{if } \sqrt{\text{Var}(\text{FFN}(x_i'))} > 1
    \end{cases}
\label{eqn:ffn_use_cases_applied}
\end{equation}

Further solving Eq.~\eqref{eqn:ffn_use_cases_applied}, we obtain the following:

\begin{equation}
0<\left|1 - \sqrt{\text{Var}(\text{FFN}(x_i'))} \right| \leq 1 \Rightarrow
    \begin{cases}
    0 \leq \sqrt{\text{Var}(\text{FFN}(x_i'))} < 1, & \text{if } \sqrt{\text{Var}(\text{FFN}(x_i'))} < 1 \\
    1 < \sqrt{\text{Var}(\text{FFN}(x_i'))} \leq 2, & \text{if } \sqrt{\text{Var}(\text{FFN}(x_i'))} > 1
    \end{cases}
\label{eqn:ffn_use_cases_applied_rearranged}
\end{equation}

Hence, the inequality \( 0 < \left|1 - \sqrt{\text{Var}(\text{FFN}(x_i'))} \right| \leq 1 \) holds true when

\begin{equation}
    \sigma_{\text{FFN}_i} = \sqrt{\text{Var}(\text{FFN}(x_i'))} \in [0, 2] - \{1\}
    \label{eqn:ffn_condition}
\end{equation}

Likewise, the inequality \( 0 < \left|1 - \sqrt{\text{Var}(\text{MHSA}(x_{i+1}))} \right| \leq 1 \) holds true when

\begin{equation}
    \sigma_{\text{MHSA}_{i+1}} = \sqrt{\text{Var}(\text{MHSA}(x_{i+1}))} \in [0, 2] - \{1\}
    \label{eqn:mhsa_condition}
\end{equation}
Hence, from Eq.~\eqref{eqn:ffn_condition} \& Eq.~\eqref{eqn:mhsa_condition}, we prove that $\frac{(1+s_\text{max}(J_\text{FFN}^{x_{i+1}'}))(1+s_\text{max}(J_\text{MHSA}^{x_{i+1}}))}{\left|1 - \sqrt{\text{Var}(\text{FFN}(x_{i}'))} \right| \left|1 - \sqrt{\text{Var}(\text{MHSA}(x_{i+1}))} \right|}  \geq 1$, further proving that $\text{UB}(\| g_{z_{i}'} \|_2) \geq \text{UB}(\| g_{z_{i+1}'} \|_2)$.

Consequently, we prove that \textbf{the upper bound of L2-norm of gradients for Early Layers $\text{LN}_2$ are higher than the one of Later Layers $\text{LN}_2$ in Post-LN models}, formally represented as follows:
\begin{equation} 
\begin{aligned}
\text{UB}(\|g_{z_1'}\|_2) \geq \text{UB}(\|g_{z_2'}\|_2) \geq \dots \geq \text{UB}(\|g_{z_N'}\|_2), 
\end{aligned}
\label{eqn:early_layers_grads_higher_post_ln2}
\end{equation}
when \( 0 < \left|1 - \sqrt{\text{Var}(\text{FFN}(x_i'))} \right| \leq 1 \) and \( 0 < \left|1 - \sqrt{\text{Var}(\text{MHSA}(x_{i+1}))} \right| \leq 1 \).

\subsubsection{Analysis for $\text{LN}_1$ $(g_{z_{i}})$:} We need to prove that $\text{UB}(\| g_{z_{i}} \|_2) \geq \text{UB}(\| g_{z_{i+1}} \|_2)$, corresponding to $i^\text{th}$ and $(i+1)^\text{th}$ layer. 

We can compute $\text{UB}(\| g_{z_{i}} \|_2)$ and $\text{UB}(\| g_{z_{i+1}} \|_2)$ using Eq~\eqref{eqn:post_ln_z1_grad_jacobian_l2_norm_step4_rearranged_var} as follows:

\begin{equation}
\begin{aligned}
    \text{UB}(\| g_{z_{i}} \|_2) = & s_\text{max}(P_1) \cdot  \left( \frac{1}{\prod_{j=i}^{N} \left|1 - \sqrt{\text{Var}(\text{FFN}(x_{j}')})\right| \left|1 - \sqrt{\text{Var}(\text{MHSA}(x_{j})})\right|} \right) \cdot \\ & \cdot \prod_{j=i}^{N} \left( 1 + s_{\text{max}}(J_{\text{FFN}}^{x_{j}'}) \right) \cdot \prod_{j=i+1}^{N} \left( 1 + s_{\text{max}}(J_{\text{MHSA}}^{x_{j}}) \right)
\end{aligned}
\end{equation}

\begin{equation}
\begin{aligned}
    \text{UB}(\| g_{z_{i+1}} \|_2) = & s_\text{max}(P_1) \cdot  \left( \frac{1}{\prod_{j=i+1}^{N} \left|1 - \sqrt{\text{Var}(\text{FFN}(x_{j}')})\right| \left|1 - \sqrt{\text{Var}(\text{MHSA}(x_{j})})\right|} \right) \cdot \\ & \cdot \prod_{j=i+1}^{N} \left( 1 + s_{\text{max}}(J_{\text{FFN}}^{x_{j}'}) \right) \cdot \prod_{j=i+2}^{N} \left( 1 + s_{\text{max}}(J_{\text{MHSA}}^{x_{j}}) \right)
\end{aligned}
\end{equation}

We then substitute these expressions in the inequality $\text{UB}(\| g_{z_{i}} \|_2) \geq \text{UB}(\| g_{z_{i+1}} \|_2)$, to prove that early layers have higher gradients in comparison to later layers.

\begin{equation}
\begin{split}
    s_\text{max}(P_1) \cdot 
    \left( \frac{1}{\prod_{j=i}^{N} 
        \left|1 - \sqrt{\text{Var}(\text{FFN}(x_{j}'))} \right| 
        \cdot \left|1 - \sqrt{\text{Var}(\text{MHSA}(x_{j}))} \right|} 
    \right) \\
    \cdot \prod_{j=i}^{N} \left( 1 + s_{\text{max}}(J_{\text{FFN}}^{x_{j}'}) \right)
    \cdot \prod_{j=i+1}^{N} \left( 1 + s_{\text{max}}(J_{\text{MHSA}}^{x_{j}}) \right) \\
    \geq\;
    s_\text{max}(P_1) \cdot 
    \left( \frac{1}{\prod_{j=i+1}^{N} 
        \left|1 - \sqrt{\text{Var}(\text{FFN}(x_{j}'))} \right| 
        \cdot \left|1 - \sqrt{\text{Var}(\text{MHSA}(x_{j}))} \right|} 
    \right) \\
    \cdot \prod_{j=i+1}^{N} \left( 1 + s_{\text{max}}(J_{\text{FFN}}^{x_{j}'}) \right) \cdot \prod_{j=i+2}^{N} \left( 1 + s_{\text{max}}(J_{\text{MHSA}}^{x_{j}}) \right)
\end{split}
\label{eqn:post_ln_z1_grad_jacobian_l2_norm_i_i+1}
\end{equation}

This can be further rewritten as follows:

\begin{equation}
        \frac{(1+s_\text{max}(J_\text{FFN}^{x_{i}'}))(1+s_\text{max}(J_\text{MHSA}^{x_{i+1}}))}{\left|1 - \sqrt{\text{Var}(\text{FFN}(x_{i}'))} \right| \left|1 - \sqrt{\text{Var}(\text{MHSA}(x_{i}))} \right|}  \geq 1
\label{eqn:post_ln_z1_grad_jacobian_l2_norm_i_i+1_rearranged}
\end{equation}

We already know that $(1 + s_{\text{max}}(J_{\text{MHSA}}^{x_{i+1}})) \geq 1 \quad \mbox{and} \quad (1 + s_{\text{max}}(J_{\text{FFN}}^{x_{i}'})) \geq 1$ from Eq.~\eqref{eqn:post_ln_mhsa_ffn_bounds}. 

Now, to prove Eq.~\eqref{eqn:post_ln_z1_grad_jacobian_l2_norm_i_i+1_rearranged} to be true, we need to prove that 

\begin{equation}
    0<\left|1 - \sqrt{\text{Var}(\text{FFN}(x_{i}'))} \right| \leq 1 \quad \& \quad 0<\left|1 - \sqrt{\text{Var}(\text{MHSA}(x_{i}))} \right| \leq 1
    \label{eqn:var_mhsa_ffn_bound_z1}
\end{equation}

We do not consider the scenario where either $\left|1 - \sqrt{\text{Var}(\text{FFN}(x_{i}'))} \right|=0$ or $\left|1 - \sqrt{\text{Var}(\text{MHSA}(x_{i}))} \right|=0$, because if either/both of them becomes 0 then the gradient norm would go infinity, which we do not observe in real-world models either.

From Eq.~\eqref{eqn:ffn_condition} \& Eq.~\eqref{eqn:mhsa_condition}, we know Eq.~\eqref{eqn:var_mhsa_ffn_bound_z1} holds true under the defined conditions.

Hence, we prove $\frac{(1+s_\text{max}(J_\text{FFN}^{x_{i}'}))(1+s_\text{max}(J_\text{MHSA}^{x_{i+1}}))}{\left|1 - \sqrt{\text{Var}(\text{FFN}(x_{i}'))} \right| \left|1 - \sqrt{\text{Var}(\text{MHSA}(x_{i}))} \right|}  \geq 1$, further proving that $\text{UB}(\| g_{z_{i}} \|_2) \geq \text{UB}(\| g_{z_{i+1}} \|_2)$. This then inductively proves that \textbf{the upper bound of L2-norm of gradients for Early Layers $\text{LN}_1$ are greater than the one of Later Layers $\text{LN}_1$ in Post-LN models}, formally represented as follows:
\begin{equation} 
\begin{aligned}
\text{UB}(\|g_{z_1}\|_2) \geq \text{UB}(\|g_{z_2}\|_2) \geq \dots \geq \text{UB}(\|g_{z_N}\|_2), 
\end{aligned}
\label{eqn:early_layers_grads_higher_post_ln1}
\end{equation}
when \( 0 < \left|1 - \sqrt{\text{Var}(\text{FFN}(x_i'))} \right| \leq 1 \) and \( 0 < \left|1 - \sqrt{\text{Var}(\text{MHSA}(x_{i}))} \right| \leq 1 \).

\section{For Pre-LN model:} 

\subsubsection{Analysis for $\text{LN}_2$ $(g_{x_{i}'})$:} We need to prove that $\text{UB}(\| g_{x_{i}'} \|_2) \geq \text{UB}(\| g_{x_{i+1}'} \|_2)$, corresponding to $i^\text{th}$ and $(i+1)^\text{th}$ layer. 

We can compute $\text{UB}(\| g_{x_{i}'} \|_2)$ and $\text{UB}(\| g_{x_{i+1}'} \|_2)$ using Eq.~\eqref{eqn:pre_xl'_step5_l2_norm_upper_bound} as follows:

\begin{equation}
\begin{aligned}
\text{UB}(\|g_{x_{i}'}\|_2) =  s_\text{max}(P_2) \cdot \prod_{j=i}^{N} \left(1 + s_\text{max}(J_{\text{FFN}}^{\text{LN}_2(x_{j}')} J_{\text{LN}_2}^{x_{j}'})\right) \cdot \prod_{j=i+1}^{N} \left(1 + s_\text{max}(J_{\text{MHSA}}^{\text{LN}_1(x_{j})} J_{\text{LN}_1}^{x_{j}})\right)
\end{aligned}
\end{equation}

\begin{equation}
\begin{aligned}
\text{UB}(\|g_{x_{i+1}'}\|_2) =  s_\text{max}(P_2) \cdot \prod_{j=i+1}^{N} \left(1 + s_\text{max}(J_{\text{FFN}}^{\text{LN}_2(x_{j}')} J_{\text{LN}_2}^{x_{j}'})\right) \cdot \prod_{j=i+2}^{N} \left(1 + s_\text{max}(J_{\text{MHSA}}^{\text{LN}_1(x_{j})} J_{\text{LN}_1}^{x_{j}})\right)
\end{aligned}
\end{equation}

We then substitute these expressions in the inequality $\text{UB}(\| g_{x_{i}'} \|_2) \geq \text{UB}(\| g_{x_{i+1}'} \|_2)$, to prove that early layers have higher gradients in comparison to later layers.

\begin{equation}
\begin{aligned}
&  s_\text{max}(P_2) \cdot \prod_{j=i}^{N} \left(1 + s_\text{max}\left(J_{\text{FFN}}^{\text{LN}_2(x_j')} J_{\text{LN}_2}^{x_j'}\right)\right)
\cdot
\prod_{j=i+1}^{N} \left(1 + s_\text{max}\left(J_{\text{MHSA}}^{\text{LN}_1(x_j)} J_{\text{LN}_1}^{x_j}\right)\right) \\
&\geq
 s_\text{max}(P_2) \cdot \prod_{j=i+1}^{N} \left(1 + s_\text{max}\left(J_{\text{FFN}}^{\text{LN}_2(x_j')} J_{\text{LN}_2}^{x_j'}\right)\right)
\cdot
\prod_{j=i+2}^{N} \left(1 + s_\text{max}\left(J_{\text{MHSA}}^{\text{LN}_1(x_j)} J_{\text{LN}_1}^{x_j}\right)\right)
\end{aligned}
\end{equation}

This can be further re-written as follows:

\begin{equation}
\left(1 + s_\text{max}\left(J_{\text{FFN}}^{\text{LN}_2(x_i')} J_{\text{LN}_2}^{x_i'}\right)\right)
\cdot
\left(1 + s_\text{max}\left(J_{\text{MHSA}}^{\text{LN}_1(x_{i+1})} J_{\text{LN}_1}^{x_{i+1}}\right)\right)
\geq 1
\label{eqn:pre_ln_x2_grad_jacobian_l2_norm_i_i+1_rearranged}
\end{equation}

We know that every singular value of a matrix A is greater than or equal to 0, i.e., $s_k(A) \geq 0 \quad \forall k$. Hence, for every transformer layer,
\begin{equation}
    \begin{aligned}
    \left(1 + s_\text{max}\left(J_{\text{FFN}}^{\text{LN}_2(x_i')} J_{\text{LN}_2}^{x_i'}\right)\right)\geq 1 \quad \mbox{and} \quad \left(1 + s_\text{max}\left(J_{\text{MHSA}}^{\text{LN}_1(x_{i})} J_{\text{LN}_1}^{x_{i}}\right)\right) \geq 1
    \end{aligned}
    \label{eqn:pre_ln_mhsa_ffn_bounds}
\end{equation}

This proves that Eq.~\eqref{eqn:pre_ln_x2_grad_jacobian_l2_norm_i_i+1_rearranged} is true, hence also proving that  $\text{UB}(\| g_{x_{i}'} \|_2) \geq \text{UB}(\| g_{x_{i+1}'} \|_2) \quad \forall i$.

This then inductively proves that \textbf{the upper bound of L2-norm of gradients for Early Layers $\text{LN}_2$ are greater than the one of Later Layers $\text{LN}_2$ in Pre-LN models}, formally represented as follows:
\begin{equation} 
\begin{aligned}
\text{UB}(\|g_{x_1'}\|_2) \geq \text{UB}(\|g_{x_2'}\|_2) \geq \dots \geq \text{UB}(\|g_{x_N'}\|_2)
\end{aligned}
\label{eqn:early_layers_grads_higher_pre_ln2}
\end{equation}

\subsubsection{Analysis for $\text{LN}_1$ $(g_{x_{i}})$:} We need to prove that $\text{UB}(\| g_{x_{i}} \|_2) \geq \text{UB}(\| g_{x_{i+1}} \|_2)$, corresponding to $i^\text{th}$ and $(i+1)^\text{th}$ layer. 

We can compute $\text{UB}(\| g_{x_{i}} \|_2)$ and $\text{UB}(\| g_{x_{i+1}} \|_2)$ using Eq.~\eqref{eqn:pre_xl_step5_l2_norm_upper_bound} as follows:

\begin{equation}
\begin{aligned}
\text{UB}(\|g_{x_{i}}\|_2) =  s_\text{max}(P_2) \cdot \prod_{j=i}^{N} \left(1 + s_\text{max}(J_{\text{FFN}}^{\text{LN}_2(x_{j}')} J_{\text{LN}_2}^{x_{j}'})\right) \cdot \prod_{j=i}^{N} \left(1 + s_\text{max}(J_{\text{MHSA}}^{\text{LN}_1(x_{j})} J_{\text{LN}_1}^{x_{j}})\right)
\end{aligned}
\end{equation}

\begin{equation}
\begin{aligned}
\text{UB}(\|g_{x_{i+1}}\|_2) =  s_\text{max}(P_2) \cdot \prod_{j=i+1}^{N} \left(1 + s_\text{max}(J_{\text{FFN}}^{\text{LN}_2(x_{j}')} J_{\text{LN}_2}^{x_{j}'})\right) \cdot \prod_{j=i+1}^{N} \left(1 + s_\text{max}(J_{\text{MHSA}}^{\text{LN}_1(x_{j})} J_{\text{LN}_1}^{x_{j}})\right)
\end{aligned}
\end{equation}

We then substitute these expressions in the inequality $\text{UB}(\| g_{x_{i}} \|_2) \geq \text{UB}(\| g_{x_{i+1}} \|_2)$, to prove that early layers have higher gradients in comparison to later layers.

\begin{equation}
\begin{aligned}
&  s_\text{max}(P_2) \cdot \prod_{j=i}^{N} \left(1 + s_\text{max}\left(J_{\text{FFN}}^{\text{LN}_2(x_j')} J_{\text{LN}_2}^{x_j'}\right)\right)
\cdot
\prod_{j=i}^{N} \left(1 + s_\text{max}\left(J_{\text{MHSA}}^{\text{LN}_1(x_j)} J_{\text{LN}_1}^{x_j}\right)\right) \\
&\geq
 s_\text{max}(P_2) \cdot \prod_{j=i+1}^{N} \left(1 + s_\text{max}\left(J_{\text{FFN}}^{\text{LN}_2(x_j')} J_{\text{LN}_2}^{x_j'}\right)\right)
\cdot
\prod_{j=i+1}^{N} \left(1 + s_\text{max}\left(J_{\text{MHSA}}^{\text{LN}_1(x_j)} J_{\text{LN}_1}^{x_j}\right)\right)
\end{aligned}
\end{equation}

This can be further re-written as follows:

\begin{equation}
\left(1 + s_\text{max}\left(J_{\text{FFN}}^{\text{LN}_2(x_i')} J_{\text{LN}_2}^{x_i'}\right)\right)
\cdot
\left(1 + s_\text{max}\left(J_{\text{MHSA}}^{\text{LN}_1(x_{i})} J_{\text{LN}_1}^{x_{i}}\right)\right)
\geq 1
\label{eqn:pre_ln_x1_grad_jacobian_l2_norm_i_i+1_rearranged}
\end{equation}

We know that $\left(1 + s_\text{max}\left(J_{\text{FFN}}^{\text{LN}_2(x_i')} J_{\text{LN}_2}^{x_i'}\right)\right)\geq 1 \quad \mbox{and} \quad \left(1 + s_\text{max}\left(J_{\text{MHSA}}^{\text{LN}_1(x_{i})} J_{\text{LN}_1}^{x_{i}}\right)\right) \geq 1$ from Eq.~\eqref{eqn:pre_ln_mhsa_ffn_bounds}.

This proves that Eq.~\eqref{eqn:pre_ln_x1_grad_jacobian_l2_norm_i_i+1_rearranged} is true, hence also proving that  $\text{UB}(\| g_{x_{i}} \|_2) \geq \text{UB}(\| g_{x_{i+1}} \|_2) \quad \forall i$.

This then consequently proves that \textbf{the upper bound of L2-norm of gradients for Early Layers $\text{LN}_1$ are greater than the one of Later Layers $\text{LN}_1$ in Pre-LN models}, formally represented as follows:
\begin{equation} 
\begin{aligned}
\text{UB}(\|g_{x_1}\|_2) \geq \text{UB}(\|g_{x_2}\|_2) \geq \dots \geq \text{UB}(\|g_{x_N}\|_2)
\end{aligned}
\label{eqn:early_layers_grads_higher_pre_ln1}
\end{equation}

\hfill \qedsymbol{}

\section{Training Details} \label{sec:training_details}
This section outlines the detailed configurations of the datasets and models used in our experiments, including dataset splits, pre-processing steps, model architectures, and training settings.

\subsection{Datasets}
Below we discuss the 6 datasets used in our paper.

\paragraph{Emotions} dataset, proposed in \citet{saravia-etal-2018-carer}, consists of 16,000 train, 2,000 validation and 2,000 test samples, each sample belonging to one of the 6 classes (class 0-5): sadness, joy, love, anger, fear and surprise. To induce the notion of noisy labels, we randomly flip labels of 1\% of class 5 train samples to any another random class label and let the model train till we reach 100\% train accuracy (memorizing the noisy labels).

\paragraph{TweetTopic} dataset, developed in \citet{dimosthenis-etal-2022-twitter}, consists of 2,858 train, 352 validation, and 376 test samples, spanning across 6 classes (class 0-5): arts\_\&\_culture, business\_\&\_entrepreneurs, pop\_culture, science\_\&\_technology, sports\_\&\_gaming, daily\_life. To introduce memorization of noisy labels, we flip labels of 1\% of class 3 train samples to any another random class label while training the model till it reaches 100\% train accuracy.

\paragraph{News} dataset, proposed in \citet{okite97_news_data}, consists of 4,686 train and 828 test samples, spanning across 6 classes (class 0-5): business, sports, politics, health, entertainment and tech. We then split the train set to train \& validation using 90:10 stratified split over the class labels. Also, we introduce noisy labels, by flipping labels of 1\% of class 5 train samples to any another random class label, while letting the model overfit to 100\% train accuracy. 

\paragraph{CIFAR10} dataset, first introduced in \citet{krizhevsky2009learning}, consists of 60,000 samples, with each belonging to one of the 10 classes (class 0-9): airplane, automobile, bird, cat, deer, dog, frog, horse, ship, and truck. In our setup we consider a subset of - 16,000 train, 4,000 validation and 10,000 test samples. We also resize the images to size (224x224x3) for training the transformer model. To introduce noisy labels, we flip labels of 1\% of class 9 train samples to any another random class label, and train the model till it reaches 100\% train accuracy.

\paragraph{UTK-Face} dataset, proposed in \citet{zhang2017age}, consists of 23,705 samples and 5 classes (classs 0-4) depicting ethnicity: white, black, asian, indian and others. We split the dataset into train, validation and testing using 65:15:20 stratified split. We also resize the images to size (224x224x3) for training the transformer model. We then introduce noisy labels, by flipping labels of 1\% of class 2 train samples to any another random class label, and train the model till it achieves 100\% train accuracy.

\paragraph{NICO++} dataset \citep{zhang2023nico++} consists of approximately 60,000 images distributed across 60 categories of everyday objects. In this study, we select a subset of 15 object categories, including car, flower, penguin, camel, chair, monitor, truck, wheat, sword, seal, lion, fish, dolphin, lifeboat, and tank. The dataset is partitioned into training (80\%), validation (10\%), and testing (10\%) sets using stratified sampling. Images are resized to (224x224x3) for consistency with the model input requirements. We then introduce noisy labels, by flipping labels of 1\% of class 6 train samples to any another random class label, and train the model till it achieves 100\% train accuracy.

\subsection{Models}
Below we discuss the 13 transformer models (6 Post-LN and 7 Pre-LN) considered as part of our paper. We utilize the Sequence Classification variant of all the models available on Huggingface\footnote{\url{https://huggingface.co/docs/transformers/index}}.

\begin{table}[ht]
\centering
\begin{tabular}{|p{4cm}|p{9cm}|}
\hline
\textbf{Post-LN Models} & \textbf{Description} \\
\hline
BERT \citep{devlin2019bert} & 12-layer bidirectional transformer for masked language modeling and next sentence prediction. \\
DeBERTa \citep{he2020deberta} & 12-layer model with disentangled position/content embeddings and decoding-enhanced attention. \\
RoBERTa \citep{yinhan2019roberta} & 12-layer robustly optimized BERT variant trained with dynamic masking and more data. \\
ELECTRA \citep{clark2020electra} & 12-layer model using replaced token detection for sample-efficient pre-training. \\
DistillBERT \citep{sanh2019distilbert} & 6-layer distilled BERT that is smaller and faster variant of BERT, retaining strong performance. \\
Longformer \citep{beltagy2020longformer} & 12-layer model with sparse attention for efficient long-sequence processing. \\
\hline
\textbf{Pre-LN Models} & \textbf{Description} \\
\hline
GPT2-Medium \citep{radford2019language} & 24-layer unidirectional transformer trained for causal language modeling. \\
GPTNeo-125M \citep{black2022gpt} & 12-layer open-source causal language model trained on The Pile dataset \\
Qwen2-0.5B \citep{yang2024qwen2} & 24-layer efficient LLM optimized for generative tasks, using RMSNorm \citep{zhang2019root} \\
RoBERTA-PreLayerNorm \citep{ott-etal-2019-fairseq} & 24-layer variant of RoBERTa with Pre-LN setting for improved training stability. \\
ViT-B \citep{alexey2020image} & 12-layer Vision Transformer Base model for image classification. \\
ViT-S \citep{assran2022masked} & 12-layer smaller ViT variant trained with Masked Siamese Networks (MSN). \\
DeiT \citep{touvron2021training} & 12-layer Data-efficient Image Transformer trained with distillation, without external data. \\
\hline
\end{tabular}
\caption{Overview of the 13 transformer models categorized into 6 Post-LN and 7 Pre-LN architectures.}
\label{tab:transformer-models}
\end{table}

\newpage

\subsection{Training Settings \& Hyperparameters}
In our study, we explore various combinations of different models and datasets across our experiments, as follows - (1) Emotions dataset with BERT (Post-LN), DeBERTa (Post-LN), DistillBERT (Post-LN), GPTNeo (Pre-LN), (2) News dataset with ELECTRA (Post-LN), Longformer (Post-LN), Qwen2 (Pre-LN), (3) Tweets dataset with RoBERTa (Post-LN), GPT2 (Pre-LN), RoBERTa-PreLayerNorm (Pre-LN), (4) CIFAR10 dataset with ViT-B (Pre-LN), (5) UTK-Face dataset with DeiT (Pre-LN), and (6) NICO++ dataset with ViT-S (Pre-LN). We fully unfreeze all layers of the pre-trained models during training.

Regarding the hyperparameters, we consider a learning rate of 2e-5 and a batch size of 16 for all models. Then, we train the Post-LN models for 40 epochs and Pre-LN models for 70 epochs. We run 70 epochs for Pre-LN models, because after removal of LN parameters, the learning accuracy is impacted significantly in Pre-LN models from the start. Hence, we examine if the accuracy would increase by letting the model train more. In addition to that, we do not use any data augmentation in our training procedures. We also provide the code for our experiments in the supplementary file.

\subsection{Grouping of Early, Middle, and Later Layers}\label{sec:grouping_early_middle_later}

\begin{figure}[h]
    \centering
    \begin{subfigure}{0.52\textwidth}
        \centering
        \includegraphics[width=\textwidth]{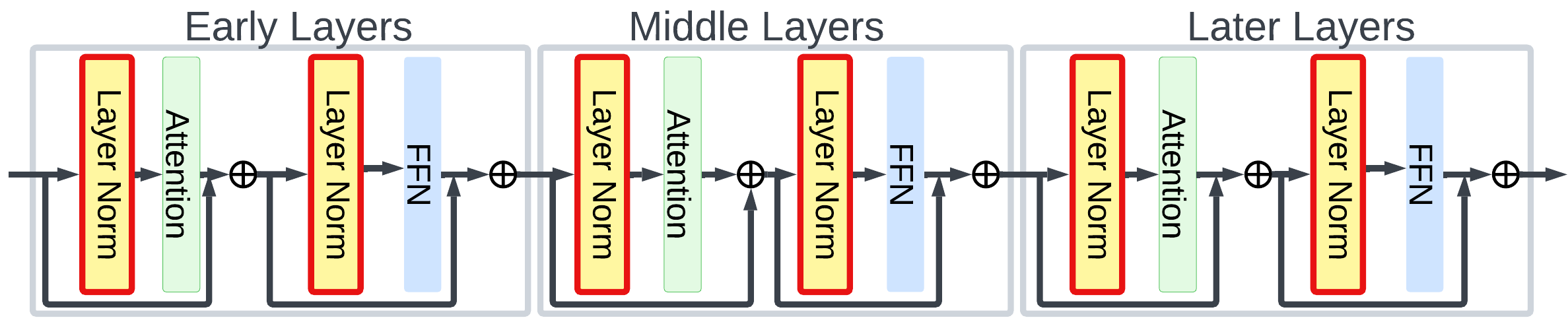}
        \caption{Pre-LN architecture}
        \label{fig:file1}
    \end{subfigure}
    \hfill
    \begin{subfigure}{0.47\textwidth}
        \centering
        \includegraphics[width=\textwidth]{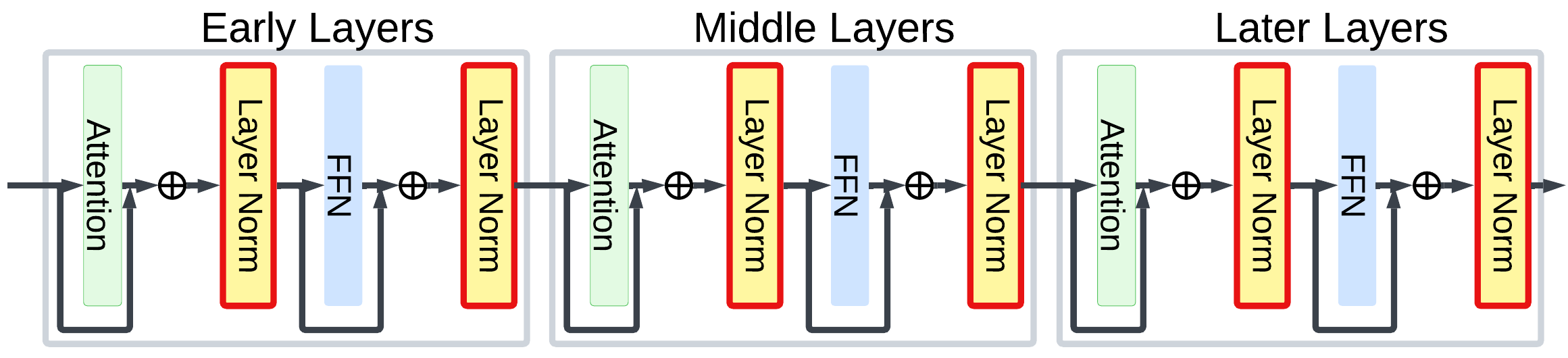}
        \caption{Post-LN architecture}
        \label{fig:file2}
    \end{subfigure}
    \caption{Pre-LN vs. Post-LN architectures depicting LN placement and categorization of early, middle and later layers}
    \label{fig:Pre_Post_Arch_Diff}
\end{figure}

To investigate which group of layers' Layer Normalization (LN) contributes most to memorization and learning, we divide the transformer layers into three subsets: \textbf{Early}, \textbf{Middle}, and \textbf{Later} layers as shown in Fig.~\ref{fig:Pre_Post_Arch_Diff}.

Formally, for a transformer model with $N$ layers (where $N$ is divisible by 3), we define:

\begin{equation}
\begin{aligned}
\text{Early Layers} &= \{1, 2, \ldots, \tfrac{N}{3} \} \\
\text{Middle Layers} &= \left\{ \tfrac{N}{3} + 1, \ldots, \tfrac{2N}{3} \right\} \\
\text{Later Layers} &= \left\{ \tfrac{2N}{3} + 1, \ldots, N\right\}
\end{aligned}
\end{equation}

This grouping helps separately examine which set of layers most significantly influences learning and memorization in transformers.

\section{Additional Experiments \& Results}

This section presents additional experiments (on top of the ones discussed in Sec.~\ref{sec:discrepancy_post_pre_ln}, Sec.~\ref{sec:early_layers_imp} and Sec.~\ref{sec:learn_mem_grad})and results on supplementary datasets and models, expanding on the analyses in Sec.~\ref{sec:ln_impact_mem_learn}, \ref{sec:significance_early_layers_ln_appendix}, and \ref{sec:gradients_explain_appendix}. These experiments aim to: (1) establish the distinctive impact of Layer Normalization (LN) on memorization and learning in Pre-LN vs. Post-LN models, (2) assess the role of LN in early layers, and (3) investigate how gradient behavior accounts for the observed phenomena.

\subsection{Impact of LN on Memorization \& Learning in Pre- and Post-LN models} \label{sec:ln_impact_mem_learn}
In this section, we present the results corresponding to the distinctive impact of LN of memorization and learning for the remaining Pre and Post LN models spanning multiple datasets. These results further corroborate our finding that removal of LN parameters in pre-LN models critically destabilizes learning while exacerbating overfitting, while for Post-LN models, LN parameters removal, suppresses memorization and facilitates true label recovery without impacting learning.

\begin{figure}[htbp]
    \centering
    \begin{subfigure}[t]{0.35\textwidth}
        \centering
        \includegraphics[width=\textwidth]{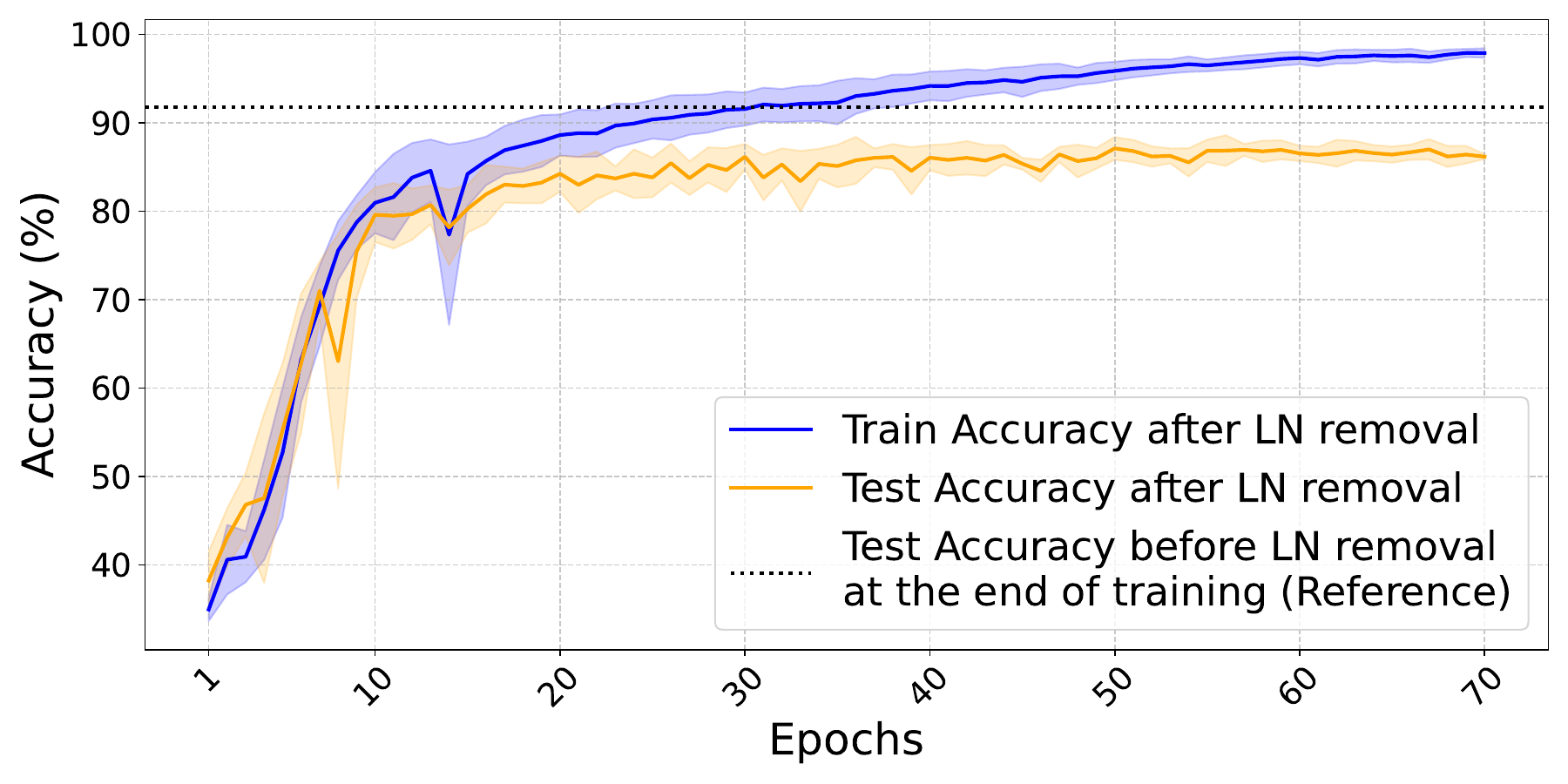}
        \caption{Learning (Test) accuracy over epochs for Pre-LN Model (GPTNeo)}
        \label{fig:pre_ln_learning_epochs_gpt_neo}
    \end{subfigure}
    \hspace{5pt}
    \begin{subfigure}[t]{0.35\textwidth}
        \centering
        \includegraphics[width=\textwidth]{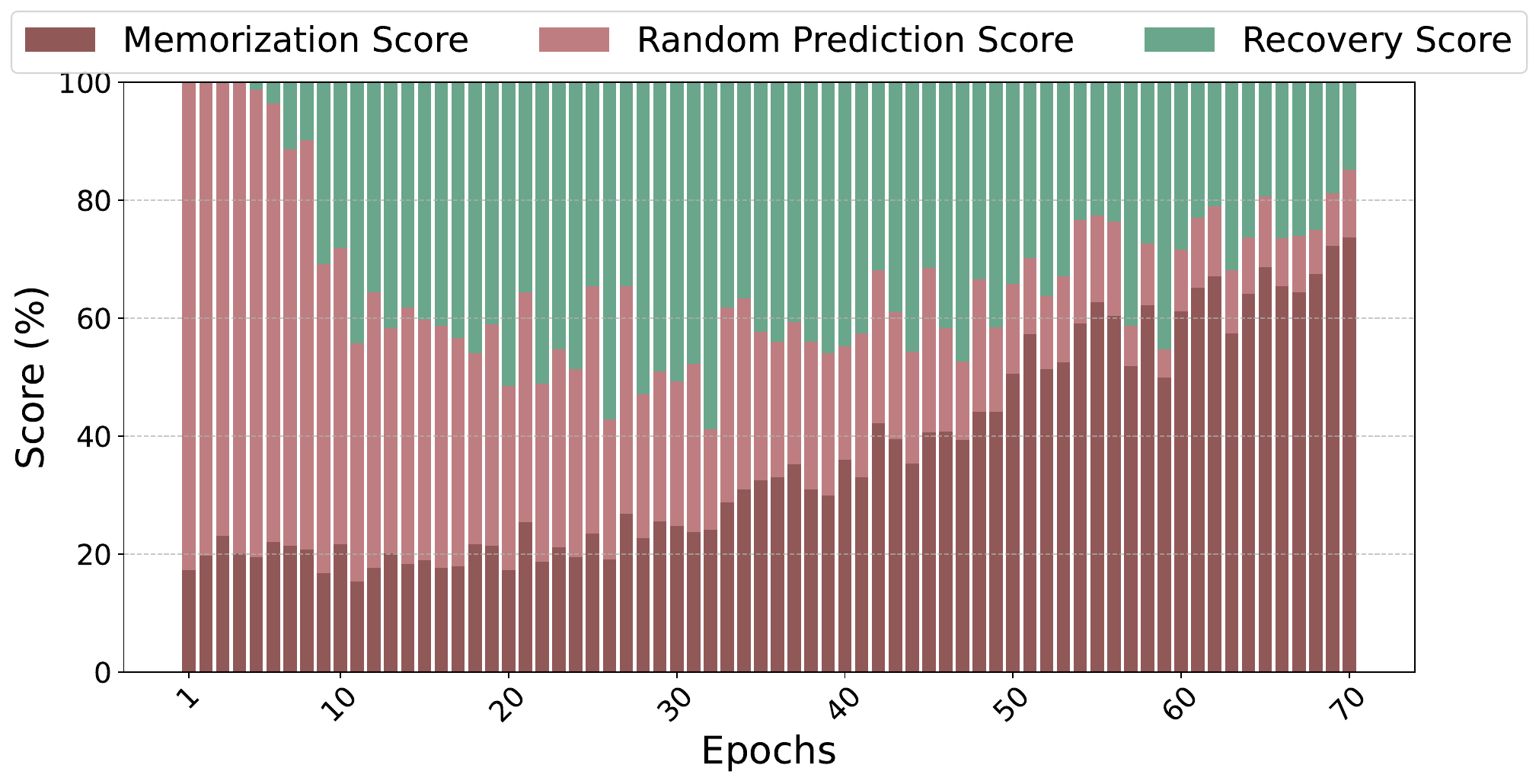}
        \caption{Memorization, Recovery and Random Predictions over epochs for Pre-LN Model (GPTNeo)}
        \label{fig:pre_ln_mem_epochs_gpt_neo}
    \end{subfigure}
    \hspace{5pt}
    \begin{subfigure}[t]{0.25\textwidth}
        \centering
        \includegraphics[width=\textwidth]{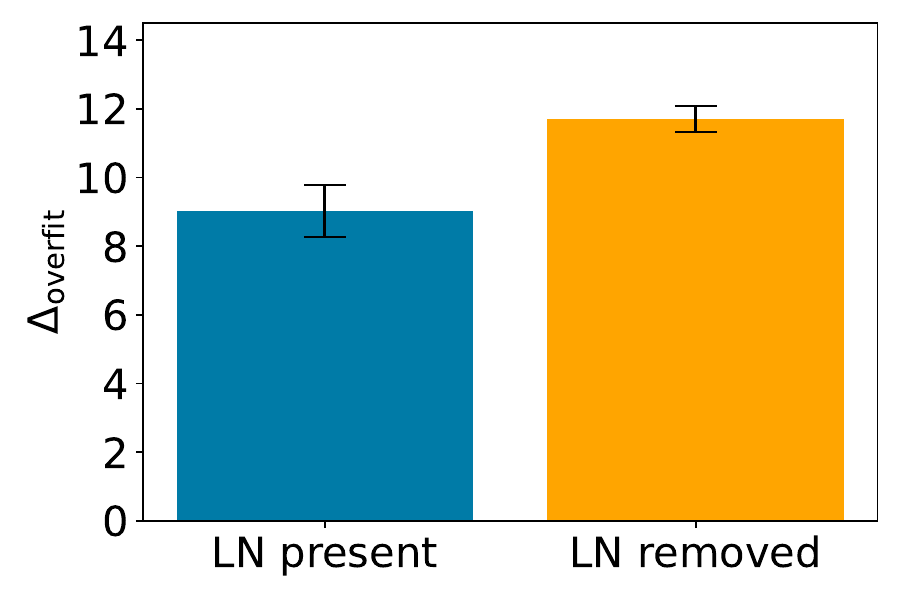}
        \caption{Overfitting gap for Pre-LN Model (GPTNeo)}
        \label{fig:pre_ln_overfit_gap_gpt_neo}
    \end{subfigure}
    \caption{\textbf{LN removal destabilizes learning in Pre-LN model - GPTNeo, Emotions Dataset:} LN removal critically affects learning while memorization still persists in GPTNeo. This further exacerbates overfitting, explained by increasing train-test accuracy gap when LN is removed, due to the drop in test-accuracy.}
    \label{fig:learning_destabilized_gptneo}
\end{figure}

\begin{figure}[htbp]
    \centering
    \begin{subfigure}[t]{0.35\textwidth}
        \centering
        \includegraphics[width=\textwidth]{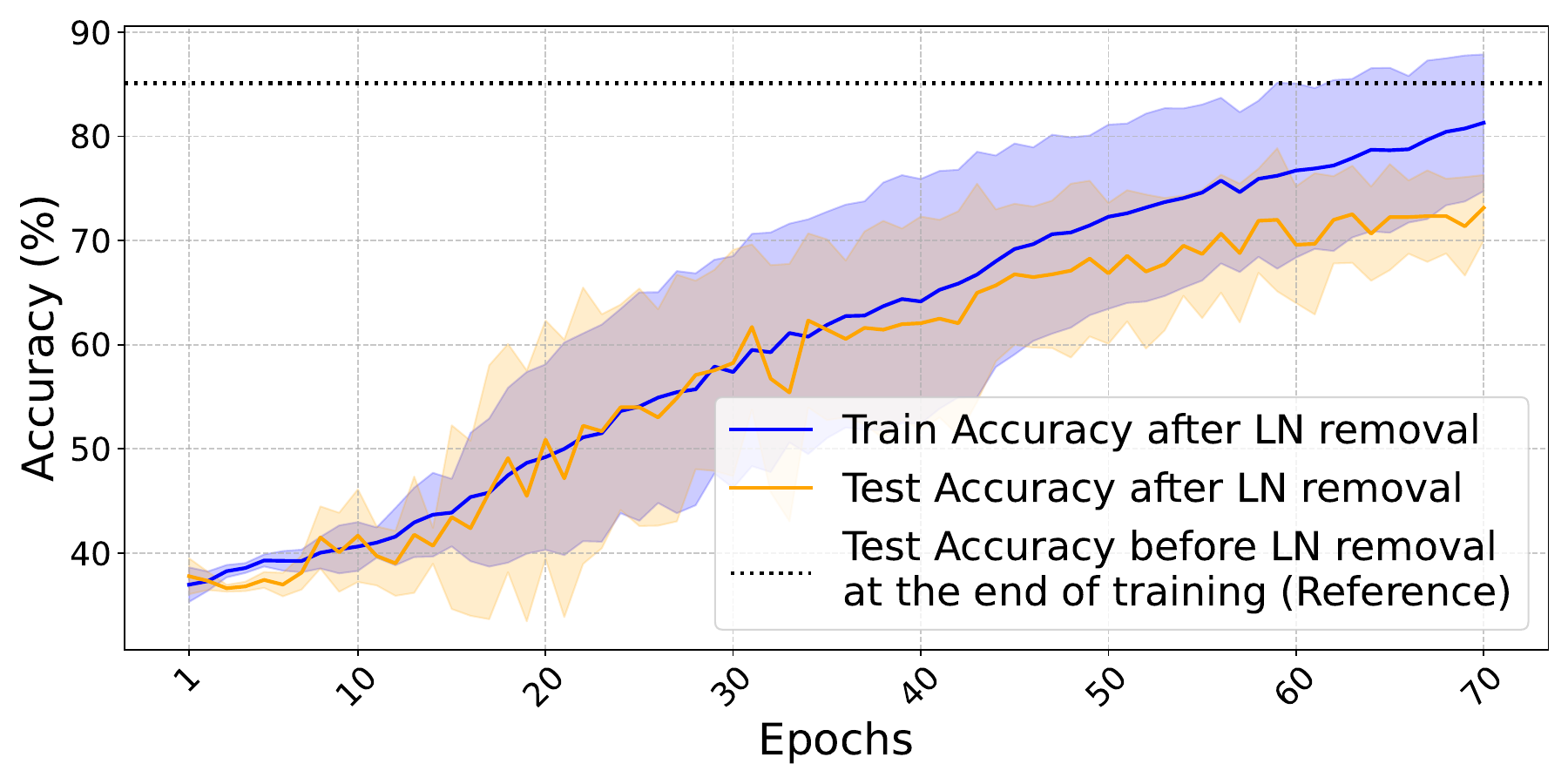}
        \caption{Learning (Test) accuracy over epochs for Pre-LN Model (GPT2)}
        \label{fig:pre_ln_learning_epochs_gpt2}
    \end{subfigure}
    \hspace{5pt}
    \begin{subfigure}[t]{0.35\textwidth}
        \centering
        \includegraphics[width=\textwidth]{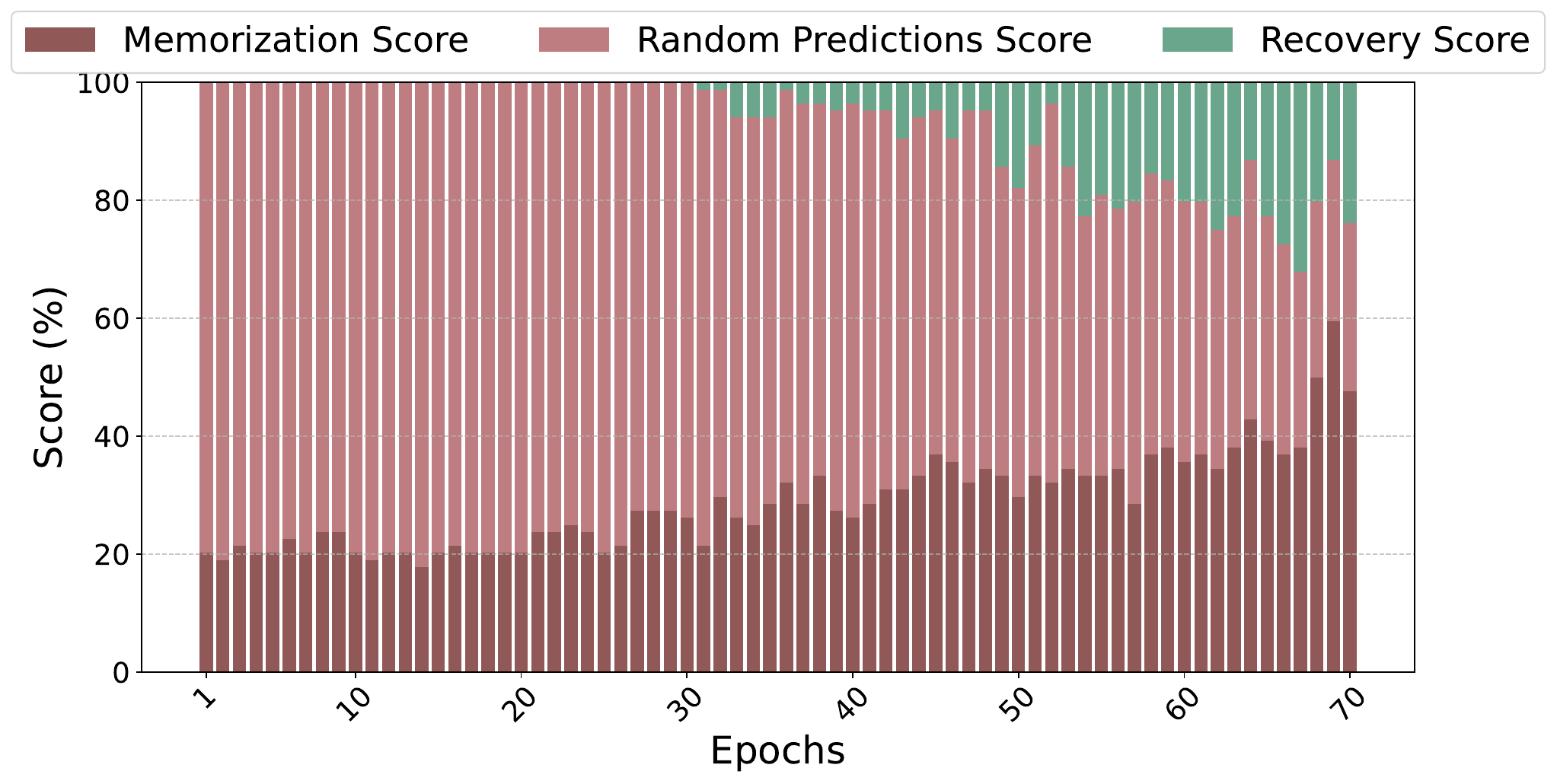}
        \caption{Memorization, Recovery and Random Predictions over epochs for Pre-LN Model (GPT2)}
        \label{fig:pre_ln_mem_epochs_gpt2}
    \end{subfigure}
    \hspace{5pt}
    \begin{subfigure}[t]{0.25\textwidth}
        \centering
        \includegraphics[width=\textwidth]{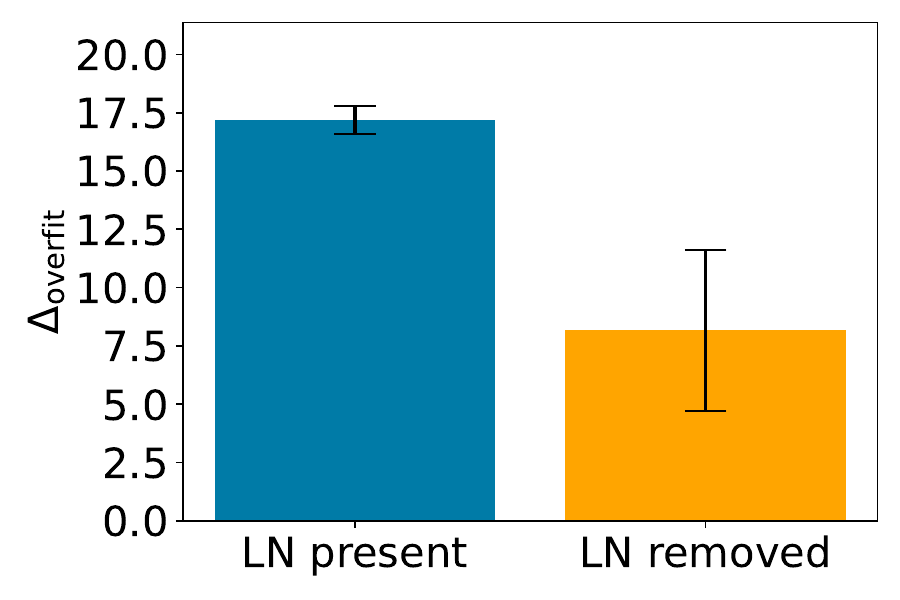}
        \caption{Overfitting gap for Pre-LN Model (GPT2)}
        \label{fig:pre_ln_overfit_gap_gpt2}
    \end{subfigure}
    \caption{\textbf{LN removal destabilizes learning in Pre-LN model - GPT2, TweetTopic Dataset:} LN removal critically affects learning while memorization still persists in GPT2. For GPT2, the overfitting gap decreased after LN removal, because the model could not even stabilize during training due to the destabilization of learning. Hence both train and test accuracies remain low and comparable. However, the learning accuracy still remains low when LN is absent in comparison to when LN was present, and struggle with high memorization and random predictions (\textcolor{BrickRed}{red}-color family bars).}
    \label{fig:learning_destabilized_gpt2}

\end{figure}

\begin{figure}[htbp]
    \centering
    \begin{subfigure}[t]{0.35\textwidth}
        \centering
        \includegraphics[width=\textwidth]{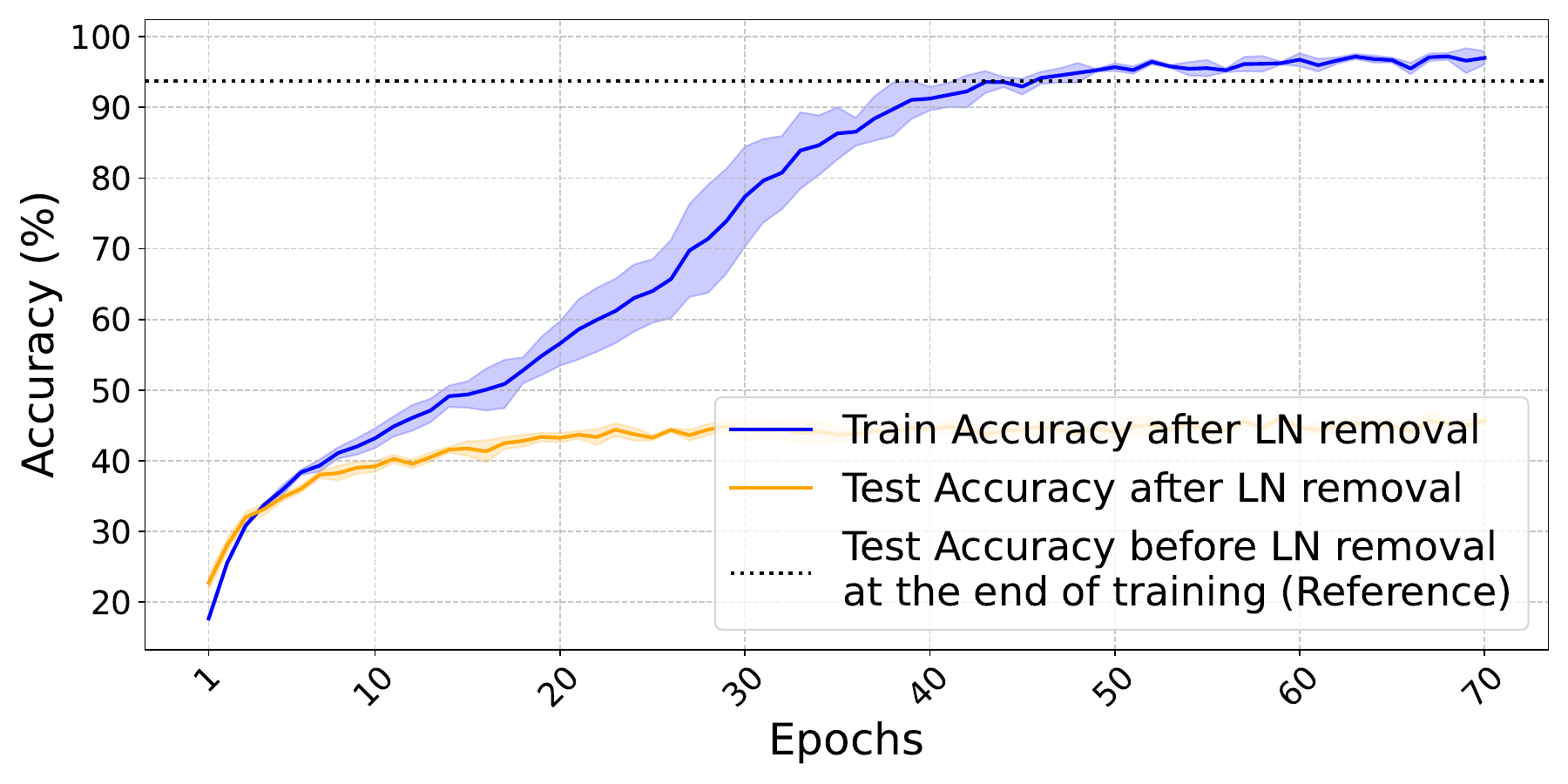}
        \caption{Learning (Test) accuracy over epochs for Pre-LN Model (ViT-B)}
        \label{fig:pre_ln_learning_epochs_vit_base}
    \end{subfigure}
    \hspace{5pt}
    \begin{subfigure}[t]{0.35\textwidth}
        \centering
        \includegraphics[width=\textwidth]{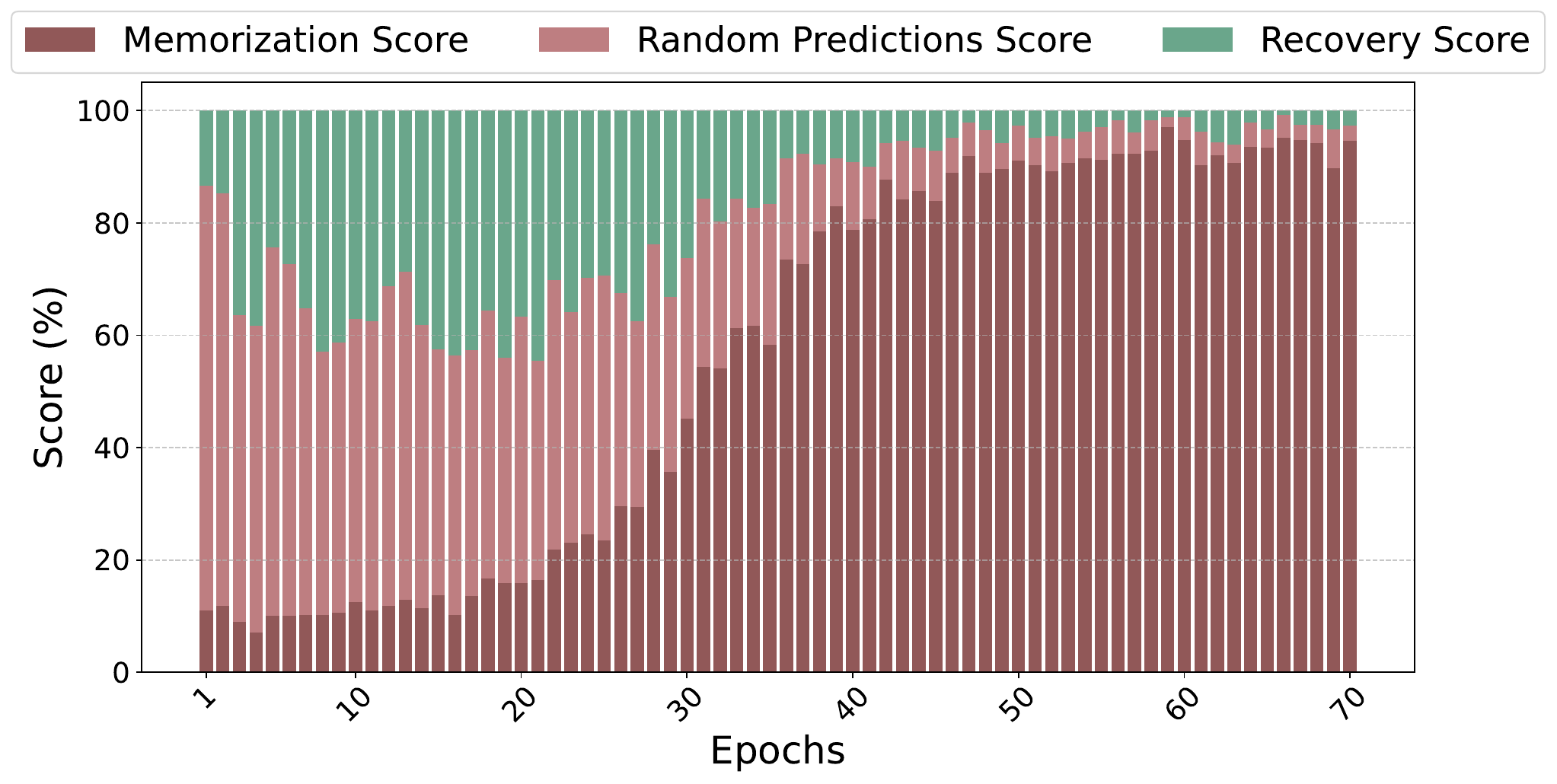}
        \caption{Memorization, Recovery and Random Predictions over epochs for Pre-LN Model (ViT-B)}
        \label{fig:pre_ln_mem_epochs_vit_base}
    \end{subfigure}
    \hspace{5pt}
    \begin{subfigure}[t]{0.25\textwidth}
        \centering
        \includegraphics[width=\textwidth]{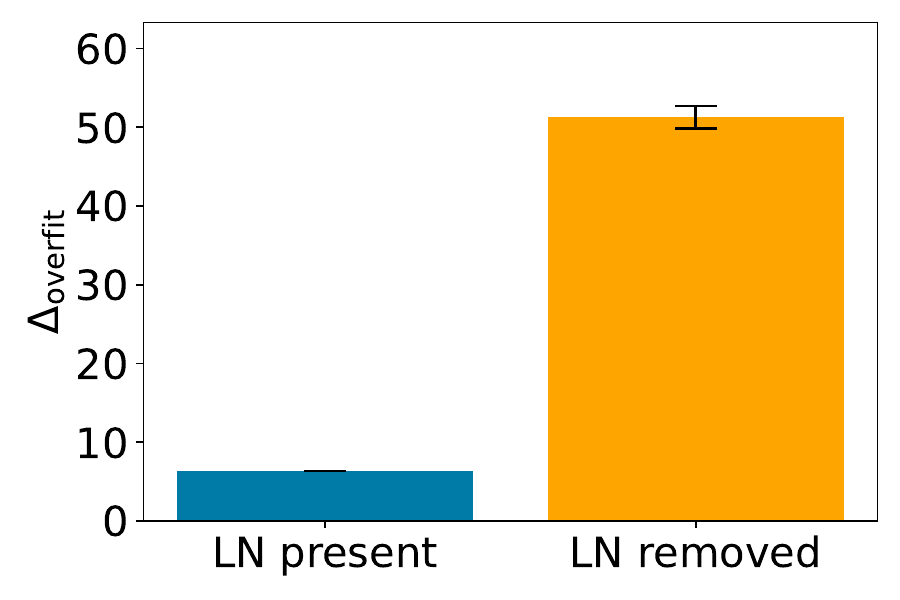}
        \caption{Overfitting gap for Pre-LN Model (ViT-B)}
        \label{fig:pre_ln_overfit_gap_vit_base}
    \end{subfigure}
    \caption{\textbf{LN removal destabilizes learning in Pre-LN model - ViT-B, CIFAR10 Dataset:} LN removal critically affects learning while memorization still persists in ViT-B. This further exacerbates overfitting seen by increasing train-test accuracy gap when LN is removed, due to the drop in test-accuracy.}
    \label{fig:learning_destabilized_vit_base}

\end{figure}

\begin{figure}[htbp]
    \centering
    \begin{subfigure}[t]{0.35\textwidth}
        \centering
        \includegraphics[width=\textwidth]{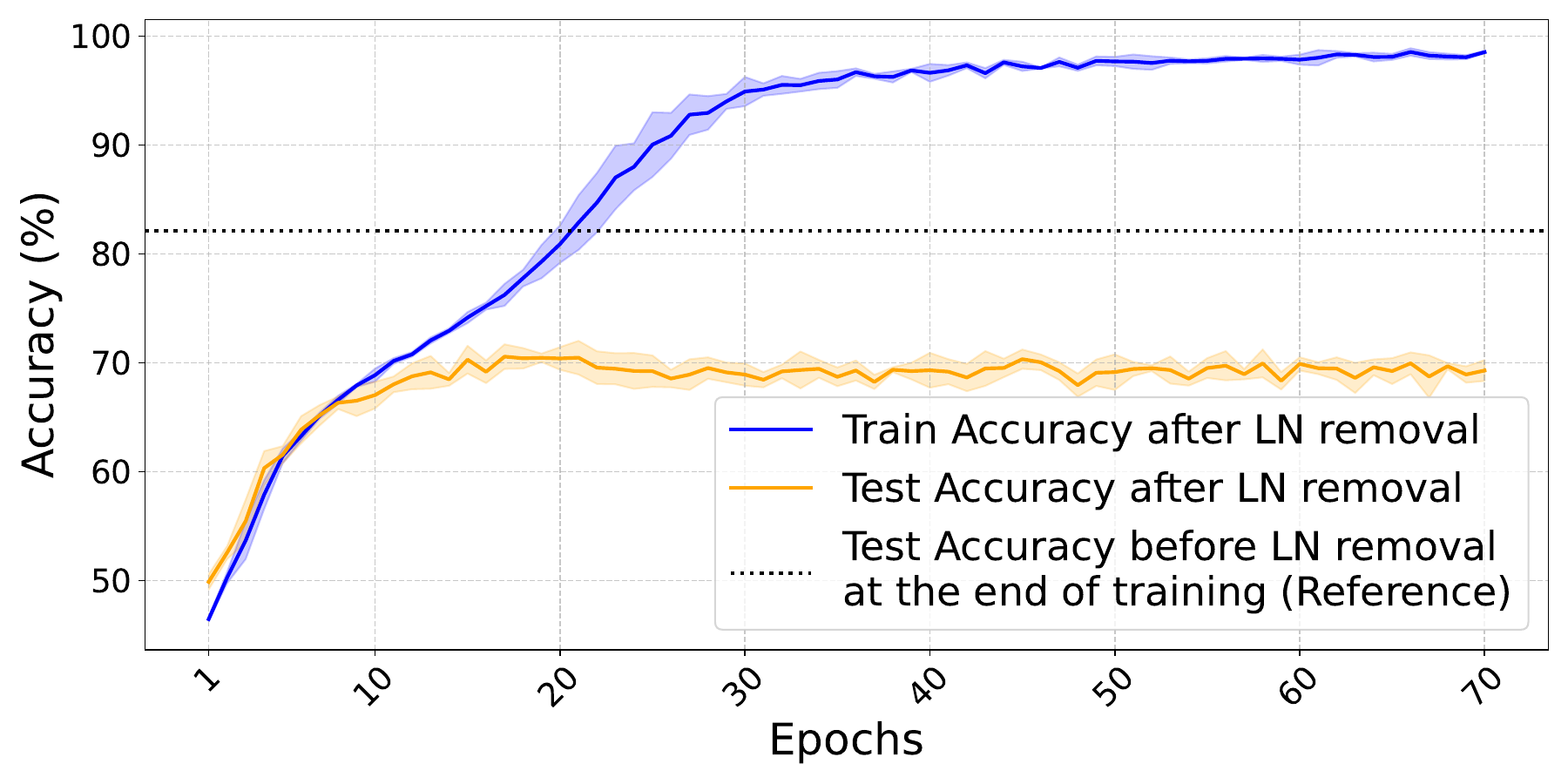}
        \caption{Learning (Test) accuracy over epochs for Pre-LN Model (DeiT)}
        \label{fig:pre_ln_learning_epochs_deit}
    \end{subfigure}
    \hspace{5pt}
    \begin{subfigure}[t]{0.35\textwidth}
        \centering
        \includegraphics[width=\textwidth]{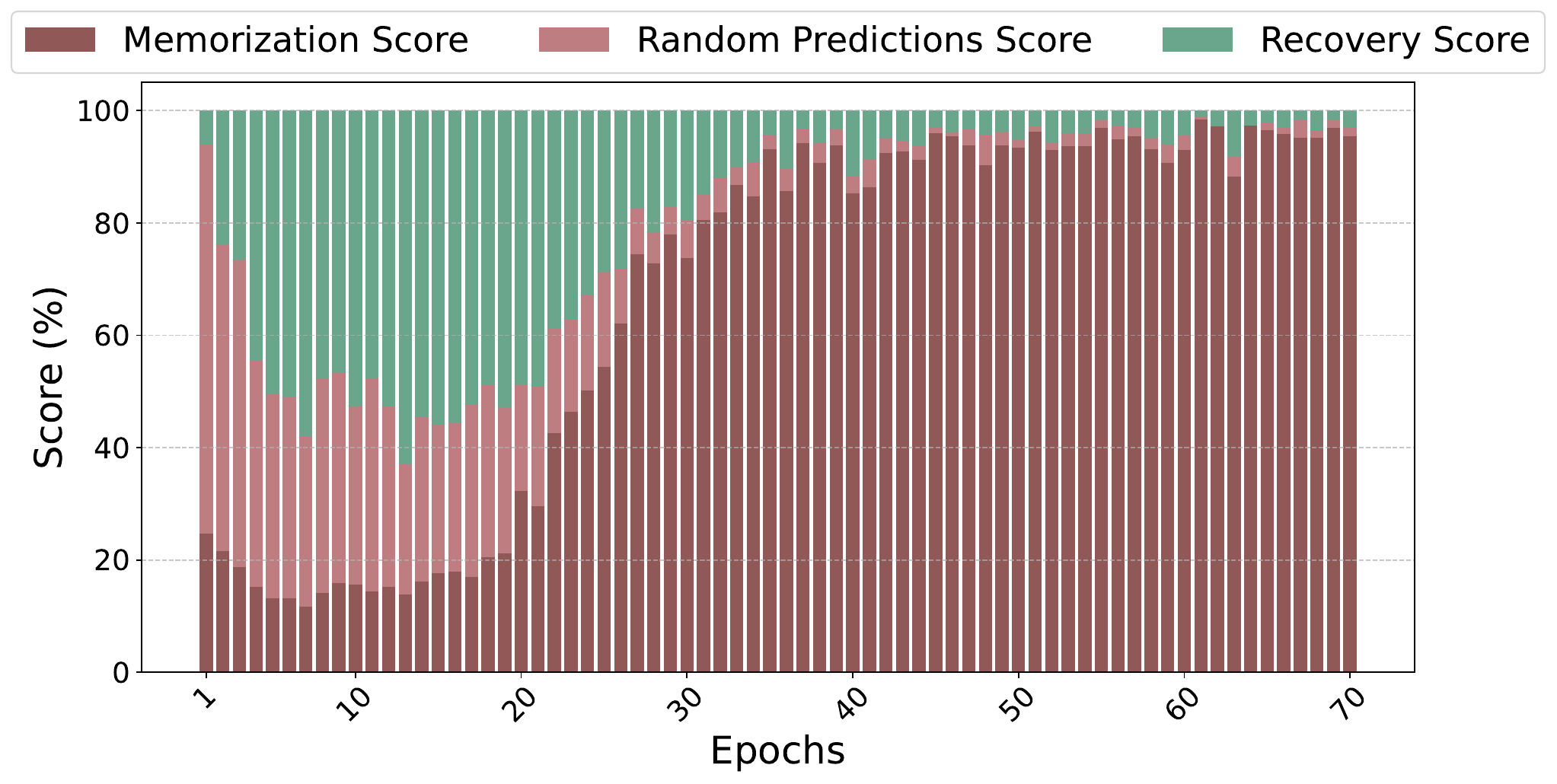}
        \caption{Memorization, Recovery and Random Predictions over epochs for Pre-LN Model (DeiT)}
        \label{fig:pre_ln_mem_epochs_deit}
    \end{subfigure}
    \hspace{5pt}
    \begin{subfigure}[t]{0.25\textwidth}
        \centering
        \includegraphics[width=\textwidth]{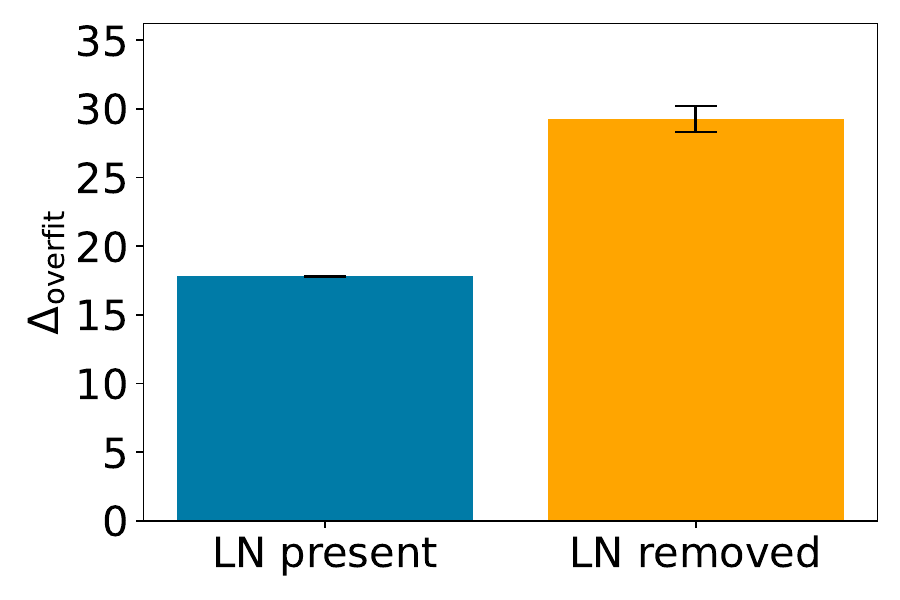}
        \caption{Overfitting gap for Pre-LN Model (DeiT)}
        \label{fig:pre_ln_overfit_gap_deit}
    \end{subfigure}
    \caption{\textbf{LN removal destabilizes learning in Pre-LN model - DeiT, UTK-Face Dataset:} LN removal critically affects learning while memorization still persists in DeiT. This further exacerbates overfitting seen by increasing train-test accuracy gap when LN is removed, due to the drop in test-accuracy.}
    \label{fig:learning_destabilized_deit}

\end{figure}

\begin{figure}[htbp]
    \centering
    \begin{subfigure}[t]{0.35\textwidth}
        \centering
        \includegraphics[width=\textwidth]{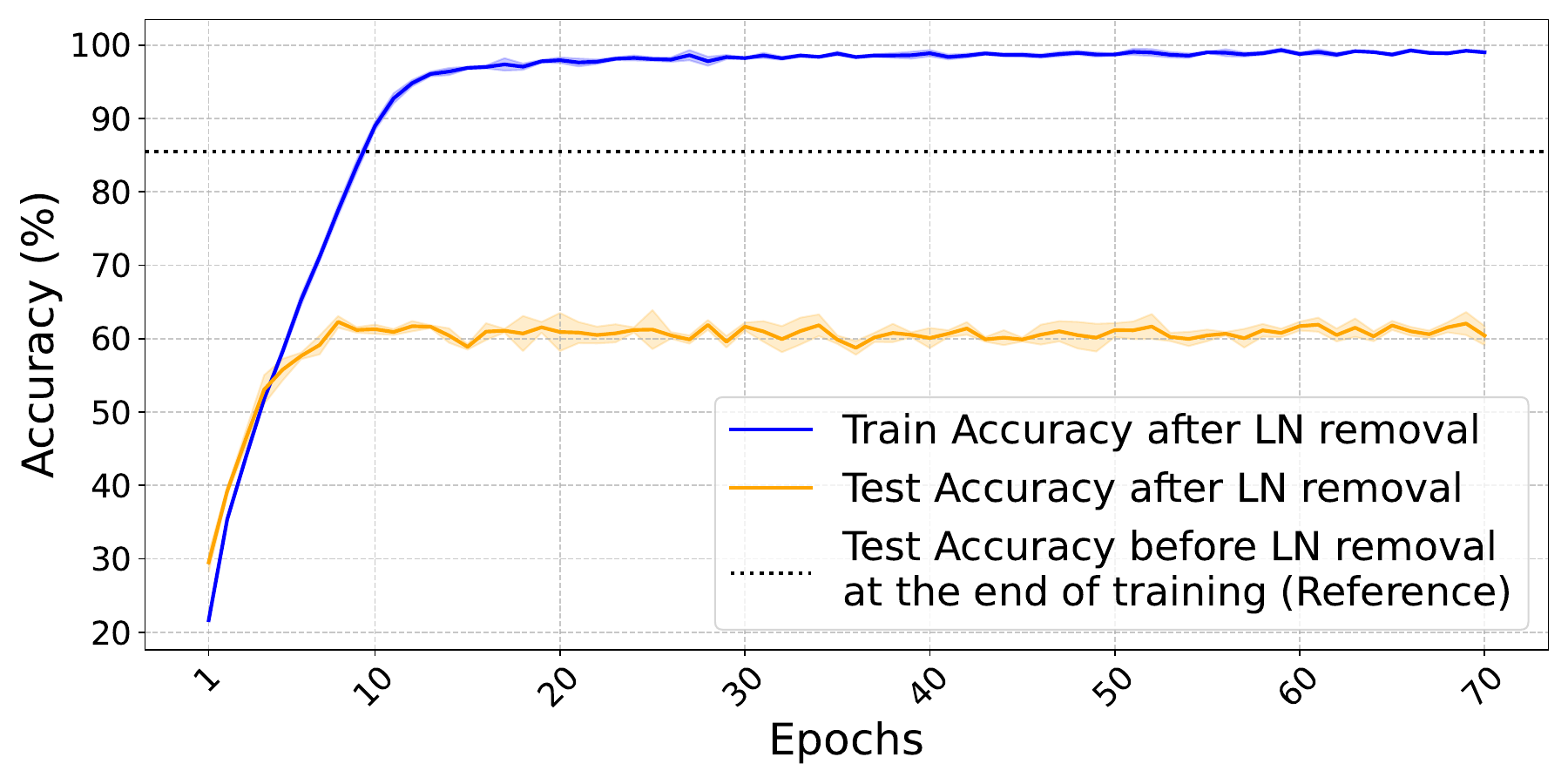}
        \caption{Learning (Test) accuracy over epochs for Pre-LN Model (ViT-S)}
        \label{fig:pre_ln_learning_epochs_vit_small}
    \end{subfigure}
    \hspace{5pt}
    \begin{subfigure}[t]{0.35\textwidth}
        \centering
        \includegraphics[width=\textwidth]{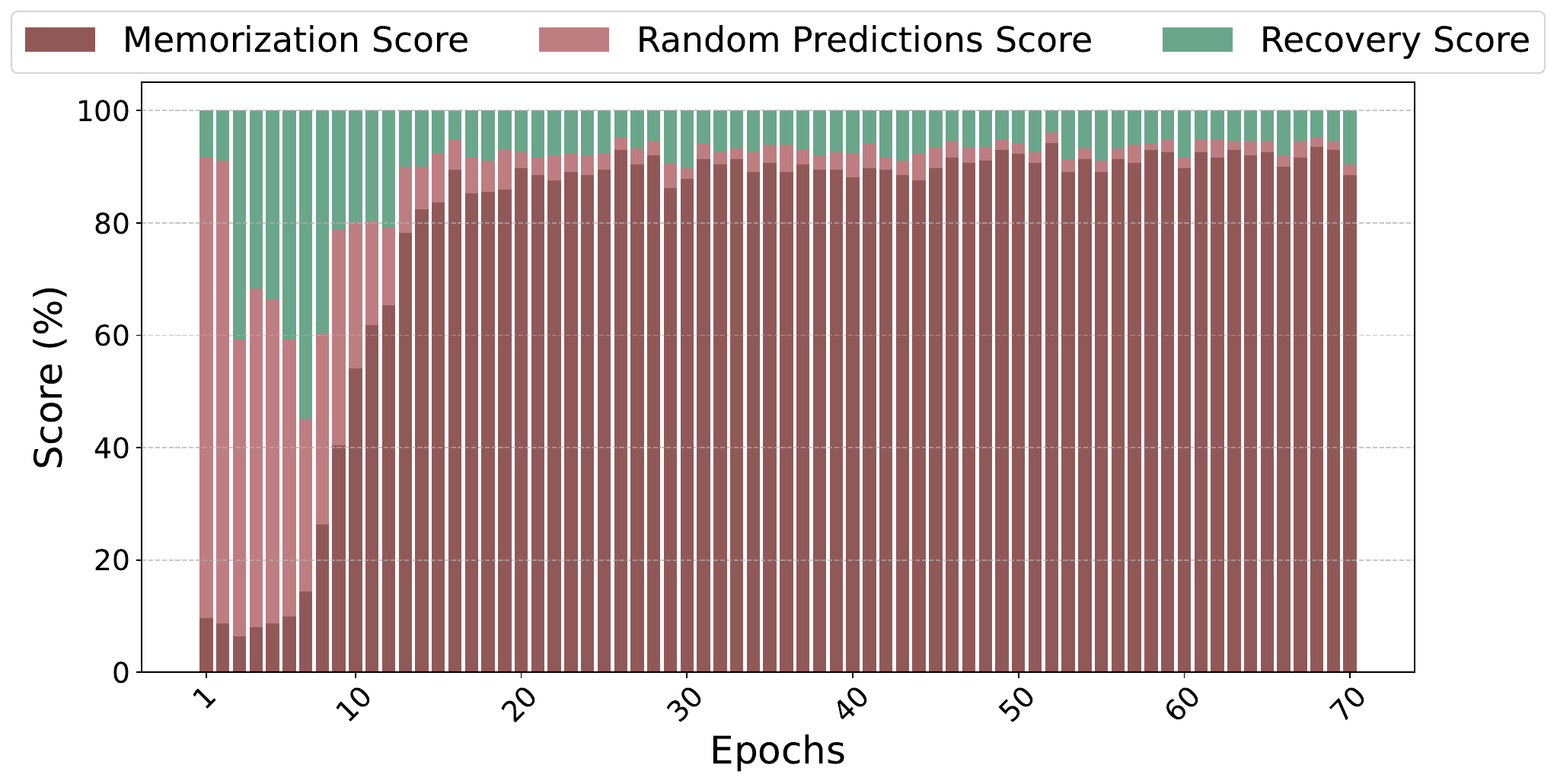}
        \caption{Memorization, Recovery and Random Predictions over epochs for Pre-LN Model (ViT-S)}
        \label{fig:pre_ln_mem_epochs_vit_small}
    \end{subfigure}
    \hspace{5pt}
    \begin{subfigure}[t]{0.25\textwidth}
        \centering
        \includegraphics[width=\textwidth]{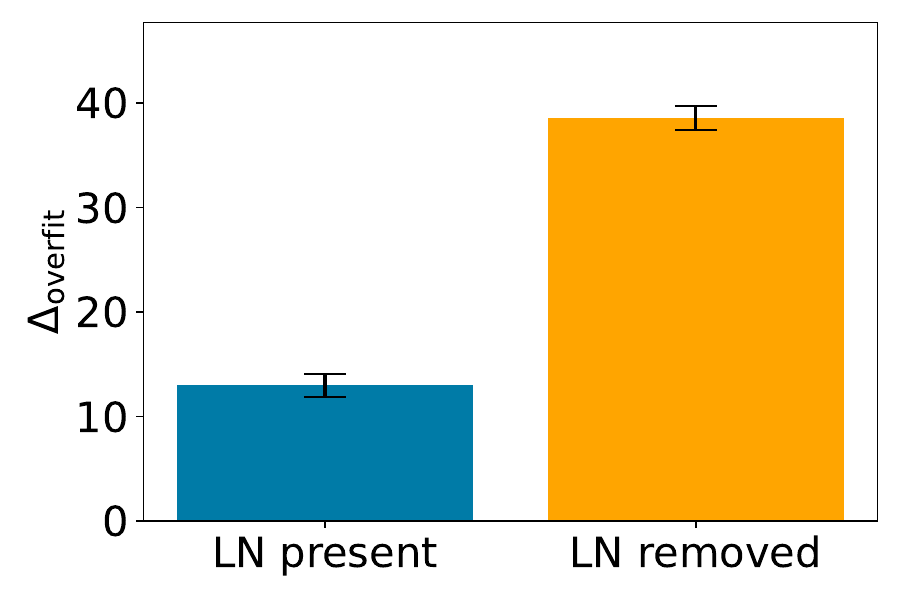}
        \caption{Overfitting gap for Pre-LN Model (ViT-S)}
        \label{fig:pre_ln_overfit_gap_vit_small}
    \end{subfigure}
    \caption{\textbf{LN removal destabilizes learning in Pre-LN model - ViT-S, NICO++ Dataset:} LN removal critically affects learning while memorization still persists in ViT-S. This further exacerbates overfitting seen by increasing train-test accuracy gap when LN is removed, due to the drop in test-accuracy.}
    \label{fig:learning_destabilized_vit_small}

\end{figure}

\begin{figure}[htbp]
    \centering
    \begin{subfigure}[t]{0.35\textwidth}
        \centering
        \includegraphics[width=\textwidth]{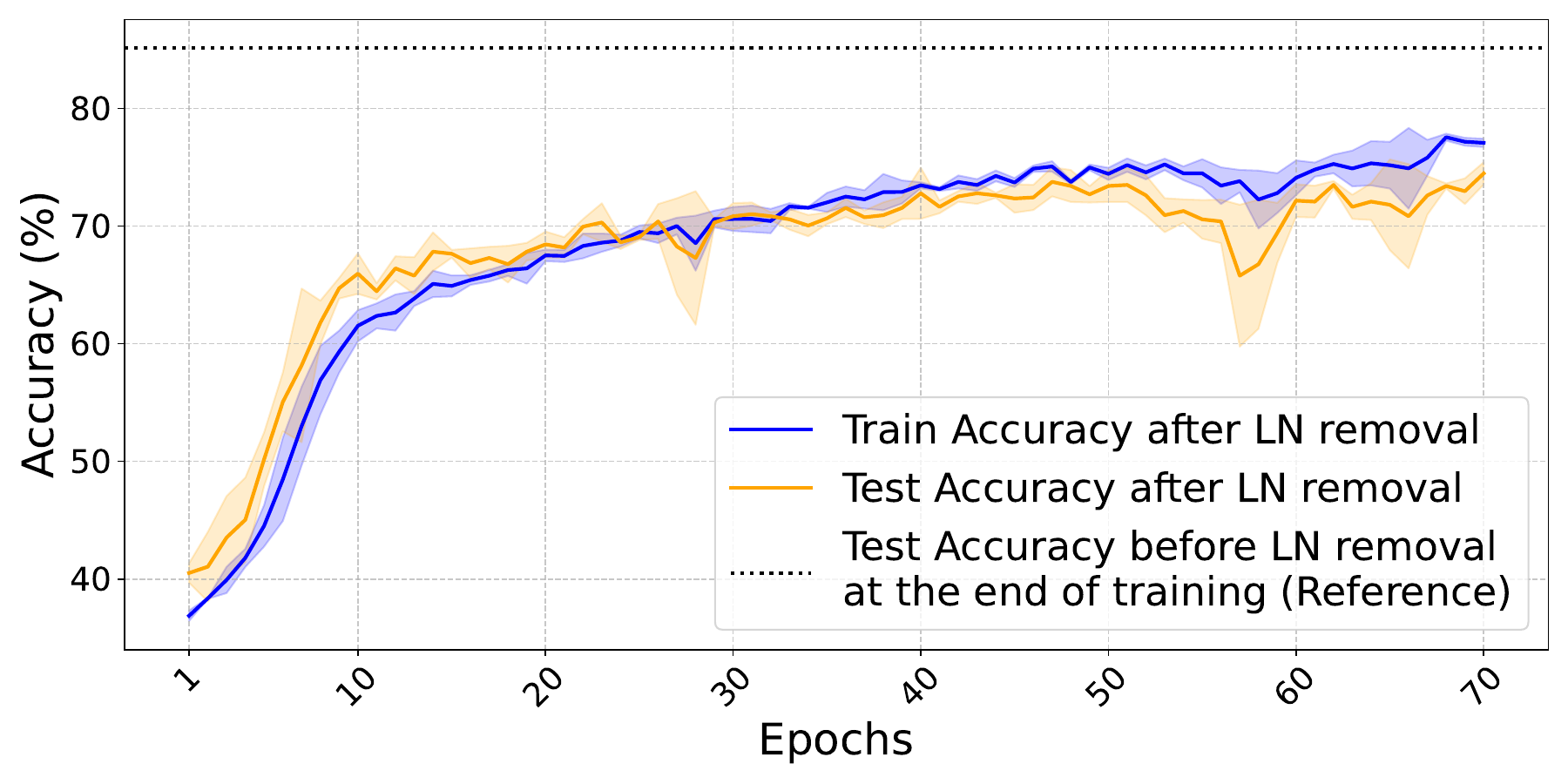}
        \caption{Learning (Test) accuracy over epochs for Pre-LN Model (RoBERTa-PreLayerNorm)}
        \label{fig:pre_ln_learning_epochs_roberta_preln}
    \end{subfigure}
    \hspace{5pt}
    \begin{subfigure}[t]{0.35\textwidth}
        \centering
        \includegraphics[width=\textwidth]{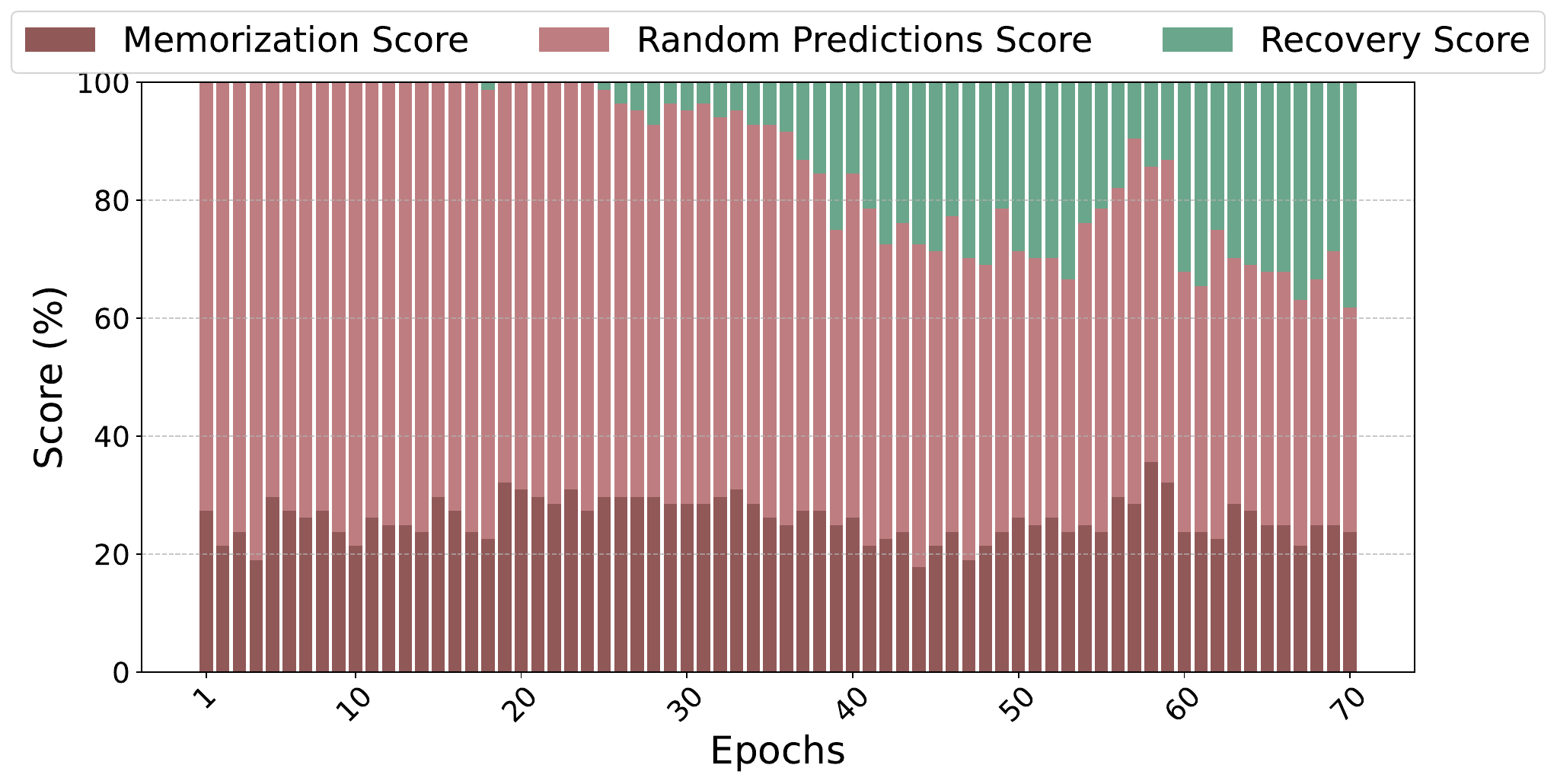}
        \caption{Memorization, Recovery and Random Predictions over epochs for Pre-LN Model (RoBERTa-PreLayerNorm)}
        \label{fig:pre_ln_mem_epochs_roberta_preln}
    \end{subfigure}
    \hspace{5pt}
    \begin{subfigure}[t]{0.25\textwidth}
        \centering
        \includegraphics[width=\textwidth]{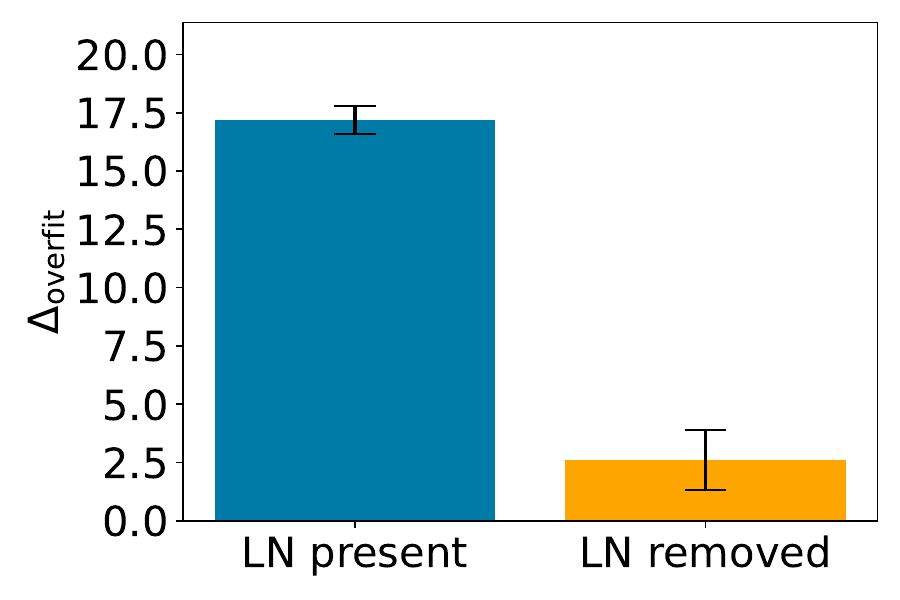}
        \caption{Overfitting gap for Pre-LN Model (RoBERTa-PreLayerNorm)}
        \label{fig:pre_ln_overfit_gap_roberta_preln}
    \end{subfigure}
    \caption{\textbf{LN removal destabilizes learning in Pre-LN model - RoBERTa-PreLayerNorm, TweetTopic Dataset:} LN removal critically affects learning while memorization still persists in RoBERTa-PreLayerNorm. For RoBERTa-PreLayerNorm, the overfitting gap decreased after LN removal, because the model could not even stabilize during training due to the destabilization of learning. Hence both train and test accuracies remain low and comparable. However, the learning accuracy still remains low when LN is absent in comparison to when LN is present, and struggles with high memorization and random predictions (\textcolor{BrickRed}{red}-color family bars).}
    \label{fig:learning_destabilized_roberta_preln}

\end{figure}

\begin{figure}[htbp]
    \centering
    \begin{subfigure}[t]{0.35\textwidth}
        \centering
        \includegraphics[width=\textwidth]{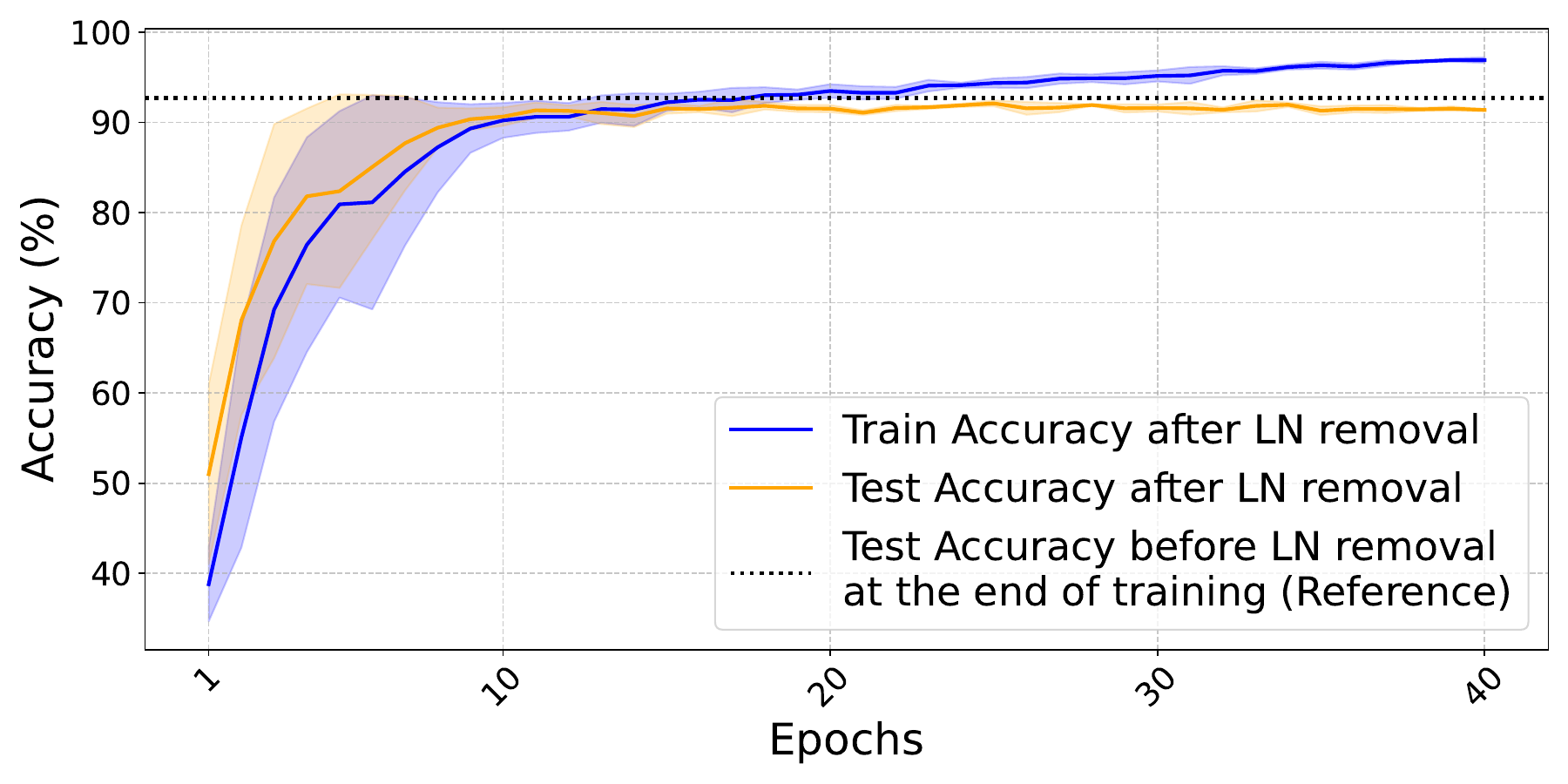}
        \caption{Learning (Test) accuracy over epochs for Post-LN Model (BERT)}
        \label{fig:post_ln_learning_epochs_bert}
    \end{subfigure}
    \hspace{5pt}
    \begin{subfigure}[t]{0.35\textwidth}
        \centering
        \includegraphics[width=\textwidth]{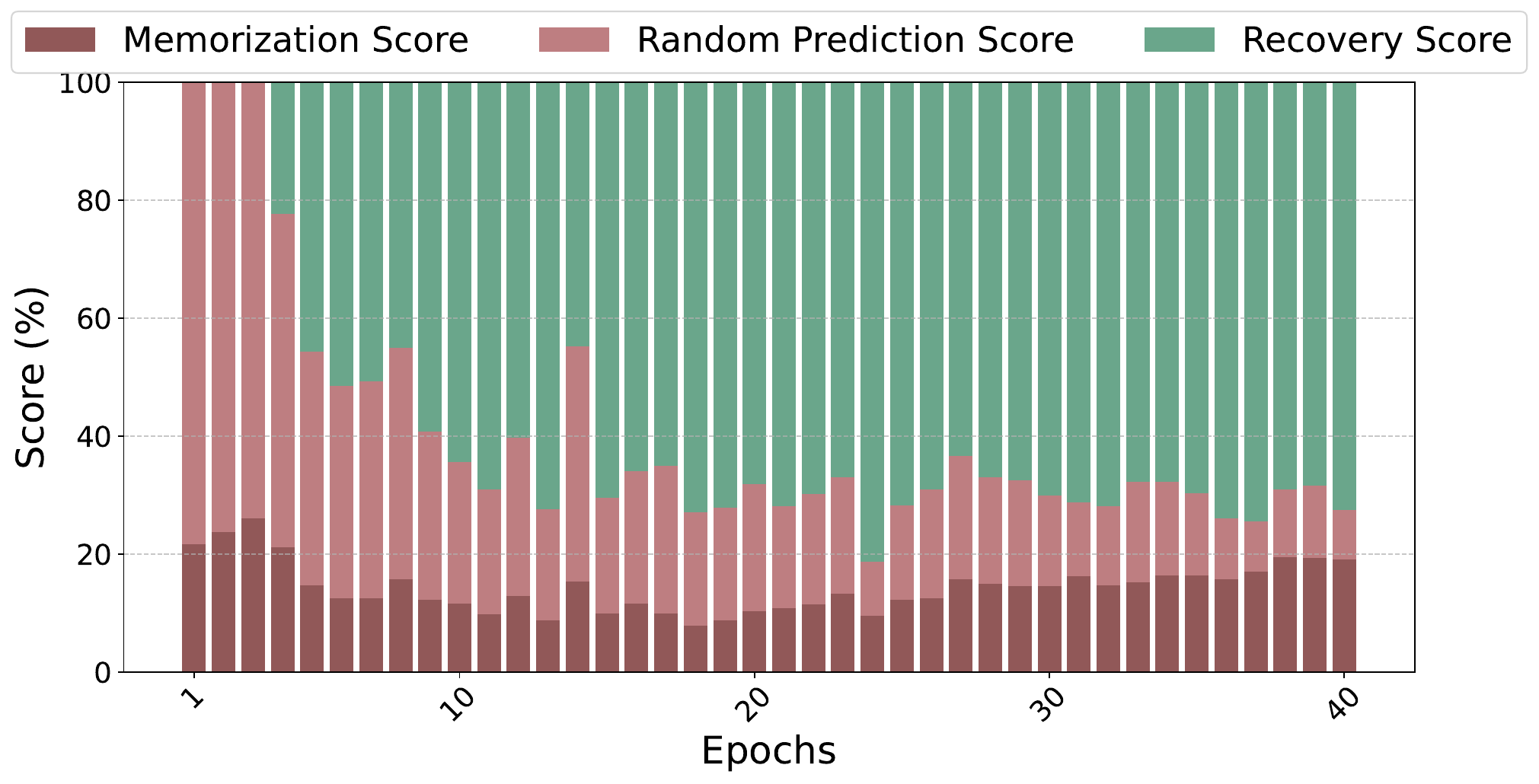}
        \caption{Memorization, Recovery and Random Predictions over epochs for Post-LN Model (BERT)}
        \label{fig:post_ln_mem_epochs_bert}
    \end{subfigure}
    \hspace{5pt}
    \begin{subfigure}[t]{0.25\textwidth}
        \centering
        \includegraphics[width=\textwidth]{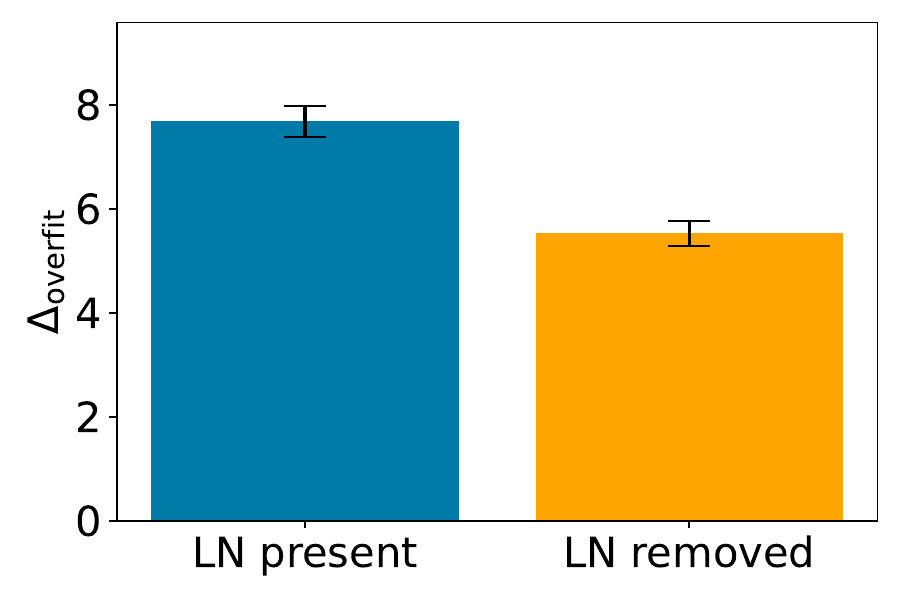}
        \caption{Overfitting gap for Post-LN Model (BERT)}
        \label{fig:post_ln_overfit_gap_bert}
    \end{subfigure}
    \caption{\textbf{LN removal suppresses memorization \& facilitates true label recovery in Post-LN model - BERT, Emotions Dataset:} LN removal in BERT suppresses memorization and facilitates true label recovery, while keeping learning intact. This further reduces overfitting seen by decreasing train-test accuracy gap when LN is removed.}
    \label{fig:mem_suppressed_bert}
\end{figure}

\begin{figure}[htbp]
    \centering
    \begin{subfigure}[t]{0.35\textwidth}
        \centering
        \includegraphics[width=\textwidth]{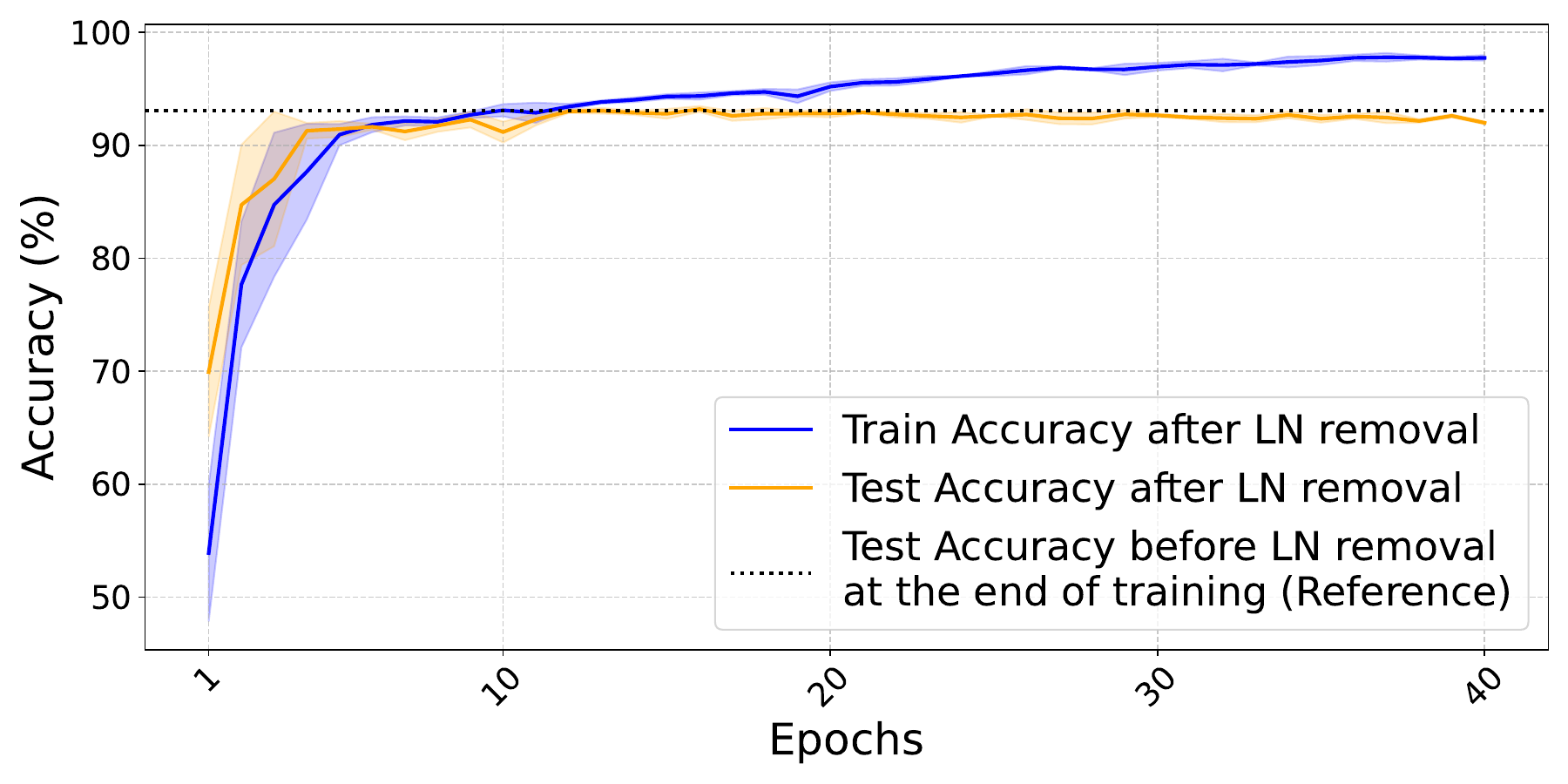}
        \caption{Learning (Test) accuracy over epochs for Post-LN Model (DeBERTa)}
        \label{fig:post_ln_learning_epochs_deberta}
    \end{subfigure}
    \hspace{5pt}
    \begin{subfigure}[t]{0.35\textwidth}
        \centering
        \includegraphics[width=\textwidth]{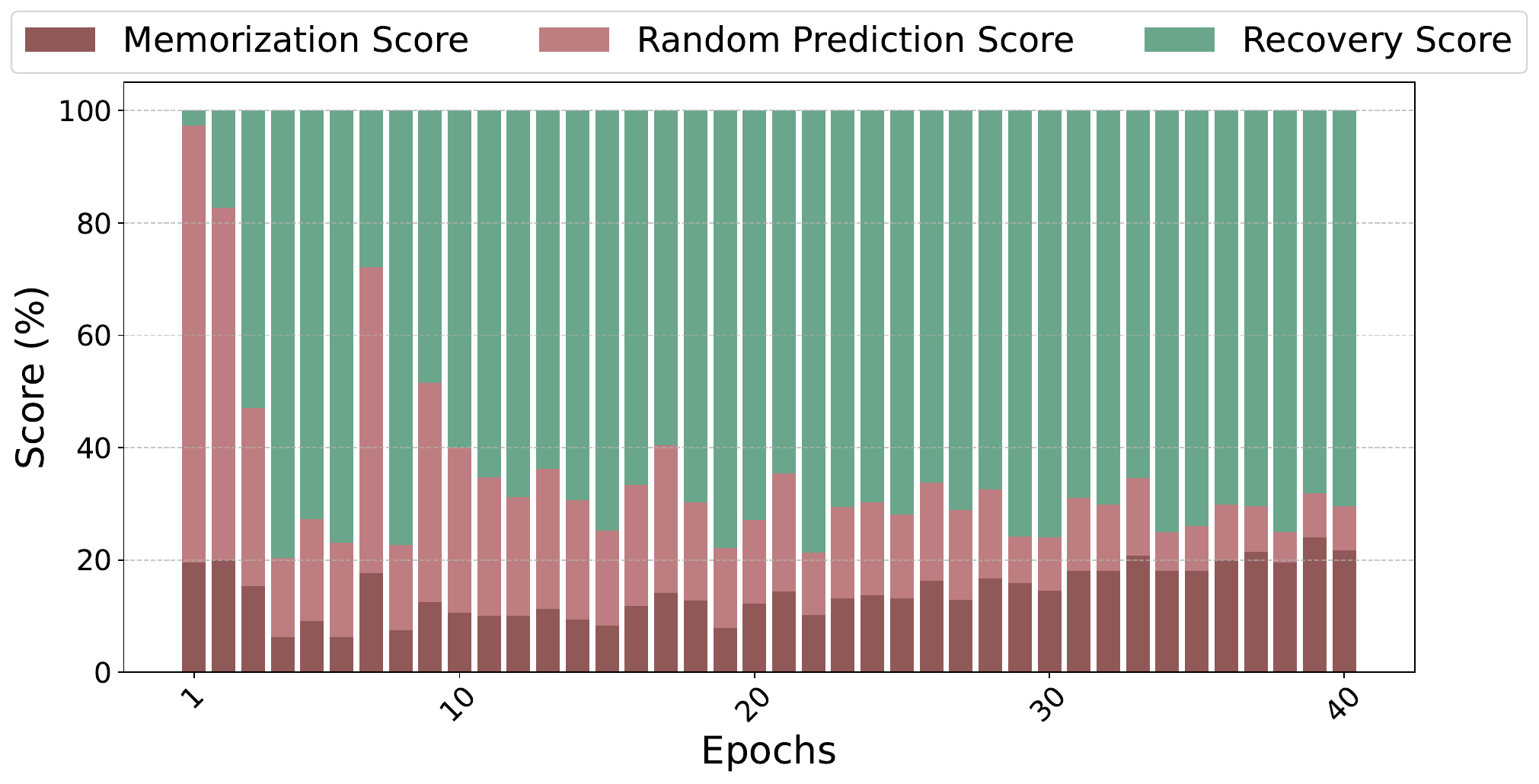}
        \caption{Memorization, Recovery and Random Predictions over epochs for Post-LN Model (DeBERTa)}
        \label{fig:post_ln_mem_epochs_deberta}
    \end{subfigure}
    \hspace{5pt}
    \begin{subfigure}[t]{0.25\textwidth}
        \centering
        \includegraphics[width=\textwidth]{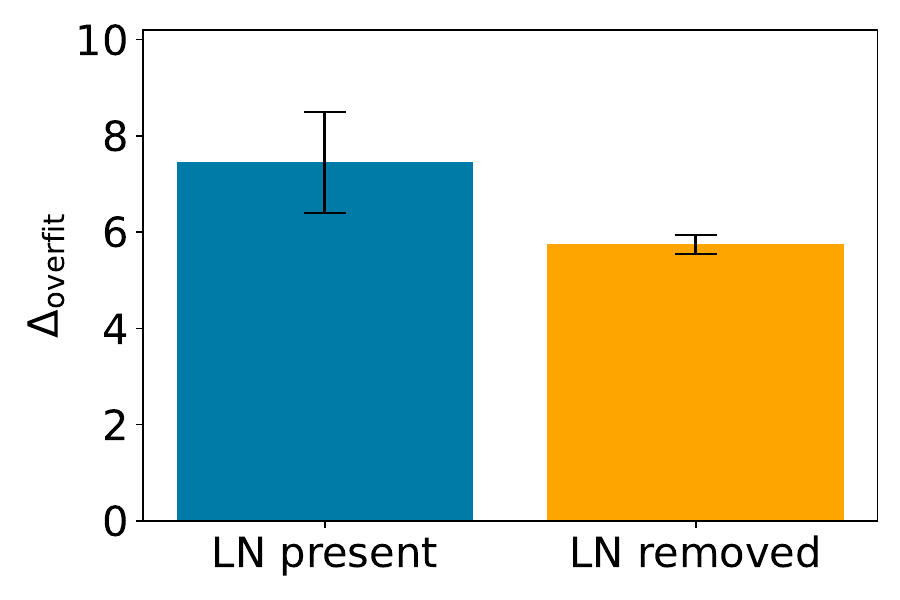}
        \caption{Overfitting gap for Post-LN Model (DeBERTa)}
        \label{fig:post_ln_overfit_gap_deberta}
    \end{subfigure}
    \caption{\textbf{LN removal suppresses memorization \& facilitates true label recovery in Post-LN model - DeBERTa, Emotions Dataset:} LN removal in DeBERTa suppresses memorization and facilitates true label recovery, while keeping learning intact. This further reduces overfitting seen by decreasing train-test accuracy gap when LN is removed.}
    \label{fig:mem_suppressed_deberta}
\end{figure}

\begin{figure}[htbp]
    \centering
    \begin{subfigure}[t]{0.35\textwidth}
        \centering
        \includegraphics[width=\textwidth]{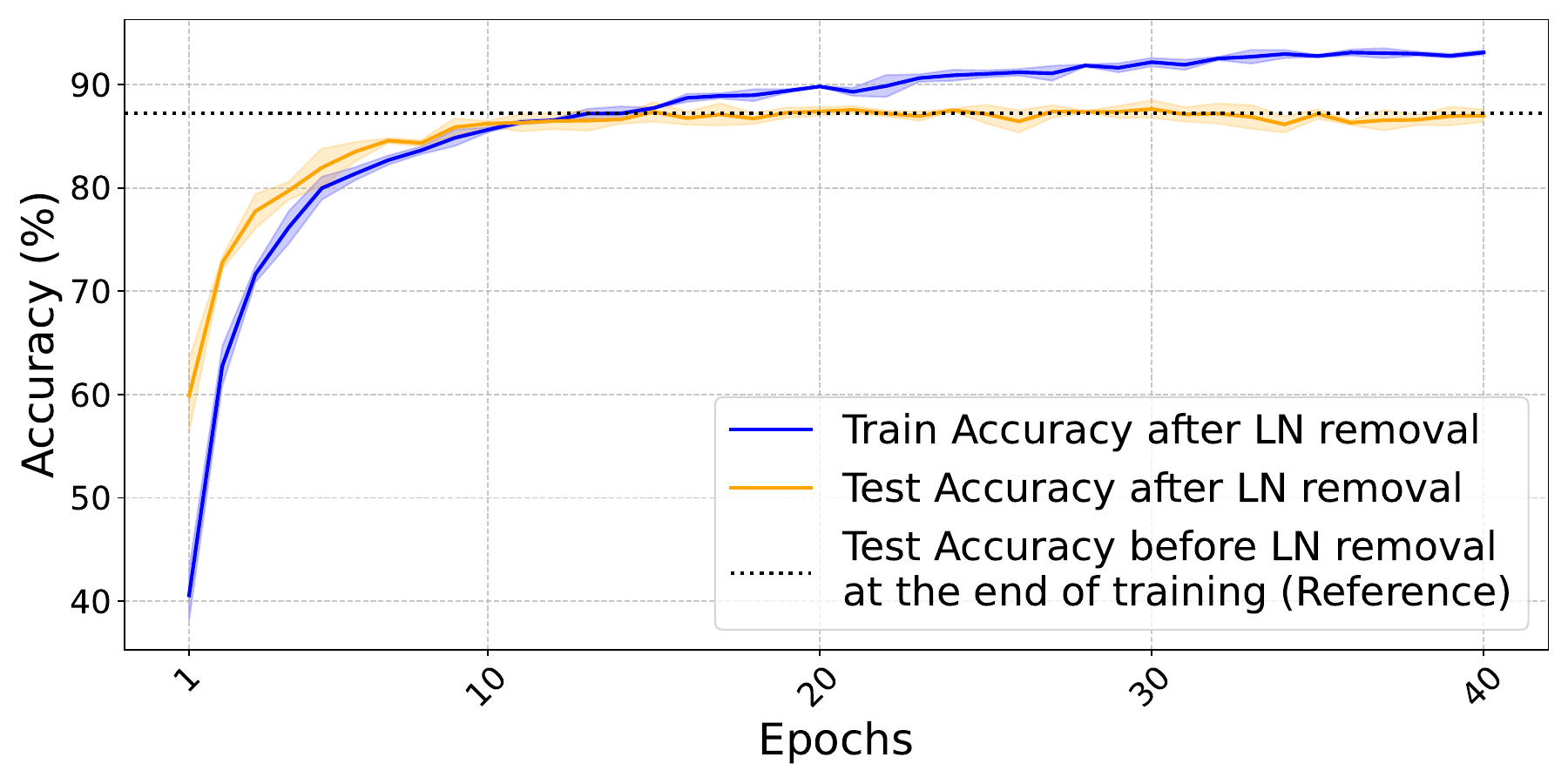}
        \caption{Learning (Test) accuracy over epochs for Post-LN Model (Longformer)}
        \label{fig:post_ln_learning_epochs_longformer}
    \end{subfigure}
    \hspace{5pt}
    \begin{subfigure}[t]{0.35\textwidth}
        \centering
        \includegraphics[width=\textwidth]{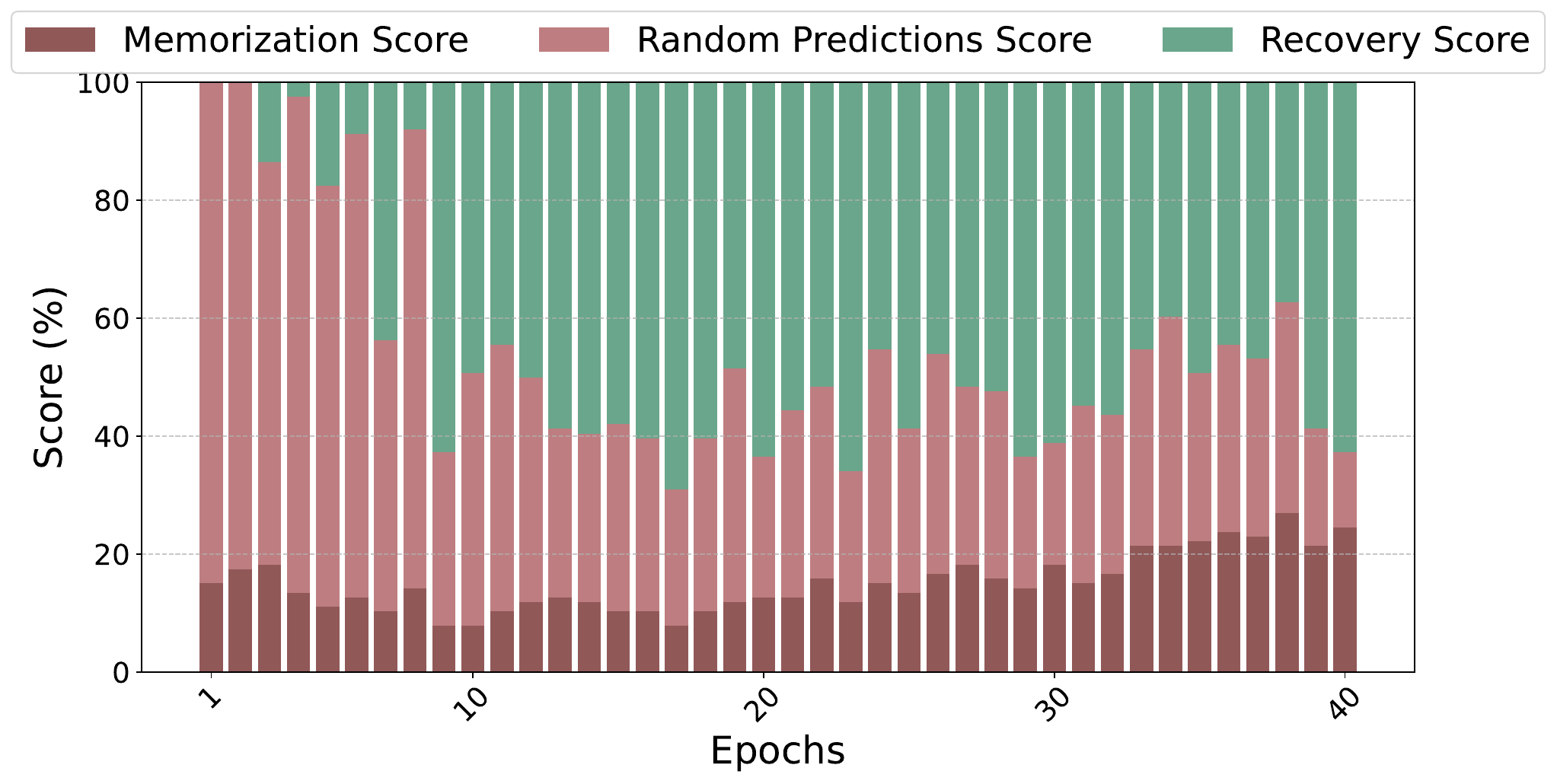}
        \caption{Memorization, Recovery and Random Predictions over epochs for Post-LN Model (Longformer)}
        \label{fig:post_ln_mem_epochs_longformer}
    \end{subfigure}
    \hspace{5pt}
    \begin{subfigure}[t]{0.25\textwidth}
        \centering
        \includegraphics[width=\textwidth]{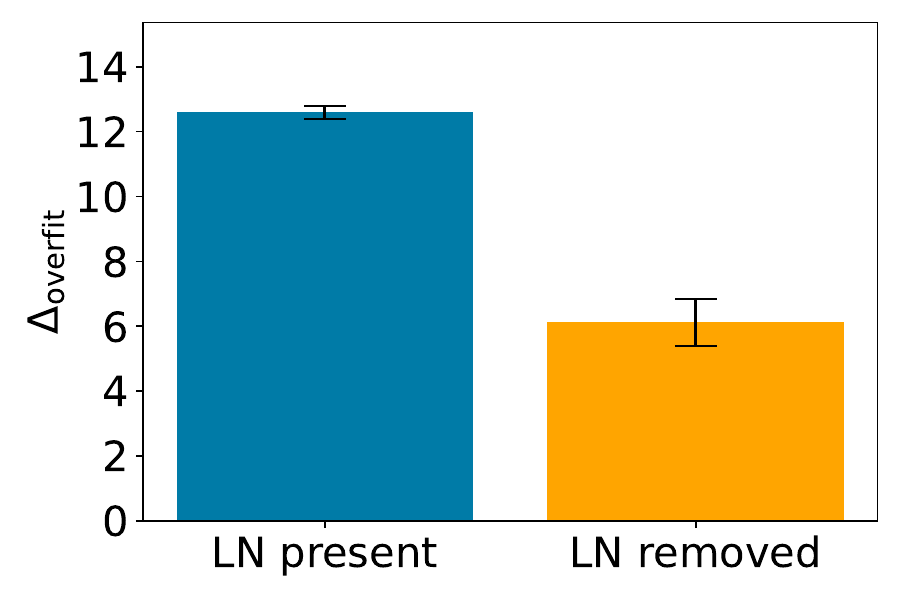}
        \caption{Overfitting gap for Post-LN Model (Longformer)}
        \label{fig:post_ln_overfit_gap_longformer}
    \end{subfigure}
    \caption{\textbf{LN removal suppresses memorization \& facilitates true label recovery in Post-LN model - Longformer, News Dataset:} LN removal in Longformer suppresses memorization and facilitates true label recovery, while keeping learning intact. This further reduces overfitting seen by decreasing train-test accuracy gap when LN is removed.}
    \label{fig:mem_suppressed_longformer}
\end{figure}

\begin{figure}[htbp]
    \centering
    \begin{subfigure}[t]{0.35\textwidth}
        \centering
        \includegraphics[width=\textwidth]{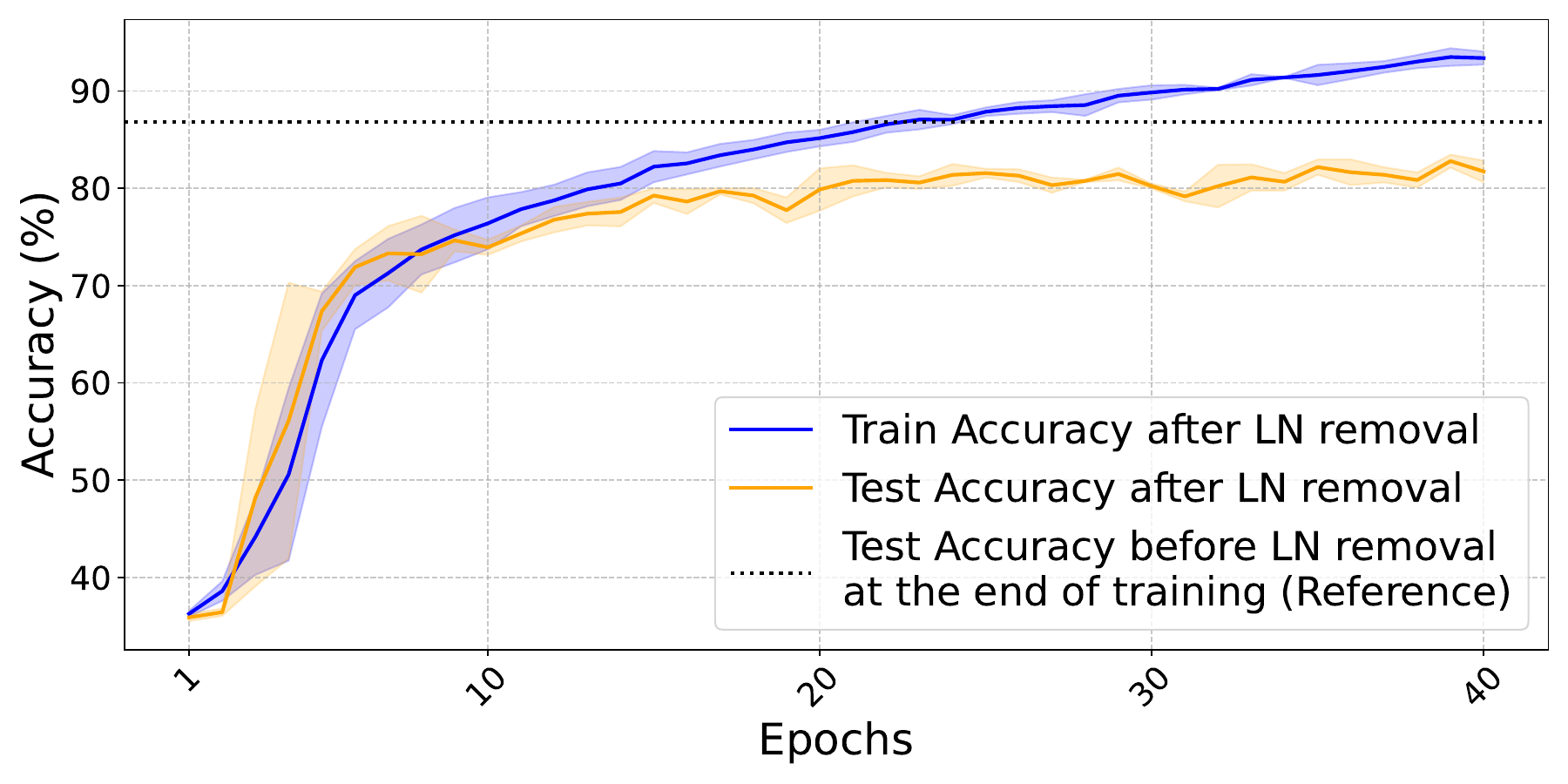}
        \caption{Learning (Test) accuracy over epochs for Post-LN Model (RoBERTa)}
        \label{fig:post_ln_learning_epochs_roberta}
    \end{subfigure}
    \hspace{5pt}
    \begin{subfigure}[t]{0.35\textwidth}
        \centering
        \includegraphics[width=\textwidth]{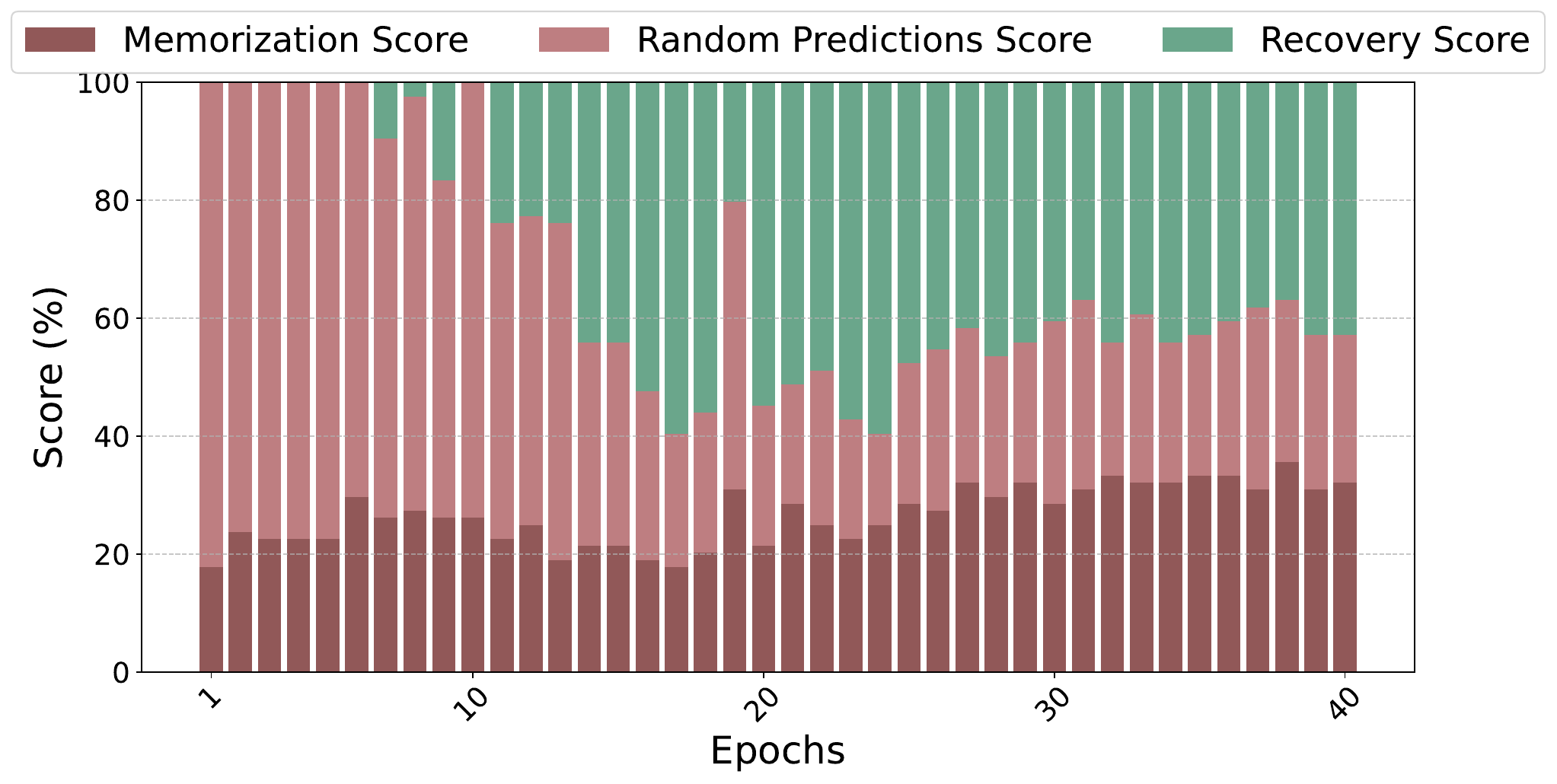}
        \caption{Memorization, Recovery and Random Predictions over epochs for Post-LN Model (RoBERTa)}
        \label{fig:post_ln_mem_epochs_roberta}
    \end{subfigure}
    \hspace{5pt}
    \begin{subfigure}[t]{0.25\textwidth}
        \centering
        \includegraphics[width=\textwidth]{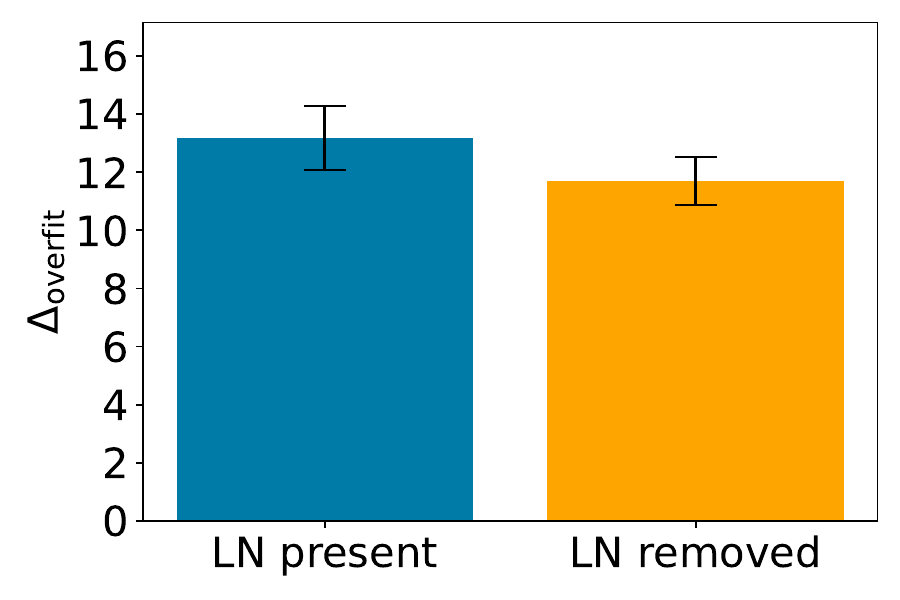}
        \caption{Overfitting gap for Post-LN Model (RoBERTa)}
        \label{fig:post_ln_overfit_gap_roberta}
    \end{subfigure}
    \caption{\textbf{LN removal suppresses memorization \& facilitates true label recovery in Post-LN model - RoBERTa, TweetTopic Dataset:} LN removal in Longformer suppresses memorization and facilitates true label recovery, while minimal drop in learning. This further reduces overfitting seen by decreasing train-test accuracy gap when LN is removed. \emph{Note: For RoBERTa we see a slight drop in learning accuracy which is much lower in comparison to the huge drop which happens in GPT2 setup in Fig.~\ref{fig:learning_destabilized_gpt2}. Furthermore, memorization is still suppressed in comparison to GPT2 where memorization still persisted.}}
    \label{fig:mem_suppressed_roberta}
\end{figure}

\begin{figure}[htbp]
    \centering
    \begin{subfigure}[t]{0.35\textwidth}
        \centering
        \includegraphics[width=\textwidth]{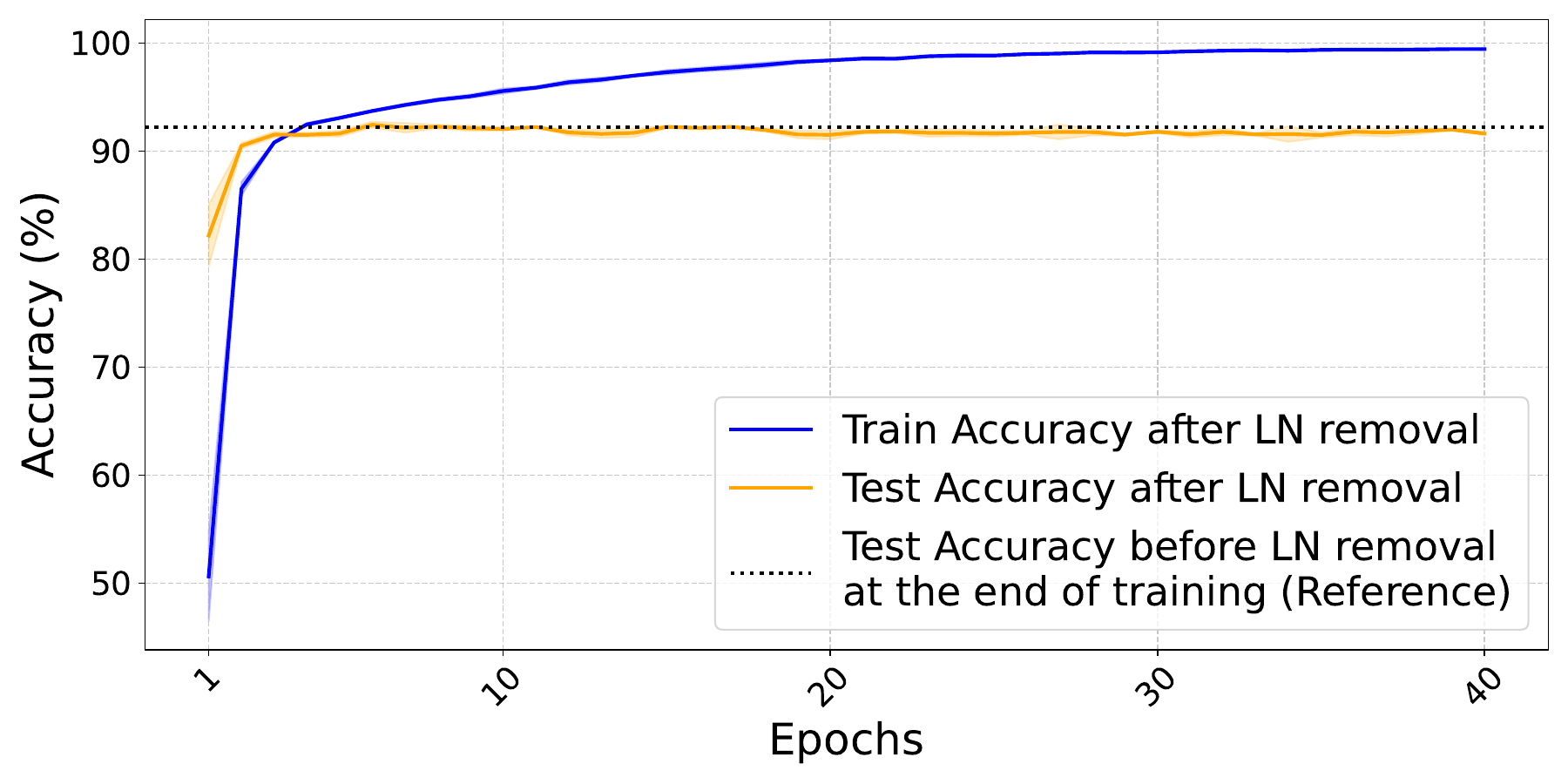}
        \caption{Learning (Test) accuracy over epochs for Post-LN Model (DistilBERT)}
        \label{fig:post_ln_learning_epochs_distilbert}
    \end{subfigure}
    \hspace{5pt}
    \begin{subfigure}[t]{0.35\textwidth}
        \centering
        \includegraphics[width=\textwidth]{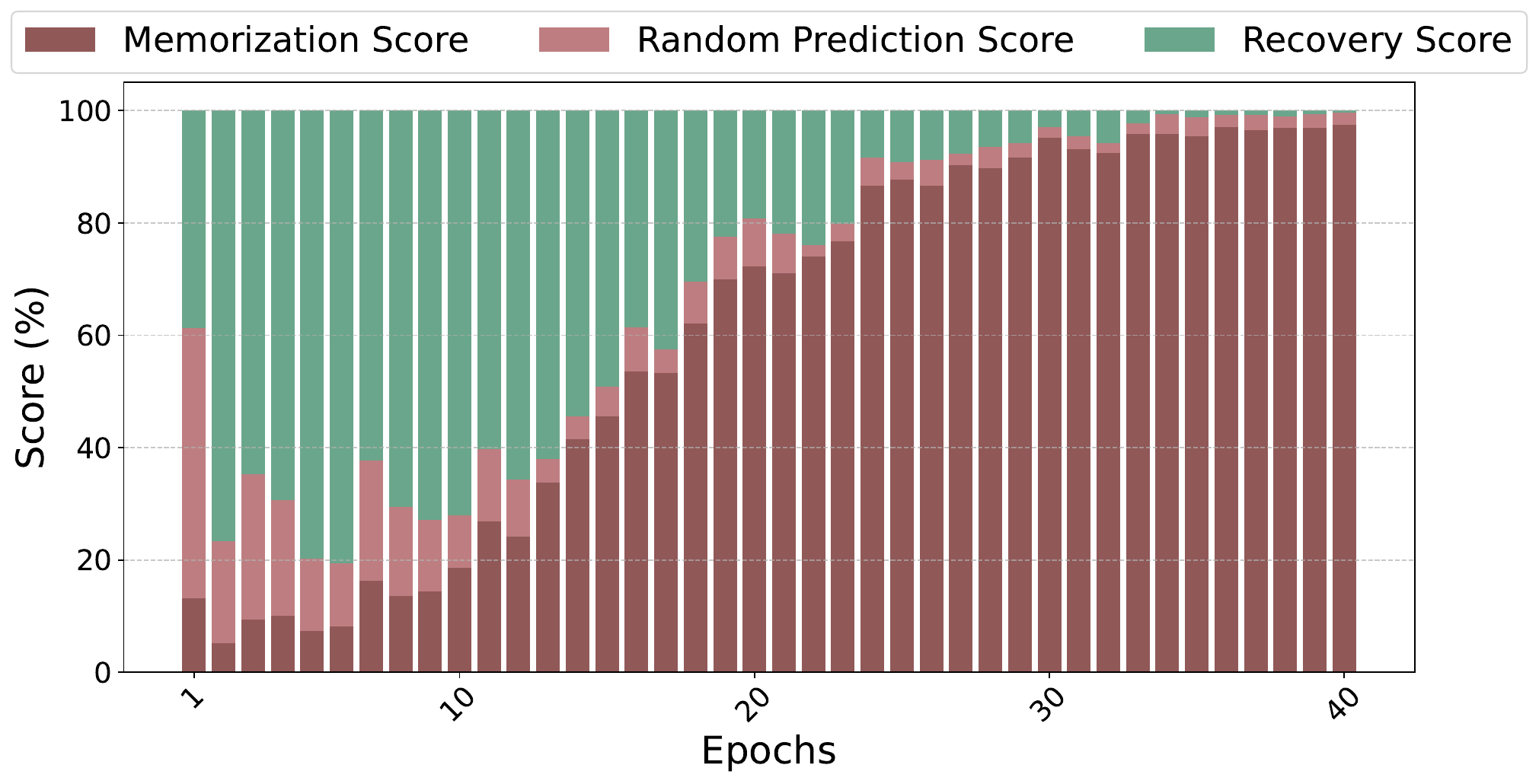}
        \caption{Memorization, Recovery and Random Predictions over epochs for Post-LN Model (DistilBERT)}
        \label{fig:post_ln_mem_epochs_distilbert}
    \end{subfigure}
    \hspace{5pt}
    \begin{subfigure}[t]{0.25\textwidth}
        \centering
        \includegraphics[width=\textwidth]{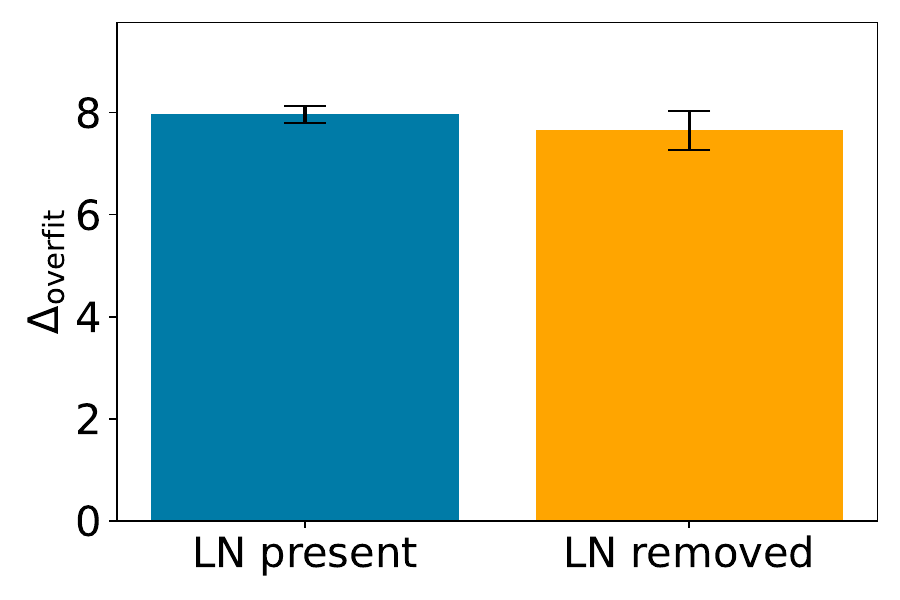}
        \caption{Overfitting gap for Post-LN Model (DistilBERT)}
        \label{fig:post_ln_overfit_gap_distilbert}
    \end{subfigure}
    \caption{\textbf{LN removal does not suppress memorization in DistilBERT (Emotions Dataset) but does not affect learning:} LN removal in DistilBERT does not suppress memorization (but it does not impact learning) as we observe for all other 5 Post-LN models. We think other components in the Transformer architecture might have a more profound impact on memorization in DistilBERT. But, since this work is primarily about LN impact, we refrain from dealing with other components, making it an interesting future work.}
    \label{fig:mem_suppressed_distilbert}
\end{figure}

The results are verified against Pre-LN models - GPTNeo, GPT2, ViT-B, DeiT, ViT-S, RoBERTa-PreLayernorm and Post-LN models - BERT, DeBERTa, Longformer, RoBERTa, DistilBERT spanning across multiple language and vision datasets - Emotions, News, Tweets, CIFAR10, NICO++, UTK-Face, are provided in Figs.~\ref{fig:learning_destabilized_gptneo}, \ref{fig:learning_destabilized_gpt2}, \ref{fig:learning_destabilized_vit_base}, \ref{fig:learning_destabilized_deit}, \ref{fig:learning_destabilized_vit_small}, 
\ref{fig:learning_destabilized_roberta_preln}, \ref{fig:mem_suppressed_bert}, \ref{fig:mem_suppressed_deberta}, \ref{fig:mem_suppressed_longformer},\ref{fig:mem_suppressed_roberta}. and
\ref{fig:mem_suppressed_distilbert}.

\subsubsection{Pre-LN models - Learning Destabilized}


\clearpage

\subsubsection{Post-LN models - Suppression of Memorization \& True Labels Recovery}

\newpage
\subsection{Significance of Early Layers LN} \label{sec:significance_early_layers_ln_appendix}
In this section, we illustrate the results corresponding to the significance of Early Layers LN in impacting learning and suppressing memorization for remaining Pre- and Post-LN models, respectively, across multiple datasets.

\begin{figure}[htbp]
    \centering
    \includegraphics[width=\textwidth]{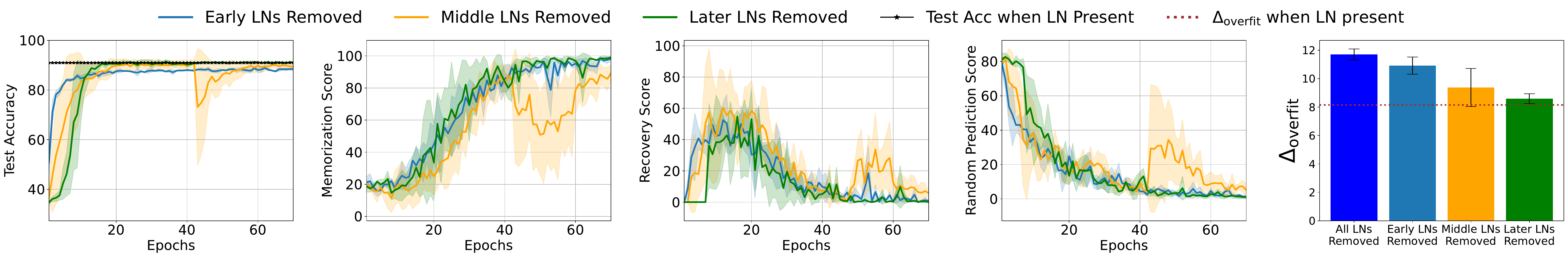}
    \caption{\textbf{Pivotal impact of early LNs for learning in Pre-LN model (GPTNeo, Emotions Dataset).} We can clearly observe the impact of early layers LN removal on destabilizing learning in GPTNeo, accompanied with higher train-test-accuracy gap, $\Delta_{\text{overfit}}^{\text{Pre, early}}$, than for later layers, $\Delta_{\text{overfit}}^{\text{Pre, later}}$, and poor memorization suppression.}
    \label{fig:pre-ln-early-middle-later_gpt_neo}
\end{figure}

\begin{figure}[htbp]
    \centering
    \includegraphics[width=\textwidth]{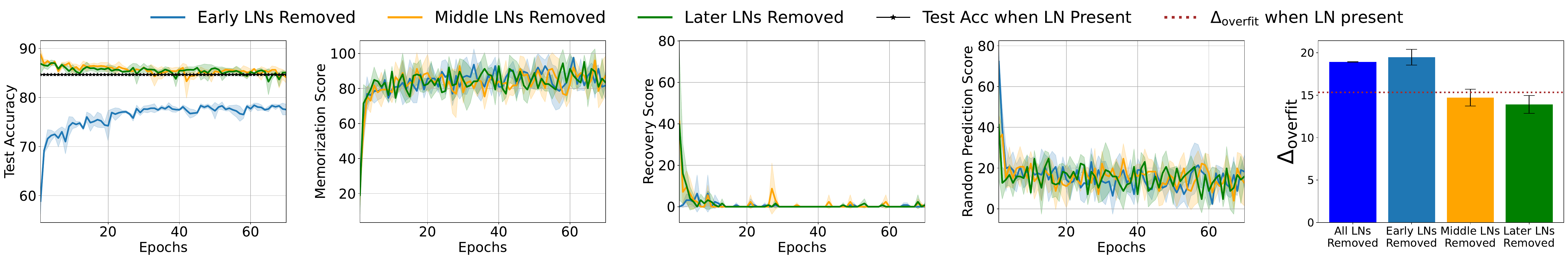}
    \caption{\textbf{Pivotal impact of early LNs for learning in Pre-LN model (Qwen2, News Dataset).} We can clearly observe the impact of early layers LN removal on destabilizing learning in Qwen2, accompanied with higher train-test-accuracy gap, $\Delta_{\text{overfit}}^{\text{Pre, early}}$, than for later layers, $\Delta_{\text{overfit}}^{\text{Pre, later}}$, and poor memorization suppression.}
    \label{fig:pre-ln-early-middle-later_qwen2}
\end{figure}

\begin{figure}[htbp]
    \centering
    \includegraphics[width=\textwidth]{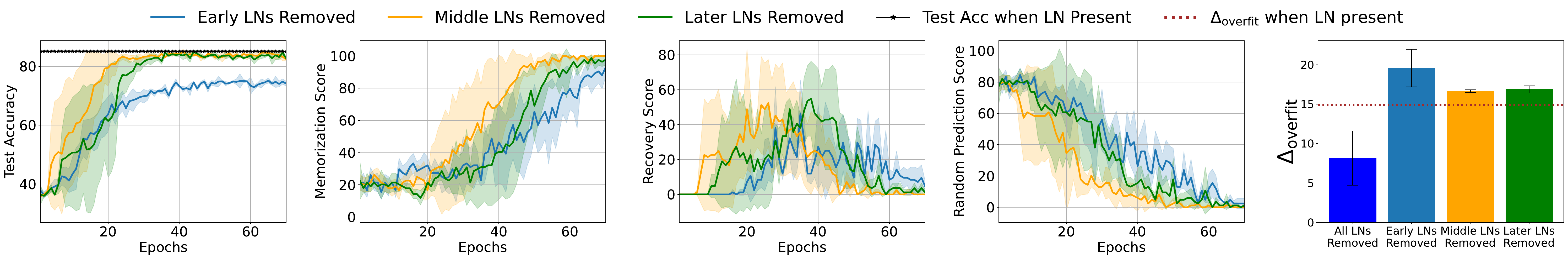}
    \caption{\textbf{Pivotal impact of early LNs for learning in Pre-LN model (GPT2, TweetTopic Dataset).} We can clearly observe the impact of early layers LN removal on destabilizing learning in GPT2, accompanied with higher train-test-accuracy gap, $\Delta_{\text{overfit}}^{\text{Pre, early}}$, than for later layers, $\Delta_{\text{overfit}}^{\text{Pre, later}}$, and poor memorization suppression. \emph{Note: When all layers LNs are removed, $\Delta_{\text{overfit}}^{\text{Pre, all}}$ is low because when we removed all LNs from GPT2 then the model could not stabilize in training, reaching very low train accuracy, similar train-test accuracy as seen in Fig.~\ref{fig:pre_ln_learning_epochs_gpt2}, hence low $\Delta_{\text{overfit}}^{\text{Pre, all}}$. However, when we just removed early LNs, the model is able to converge to a sufficiently high train accuracy, but the learning is impacted much severely in comparison to later layers. }}
    \label{fig:pre-ln-early-middle-later_gpt2}
\end{figure}

\begin{figure}[htbp]
    \centering
    \includegraphics[width=\textwidth]{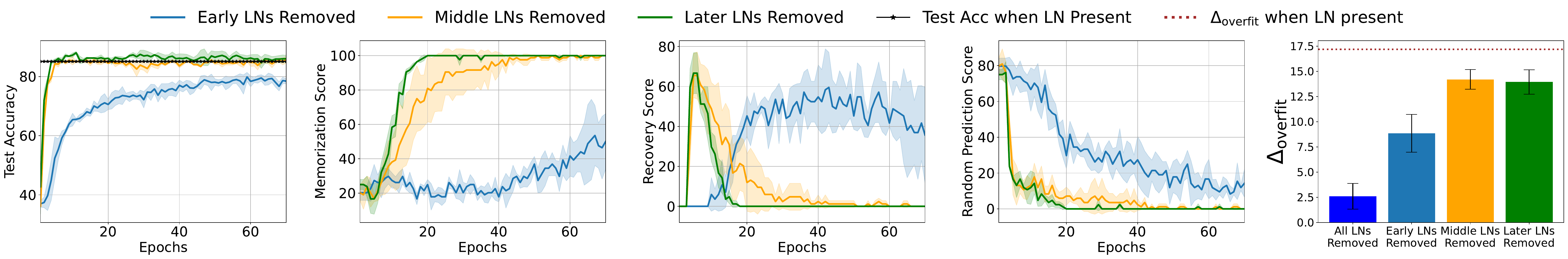}
    \caption{\textbf{Pivotal impact of early LNs for learning in Pre-LN model (RoBERTa-PreLayerNorm, TweetTopic Dataset).} We observe that for RoBERTa-PreLayerNorm, removing early layers LN impacts learning, however, memorization also seems to get suppressed, when compared with removing all LNs where we observe persistent memorization and destabilized learning (Fig.~\ref{fig:pre_ln_learning_epochs_roberta_preln}). Despite this unique trend of RoBERTa-PreLayernorm (not following the consistent trend just as in the other 6 Pre-LN models), early layers LN still remain the most significant.}
    \label{fig:pre-ln-early-middle-later_roberta_preln}
\end{figure}

\begin{figure}[htbp]
    \centering
    \includegraphics[width=\textwidth]{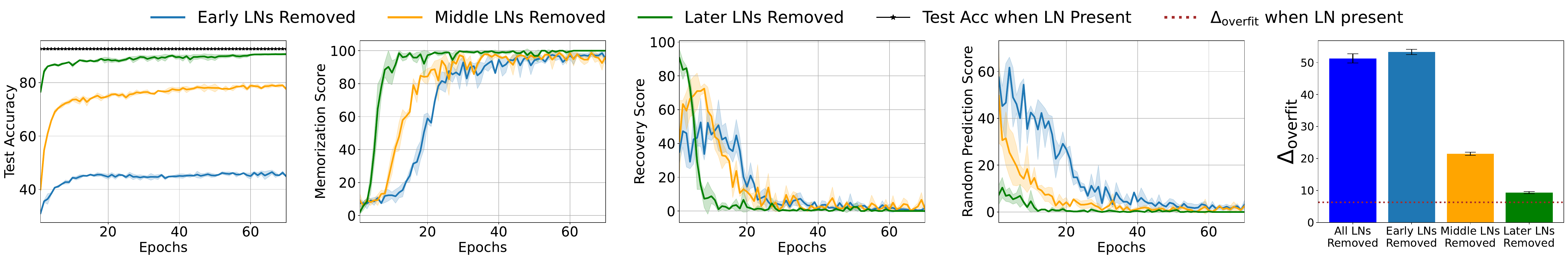}
    \caption{\textbf{Pivotal impact of early LNs for learning in Pre-LN model (ViT-B, CIFAR10 Dataset).} We can clearly observe the impact of early layers LN removal on destabilizing learning in ViT-B, accompanied with higher train-test-accuracy gap, $\Delta_{\text{overfit}}^{\text{Pre, early}}$, than for later layers, $\Delta_{\text{overfit}}^{\text{Pre, later}}$, and poor memorization suppression.}
    \label{fig:pre-ln-early-middle-later_vit_base}
\end{figure}

\begin{figure}[htbp]
    \centering
    \includegraphics[width=\textwidth]{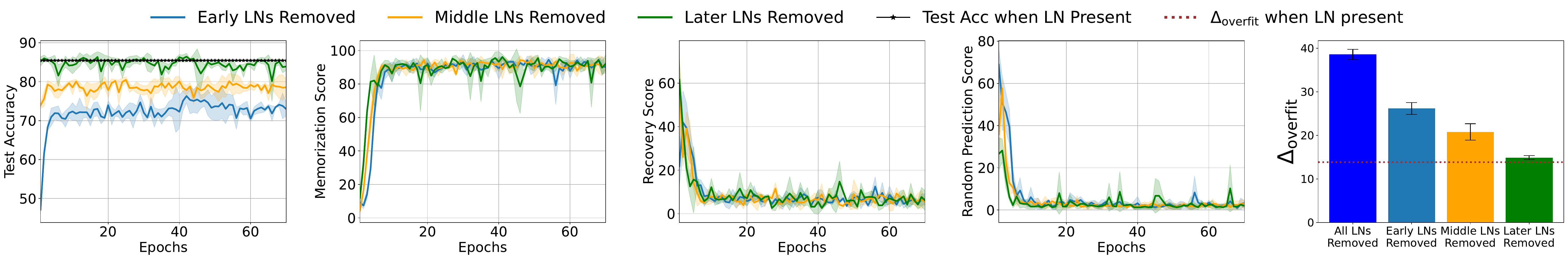}
    \caption{\textbf{Pivotal impact of early LNs for learning in Pre-LN model (CIFAR100, ViT-S Dataset).} We can clearly observe the impact of early layers LN removal on destabilizing learning in ViT-S, accompanied with higher train-test-accuracy gap, $\Delta_{\text{overfit}}^{\text{Pre, early}}$, than for later layers, $\Delta_{\text{overfit}}^{\text{Pre, later}}$, and poor memorization suppression.}
    \label{fig:pre-ln-early-middle-later_vit_small}
\end{figure}

\begin{figure}[htbp]
    \centering
    \includegraphics[width=\textwidth]{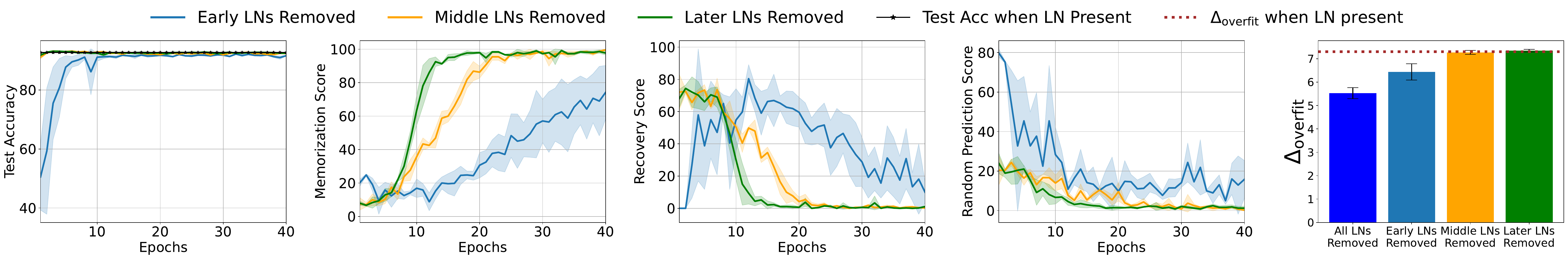}
    \caption{\textbf{Pivotal impact of early LNs on memorization in Post-LN model (BERT, Emotions Dataset).} We can clearly observe the impact of early layers LN removal on suppressing memorization \& achieving true label recovery in BERT, accompanied with higher train-test-accuracy gap, $\Delta_{\text{overfit}}^{\text{Pre, early}}$, than for later layers, $\Delta_{\text{overfit}}^{\text{Pre, later}}$, while learning being intact.}
    \label{fig:post-ln-early-middle-later_bert}
\end{figure}

\begin{figure}[htbp]
    \centering
    \includegraphics[width=\textwidth]{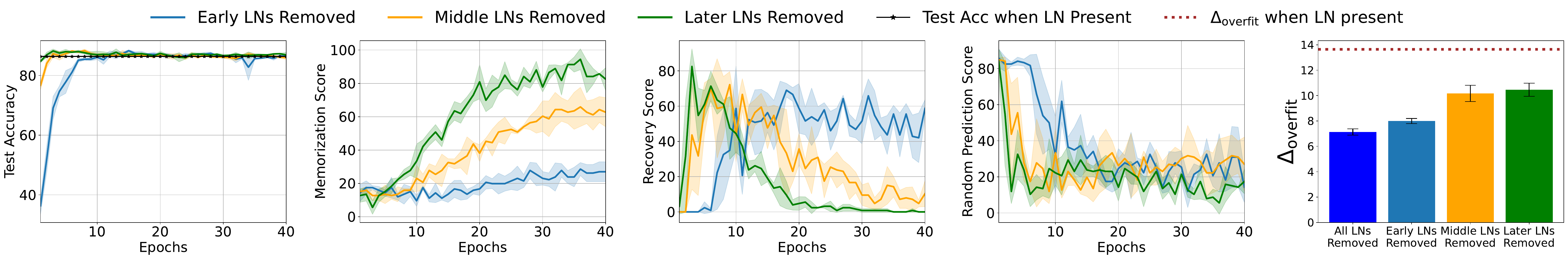}
    \caption{\textbf{Pivotal impact of early LNs on memorization in Post-LN model (ELECTRA, News Dataset).} We can clearly observe the impact of early layers LN removal on suppressing memorization \& achieving true label recovery in ELECTRA, accompanied with higher train-test-accuracy gap, $\Delta_{\text{overfit}}^{\text{Pre, early}}$, than for later layers, $\Delta_{\text{overfit}}^{\text{Pre, later}}$, while learning being intact.}
    \label{fig:post-ln-early-middle-later_electra}
\end{figure}

\begin{figure}[htbp]
    \centering
    \includegraphics[width=\textwidth]{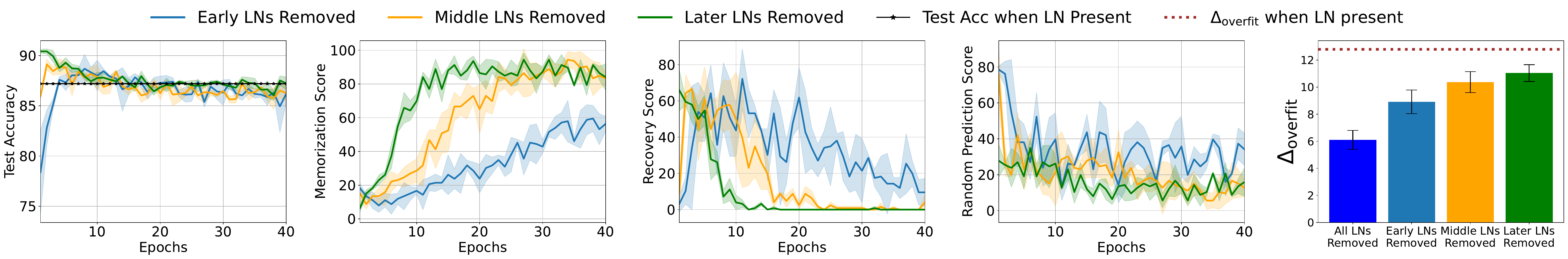}
    \caption{\textbf{Pivotal impact of early LNs on memorization in Post-LN model (Longformer, News Dataset).} We can clearly observe the impact of early layers LN removal on suppressing memorization \& achieving true label recovery in Longformer, accompanied with higher train-test-accuracy gap, $\Delta_{\text{overfit}}^{\text{Pre, early}}$, than for later layers, $\Delta_{\text{overfit}}^{\text{Pre, later}}$, while learning being intact.}
    \label{fig:post-ln-early-middle-later_longformer}
\end{figure}

\begin{figure}[htbp]
    \centering
    \includegraphics[width=\textwidth]{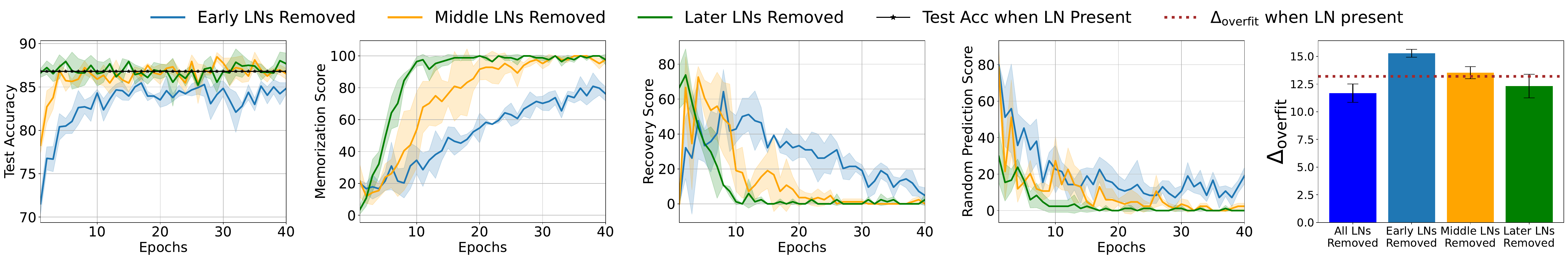}
    \caption{\textbf{Pivotal impact of early LNs on memorization in Post-LN model (RoBERTa, TweetTopic Dataset).} We can clearly observe the impact of early layers LN removal on suppressing memorization \& achieving true label recovery in RoBERTa, while learning being minimally impacted. \emph{Note: In the case of RoBERTa, $\Delta_{\text{overfit}}^{\text{Pre, early}}$ is slightly higher than $\Delta_{\text{overfit}}^{\text{Pre, early}}$, because even though removing early LNs, led to a greater memorization suppression than later LNs, but learning also got affected slightly, hence, the overfitting gap increased. However, we need to compare this Post-LN result with its Pre-LN counterpart of GPT2 (Fig.~\ref{fig:pre-ln-early-middle-later_gpt2}), where early LNs removal, affected learning even more, without suppressing memorization.}}
    \label{fig:post-ln-early-middle-later_roberta}
\end{figure}

\begin{figure}[htbp]
    \centering
    \includegraphics[width=\textwidth]{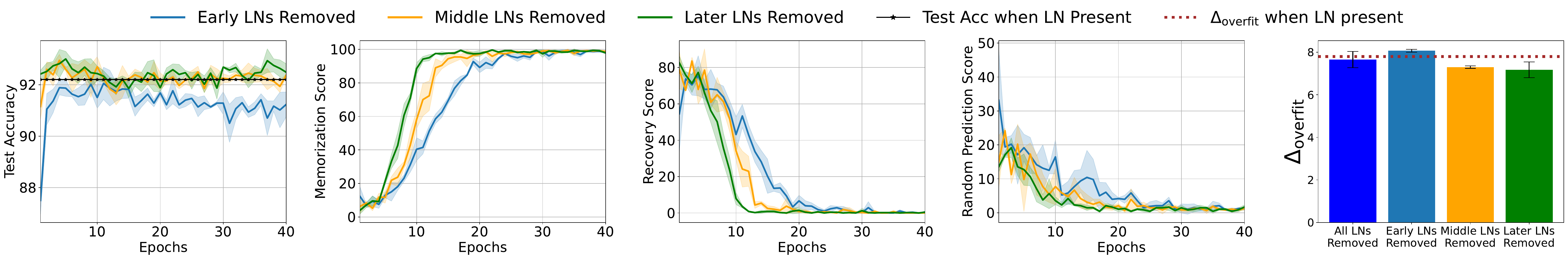}
    \caption{\textbf{Pivotal impact of early LNs on memorization in Post-LN model (DistilBERT, Emotions Dataset).} For the case of DistilBERT, we observe that just like when all LNs removal could not suppress memorization, the same happens with early LNs removal. However, we want to draw attention to the delay in achieving memorization, i.e., early LNs removal results in the highest delay in memorization compared to Middle/Later LNs removal. This shows that Early LNs are still the most significant.}
    \label{fig:post-ln-early-middle-later_distilbert}
\end{figure}

The results verified against Pre-LN models - GPTNeo, Qwen2, GPT2, RoBERTa-PreLayernorm, ViT-B, ViT-S, and Post-LN models - BERT, ELECTRA, Longformer, RoBERTa, DistilBERT are shown in Figs.~\ref{fig:pre-ln-early-middle-later_gpt_neo}, \ref{fig:pre-ln-early-middle-later_qwen2}, \ref{fig:pre-ln-early-middle-later_gpt2}, \ref{fig:pre-ln-early-middle-later_roberta_preln}, \ref{fig:pre-ln-early-middle-later_vit_base}, \ref{fig:pre-ln-early-middle-later_vit_small}, \ref{fig:post-ln-early-middle-later_bert}, \ref{fig:post-ln-early-middle-later_electra}, \ref{fig:post-ln-early-middle-later_longformer}, \ref{fig:post-ln-early-middle-later_roberta}, and \ref{fig:post-ln-early-middle-later_distilbert}.

\subsubsection{Pre-LN Models: Early Layers LN drives learning}

\newpage
\subsubsection{Post-LN Models: Early Layers LN suppresses memorization}

\newpage

\subsection{Gradients explain the impact of LN on Memorization \& Learning} \label{sec:gradients_explain_appendix}

In this section, we provide additional results to understand the distinctive impact of LN on memorization and learning through the lens of gradients. These results provide a deeper understanding of why (1) LN removal destabilizes learning in Pre-LN models, and suppresses memorization in Post-LN models, (2) Early Layers LN are more significant than later layers LN in driving these phenomena.

\subsubsection{Language Datasets Results}

\begin{figure}[htbp]
    \centering

    \begin{subfigure}{0.5\textwidth}
        \centering
        \includegraphics[width=\textwidth]{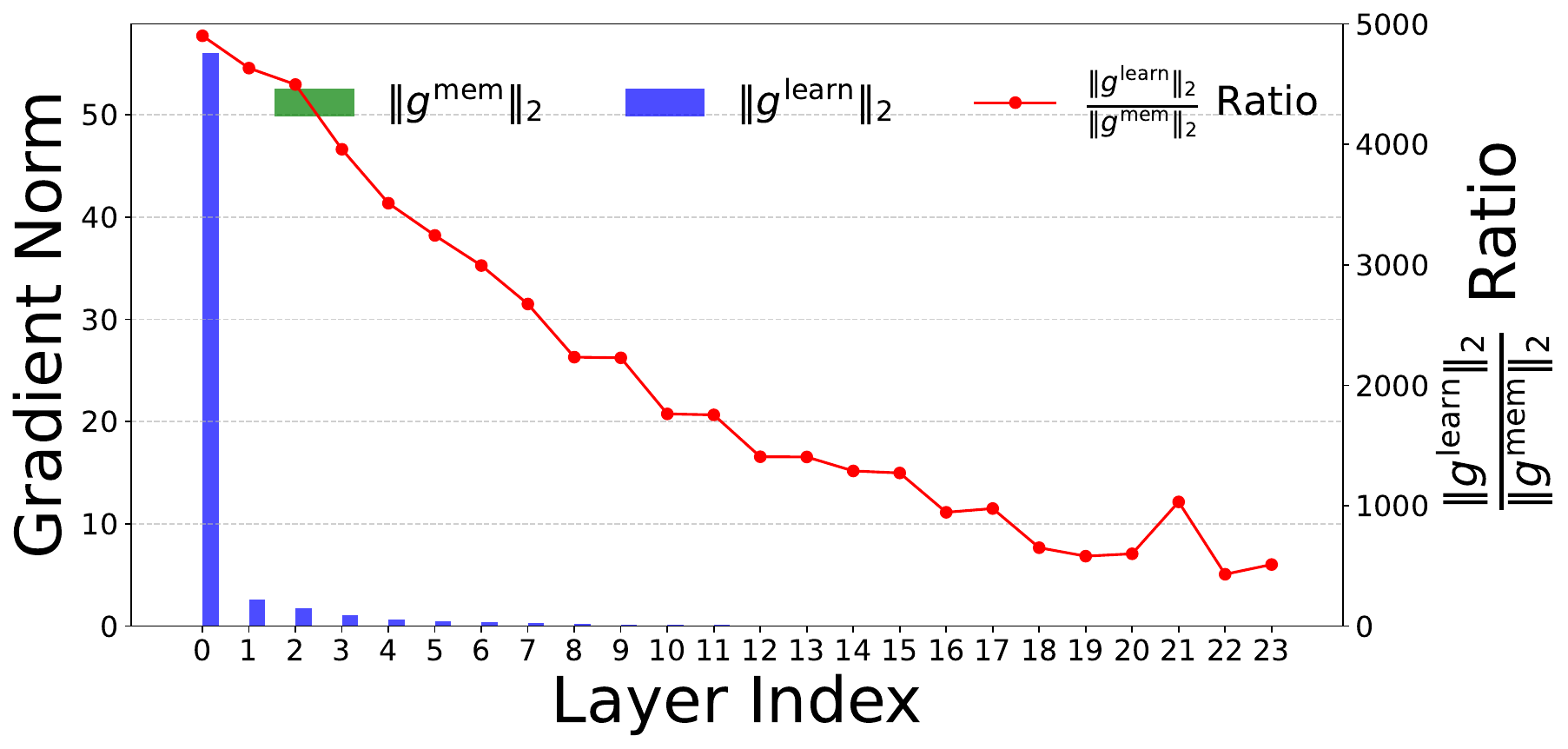}
        \caption{Pre-LN (Qwen2) $\left\lVert g_x^{\text{learn}} \right\rVert_2$ \& $\left\lVert g_x^{\text{mem}} \right\rVert_2$ analysis across layers}
        \label{fig:Pre_LN_gradients_qwen2_lhs}
    \end{subfigure}
    
    \begin{subfigure}{0.45\textwidth}
        \centering
        \includegraphics[width=\textwidth]{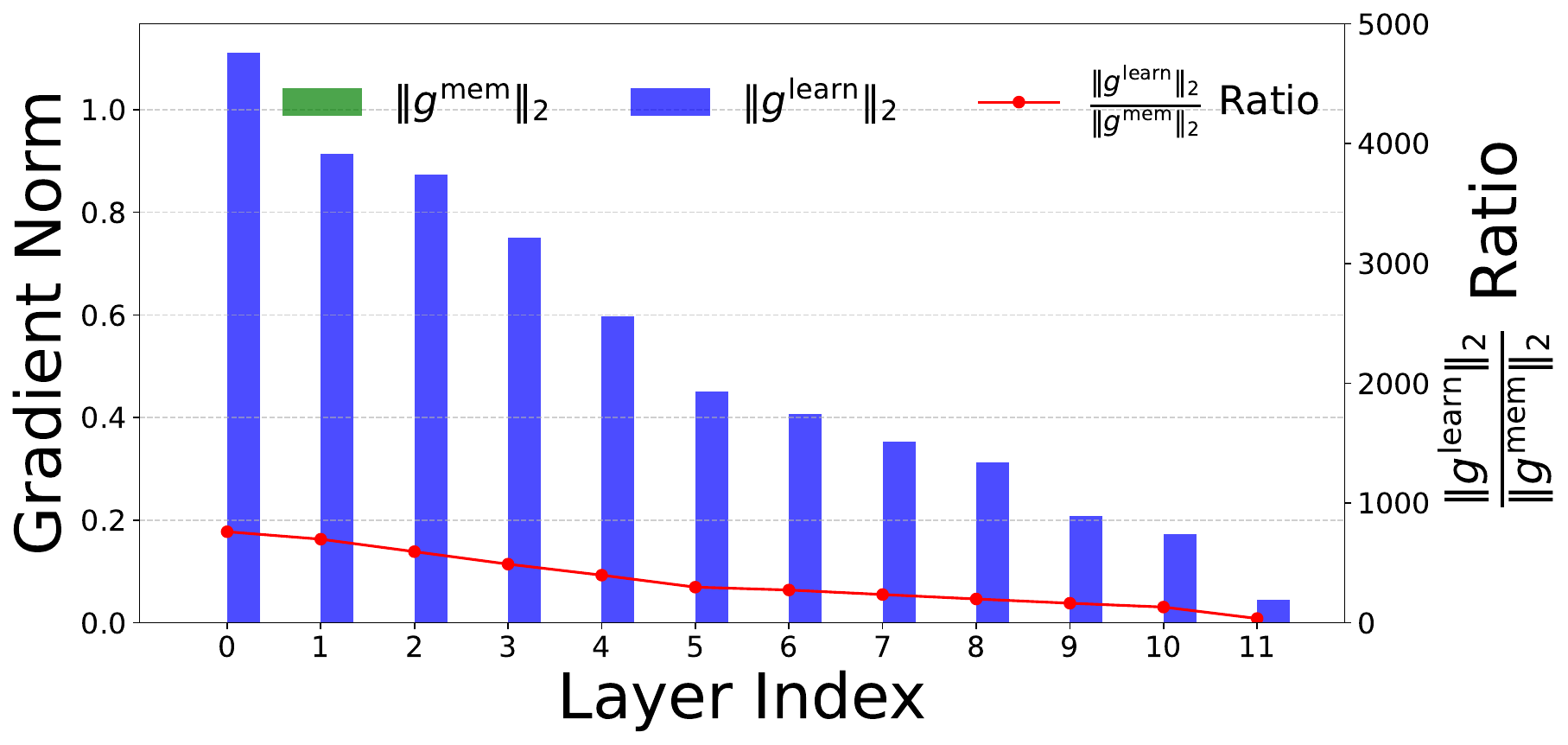}
        \caption{Post-LN (ELECTRA) $\left\lVert g_x^{\text{learn}} \right\rVert_2$ \& $\left\lVert g_x^{\text{mem}} \right\rVert_2$ analysis across layers}
        \label{fig:Post_LN_gradients_electra_lhs}
    \end{subfigure}
    \hspace{1mm}
    \begin{subfigure}{0.45\textwidth}
        \centering
        \includegraphics[width=\textwidth]{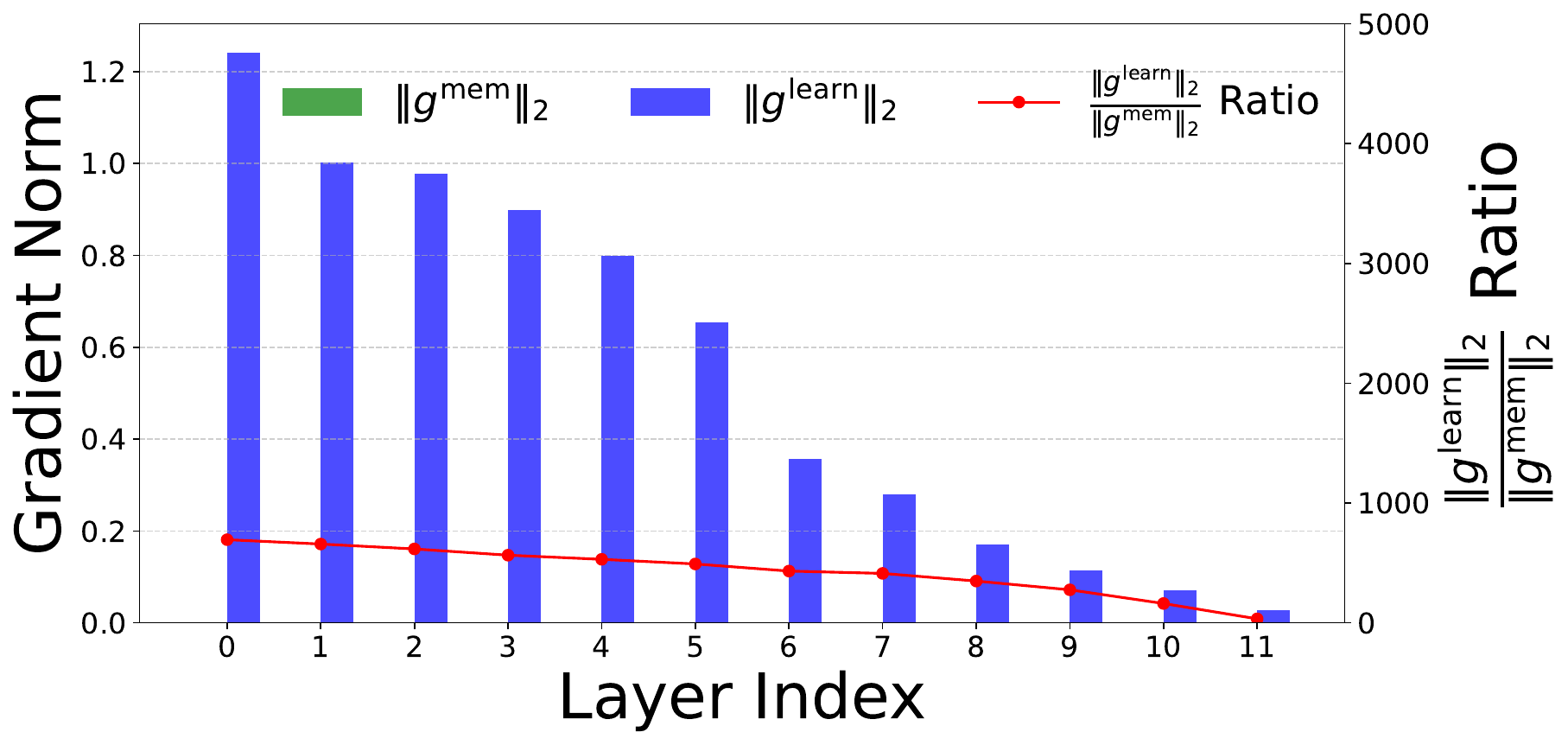}
        \caption{Post-LN (Longformer) $\left\lVert g_x^{\text{learn}} \right\rVert_2$ \& $\left\lVert g_x^{\text{mem}} \right\rVert_2$ analysis across layers}
        \label{fig:Post_LN_gradients_longformer_lhs}
    \end{subfigure}
    
    \caption{\textbf{Learning vs. Memorization Gradients in Pre- and Post-LN Models (News Dataset):} Results clearly exhibit high gradient norms of early layers LNs than later layers for both learning and memorization in Pre-LN (Qwen2) and Post-LN (ELECTRA, Longformer) models. Importantly, the learning gradient norms ($\left\lVert g_x^{\text{learn}} \right\rVert_2$) are consistently higher than the memorization gradient norms ($\left\lVert g_x^{\text{mem}} \right\rVert_2$) across all layers. Furthermore, the ratio $\left\lVert g_x^{\text{learn}} \right\rVert_2 \big/ \left\lVert g_x^{\text{mem}} \right\rVert_2$ is significantly higher in Pre-LN models compared to Post-LN models.}
    \label{fig:Pre_Post_Gradients_news_electra_longformer_qwen2}
\end{figure}

\begin{figure}[htbp]
    \centering
    \begin{subfigure}{0.45\textwidth}
        \centering
        \includegraphics[width=\textwidth]{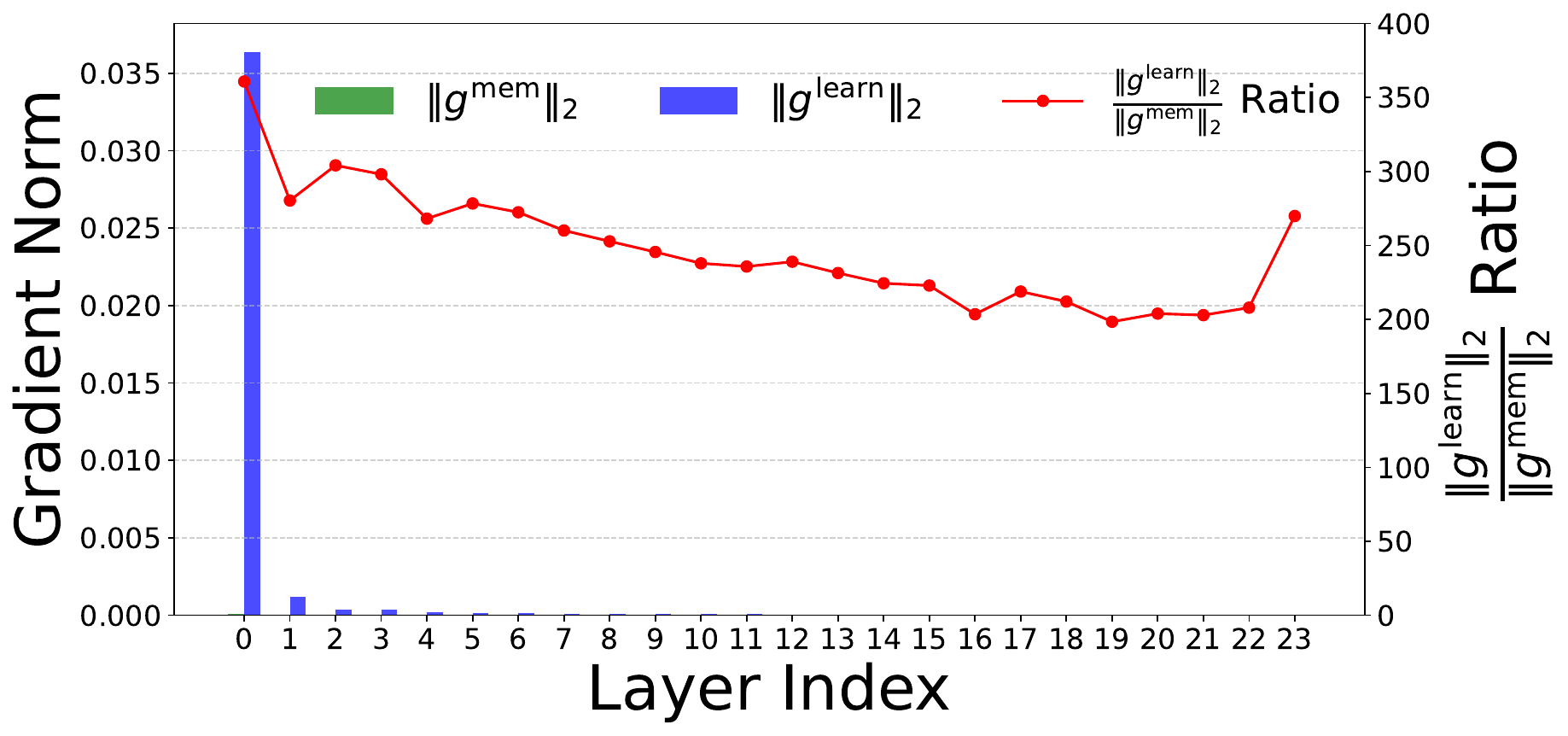}
        \caption{Pre-LN (GPT2) $\left\lVert g_x^{\text{learn}} \right\rVert_2$ \& $\left\lVert g_x^{\text{mem}} \right\rVert_2$ analysis across layers}
        \label{fig:Pre_LN_gradients_gpt2_lhs}
    \end{subfigure}
    \hspace{5mm}
    \begin{subfigure}{0.45\textwidth}
        \centering
        \includegraphics[width=\textwidth]{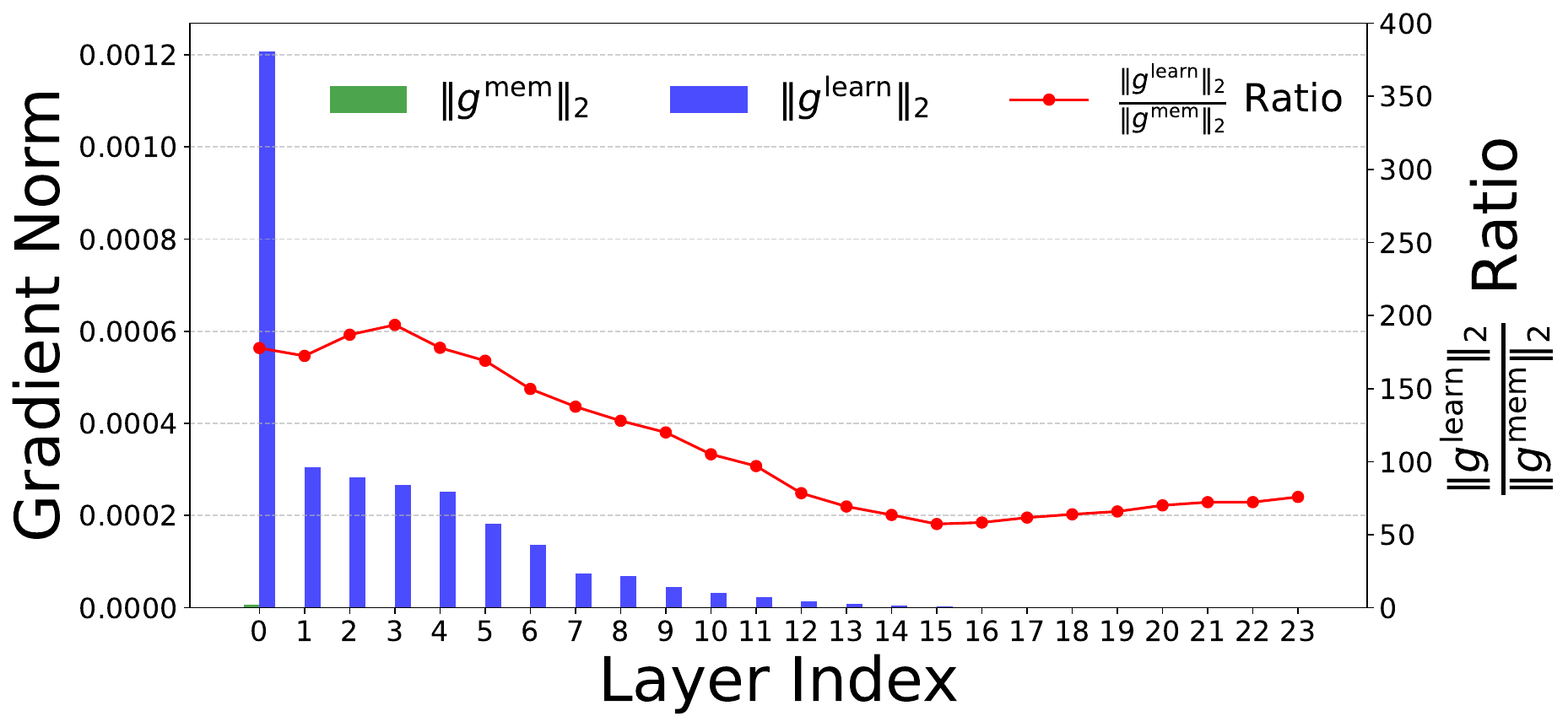}
        \caption{Pre-LN (RoBERTa-PreLayerNorm) $\left\lVert g_x^{\text{learn}} \right\rVert_2$ \& $\left\lVert g_x^{\text{mem}} \right\rVert_2$ analysis across layers}
        \label{fig:Pre_LN_gradients_roberta_preln_lhs}
    \end{subfigure}
    \begin{subfigure}{0.5\textwidth}
        \centering
        \includegraphics[width=\textwidth]{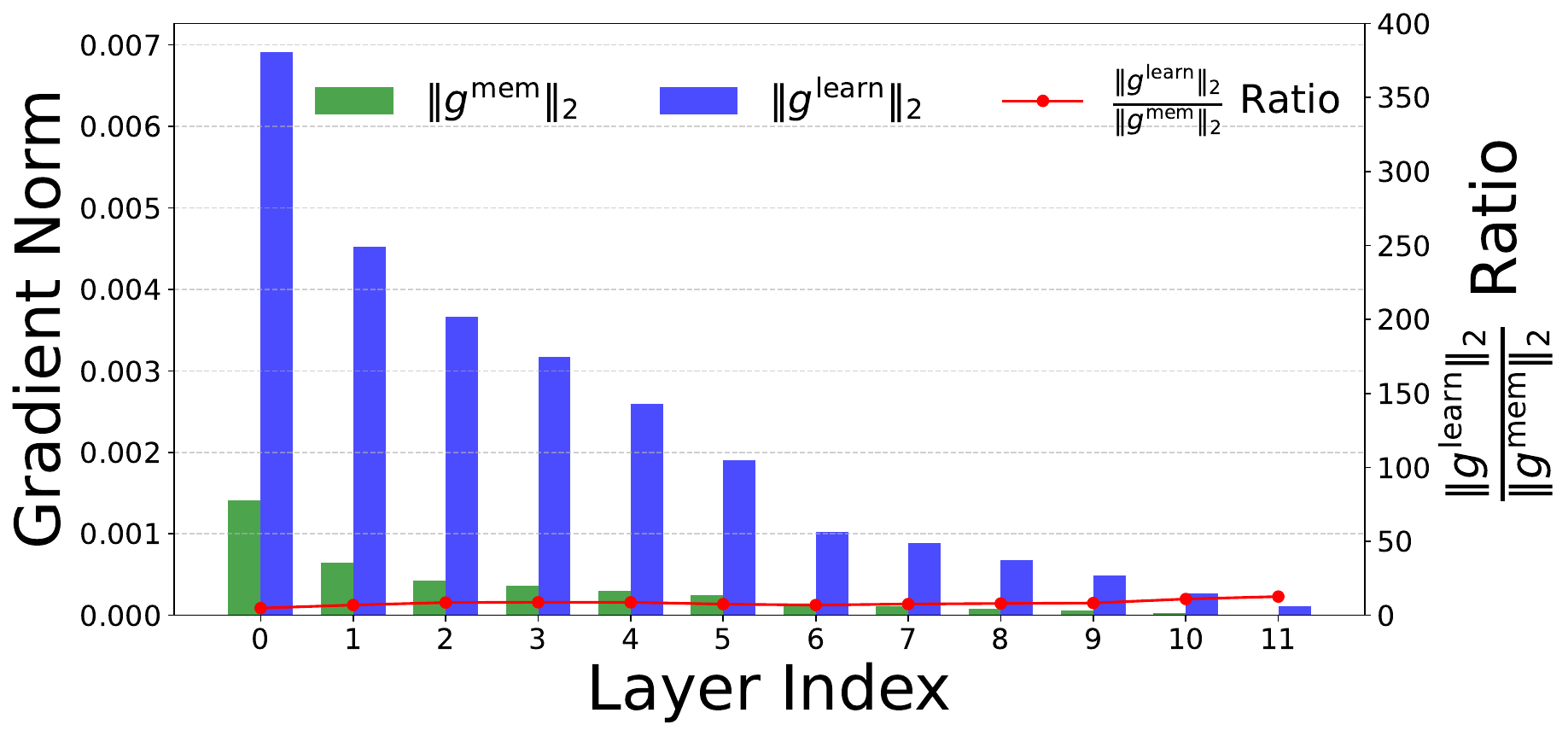}
        \caption{Post-LN (RoBERTa) $\left\lVert g_x^{\text{learn}} \right\rVert_2$ \& $\left\lVert g_x^{\text{mem}} \right\rVert_2$ analysis across layers}
        \label{fig:Post_LN_gradients_tweets_lhs}
    \end{subfigure}
    \caption{\textbf{Learning vs. Memorization Gradients in Pre- and Post-LN Models (TweetTopic Dataset):} Results clearly exhibit high gradient norms of early layers LNs than later layers for both learning and memorization in Pre-LN (GPT2, RoBERTA-PreLayerNorm) and Post-LN (RoBERTa) models. Importantly, the learning gradient norms ($\left\lVert g_x^{\text{learn}} \right\rVert_2$) are consistently higher than the memorization gradient norms  ($\left\lVert g_x^{\text{mem}} \right\rVert_2$) across all layers. Furthermore, the ratio $\left\lVert g_x^{\text{learn}} \right\rVert_2 \big/ \left\lVert g_x^{\text{mem}} \right\rVert_2$ is significantly higher in Pre-LN models compared to Post-LN models.}
    \label{fig:Pre_Post_Gradients_tweets_roberta_gpt2}
\end{figure}

\begin{figure}[htbp]
    \centering
    \begin{subfigure}{0.45\textwidth}
        \centering
        \includegraphics[width=\textwidth]{plots/Pre_LN/emotions_gpt_neo_gradients_lhs.pdf}
        \caption{Pre-LN (GPTNeo) $\left\lVert g_x^{\text{learn}} \right\rVert_2$ \& $\left\lVert g_x^{\text{mem}} \right\rVert_2$ analysis across layers}
        \label{fig:Pre_LN_gradients_gpt_neo_lhs_app}
    \end{subfigure}
    
    \begin{subfigure}{0.45\textwidth}
        \centering

        \includegraphics[width=\textwidth]{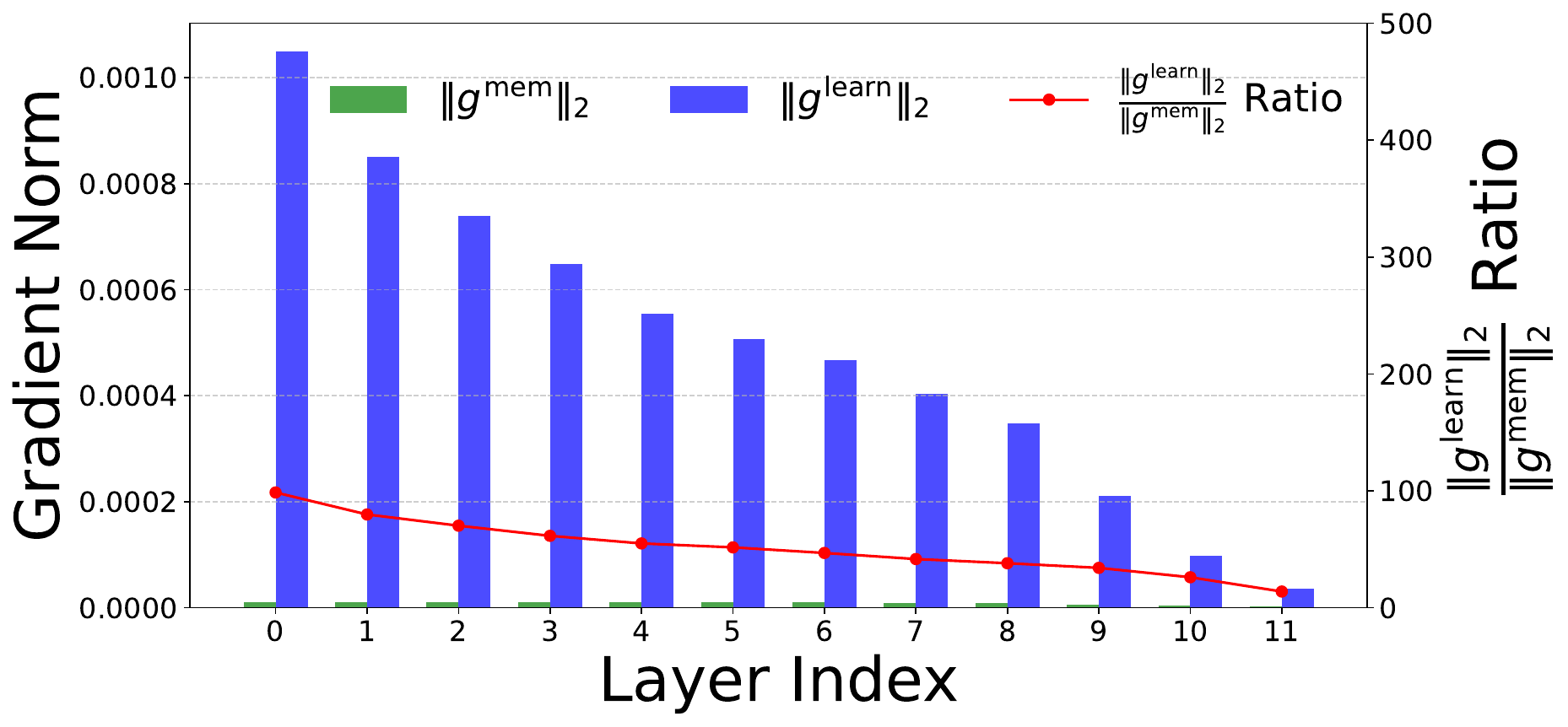}
        \caption{Post-LN (BERT) $\left\lVert g_x^{\text{learn}} \right\rVert_2$ \& $\left\lVert g_x^{\text{mem}} \right\rVert_2$ analysis across layers}
        \label{fig:Post_LN_gradients_bert_lhs}
    \end{subfigure}
    \hspace{5mm}
    \begin{subfigure}{0.45\textwidth}
        \centering

        \includegraphics[width=\textwidth]{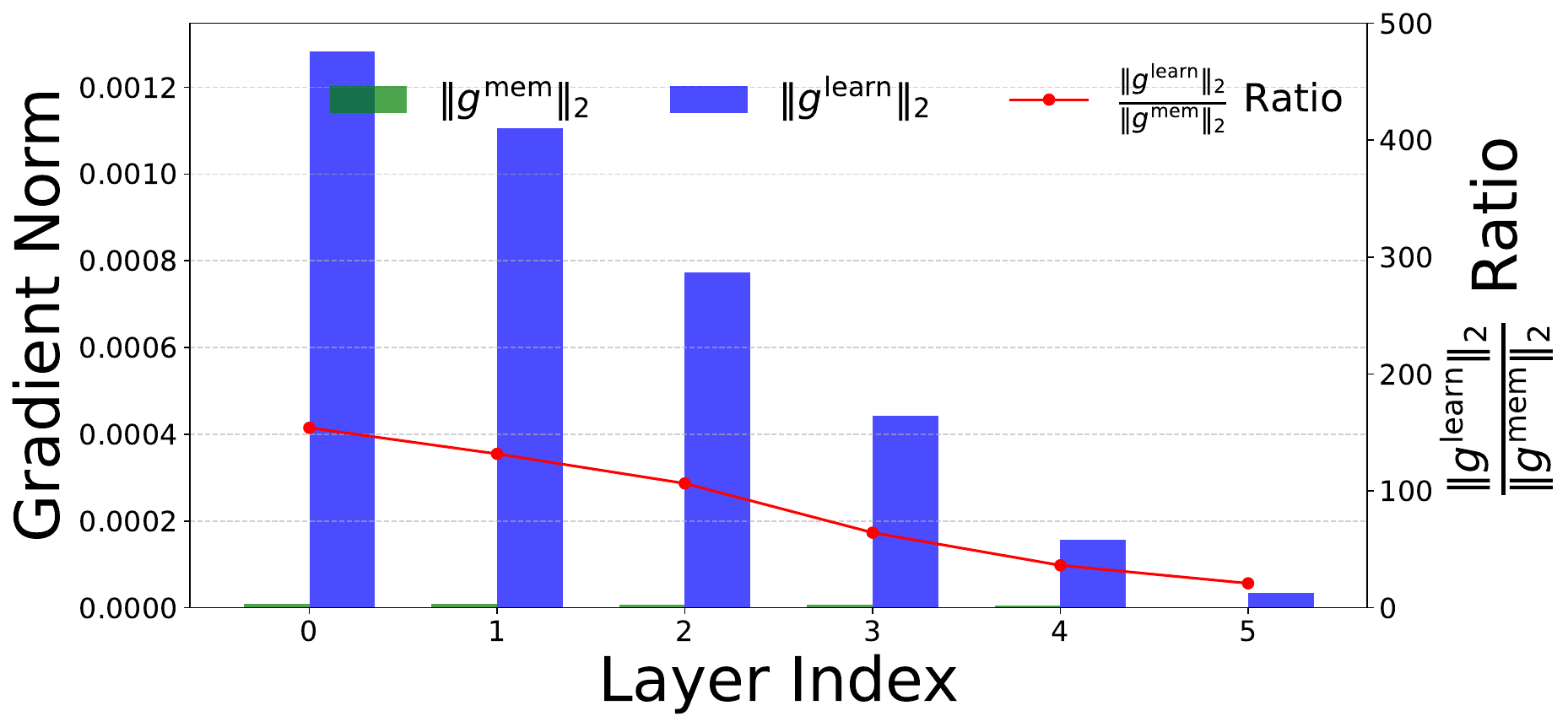}
        \caption{Post-LN (DistilBERT) $\left\lVert g_x^{\text{learn}} \right\rVert_2$ \& $\left\lVert g_x^{\text{mem}} \right\rVert_2$ analysis across layers}
        \label{fig:Post_LN_gradients_distilbert_lhs}
    \end{subfigure}

    \caption{\textbf{Learning vs. Memorization Gradients in Pre- and Post-LN Models (Emotions Dataset):} Results clearly exhibit high gradient norms of early layers LNs than later layers for both learning and memorization in Pre-LN (GPTNeo) and Post-LN (BERT, DistilBERT) models. Importantly, the learning gradient norms ($\left\lVert g_x^{\text{learn}} \right\rVert_2$) are consistently higher than the memorization gradient norms  ($\left\lVert g_x^{\text{mem}} \right\rVert_2$) across all layers. Furthermore, the ratio $\left\lVert g_x^{\text{learn}} \right\rVert_2 \big/ \left\lVert g_x^{\text{mem}} \right\rVert_2$ is significantly higher in Pre-LN models compared to Post-LN models.}
    \label{fig:Pre_Post_Gradients_emotions_bert_gptneo}
\end{figure}

In the language modality, we experimented with additional Pre-LN (GPTNeo, GPT2, Qwen2) and Post-LN models (BERT, RoBERTa, Longformer, ELECTRA) across Emotions, News, and Tweets Datasets. The results for the same are depicted in Figs.~\ref{fig:Pre_Post_Gradients_news_electra_longformer_qwen2}, \ref{fig:Pre_Post_Gradients_tweets_roberta_gpt2}, and \ref{fig:Pre_Post_Gradients_emotions_bert_gptneo}.

\newpage

\subsubsection{Vision Datasets Results}

\begin{figure}[htbp]
    \centering
    \begin{subfigure}{0.5\textwidth}
        \centering
        \includegraphics[width=\textwidth]{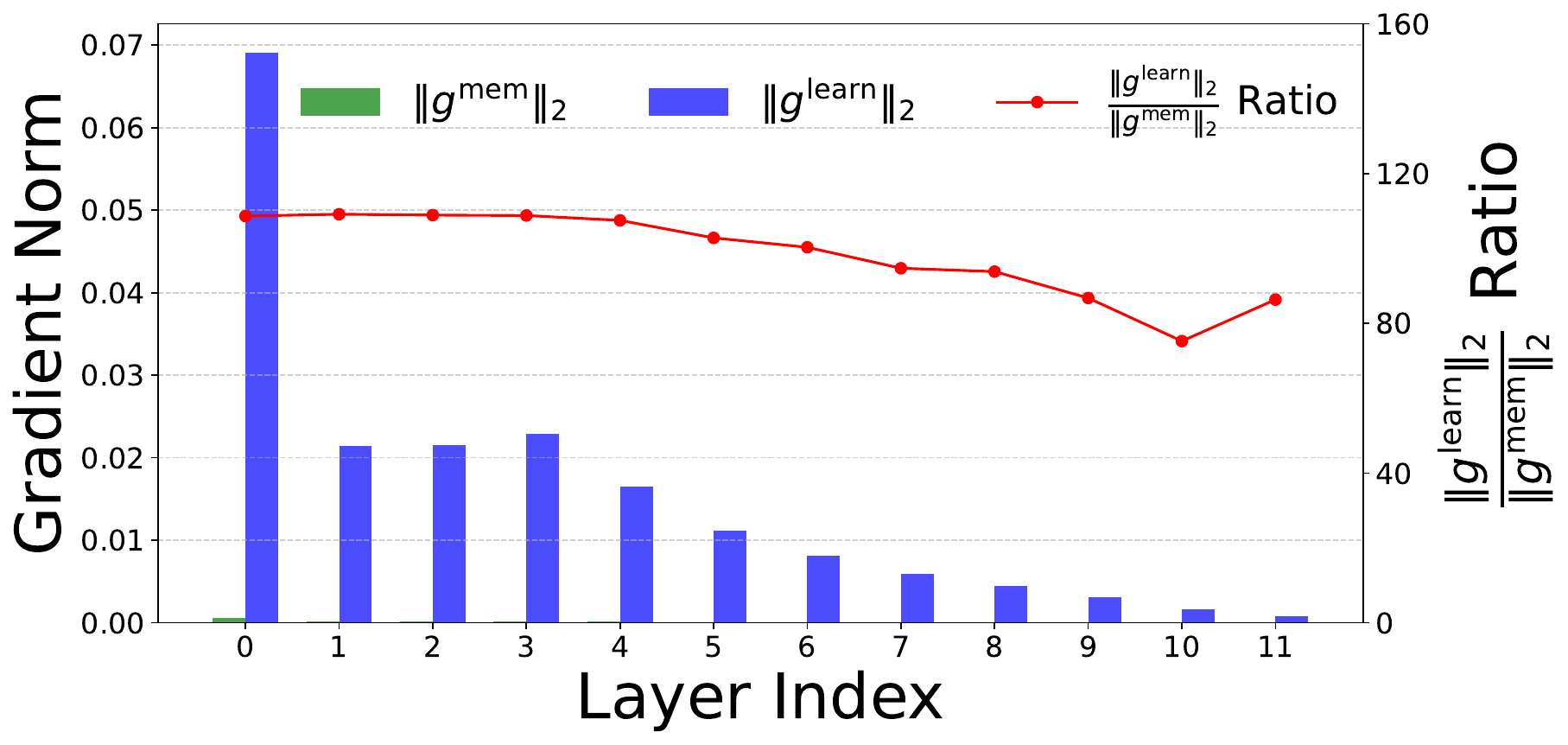}
        \caption{Pre-LN (ViT-B) $\left\lVert g_x^{\text{learn}} \right\rVert_2$ \& $\left\lVert g_x^{\text{mem}} \right\rVert_2$ analysis across layers}
        \label{fig:Pre_LN_gradients_vit_base_lhs}
    \end{subfigure}
    \caption{\textbf{Learning vs. Memorization Gradients in Pre-LN Models (CIFAR10 Dataset):} Results clearly exhibit high gradient norms of early layers LNs than later layers for both learning and memorization in Pre-LN (ViT-B) models. Importantly, the learning gradient norms ($\left\lVert g_x^{\text{learn}} \right\rVert_2$) are consistently higher than the memorization gradient norms  ($\left\lVert g_x^{\text{mem}} \right\rVert_2$) across all layers.}
    \label{fig:Pre_Post_Gradients_utk_face_vit_base}
\end{figure}

\begin{figure}[htbp]
    \centering
    \begin{subfigure}{0.5\textwidth}
        \centering
        \includegraphics[width=\textwidth]{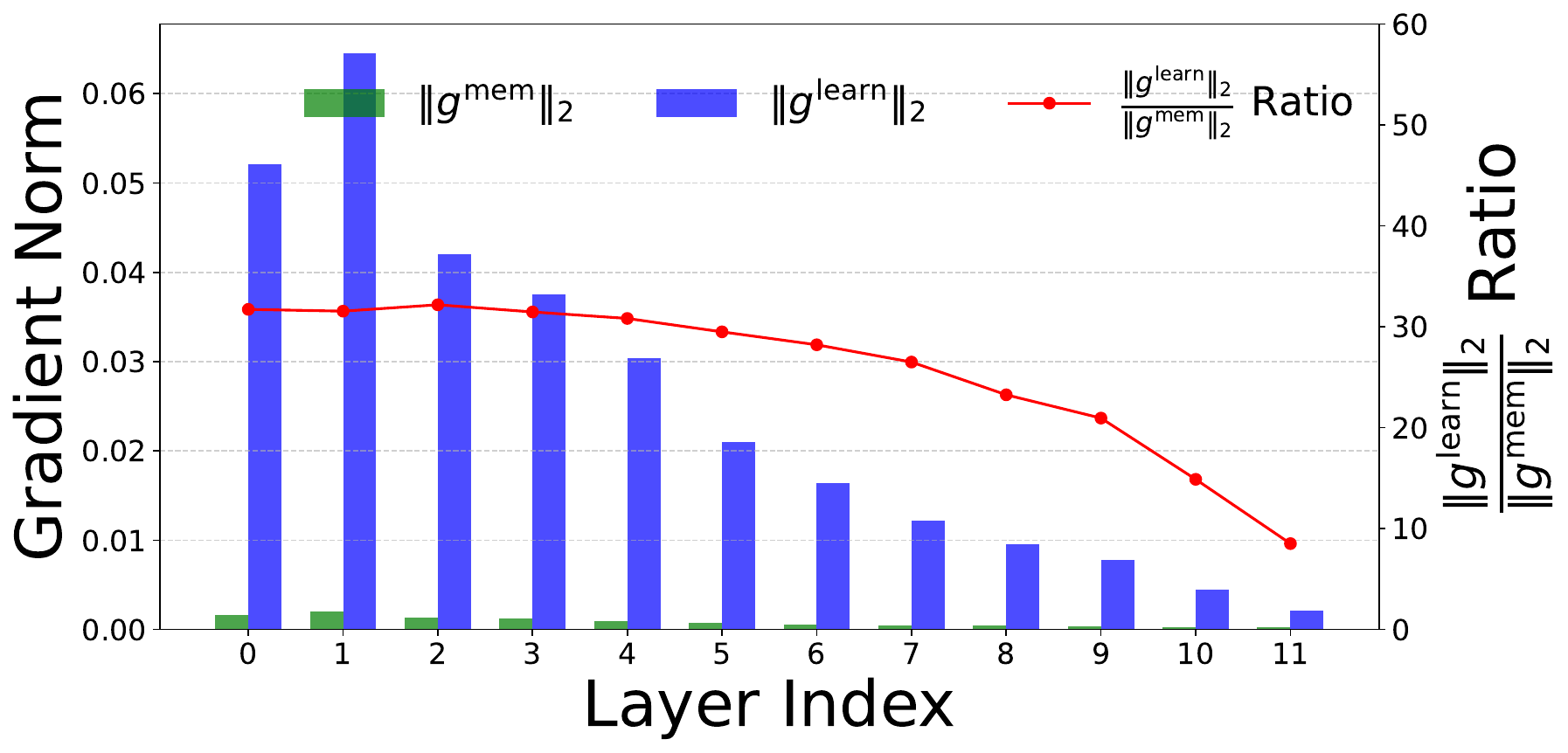}
        \caption{Pre-LN (ViT-S) $\left\lVert g_x^{\text{learn}} \right\rVert_2$ \& $\left\lVert g_x^{\text{mem}} \right\rVert_2$ analysis across layers}
        \label{fig:Pre_LN_gradients_vit_small_lhs}
    \end{subfigure}
    \caption{\textbf{Learning vs. Memorization Gradients in Pre-LN Models (NICO++ Dataset):} Results clearly exhibit high gradient norms of early layers LNs than later layers for both learning and memorization in Pre-LN (ViT-S) models. Importantly, the learning gradient norms ($\left\lVert g_x^{\text{learn}} \right\rVert_2$) are consistently higher than the memorization gradient norms ($\left\lVert g_x^{\text{mem}} \right\rVert_2$) across all layers.}
    \label{fig:Pre_Post_Gradients_utk_face_vit_small}
\end{figure}

\begin{figure}[htbp]
    \centering
    \begin{subfigure}{0.5\textwidth}
        \centering
        \includegraphics[width=\textwidth]{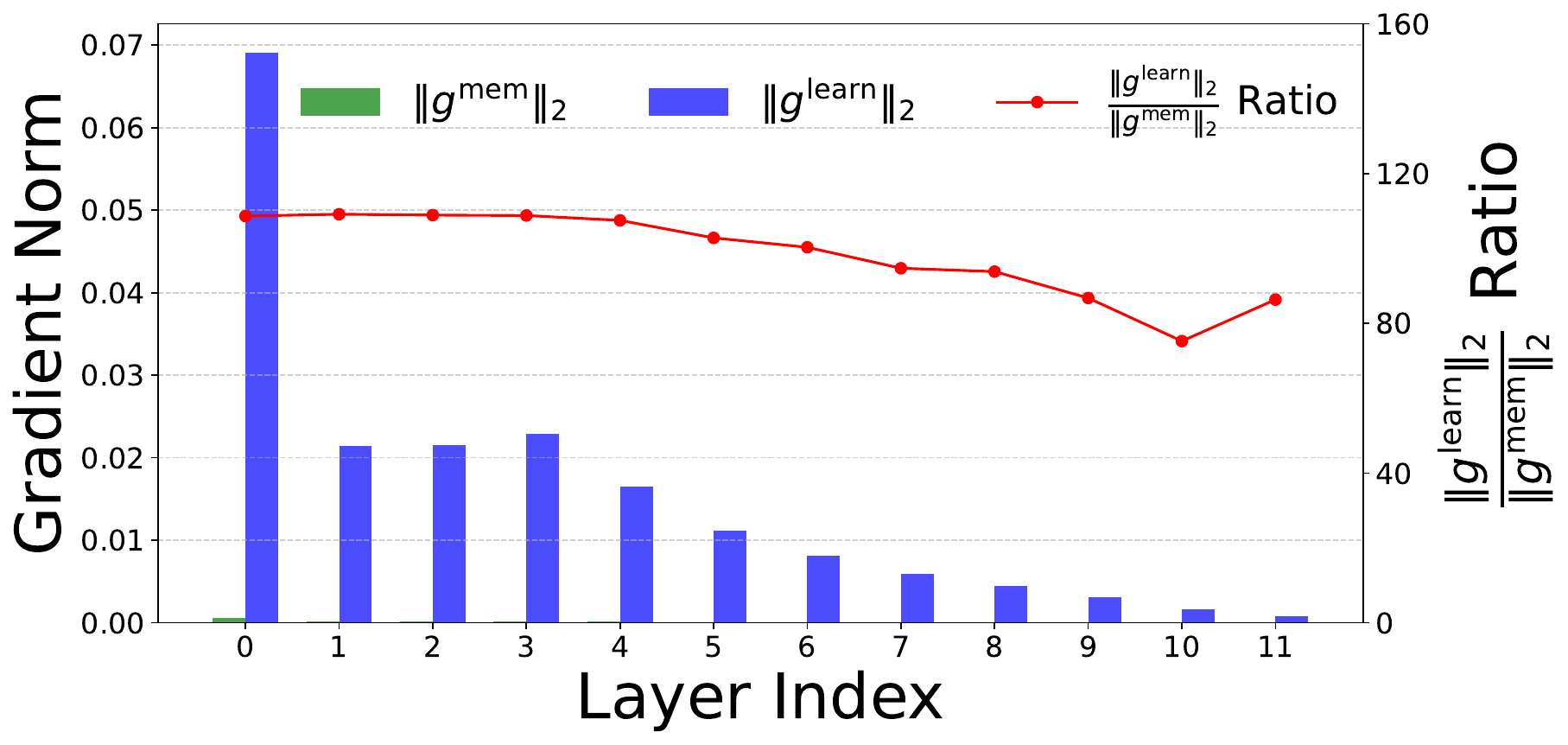}
        \caption{Pre-LN (DeiT) $\left\lVert g_x^{\text{learn}} \right\rVert_2$ \& $\left\lVert g_x^{\text{mem}} \right\rVert_2$ analysis across layers}
        \label{fig:Pre_LN_gradients_deit_lhs}
    \end{subfigure}
    \caption{\textbf{Learning vs. Memorization Gradients in Pre-LN Models (UTK-Face Dataset):} Results clearly exhibit high gradient norms of early layers LNs than later layers for both learning and memorization in Pre-LN (DeiT) models. Importantly, the learning gradient norms ($\left\lVert g_x^{\text{learn}} \right\rVert_2$) are consistently higher than the memorization gradient norms ($\left\lVert g_x^{\text{mem}} \right\rVert_2$) across all layers.}
    \label{fig:Pre_Post_Gradients_utk_face_deit}
\end{figure}

For the vision modality, it needs to be acknowledged that Post-LN architectures are not available in practice/literature, and only Pre-LN models are available. Hence, we provide additional experiments for Pre-LN models - ViT-B, ViT-S, and DeiT using multiple datasets - CIFAR10, CIFAR100, UTK-Face. The results for the same are presented in Figs.~\ref{fig:Pre_Post_Gradients_utk_face_vit_base}, \ref{fig:Pre_Post_Gradients_utk_face_vit_small}, and \ref{fig:Pre_Post_Gradients_utk_face_deit}.

\section{Analysis across multiple Noisy Label Ratios and Optimizer}

We extend our analysis to evaluate whether the claim—\emph{removing LN parameters mitigates memorization in Post-LN models while impairing generalization in Pre-LN models}—holds consistently across higher noisy label ratios (2\% and 5\%) in the training dataset, as shown in Table~\ref{tab:higher_noise_ratios}.

\begin{table}[h!]
\centering
\begin{tabular}{ccccccc}
\toprule
\textbf{Noise} & \textbf{Model} & \textbf{Setting} & \textbf{Learning (↑)} & \textbf{Memorization (↓)} & \textbf{Recovery (↑)} & \textbf{Random} \\
& & & & & & \textbf{Prediction (↓)} \\
\midrule
\multirow{4}{*}{2\%} 
& \textbf{Post-LN (BERT)} & Before & 91.70 & 100.00 & 0.00 & 0.00 \\
&  & After & 92.00 & 20.62 & 76.25 & 3.12 \\
& \textbf{Pre-LN (GPT-Neo)} & Before & 91.35 & 100.00 & 0.00 & 0.00 \\
&  & After & 84.85 & 66.87 & 16.56 & 16.56 \\
\midrule
\multirow{4}{*}{5\%} 
& \textbf{Post-LN (BERT)} & Before & 90.35 & 100.00 & 0.00 & 0.00 \\
&  & After & 91.25 & 27.00 & 66.88 & 6.12 \\
& \textbf{Pre-LN (GPT-Neo)} & Before & 90.35 & 100.00 & 0.00 & 0.00 \\
&  & After & 82.60 & 67.50 & 11.00 & 21.50 \\
\bottomrule
\end{tabular}
\caption{Results for higher noisy label ratios (2\% and 5\%) on the Emotions dataset.}
\label{tab:higher_noise_ratios}
\end{table}

Additionally, we assess the robustness of this finding under a different optimization setting by experimenting with Muon optimizer other than Adam, which was used in the primary experiments, as shown in Table~\ref{tab:muon_opt}.

\begin{table}[h!]
\centering
\begin{tabular}{lccccc}
\toprule
\textbf{Model} & \textbf{Setting} & \textbf{Learning (↑)} & \textbf{Memorization (↓)} & \textbf{Recovery (↑)} & \textbf{Random} \\
& & & & & \textbf{Prediction (↓)} \\
\midrule
\textbf{Post-LN (BERT)} & Before & 91.95 & 100.00 & 0.00 & 0.00 \\
 & After & 92.00 & 25.00 & 62.50 & 12.50 \\
\textbf{Pre-LN (GPT-Neo)} & Before & 91.70 & 100.00 & 0.00 & 0.00 \\
 & After & 85.55 & 32.50 & 29.38 & 38.12 \\
\bottomrule
\end{tabular}
\caption{Results with the Muon optimizer with Emotions Dataset.}
\label{tab:muon_opt}
\end{table}

Overall, the results demonstrate that our observations are agnostic to optimizer and noise-label ratios.

\section{Loss and Gradient Norms Analysis across Epochs}

\begin{figure}[htbp]
    \centering

    \begin{subfigure}[t]{0.48\textwidth}
        \centering
        \includegraphics[width=\linewidth]{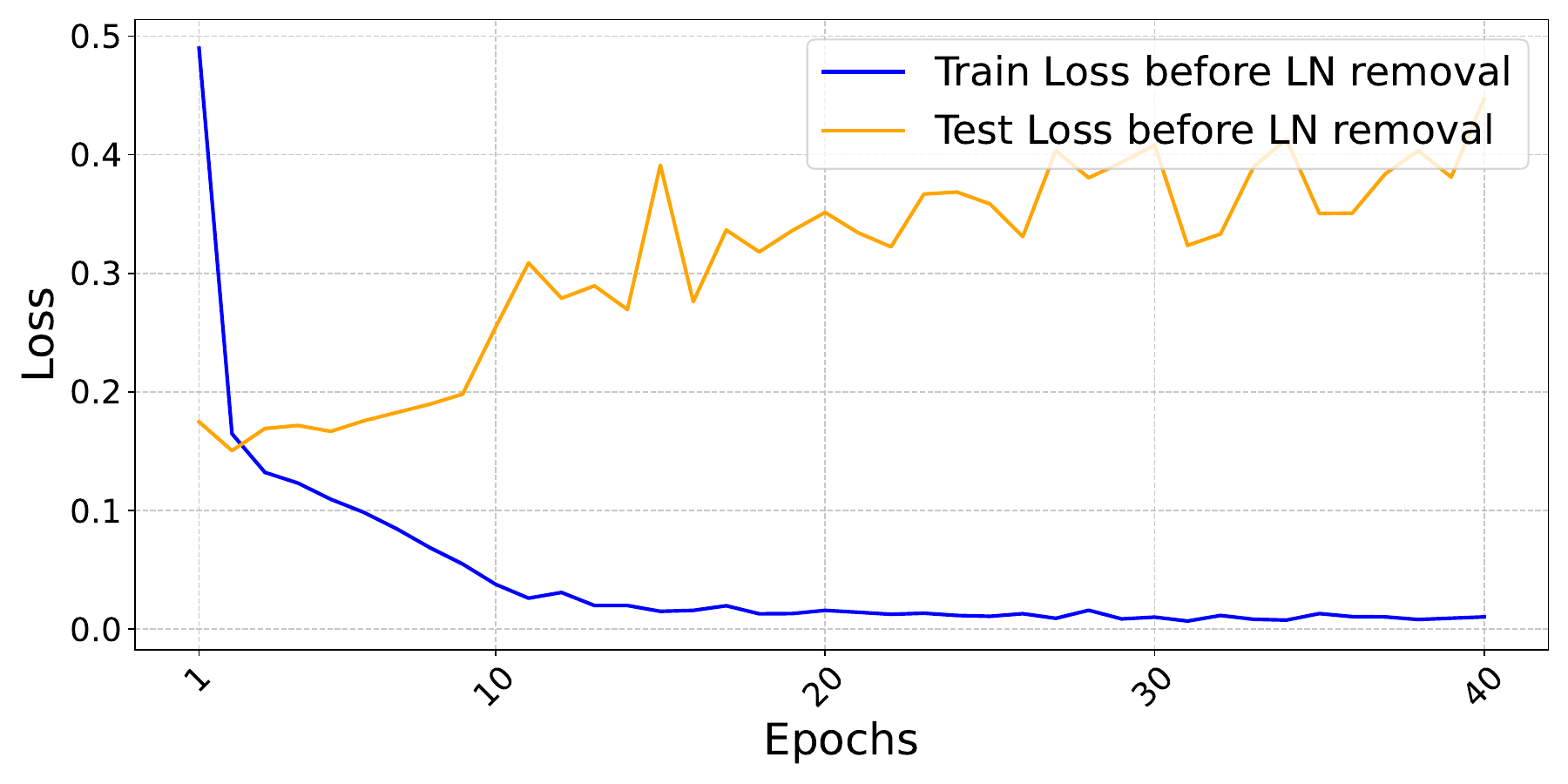}
        \caption{BERT (Post-LN) before LN removal.}
        \label{fig:bert_loss_before}
    \end{subfigure}
    \hfill
    \begin{subfigure}[t]{0.48\textwidth}
        \centering
        \includegraphics[width=\linewidth]{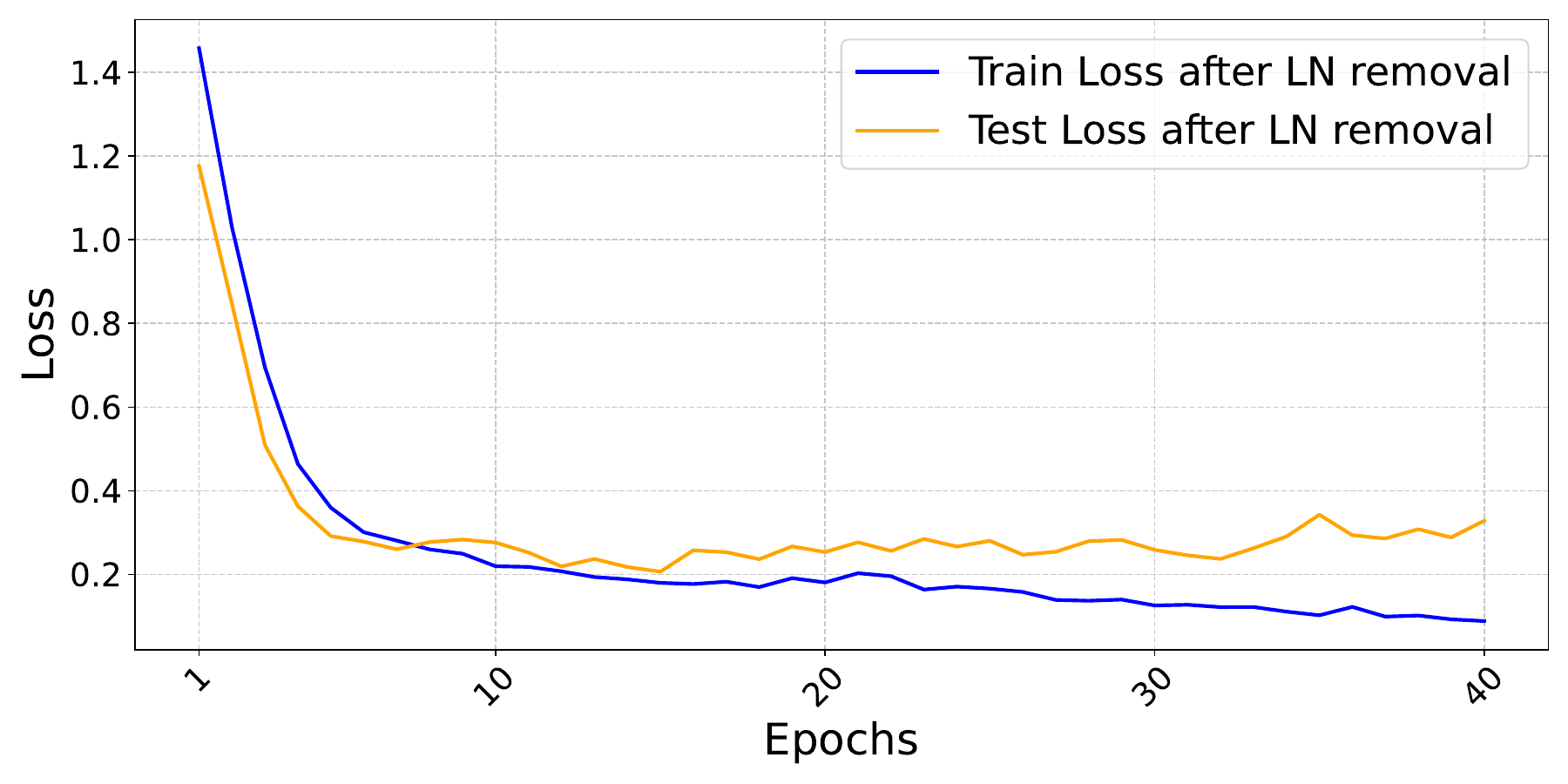}
        \caption{BERT (Post-LN) after LN removal.}
        \label{fig:bert_loss_after}
    \end{subfigure}

    \vspace{0.5em}

    \begin{subfigure}[t]{0.48\textwidth}
        \centering
        \includegraphics[width=\linewidth]{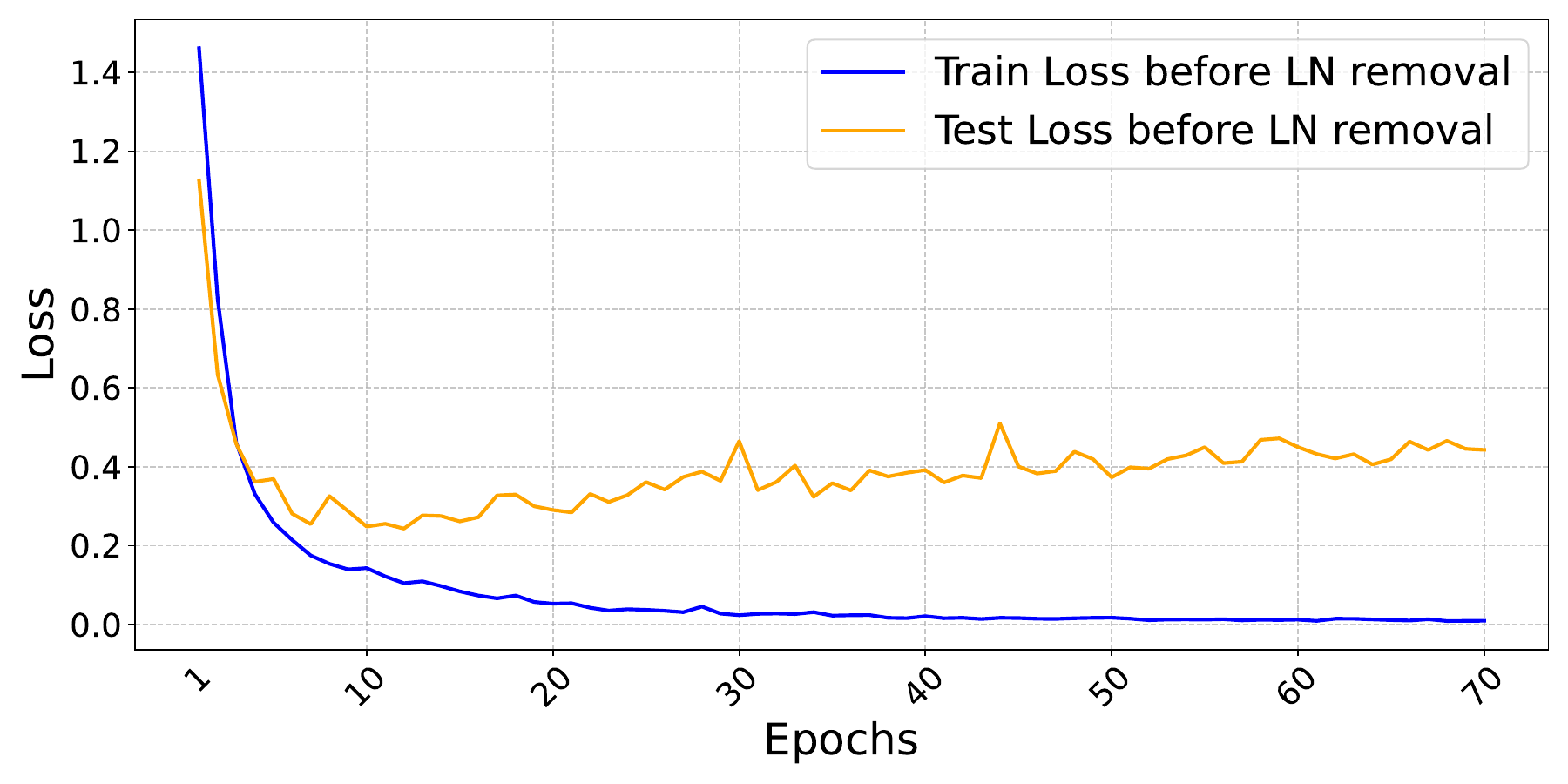}
        \caption{GPT-Neo (Pre-LN) before LN removal.}
        \label{fig:gptneo_loss_before}
    \end{subfigure}
    \hfill
    \begin{subfigure}[t]{0.48\textwidth}
        \centering
        \includegraphics[width=\linewidth]{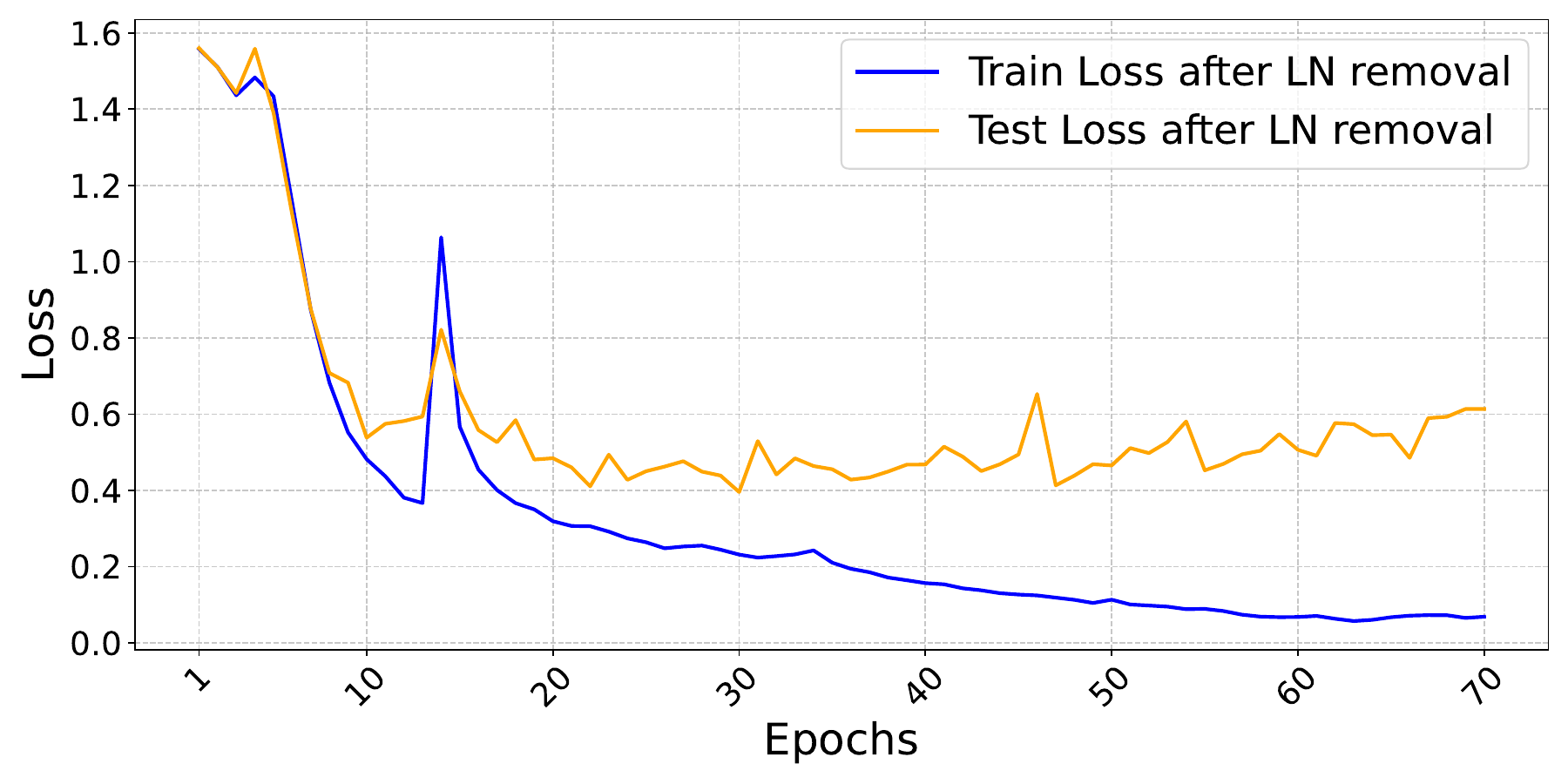}
        \caption{GPT-Neo (Pre-LN) after LN removal.}
        \label{fig:gptneo_loss_after}
    \end{subfigure}

    \caption{
    Train and test loss trends across epochs for Post-LN (BERT) and Pre-LN (GPT-Neo) models before and after LayerNorm (LN) removal. 
    In Post-LN models, LN removal reduces the overfitting gap (difference between train and test losses), mitigating memorization. 
    In Pre-LN models, LN removal widens the overfitting gap, indicating impaired generalization.
    }
    \label{fig:loss_comparison}
\end{figure}

In Sec.~\ref{sec:discrepancy_post_pre_ln}, we showed how on LN parameters removal in Post-LN models, memorization is mitigated over epochs, while in Pre-LN models, the testing accuracy drops as training progresses, without recovering at any time. To further corroborate this claim, we plot the train-test loss functions for Post-LN and Pre-LN models, where the overfitting gap decreases in Post-LN models, while for Pre-LN models it increases, upon LN removal, as shown in Fig.~\ref{fig:loss_comparison}.

\begin{figure}[htbp]
    \centering
    \begin{subfigure}[t]{0.48\textwidth}
        \centering
        \includegraphics[width=\linewidth]{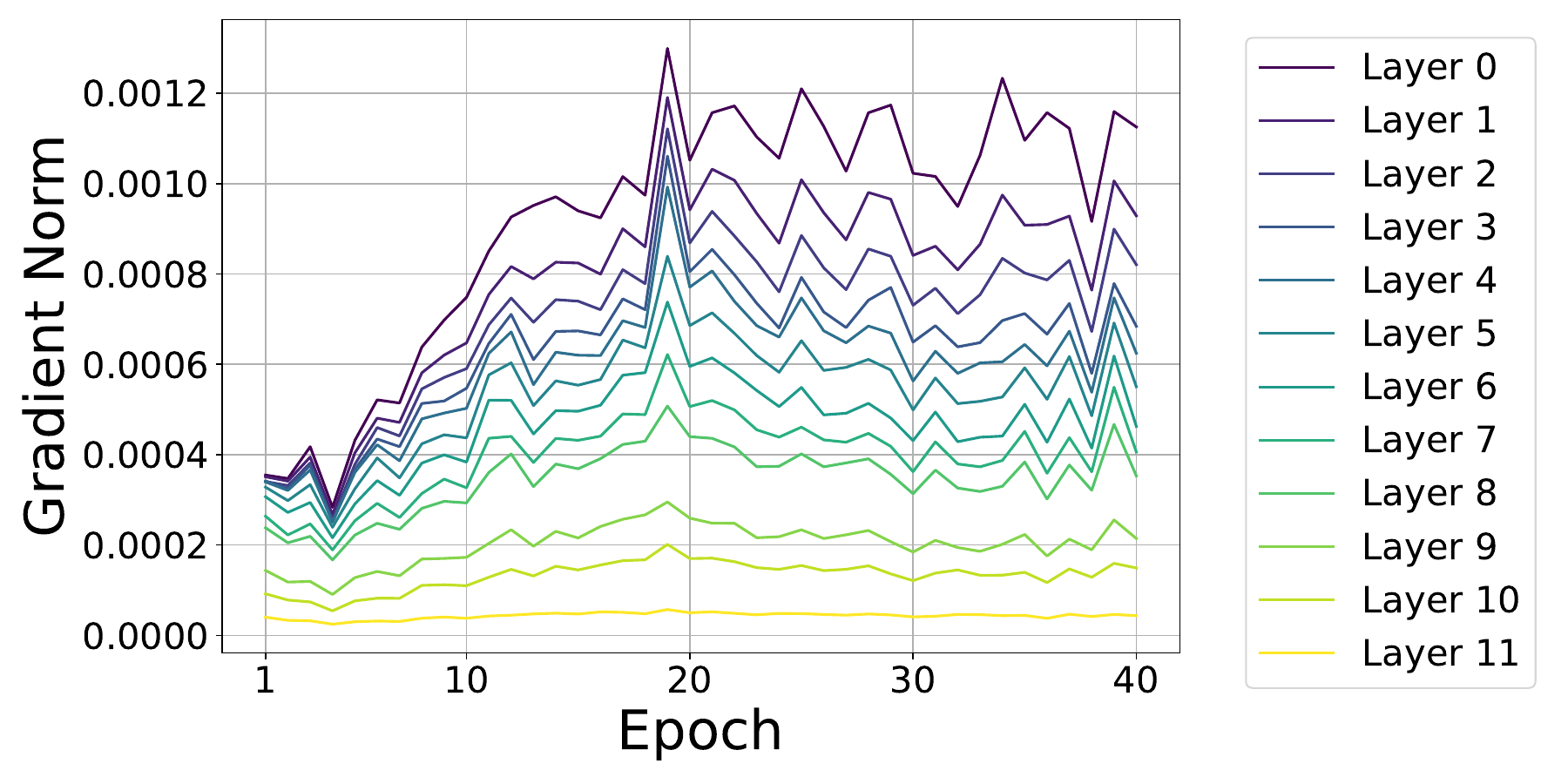}
        \caption{Gradient norms for BERT}
        \label{fig:bert_grad_norms}
    \end{subfigure}
    \hfill
    \begin{subfigure}[t]{0.48\textwidth}
        \centering
        \includegraphics[width=\linewidth]{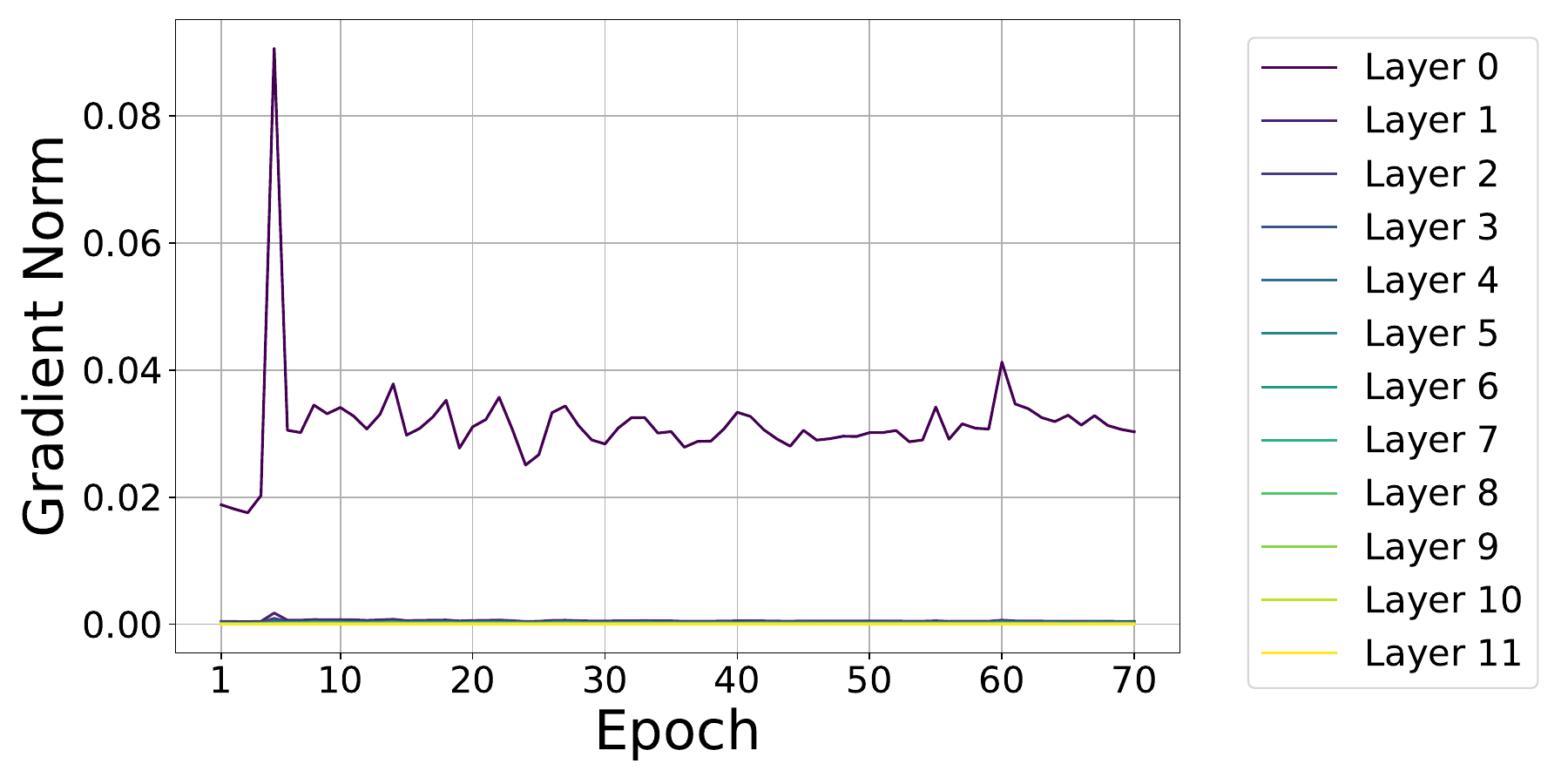}
        \caption{Gradient norms for GPT-Neo}
        \label{fig:gptneo_grad_norms}
    \end{subfigure}
    \caption{Gradient norms gradually increase across epochs for both Pre-LN and Post-LN models, with early layers exhibiting the highest gradient norms.}
    \label{fig:grad_norms_epochs}
\end{figure}

Apart from the loss curves, we also provide how the gradient norm evolves across epochs for both Post-LN and Pre-LN models in Fig.~\ref{fig:grad_norms_epochs}.

\newpage
\section{Validity of Results for Generative Modeling Task}

To verify the robustness of our findings in a generative modeling context, we conduct a Next Token Prediction (NTP) experiment on the Emotions dataset using both BERT (Post-LN) and GPT-Neo (Pre-LN) models.

Each input sample is reformulated as:
\[
\text{original text} + \text{``This emotion is [type]''}
\]
where the model predicts the token corresponding to the emotion label \([type]\), one of six possible emotion types. To introduce \textbf{noisy labels}, we randomly replace the \([type]\) token for 1\% of the training samples with a different emotion label. Two different model configurations are trained and evaluated: (i) before LN parameters removal, and (ii) after LN parameters removal.

\begin{table}[htbp]
\centering
\caption{Results on Next Token Prediction task with noisy labels.}
\begin{tabular}{lccccc}
\toprule
\textbf{Model} & \textbf{Setting} & \textbf{Learning (↑)} & \textbf{Memorization (↓)} & \textbf{Recovery (↑)} & \textbf{Random} \\
& & & & & \textbf{Prediction (↓)} \\
\midrule
\textbf{Post-LN (BERT)} & Before & 92.14 & 100.00 & 0.00 & 0.00 \\
 & After & 91.95 & 28.12 & 60.62 & 11.25 \\
\textbf{Pre-LN (GPT-Neo)} & Before & 91.70 & 100.00 & 0.00 & 0.00 \\
 & After & 85.55 & 51.25 & 19.38 & 29.38 \\
\bottomrule
\end{tabular}
\label{tab:gen_ai}
\end{table}

As shown in Table~\ref{tab:gen_ai}, removing LN parameters substantially reduces memorization in Post-LN models (BERT), while impairing learning and generalization in Pre-LN models (GPT-Neo). This confirms that the trends reported in the main paper hold consistently for generative modeling tasks as well.

\section{Broader Impacts and Limitations} \label{sec:broader_impacts}
Our study reveals that LayerNorm (LN) affects memorization and learning differently across transformer variants: disabling LN parameters suppresses memorization and aids label recovery in Post-LN models, but destabilizes learning in Pre-LN models. These insights can inform architecture design and robust training in noisy labels settings, where memorization needs to be controlled. While we focus on LN due to its central role and controllability, other components—like residual paths, attention, and feedforward layers—also influence memorization and merit further investigation. We hope this work can provide insights to the community to encourage further follow-up studies.


\newpage
\section*{NeurIPS Paper Checklist}

\begin{enumerate}

\item {\bf Claims}
    \item[] Question: Do the main claims made in the abstract and introduction accurately reflect the paper's contributions and scope?
    \item[] Answer: \answerYes{} 
    \item[] Justification: We have systematically demonstrated the distinct impact of LN on memorization and learning across Pre- and Post-LN models using various experiments and results in Sec.~\ref{sec:discrepancy_post_pre_ln}, \ref{sec:early_layers_imp}, and \ref{sec:learn_mem_grad}.

\item {\bf Limitations}
    \item[] Question: Does the paper discuss the limitations of the work performed by the authors?
    \item[] Answer: \answerYes{} 
    \item[] Justification: Yes, the paper reflects the limitations of prior work, while emphasizing the novel results of the distinctive impact of LN on memorization and learning.

\item {\bf Theory assumptions and proofs}
    \item[] Question: For each theoretical result, does the paper provide the full set of assumptions and a complete (and correct) proof?
    \item[] Answer: \answerYes{} 
    \item[] Justification: Yes, complete and correct proofs are provided for the 3 Theorems provided in the paper.

    \item {\bf Experimental result reproducibility}
    \item[] Question: Does the paper fully disclose all the information needed to reproduce the main experimental results of the paper to the extent that it affects the main claims and/or conclusions of the paper (regardless of whether the code and data are provided or not)?
    \item[] Answer: \answerYes{} 
    \item[] Justification: Yes, the paper fully discloses all the important information in replicating the experiments done in it.

\item {\bf Open access to data and code}
    \item[] Question: Does the paper provide open access to the data and code, with sufficient instructions to faithfully reproduce the main experimental results, as described in supplemental material?
    \item[] Answer: \answerYes{} 
    \item[] Justification: Yes, code corresponding to the results is provided in the supplementary file.

\item {\bf Experimental setting/details}
    \item[] Question: Does the paper specify all the training and test details (e.g., data splits, hyperparameters, how they were chosen, type of optimizer, etc.) necessary to understand the results?
    \item[] Answer: \answerYes{} 
    \item[] Justification: Yes, proper training details (data splits, hyper-parameters, optimizer, etc.) are provided in the Appendix. Furthermore, the code corresponding to the results is also provided in the supplementary file.

\item {\bf Experiment statistical significance}
    \item[] Question: Does the paper report error bars suitably and correctly defined or other appropriate information about the statistical significance of the experiments?
    \item[] Answer: \answerYes{} 
    \item[] Justification: Yes, the experiments are done across 3 random seeds for robustness in results, while providing necessary error bars.

\item {\bf Experiments compute resources}
    \item[] Question: For each experiment, does the paper provide sufficient information on the computer resources (type of compute workers, memory, time of execution) needed to reproduce the experiments?
    \item[] Answer: \answerYes{} 
    \item[] Justification: Yes, necessary compute resources specification are provided in the Appendix.
    
\item {\bf Code of ethics}
    \item[] Question: Does the research conducted in the paper conform, in every respect, with the NeurIPS Code of Ethics \url{https://neurips.cc/public/EthicsGuidelines}?
    \item[] Answer: \answerYes{} 
    \item[] Justification: Yes, the paper conforms, in every respect, with the NeurIPS Code of Ethics

\item {\bf Broader impacts}
    \item[] Question: Does the paper discuss both potential positive societal impacts and negative societal impacts of the work performed?
    \item[] Answer: \answerYes{} 
    \item[] Justification: Yes, we have added a seperate section on discussing broader impacts and limitations of our work.
    
\item {\bf Safeguards}
    \item[] Question: Does the paper describe safeguards that have been put in place for responsible release of data or models that have a high risk for misuse (e.g., pretrained language models, image generators, or scraped datasets)?
    \item[] Answer: \answerNA{} 
    \item[] Justification: \answerNA{}

\item {\bf Licenses for existing assets}
    \item[] Question: Are the creators or original owners of assets (e.g., code, data, models), used in the paper, properly credited and are the license and terms of use explicitly mentioned and properly respected?
    \item[] Answer: \answerYes{} 
    \item[] Justification: Yes, all existing datasets are properly cited in the paper. 

\item {\bf New assets}
    \item[] Question: Are new assets introduced in the paper well documented and is the documentation provided alongside the assets?
    \item[] Answer: \answerYes{} 
    \item[] Justification: Yes, proper documented code is provided for the new experiments done in the paper.

\item {\bf Crowdsourcing and research with human subjects}
    \item[] Question: For crowdsourcing experiments and research with human subjects, does the paper include the full text of instructions given to participants and screenshots, if applicable, as well as details about compensation (if any)? 
    \item[] Answer: \answerNA{} 
    \item[] Justification: \answerNA{}

\item {\bf Institutional review board (IRB) approvals or equivalent for research with human subjects}
    \item[] Question: Does the paper describe potential risks incurred by study participants, whether such risks were disclosed to the subjects, and whether Institutional Review Board (IRB) approvals (or an equivalent approval/review based on the requirements of your country or institution) were obtained?
    \item[] Answer: \answerNA{} 
    \item[] Justification: \answerNA{}

\item {\bf Declaration of LLM usage}
    \item[] Question: Does the paper describe the usage of LLMs if it is an important, original, or non-standard component of the core methods in this research? Note that if the LLM is used only for writing, editing, or formatting purposes and does not impact the core methodology, scientific rigorousness, or originality of the research, declaration is not required.
    \item[] Answer: \answerNA{} 
    \item[] Justification: \answerNA{}

\end{enumerate}

\end{document}